\documentclass[oneeqnum, onethmnum, onefignum, onetabnum]{article}

\usepackage[english]{babel}
\usepackage{amssymb,fancyhdr,amsfonts,amssymb,amsthm,amsmath,epsfig,enumerate}
\usepackage{amsmath, amssymb, graphicx, multirow, subcaption, caption, hyperref, rotating,color}

\usepackage[top=1.5in, bottom=1.25in, left=1.25in, right=1.25in]{geometry}

\def\R{\mathbb{R}}

\def\u{\mathbf{u}}

\def\x{\mathbf{x}}
\def\y{\mathbf{y}}

\def\p{\mathbf{p}}
\def\q{\mathbf{q}}
\def\t{\mathbf{t}}

\def\div{\text{div}}

\graphicspath{{./figures/}}

\title{A Survey of Pansharpening Methods with A New Band-Decoupled Variational Model\thanks{This work was supported by the Ministerio de Ciencia e Innovaci\'on under grants TIN2011-27539  and TIN2014-53772-R, and by the Centre National d'{\'E}tudes Spatiales through the project ``Optimisation bord/sol d{\'e}bruitage et d{\'e}mosa{\"i}quage'' R-S13/OT-0001-098.}}

\author{J. Duran\footnotemark[2] \and A. Buades\footnotemark[2] \and B. Coll\footnotemark[2] \and C. Sbert\footnotemark[2] \and G. Blanchet\footnotemark[3]}

\begin{document}

\maketitle

\renewcommand{\thefootnote}{\fnsymbol{footnote}}

\footnotetext[2]{Universitat de les Illes Balears, Department of Mathematics and Computer Science, Anselm Turmeda, Ctra. de Valldemossa km. 7.5, 07122 Palma de Mallorca, Spain (joan.duran@uib.es, toni.buades@uib.es, tomeu.coll@uib.es, catalina.sbert@uib.es).}
\footnotetext[3]{Centre National d'{\'E}tudes Spatiales, 18, Av.~Edouard Belin, 31055 Toulouse, France (gwendoline.blanchet@cnes.fr).}

\renewcommand{\thefootnote}{\arabic{footnote}}

\begin{abstract}
Most satellites decouple the acquisition of a panchromatic image at high spatial resolution from the acquisition of a multispectral image at lower spatial resolution. Pansharpening is a fusion technique used to increase the spatial resolution of the multispectral data while simultaneously preserving its spectral information. In this paper, we consider pansharpening as an optimization problem minimizing a cost function with a nonlocal regularization term. The energy functional which is to be minimized decouples for each band, thus permitting the application to misregistered spectral components. This requirement is achieved by dropping the, commonly used, assumption that relates the spectral and panchromatic modalities by a linear transformation. Instead, a new constraint that preserves the radiometric ratio between the panchromatic and each spectral component is introduced. An exhaustive performance comparison of the proposed fusion method with several classical and state-of-the-art pansharpening techniques illustrates its superiority in preserving spatial details, reducing color distortions, and avoiding the creation of aliasing artifacts. 
\end{abstract}

\pagestyle{myheadings}
\thispagestyle{plain}

\section{Introduction}

\begin{figure}[t!]
        \centering
        \begin{subfigure}[c]{0.4\textwidth}
                \includegraphics[width=1\textwidth]{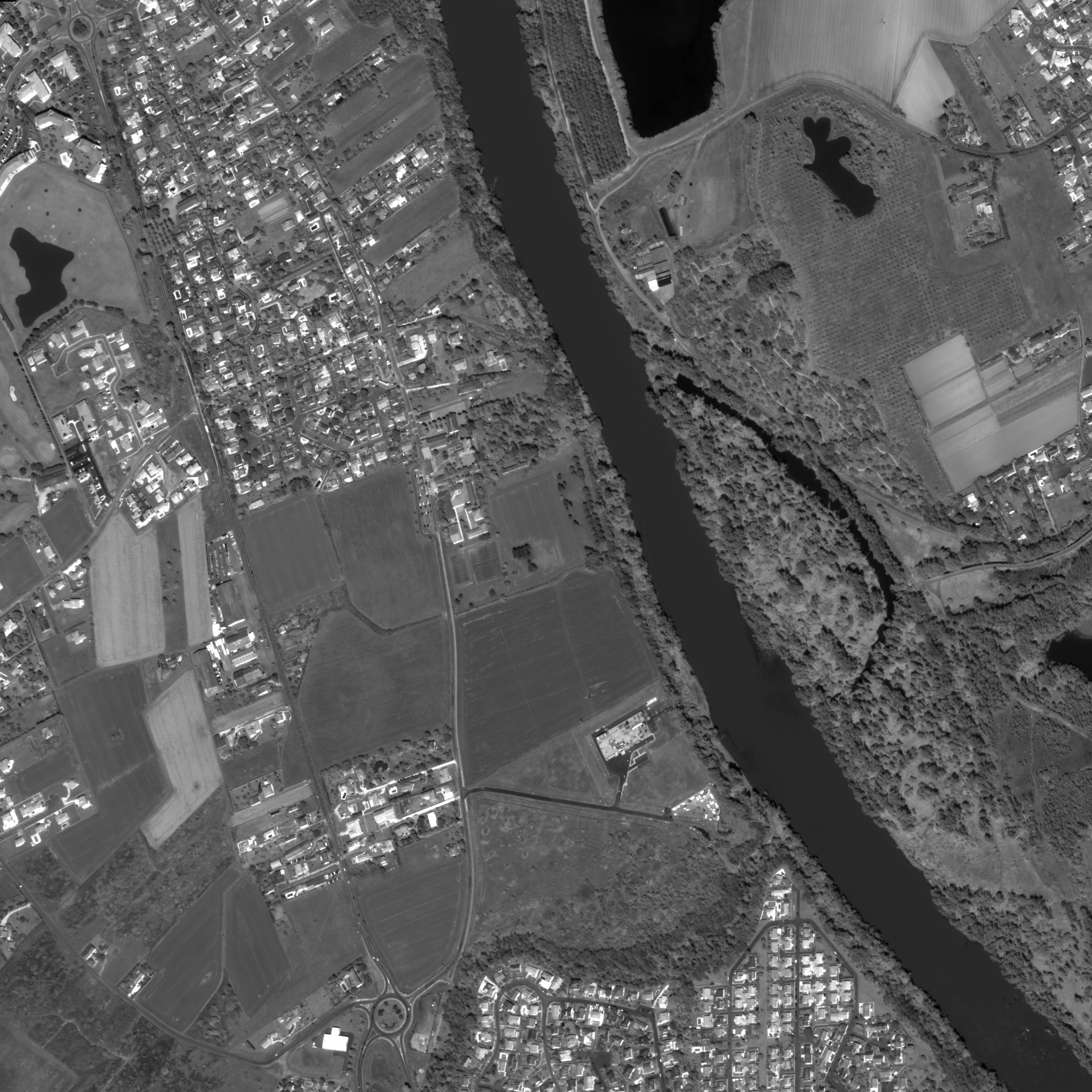}
                \caption*{Panchromatic}
        \end{subfigure}
        \hskip 0.01in
        \begin{subfigure}[c]{0.4\textwidth}
               \renewcommand{\arraystretch}{0.2}
		\begin{tabular}{c@{\hskip 0.01in}c}
			\includegraphics[width=0.498\textwidth]{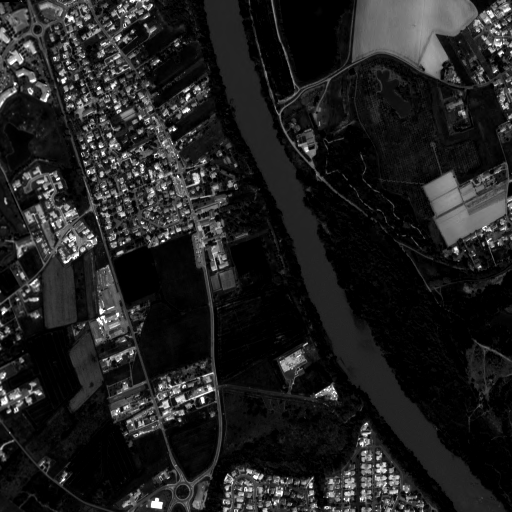} & 
			\includegraphics[width=0.498\textwidth]{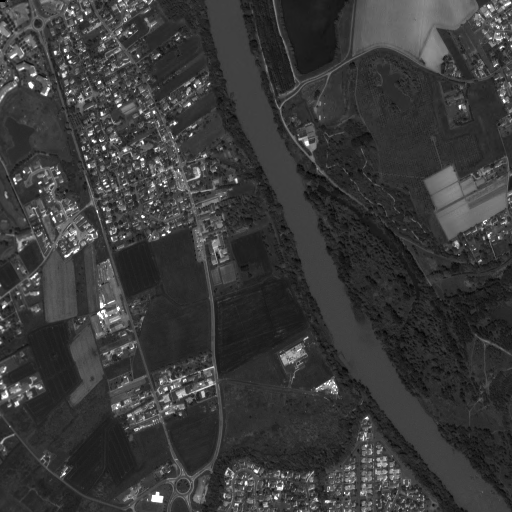} \\
			\includegraphics[width=0.498\textwidth]{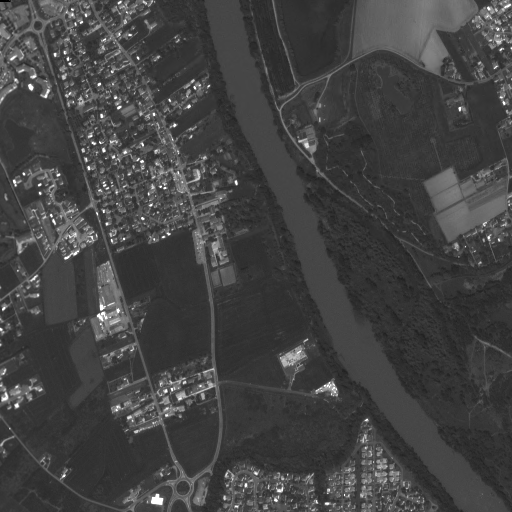} &
			\includegraphics[width=0.498\textwidth]{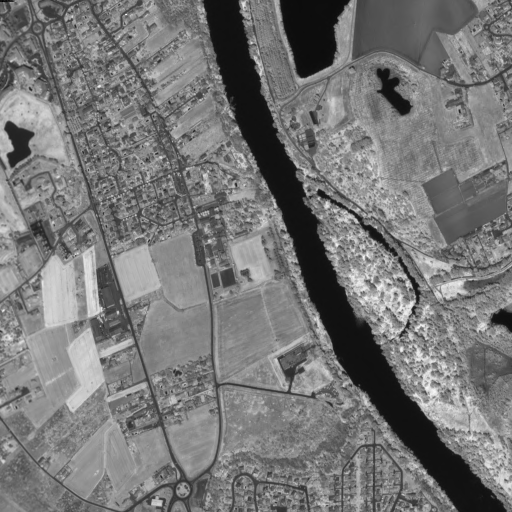}
			\\
		\end{tabular}
               \caption*{\hspace{0.25cm} Red, green, blue, near-infrared}
        \end{subfigure}
        	\caption{Pl{\'e}iades scene of Toulouse (France) provided by Centre National d'{\'E}tudes Spatiales (CNES). The spatial resolution is $70$ cm per pixel for the panchromatic and $2.8$ m per pixel for each blue, green, red, and near-infrared band.}
\label{fig:datasetPleiades}
\end{figure}

Many Earth observation satellites provide continuously growing quantities of remote sensing images useful for a wide range of both scientific and everyday tasks. Most of them, such as Ikonos, Landsat, Quickbird, and Pl{\'e}iades, decouple the acquisition of a panchromatic image at high spatial resolution from the acquisition of a multispectral image at lower spatial resolution. The wide range of wavelengths acquired by the panchromatic represents an accurate description of the geometry of the image, while each spectral component covers a reduced bandwidth range leading to a detailed color description. Spectral sensors typically produces larger pixel sizes, thus increasing the signal noise ratio of spectral images and reducing the transmission cost. As an example, Figure \ref{fig:datasetPleiades} displays the data captured by the Pl{\'e}iades satellite and furnished to us by the {\it Centre National d'{\'E}tudes Spatiales} (CNES). In this setting, pansharpening is the fusion process by which a high-resolution multispectral image is inferred.

In remote sensing, high spatial resolution is necessary to correctly detect shapes, edges and, in general, geometric structures, but different types of land are better classified using images with multiple spectral bands. Considering this trade-off, state-of-the-art techniques \cite{ThomasRanchinWald2008, PohlGenderen2015, VivoneAlparoneChanussot2015} aim at increasing the spatial resolution of the multispectral data by using the high frequencies of the companion panchromatic. In the literature, pansharpening methods are mainly labeled into two main classes, namely component substitution (CS) and multiresolution analysis (MRA). The former relies on the use of a color decorrelation transform that converts the upsampled low-resolution channels into a new color system that separates the spatial and the spectral details. Fusion occurs by partially or totally substituting the component which is supposed to contain the spatial geometry by the panchromatic and applying the transformation back. Examples of CS methods include Intensity-Hue-Saturation (IHS) transform \cite{CarperLillesand1990, TuSuShyu2001, TuHuangHungChang2004}, Principal-Component-Analysis (PCA) transform \cite{ChavezKwarteng1989, ChavezSides1991, Shettigara1992}, Gram-Schmidt (GS) orthonormalization \cite{LabenBrower2000, AiazziBarontiSelva2007}, Brovey's \cite{HalladaCox1983, GillespieKahle1987}, band-dependent spatial detail (BDSD) \cite{GarzelliNenciniCapobianco2008}, and partial replacement adaptive CS (PRACS) \cite{ChoiYuKim2011}. On the contrary, MRA-based approaches inject the high frequencies of the panchromatic into the upsampled spectral components through a multiresolution decomposition. The fusion techniques from this family mainly differ in how the low-pass version of the panchromatic is generated at each scale. Laplacian pyramid \cite{AiazziAlparone2002, AiazziAlparoneBaronti2006, LeeLee2010}, contourlet transform \cite{ShahYounan2008}, curvelet transform \cite{NenciniGarzelli2007}, discrete wavelet transform \cite{Mallat1989, Shensa1992, Yocky1995, NunezOtazuFors1999, RanchinWald2000, OtazuGonzalezFors2005, VivoneRestaino2014}, high-pass filtering (HPF) \cite{ChavezSides1991, BethuneMuller1998, Schowengerdt2006, KhanAlparone2009}, and high-pass modulation (HPM) \cite{Liu2000, WaldRanchin2002, Schowengerdt2006} are most widely used.

The main challenging task of pansharpening techniques is to get a good compromise between spatial and spectral quality. The two classes of methods described above exhibit complementary spectral-spatial quality trade-off. Although CS family is usually characterized by a high fidelity in rendering the spatial details in the final product \cite{AiazziBarontiSelva2007}, it often suffers from significant spectral distortion. This is due to the fact that the panchromatic image does not cover exactly the same wavelengths as the spectral sensors \cite{ThomasRanchinWald2008, AmroMateosVega2011, VivoneAlparoneChanussot2015}. On the contrary, MRA-based fusion aims at preserving the whole content of the low-resolution data and adding further information obtained from the panchromatic through spatial filtering  \cite{RanchinWald2000}. In contrast to CS, MRA family is more successful in spectral preservation but it often experiences spatial distortions like ringing or staircasing effects  \cite{ThomasRanchinWald2008, AmroMateosVega2011, VivoneAlparoneChanussot2015}. However, as pointed out by Aiazzi {\it et al,} \cite{AiazziAlparoneBaronti2006}, if the frequency response of the low-pass filter used in the multiscale decompostion matches the Modulation Transfer Function (MTF) of the spectral channel into which details are injected, the spatial enhancement of MRA-based methods is comparable to that of CS.

Variational techniques have recently emerged as a promising direction of research since they effectively combine aspects of different methods into a single mathematical framework. Ballester {\it et al.} \cite{BallesterCaselles2006} were the first to introduce a variational formulation for pansharpening, which they called P+XS. The authors assumed that the low-resolution channels are formed from the underlying high-resolution ones by low-pass filtering followed by subsampling. They considered a regularization term forcing the edges of each spectral band to line up with those of the panchromatic. Furthermore, P+XS functional incorporated an additional term according to which the panchromatic is a linear combination of the spectral components which are to be computed. Duran {\it et al.} \cite{DuranBuadesCollSbertSIIMS2014} proposed to keep the variational formulation introduced by Ballester {\it et al.} \cite{BallesterCaselles2006} while incorporating nonlocal regularization that takes advantage of image self-similarities and leds to a significant reduction of color artifacts. In this setting, the panchromatic image is used to derive relationships among patches describing the geometry of the desired fused image. The general idea of diffusing a color image conditionally to the geometry of any other, in particular, to the geometry of its associated grayscale intensity image, was originally proposed by Buades {\it et al.} \cite{BuadesCollLisani2007}. Several other variational models have been proposed so far \cite{LiYang2011, HeCondatChanussot2012, MoellerWittman2012, PalssonSveinsson2012, ZhuBamler2013, AlySharma2014, HeCondatBioucas2014, ZhangFang2015}.  A detailed overview of variational techniques is given in Section \ref{sec:variational}.

\begin{figure}[!t]
        \centering
        \begin{tabular}{cc}
        \includegraphics[trim= 5cm 34cm 55cm 26cm, clip=true, width=0.4\textwidth]{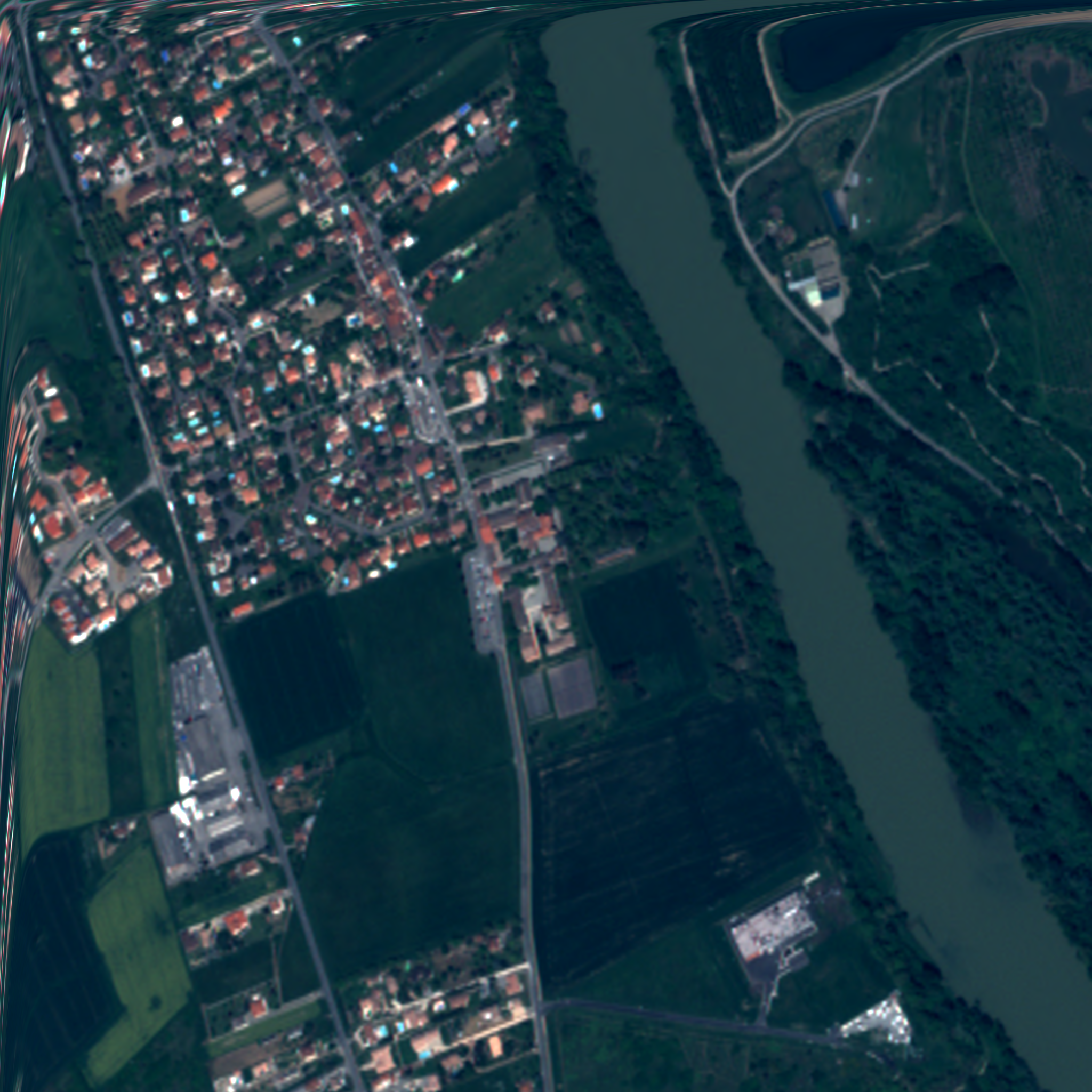}
        \includegraphics[trim= 20cm 47cm 41cm 14cm, clip=true, width=0.4\textwidth]{pleiades_bicubic.png}
        \end{tabular}
        	\caption{Upsampled spectral data, extracted from the same scene as in Figure \ref{fig:datasetPleiades}, where all bands have been registered into a common geometry. Note that strong aliasing is apparent in both images.}
	\label{fig:satellite_aliasing}
\end{figure}

Most of the pansharpening techniques previously mentioned make use of the linear combination assumption and need all data to be geometrically aligned. Unfortunately, both requirements are not satisfied by real satellite imagery, for which different spectral bands are not originally co-registered and their registration previously to pansharpening is not at all recommendable because of the strong aliasing. Indeed, the panchromatic and spectral bands are acquired according to the Push-Broom principle of CCD arrays placed in the focal plane of a telescope. The sensors are shifted within the focal plane in the direction of the satellite scrolling and the same point on the ground is not captured at the same time by all sensors or strictly under the same angle. Furthermore, one of the most relevant drawbacks of this acquisition system is the strong aliasing of the spectral bands, which usually produces jagged edges, color distortions, and stair-step effects. The MTF has low values near Nyquist for the panchromatic, thus almost avoiding undesirable aliasing effects. On the contrary, the MTF of the spectral bands having high values at Nyquist results in aliased spectral data as illustrated in Figure \ref{fig:satellite_aliasing}. Baronti {\it et al.}\cite{BarontiAiazzi2011} studied how several pansharpening methods proposed in the literature behave in the presence of misregistration and aliasing. Under general and likely assumptions, the authors proved that CS is less sensitive than MRA to these drawbacks whenever being of moderate extent.

In this paper, we propose a new nonlocal variational model for the pansharpening of real satellite images. Compared to the previous work \cite{DuranBuadesCollSbertSIIMS2014}, no assumption on the co-registration of spectral data is made. Furthermore, a new constraint imposing the preservation of the radiometric ratio between the panchromatic and each spectral band is introduced, replacing the classical linearity assumption. In practice, this energy term injects the high frequencies of the panchromatic into each high-resolution spectral component one seeks to estimate. The energy minimization can be performed independently for each channel, thus permitting the independent optimization of each spectral band and its application to misregistered and aliased spectral data. Being the functional strictly convex and quadratic, we design an efficient numerical scheme based on the gradient descent method.

The rest of the paper is organized as follows. Section \ref{sec:variational} introduces the variational formulation of pansharpening. We propose in Section \ref{sec:newmodel} a new nonlocal band-decoupled variational model that allows us to deal with misregistered and aliased spectral data. We also give detailed explanations on how to compute numerically the minimizer of the energy functional. Section \ref{sec:discussion} discusses the choices made in the design of the model, with especial attention to the validity of the linear combination and co-registration assumptions on real satellite imagery.  We perform an exhaustive comparison between the proposed model and the most relevant classical and state-of-the-art pansharpening techniques in Section \ref{sec:experiments}, followed by the conclusions in Section \ref{sec:conclusions}.

\section{Variational Formulation of Pansharpening}\label{sec:variational}

In this section, we review the variational formulations introduced in the literature for pansharpening image fusion, including the nonlocal regularization based model proposed by Duran {\it et al.} \cite{DuranBuadesCollSbertSIIMS2014}.

Let $\Omega$ be an open and bounded domain in $\mathbb{R}^M$, $M\geq 2$. We denote by $\mathbf{u}:\Omega\rightarrow\mathbb{R}^C$, with $\mathbf{u}(\x)=(u_1(\x), \ldots, u_C(\x))$ for any $\x\in\Omega$, the high-resolution image with $C$ spectral bands one seeks to estimate. In this setting, $u_k:  \Omega\rightarrow \mathbb{R}$ represents the intensity corresponding to the $k$-th spectral component. The available data from the satellite consists of a high-resolution panchromatic image $P:\Omega\rightarrow\mathbb{R}$ and a low-resolution multispectral image defined on a sampling grid $S\subseteq \Omega$ and denoted by $\u^S:\Omega\rightarrow\R^C$, with $\u^S(\x)=(u_1^S(\x), \ldots, u_C^S(\x))$ for any $\x\in S$. The purpose is to reconstruct $\mathbf{u}$  from $P$ and $\u^S$.

\subsection{Pansharpening as an Ill-Posed Inverse Problem}

The most common image formation model, pioneered by Ballester {\it et al.} \cite{BallesterCaselles2006}, assumes that the low-resolution multispectral image is formed from the high-resolution one by low-pass filtering followed by subsampling. Therefore, one has that
\begin{equation}\label{eq:spectralmodel}
u_k^{S} = (\kappa_k\ast u_k)^{\downarrow_s}+ \eta_k, \quad \forall k\in\{1,\ldots,C\},
\end{equation}
where $\downarrow_s$ denotes the subsampling operator by a factor $s$ (for most satellites, $s=4$), $\kappa_k$ is the impulse response for the $k$th spectral band, and $\eta_k$ is supposed to be i.i.d zero-mean Gaussian noise. Note that \eqref{eq:spectralmodel} is an ill-posed inverse problem in the sense that the information provided by $\mathbf{u}^S$ and the image observation model is not sufficient to ensure the existence, uniqueness, and stability of a solution $\mathbf{u}$.  These properties will be guaranteed by the introduction of a good prior and an optimization formulation. 

In view of \eqref{eq:spectralmodel}, it is necessary to assume that it is possible to evaluate $\kappa_k\ast u_k$ at any point of $S$.  For that purpose, $\kappa_k$ is considered to be the kernel of a convolution operator mapping $L^2(\Omega)$ into $\mathcal{C}(\overline{\Omega})$, that is
$$ 
\kappa_k\ast v(\y)=\int_{\Omega} \kappa_k(\y-\x)v(\x)d\x,\quad \forall k\in\{1,\ldots, C\}, \quad \forall v\in L^2(\Omega).
$$
The data-fidelity requirement based on the image formation model \eqref{eq:spectralmodel} is then written as
\begin{equation}\label{eq:spectralterm}
\sum_{k=1}^C\int_{\Omega}\Pi_S \cdot \left( \kappa_k \ast u_k(\x) - u_k^{\Omega}(\x)\right)^2 d\x.
\end{equation}
In this setting, $\Pi_S =\sum_{\x\in S} \delta_\x$ is a Dirac's comb defined by the sampling grid $S$ and $\u^{\Omega}:\Omega\rightarrow\R^C$, with $\mathbf{u}^{\Omega}(\x)=\left(u_1^{\Omega}(\x), \ldots, u_{C}^{\Omega}(\x)\right)$ for any $\x\in\Omega$, is an arbitrary extension of $\mathbf{u}^S$ as a continuous function from the sampling grid $S$ to the whole domain $\Omega$. Note that the integral of a sum of Dirac's is unambiguous as one assumes that no point of $S$ belongs to the boundary of $\Omega$. Furthermore, since the integrand term is multiplied by $\Pi_S$, the integral expression \eqref{eq:spectralterm} does not depend on the particular extension chosen in $\u^{\Omega}$.

Ballester {\it et al.} \cite{BallesterCaselles2006} further assumed that the panchromatic can be approximated by a linear combination of the different bands of the high-resolution multispectral image which is to be computed. They introduced the following constraint:
\begin{subequations}\label{eq:panconstraintterm}
\begin{equation}\label{eq:panconstraint}
P(\x)=\sum_{k=1}^C\alpha_k u_k(\x),\quad \forall \x\in \Omega,
\end{equation}
where $\{\alpha_k\}$ are mixing coefficients that give the intensity image in terms of the spectral channels, satisfying $\alpha_k\geq 0$ for all $k\in\{1,\ldots,C\}$ and $\sum_{k} \alpha_k=1$. The above constraint is equivalent to the variational formulation
\begin{equation}\label{eq:panterm}
\int_{\Omega} \left(\sum_{k=1}^C\alpha_k u_k(\x) - P(\x)\right)^2 d\x.
\end{equation}
\end{subequations}
Note that one implicitly assumes in \eqref{eq:panconstraintterm} that the panchromatic image and all spectral bands are geometrically aligned. 

Inspired by the pansharpening formulation introduced by Ballester {\it et al.} \cite{BallesterCaselles2006}, several other variational approaches have been proposed. Aly and Sharma \cite{AlySharma2014} based on \eqref{eq:spectralmodel} and \eqref{eq:panconstraintterm} but restricted the linear combination assumption \eqref{eq:panconstraint} to be only imposed on the high-pass filtered components. On the contrary, Palsson {\it et al.} \cite{PalssonSveinsson2012} redefined the image observation model \eqref{eq:spectralmodel}  by incorporating the classical constraint \eqref{eq:panconstraint} in it. He {\it et al.} \cite{HeCondatBioucas2014} used the same observation model than Palsson {\it et al.}\cite{PalssonSveinsson2012} but originally defined on the continuous reflectance spectra of the sensors. Furthermore, they relaxed the constraint given in \eqref{eq:panconstraint} by requiring a blurred version of the panchromatic image to be close to a linear combination of the blurred high-resolution channels.

M{\"o}ller {\it et al.} \cite{MoellerWittman2012} replaced the image formation model \eqref{eq:spectralmodel} by two data terms. The first one aims at preserving the chromaticity of $\u^S$ at the smooth parts of the image:
\begin{equation}\label{eq:moeller}
u_k(\x) =  \widetilde{u}_k(\x), \quad \forall \x\in\Omega\setminus\Gamma,
\end{equation}
where $\Gamma$ denotes the set of edges and texture in the panchromatic modality and, for each $k\in\{1,\ldots, C\}$, $\widetilde{u}_k:\Omega\to\R$ denotes the upsampling of the low-resolution band $u_k^S$ to the whole domain by, for instance, bicubic interpolation. If one assumes that $\widetilde{u}_k=  \kappa_k^{\top} \ast (u_k^S)^{\uparrow^s}$, where $\uparrow^s$ corresponds to the replication of each pixel $s-1$ times along horizontal and vertical directions and $\kappa^{\top}$ is the adjoint kernel to $\kappa_k$, then \eqref{eq:moeller} is nothing more than the adjoint to \eqref{eq:spectralmodel} but restricted to $\Omega\setminus\Gamma$. The second data term introduced by M{\"o}ller {\it et al.} \cite{MoellerWittman2012} consists in matching the high-level wavelet coefficients and the low-level approximation coefficients of the sought solution with those of the panchromatic image and the low-resolution multispectral bands, respectively. Importantly, M{\"o}ller\cite{MoellerWittman2012} eliminated the linearity constraint on the panchromatic. Instead, they preserved the spectral correlation by keeping the ratio of all spectral bands constant:
\begin{equation}\label{eq_moellerconstraint}
\dfrac{u_i(\x)}{u_j(\x)} = \dfrac{\widetilde{u}_i(\x)}{\widetilde{u}_j(\x)}, \quad \forall \x\in\Omega, \quad \forall i,j\in\{1,\ldots, C\}, \,i\neq j.
\end{equation}
The above constraint is equivalent to minimizing the spectral angle, which is widely used to measure the spectral distortion of fused products by means of the quality metric SAM \cite{AlparoneBaronti2004}, between each pixel frequency vector in the low-resolution and in the pansharpened multispectral images. 

Zhang {\it et al.} \cite{ZhangFang2015} considered the original observation model \eqref{eq:spectralmodel} but for which the kernel used by the satellite to aberrant the low-resolution data is not prescribed. The authors further dropped the classical assumption \eqref{eq:panconstraint} and used the constraint \eqref{eq_moellerconstraint} instead. 

It is worth noticing that the equations given in \eqref{eq:panterm} and \eqref{eq_moellerconstraint} require all spectral bands to be co-registered.

\subsection{Classical Regularization Strategies}

In their pioneering work, Ballester {\it et al.} \cite{BallesterCaselles2006} proposed to regularize the solution of the ill-posed inverse problem \eqref{eq:spectralmodel} by aligning all level lines of the panchromatic and each high-resolution multispectral band, that is,
\begin{equation}\label{eq_regBallester}
\sum_{k=1}^C\int_{\Omega} \big( |\nabla u_k(\x)| + \langle \div( \theta(\x)), u_k(\x)\rangle\big) \, d\x,
\end{equation}
where $\theta$ is the vector field that consists of all unit normal vectors of the level sets of the panchromatic image. Several other variational approaches in pansharpening \cite{MoellerWittman2012, AlySharma2014, ZhangFang2015} incorporated \eqref{eq_regBallester} as regularization term to the corresponding energy minimization based models.

Palsson {\it et al.} \cite{PalssonSveinsson2012} asked the solution arising from their observation model to have minimal total variation (TV), a prior that accounts for images having smooth transitions and which was originally proposed by Rudin, Osher and Fatemi \cite{ROF1992} for image denoising. Palsson {\it et al.} \cite{PalssonSveinsson2012} introduced the following band-decoupled regularization:
$$
\sum_{k=1}^C\int_{\Omega} \left|\nabla u_k(\x)\right|\, d\x.
$$
He {\it et al.} \cite{HeCondatChanussot2012} proposed to add the gradient of the panchromatic image into the total variation functional. Instead of penalizing the oscillations of each spectral band independently, the proposed term couples the regularization of the spectral and panchromatic components as follows:
$$
\int_{\Omega}\sqrt{\sum_{k=1}^C \left|\nabla u_k(\x)\right|^2 + \gamma^2 \left|\nabla P(\x)\right|^2} \, d\x,
$$
where $\gamma$ is a parameter that weights the contribution of the panchromatic in the regularization term. Based on the latter, He {\it et al.}\cite{HeCondatBioucas2014} exploited appropriate regularizations based on both spatial and spectral links between the panchromatic and the fused product.

Recent developments in compressive sensing have also been carried out \cite{LiYang2011, ZhuBamler2013} for the fusion problem.

\subsection{Nonlocal Regularization}

All variational techniques discussed previously describe regularity in terms of the local relationships of nearby pixels, mainly the gradient or the Laplacian of the image. The total variation \cite{ROF1992} is the most significant of such methods and pioneered as a discontinuity-preserving regularization in the sense that it assigns the same energy cost to sharp and smooth transitions. Although it is optimal to reconstruct the main geometrical shape in an image, it fails to preserve fine structures, details, and texture.

In contrast to the local case, the so-called non-local methods, make any point  interact directly with any other point in the whole domain. The closeness relationship is replaced by a similarity measure relating points having similar geometry and texture characteristics. Inspired by the success of the nonlocal-means denoising algorithm \cite{BuadesCollMorell2011}, Gilboa {\it et al.} \cite{GilboaOsher2007, GilboaOsher2008} and Kindermann {\it et al.} \cite{KindermannOsher2005} interpreted neighborhood filters as regularizations based on nonlocal operators. Nonlocal based approaches were also proposed for other applications somehow related to pansharpening, such as super-resolution \cite{EladDatsenko2009}, inpainting \cite{arias2011variational}, and demosaicking \cite{DuranBuadesTIP2014}.

Duran {\it et al.} \cite{DuranBuadesCollSbertSIIMS2014} introduced  a nonlocal regularization term taking advantage of the self-similarity principle on natural images. The corresponding energy term is  given by
\begin{equation}\label{eq:nonlocalterm}
\sum_{k=1}^C\iint_{\Omega\times\Omega} \left(u_k(\y)-u_k(\x)\right)^2\omega_P(\x,\y) \, d\y \, d\x.
\end{equation}
where the similarity distribution $\omega_P(\x,\y)$ is computed on the panchromatic image. The high-resolution panchromatic image is used to derive relationships among patches describing the geometry of the image. The weight $\omega_P:\Omega\times\Omega\rightarrow\R$ is defined as
\begin{subequations}\label{eq:weights}
\begin{equation}\label{eq:weightsdef}
\omega_P(\x, \y)=\dfrac{1}{\Upsilon(\x)} \exp\left( -\dfrac{d_{\rho}\left(P(\x)-P(\y)\right)}{h^2}\right),
\end{equation}
where
\begin{equation}\label{eq:weightsnormalization}
\Upsilon(\x)=\int_{\Omega}\exp\left(-\frac{d_{\rho}\left(P(\x), P(\y)\right)}{h^2}\right)\, d\y, \quad \forall \x\in\Omega
\end{equation}
is a normalization factor and
\begin{equation}\label{eq:weightsdist}
d_{\rho}\left(P(\x), P(\y)\right)=\int_{\Omega}G_{\rho}(\t)|P(\x+\t)-P(\y+\t)|^2\,d\t
\end{equation}
\end{subequations}
computes the distance between neighborhoods (or patches) around $\x$ and $\y$. In this framework, $G_{\rho}$ is a Gaussian kernel and $h>0$ acts as a filtering parameter. The latter controls the decay of the exponential function and, therefore, quantifies how fast the weights decrease with increasing dissimilarity of patches. In the end, the average made between very similar regions preserves the integrity of the image but reduces its small fluctuations, which contain noise. Note that the weight in \eqref{eq:weights} satisfies the usual conditions $0< \omega(\x,\y)\leq 1$ and $\int_{\Omega} \omega(\x,\y)\,d\y=1$, but the normalization using \eqref{eq:weightsnormalization} breaks down the symmetry between to given points in the image. In this regard, Duran {\it et al.} \cite{DuranBuadesCollSbertSIIMS2014} provided a rigorous vector calculus for nonlocal operators defined in terms of nonnegative and nonsymmetric weights.

Considering the fidelity term \eqref{eq:spectralterm} derived from the image formation model, the nonlocal regularization term \eqref{eq:nonlocalterm}, and the Lagrangian formulation \eqref{eq:panterm} related to the linearity constraint on the panchromatic image, Duran {\it et al.} \cite{DuranBuadesCollSbertSIIMS2014} proposed to minimize the following energy functional:
\begin{equation}\label{eq:functional1}
\begin{aligned}
J_1(\u) &= \dfrac{1}{2} \sum_{k=1}^C \iint_{\Omega\times\Omega}\left(u_k(\y)-u_k(\x)\right)^2\omega_P(\x,\y)\,d\y\,d\x \\
&+\dfrac{\lambda}{2}\int_{\Omega}\left( \sum_{k=1}^C \alpha_k u_k(\x)-P(\x)\right)^2d\x  \\
&+ \dfrac{\mu}{2} \sum_{k=1}^C\int_{\Omega} \Pi_S \cdot \left( \kappa_k\ast u_k(\x)-u^{\Omega}_k (\x)\right)^2 d\x,
\end{aligned}
\end{equation}
where $\lambda\geq 0$ and $\mu\geq 0$ are trade-off parameters controlling the contribution of each term to the whole energy. The existence and uniqueness of minimizer was guaranteed and a gradient descent method was used to compute the solution.

\section{Nonlocal Band-Decoupled Variational Model}\label{sec:newmodel}

In this section, we propose a nonlocal variational approach for dealing with misregistered spectral bands. We drop \eqref{eq:panconstraint} and incorporate a new constraint that imposes the preservation of the ratio between the panchromatic and each spectral component. For that purpose, we take into consideration the {\it Wald's protocol} \cite{WaldRanchin1997} according to which the low-frequency components of the fused product can be obtained by upsampling the low-resolution multispectral image to the high-resolution domain.

\subsection{Radiometric Constraint}

In order to preserve the geometry of the observed scene, we propose to keep the radiometric ratio between the panchromatic and each spectral band.  More concretely, we first compute the ratio between each low-resolution spectral component and a decimated panchromatic. Then, we ask to this ratio to be similar to the ratio of the original panchromatic and each band of the fused product.

For each $k\in\{1,\ldots,C\}$, let $P_k$ be the panchromatic image expressed in the same reference of $u_k$ and let $P^S_k: S\rightarrow\mathbb{R}$ be its decimation by the same downsampling process than $u_k^S$. Let $\widetilde{P}_k:\Omega\rightarrow\mathbb{R}$ and $\widetilde{u}_k:\Omega\rightarrow\mathbb{R}$ be the respective extensions of $P^S_k$ and $u^S_k$ to the whole domain by bicubic interpolation. We encourage that
\begin{subequations}\label{eq:spatialratioall}
\begin{equation}\label{eq:spatialratio}
\dfrac{u_k(\x)}{P_k(\x)} = \dfrac{\widetilde{u}_k(\x)}{\widetilde{P}_k(\x)},\quad\forall \x\in\Omega,\quad \forall k\in\{1,\ldots,C\}.
\end{equation}
It is important to emphasize that we are only assuming in the above condition that each $P_k$ is geometrically aligned with the corresponding $k$th spectral band, but nothing about the co-registration of the spectral data. Finally, casting \eqref{eq:spatialratio} in a variational framework leads to the integral expression
\begin{equation}\label{eq:spatialterm}
\int_{\Omega} \left( u_k(\x)\widetilde{P}_k(\x) - \widetilde{u}_k(\x)P_k(\x)\right)^2 d\x,\quad  \forall k\in\{1,\ldots, C\},
\end{equation}
\end{subequations}
which is to be minimized. Note that we have considered the general case in which the impulse response is different for each spectral sensor.

We observe that the expression \eqref{eq:spatialratio} can be written in the form
\begin{subequations}\label{eq:MRAterm}
\begin{equation}\label{eq:MRAterm1}
u_k(\x) = \dfrac{\widetilde{u}_k(\x)}{\widetilde{P}_k(\x)} P_k(\x), \quad \forall \x\in \Omega,\quad \forall k\in\{1,\ldots,C\}.
\end{equation}
By subtracting $\widetilde{u}_k(\x)$ to each side of the above equation, we obtain
\begin{equation}\label{eq:MRAterm2}
u_k(\x) - \widetilde{u}_k(\x) = \dfrac{\widetilde{u}_k(\x)}{\widetilde{P}_k(\x)} \left(P_k(\x) - \widetilde{P}_k(\x)\right), \quad\forall \x\in \Omega, \quad \forall k\in\{1,\ldots,C\},
\end{equation}
\end{subequations}
where $P_k-\widetilde{P}_k$ accounts for the high frequencies of the panchromatic image. Accordingly, we force the high-frequency components of each band, that is $u_k-\widetilde{u}_k$, to coincide with those of the panchromatic and, consequently, the spatial details of the panchromatic are injected into the fused product. On the other hand, the modulation coefficient $\frac{\widetilde{u}_k(\x)}{\widetilde{P}_k(\x)}$ takes the energy levels of the panchromatic and multispectral images into account, which can be different for each spectral band.

Interestingly, the constraints in \eqref{eq:MRAterm} fit in with the general formulation of MRA-based pansharpening techniques. In particular, \eqref{eq:MRAterm1} is equivalent to high-pass modulation methods \cite{Liu2000, WaldRanchin2002, Schowengerdt2006} in which the local intensity contrast of the panchromatic is reproduced in the fused product by weighting the spatial details by the ratio of the upsampled low-resolution data and the low-pass panchromatic before injection. Furthermore, since we use a single linear time-invariant low-pass filter for computing $\widetilde{P}_k$, equation \eqref{eq:MRAterm2} behaves as a high-pass filtering method \cite{ChavezSides1991, BethuneMuller1998, Schowengerdt2006, KhanAlparone2009}.

\subsection{The Energy Functional}

By taking into account the fidelity term imposed by the data generation model \eqref{eq:spectralterm} and the nonlocal regularization term \eqref{eq:nonlocalterm}-\eqref{eq:weights}, we propose to incorporate the Lagrangian formulation \eqref{eq:spatialterm} associated to the radiometric constraint \eqref{eq:spatialratio} into the final functional. Therefore, the problem consists in the minimization of the band-decoupled energy
\begin{subequations}\label{eq:functional2}
\begin{equation}
J_2(\u) = \sum_{k=1}^C J_2(u_k)
\end{equation}
such that the cost function for each spectral band $u_k$, $k\in\{1,\ldots, C\}$, is defined as
\begin{equation}
\begin{aligned}
J_2(u_k) &= \dfrac{1}{2} \iint_{\Omega\times\Omega} \left(u_k(\y)-u_k(\x)\right)^2\omega_{P_k}(\x,\y)\,d\y\,d\x \\
&+ \dfrac{\mu s^2}{2} \int_{\Omega} \Pi_S\cdot\left( \kappa_k\ast u_k(\x)-u^{\Omega}_k (\x)\right)^2 d\x\\
&+ \dfrac{\delta}{2\|P_k\|^2} \int_{\Omega} \left( u_k(\x) \widetilde{P}_k(\x) - \widetilde{u}_k(\x)P_k(\x)\right)^2 d\x,
\end{aligned}
\end{equation}
\end{subequations}
with the weight distribution in \eqref{eq:weights} being computed on each $P_k$. In this setting, $\mu\geq0$ and $\delta\geq 0$, which are respectively normalized by the sampling factor $s$ and the mean value of the panchromatic image, $\|P_k\|=\sqrt{\frac{1}{|\Omega|}\int_{\Omega}\left(P_k(\x)\right)^2 d\x}$, define the contribution of each term to the whole energy. In the case all data is co-registered, each $P_k$ denotes the original panchromatic image $P$.

Following the same arguments than Duran {\it et al.} \cite{DuranBuadesCollSbertSIIMS2014}, it can be proved that the proposed functional is proper, strictly convex, coercive, and lower semicontinuous. We can thus establish, using standard arguments in convex analysis \cite{EkelandTemam1999, Dacorogna2008}, that the optimization problem \eqref{eq:functional2} admits an unique solution in the class of $L^2(\Omega)-$weighted functions. Furthermore, if $\u=(u_1,\ldots, u_C)$ is the minimizer of \eqref{eq:functional2}, then it solves the Euler-Lagrange equation
\begin{equation}\label{eq:ELeq2}
\begin{aligned}
0 = &-\int_{\widetilde{\Omega}}\left(u_k(\y)-u_k(\x)\right)\left(\omega_{P_k}(\x,\y)+\omega_{P_k}(\y,\x)\right)\, d\y\\
&+ \mu s^2  \left(\kappa_k^{\top}\ast\left( \Pi_S\cdot\left(\kappa_k\ast u_k-u_k^{\Omega}\right)\right)\right)(\x) \\
&+\dfrac{\delta}{\|P_k\|^2} \widetilde{P}_k(\x)\left( u_k(\x)\widetilde{P}_k(\x)-\widetilde{u}_k(\x)P_k(\x)\right), \quad \forall \x\in\Omega, \quad \forall k\in\{1,\ldots,C\},
\end{aligned}
\end{equation}
where $\widetilde{\Omega}=\Omega\cup\Gamma$ denotes the domain under consideration together with a nonlocal boundary $\Gamma\in\R^M\setminus\Omega$, that is, a collar domain surrounding $\Omega$ with finite nonzero volume. The above equation allows designing an efficient optimization algorithm in next subsection.

\subsection{Numerical Minimization} \label{sec:numerics}

Let us suppose that the panchromatic is defined on a high-resolution discrete grid $I$ of size $N\times N$ pixels, and let $u_1^S,\ldots, u_C^S$ be the spectral components defined on a lower resolution grid $S$ of size $\frac{N}{s}\times \frac{N}{s}$, where $s$ is the sampling factor. Although we use the same notations than in the continuous framework, here an image has to be understood as a two-dimensional matrix in $\R^{N\times N}$ rearranged from left to right and from top to bottom into a vector of size $N^2$. Therefore, we use $u(\p)$, with $\p=(p_1,p_2)$, to denote the element in the vector $u\in\R^{N^2}$ living in the position $p_1N+p_2$.

In the discrete setting, the proposed nonlocal band-decoupled functional is given by
\begin{equation}\label{eq:discfunctional1}
\begin{aligned}
F(u_k)&= \dfrac{1}{2}  \sum_{\p,\q\in I} (u_k(\q)-u_k(\p))^2\omega_{P_k}(\p,\q) + \dfrac{\mu s^2}{2} \sum_{\p\in I} \Pi_S(\p)\left((K_k u_k)(\p)-u_k^{\Omega}(\p)\right)^2\\
&+ \dfrac{\delta}{2\|P_k\|^2} \sum_{\p\in I} \left( u_k(\p)\widetilde{P}_k(\p) - \widetilde{u}_k(\p)P_k(\p)\right)^2,
\end{aligned}
\end{equation}
where $K_k$ is the $N^2\times N^2$ matrix associated to the kernel $\kappa_k$ for each $k\in\{1,\ldots, C\}$. The Dirac's comb $\Pi_S$ is considered here as an $N\times N$ matrix such that
$$
\Pi_S(\p)=\left\lbrace\begin{array}{ll} 1 & \text{if } \p\in S, \\ 0 & \text{otherwise}, \end{array}\right. \quad \forall\p\in I.
$$
In practice, it is implemented by taking every fourth pixel to be one  along each direction. Furthermore, $\u^{\Omega}$ is an extension of $\u^S$ to the grid $I$ by means of, for instance, a simple replication by $s$ factor.

In order to minimize numerically \eqref{eq:discfunctional1}, the procedure uses a parabolic equation with time as an evolution parameter, or equivalently, the gradient descent method. For the sake of understanding, given a differentiable scalar field $f(\x)$ and an initial guess $\x^0$, the gradient descent iteratively moves to guess toward the lower values of $f$ by taking steps in the opposite direction of the gradient, $-\nabla f(\x)$. This is locally the steepest descent direction, that is, the direction that $\x$ would need to move in order to decrease the quickest. Therefore, the minimum is computed iteratively by $\x^{n+1} = \x^n - \tau \nabla f(\x^n)$, where $\tau$ accounts for the step size.

Note that the Euler-Lagrange equation \eqref{eq:ELeq2} is linear in $u_k$. This fact is an advantage to solve it since the linearity allows one to build an explicit scheme for computing the minimizer. Indeed, the solution for each spectral band is obtained pixel-by-pixel by iterating the equation
\begin{equation}\label{eq:ELequationdisc}
\begin{aligned}
u_k^{(n+1)}(\p) &= u_k^{(n)}(\p) - \tau\sum_{\q\in I} \big(u_k^{(n)}(\p) -u_k^{(n)}(\q)\big)\big(\omega_{P_k}(\p,\q)+ \omega_{P_k}(\q,\p)\big) \\ 
&- \tau\mu s^2 \left(K_k^{\top}\Pi_S  \left(K_k u_k^{(n)}-u_k^{\Omega}\right)\right)(\p)\\
&-\tau\dfrac{\delta}{\|P_k\|^2} \left(u_k^{(n)}(\p)\widetilde{P}_k(\p) - \widetilde{u}_k(\p)P_k(\p) \right)\widetilde{P}_k(\p),
\end{aligned}
\end{equation}
where $n\geq 0$ is the iteration number and $\tau$ is the artificial time step in the descent direction. Note that an initialization $\u^{(0)}=(u^{(0)}_1,\ldots,u^{(0)}_C)$ is required. 

For computational purposes, the nonlocal regularization term is limited to interact only between pixels at a certain distance (the so-called {\it search window}), that is, the weight $\omega_{P_k}(\p,\q)$ is zero for all pixels $\p$ and $\q$ such that $\|\p-\q\|_{\infty}>\nu_r$, for a certain parameter $\nu_r>0$. More precisely, we define
\begin{subequations}\label{eq:weightsdiscret}
\begin{equation}
\omega_{P_k}(\p,\q)= \left\{ \begin{array}{ll} \dfrac{1}{\Upsilon(\p)} \exp\left(-{\dfrac{1}{h^2}\displaystyle\sum_{\{\t : \|\t\|_{\infty}\leq \nu_c\}} |P_k(\p+\t)-P_k(\q+\t)|^2}\right) & \text{if } \|\p-\q\|_{\infty} \leq \nu_r, \\ 0 & \text{otherwise}, \end{array}\right.
\end{equation}
where $\nu_c>0$ determines the size of a window centered at $\mathbf{0}$ (the so-called {\it comparison window}). The weight distribution is in general sparse since only a few nonzero weights are considered. The normalization factor $\Upsilon(\p)$ is defined as
\begin{equation}
\Upsilon(\p)=\sum_{\{\q\in I : \|\q-\p\|_{\infty} \leq \nu_r\}} \exp\left({-\dfrac{1}{h^2}\displaystyle\sum_{\{\t : \|\t\|_{\infty}\leq \nu_c\}} |P_k(\p+\t)-P_k(\q+\t)|^2}\right).
\end{equation}
Note that the Gaussian kernel $G_{\rho}$ introduced in \eqref{eq:weights} is not considered in practice as it is only necessary when the size of the windows increase considerably. Finally, in order to avoid an excessive weighting of the reference pixel, $\omega_{P_k}(\p,\p)$ is set to the maximum of the weights:
\begin{equation}
\omega_{P_k}(\p,\p) = \max \{ \omega_{P_k}(\p,\q) :  \|\p-\q\|_{\infty}\leq\nu_r, \q \neq \p\}.
\end{equation}
\end{subequations}

Since the numerical scheme \eqref{eq:ELequationdisc} decouples for each spectral component, we can proceed for each $u_k$ as follows:
\begin{enumerate}
\item[i)] Superimpose the panchromatic image, which hardly contains aliasing, on the reference of $u_k$.
\item[ii)] Compute the weight function $\omega_{P_k}$ on the registered panchromatic.
\item[iii)] Solve the pansharpening problem for $u_k$ by iterating \eqref{eq:ELequationdisc} until convergence.
\item[iv)] Superimpose all obtained high-resolution spectral bands, which are supposed to be free from aliasing artifacts, on a common geometry for visualization purposes.
\end{enumerate}
By using the above procedure, we avoid resampling or re-interpolating the aliased low-resolution spectral components and the algorithm applies on the original data instead. In the case the data is co-registered, there is no need to modify the panchromatic image nor superimpose all bands after pansharpening. 

As stopping criterion we used a tolerance value of $10^{-6}$ for the relative error between two consecutive iterations, that is
\begin{equation}\label{eq:stop}
\sqrt{\dfrac{1}{|I|} \sum_{\p\in I}\left(u_k^{(n+1)}(\p)-u_k^{(n)}(\p)  \right)^2} < 10^{-6}.
\end{equation}
Anyway, we stopped the algorithm after $500$ iterations even if the tolerance was not reached. We experimentally checked that these are enough iterations for convergence, since the relative error between two consecutive steps is small enough. Finally, let us remark that we used the same parameters $\mu$, $\delta$, and $h$ for all spectral components. Therefore, one will expect slightly better results if these parameters are optimized for each band, but at some computational cost.

\section{Discussion}\label{sec:discussion}

\subsection{``Panchro-Spectral'' Constraint for Real Satellite Data}\label{sec:linearity}

The variational model \eqref{eq:functional1} led to a significant reduction of color artifacts with respect to state-of-the-art pansharpening methods as the experiments by Duran {\it et al.} \cite{DuranBuadesCollSbertSIIMS2014} demonstrated. However, the energy functional still uses the hypothesis that the panchromatic image is a linear combination of the high-resolution spectral components (called {\it panchro-spectral constraint} from here on), which is not true in a real scenario. A false linear combination can further damage the spectral quality of the data. On the other hand, the raw data captured by the satellite are geometrically misregistered and, thus, equations of the form \eqref{eq:panconstraint} cannot be directly imposed.

\begin{figure}[!htpb]
        \centering
        \begin{tabular}{c}
        \includegraphics[trim= 0cm 0.1cm 0cm 0cm, clip=true, width=0.7\textwidth]{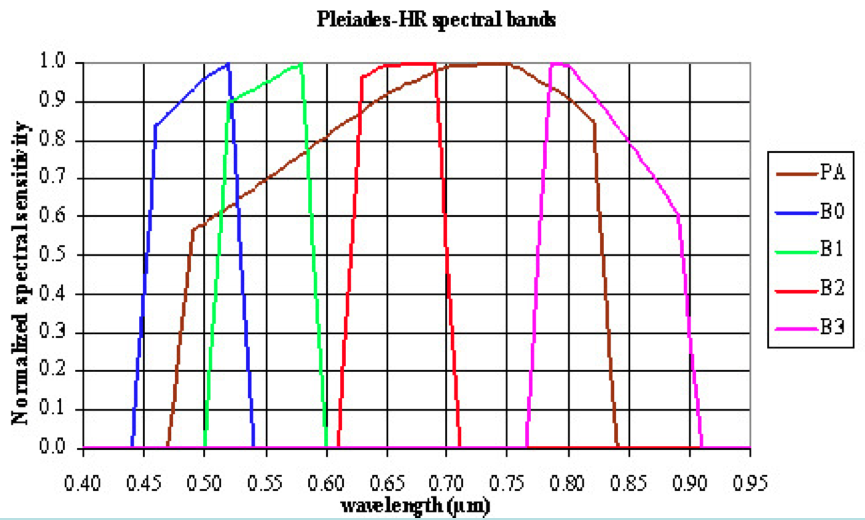}
        \end{tabular}
        	\caption{Spectral sensitivity of the blue, green, red, near-infrared, and panchromatic sensors to different wavelengths of light for the Pl{\'e}iades satellite system.}
	\label{fig:satellite_responses}
\end{figure}

Figure \ref{fig:satellite_responses} plots the spectral sensitivities of the blue, green, red, near-infrared, and panchromatic sensors to different wavelengths of light for the Pl{\'e}iades satellite system. One realizes that the assumption given in \eqref{eq:panconstraint} does not follow in general. Indeed, the panchromatic sensor covers frequencies which are not covered by any of the others, but there are also wavelengths covered by the blue and near-infrared sensors that do not fall under the scope of the panchromatic sensitivity.

We now experimentally check the panchro-spectral constraint for Pl{\'e}iades images courtesy of CNES. The original panchromatic image is compared with the intensity image associated to the low-resolution spectral components by means of the linear combination 
\begin{equation}\label{eq:linearPleiades}
I^S(\x)=\alpha_B B(\x) + \alpha_G G(\x) + \alpha_R R(\x) + \alpha_I I(\x), \quad \forall \x\in S,
\end{equation}
where the mixing coefficients $\alpha_B$, $\alpha_G$, $\alpha_R$, and $\alpha_I$ involving the blue, green, red, and near-infrared channels, respectively, were furnished to us by the CNES. These coefficients are statistically acceptable on a variety of landscapes and used for problems related to the treatment of soil. In order to make the images comparable, we also downsampled the original panchromatic to the resolution of the spectral channels by decimation.

\begin{figure}[!t]
\centering 
\begin{tabular}{c@{\hskip 0.03in}c@{\hskip 0.03in}c}
  \includegraphics[width=0.32\textwidth]{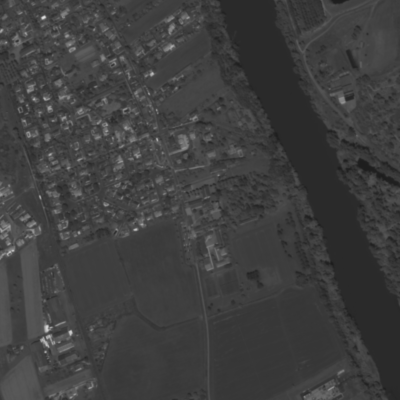} &
  \includegraphics[width=0.32\textwidth]{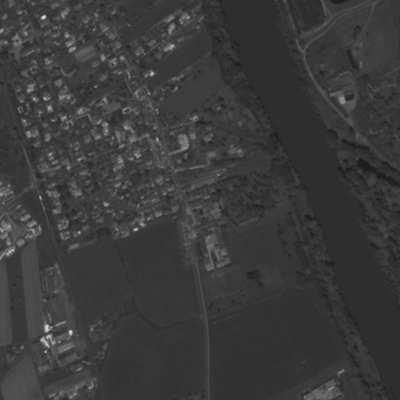} &
  \includegraphics[width=0.32\textwidth]{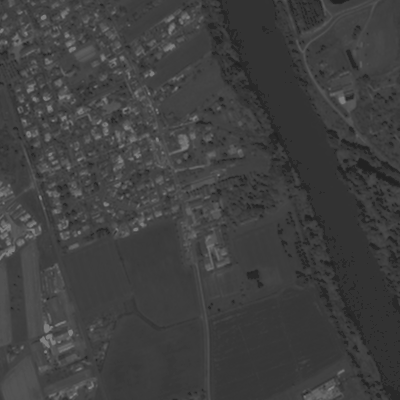} \\
  \footnotesize Decimated panchromatic & \footnotesize Intensity & \footnotesize Histogram-specified intensity
  \end{tabular}
\caption{From left to right, the original panchromatic downsampled to the resolution of the spectral bands, the intensity image obtained from the linear combination of the low-resolution components as given in \eqref{eq:linearPleiades}, and the histogram specification of this linear combination. We observe that, although the results seem to be similar at first glance, the contrast and colors are not exactly the same. See, for instance, the dark part in the river.}
\label{fig:panlinearPleiades}
\end{figure}

Figure \ref{fig:panlinearPleiades} displays the decimated panchromatic (left picture) and the intensity image (central picture) obtained from \eqref{eq:linearPleiades}. A careful inspection illustrates that, even if contrast order is quite similar, colors are not the same -- see, for example, the dark part in the river. In order to reduce these color differences, the panchromatic histogram is specified to $I^S$, the result of which is also displayed in Figure \ref{fig:panlinearPleiades} (right picture). Although the root-mean-squared error (RMSE) reduces from $5.72$ in $I^S$ to $4.13$ in the specificied variant, it is still quite meaningful in view of the fact that the range of the images is $[0,255]$. In conclusion, although the original panchromatic and $I^S$ are similar at a first glance, the numerical difference between both is too large to use \eqref{eq:panconstraint} as a constraint in a pansharpening model.

\subsection{Co-Registration of the Spectral Components}\label{sec_registrationspectral}

Most of pansharpening techniques proposed in the literature implicitly assume that the spectral components are co-registered and registered with the panchromatic, which is not the case for real satellite imagery. Furthermore, because of the aliasing in the low-resolution data (see Figure \ref{fig:satellite_aliasing}), re-interpolation into a common reference is not at all recommendable.

\begin{figure}[!t]
\centering 
\begin{tabular}{c@{\hskip 0.03in}c@{\hskip 0.03in}c}
  \includegraphics[trim= 30.5cm 30cm 12.5cm 13cm, clip=true, width=0.32\textwidth]{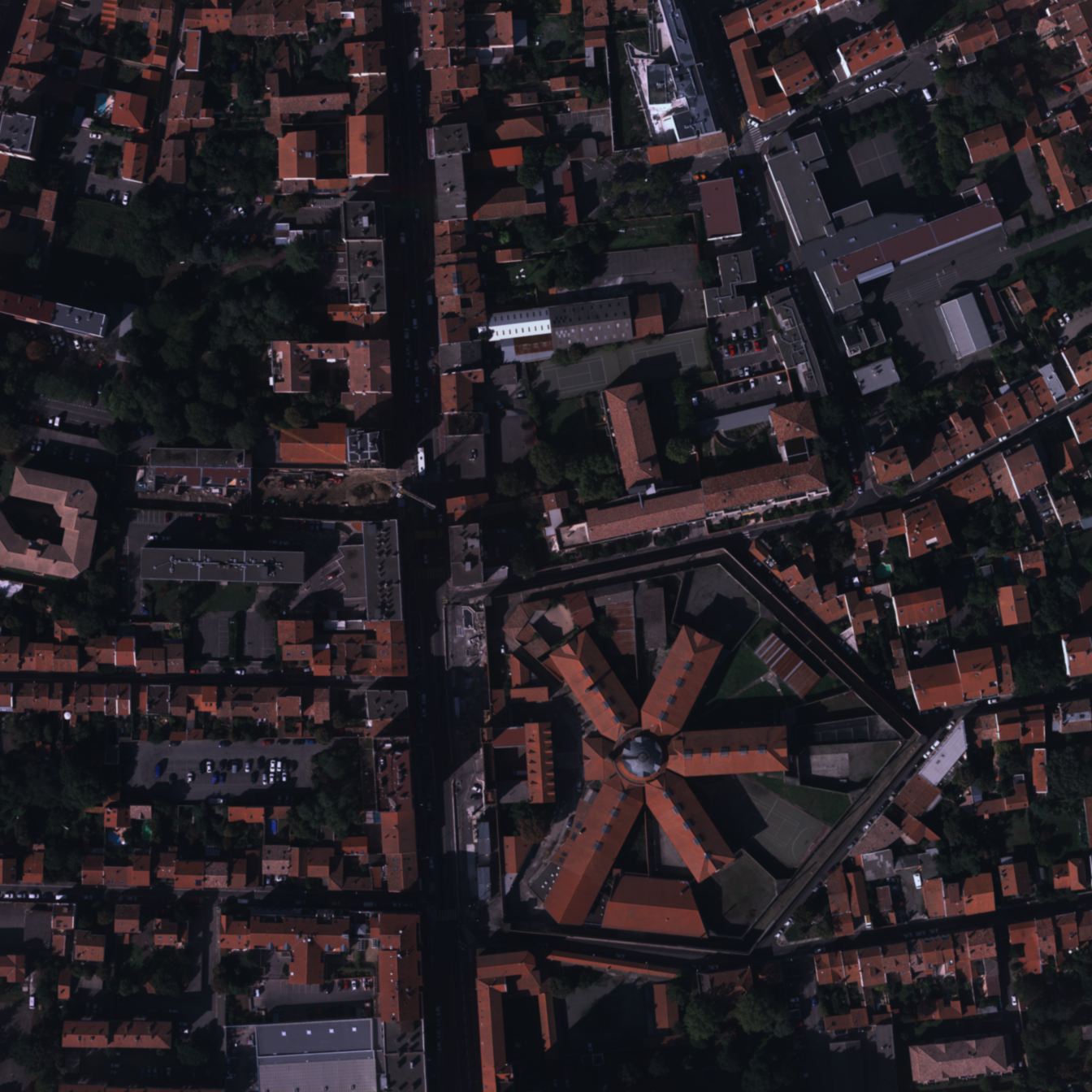} &
  \includegraphics[trim= 30.5cm 30cm 12.5cm 13cm, clip=true, width=0.32\textwidth]{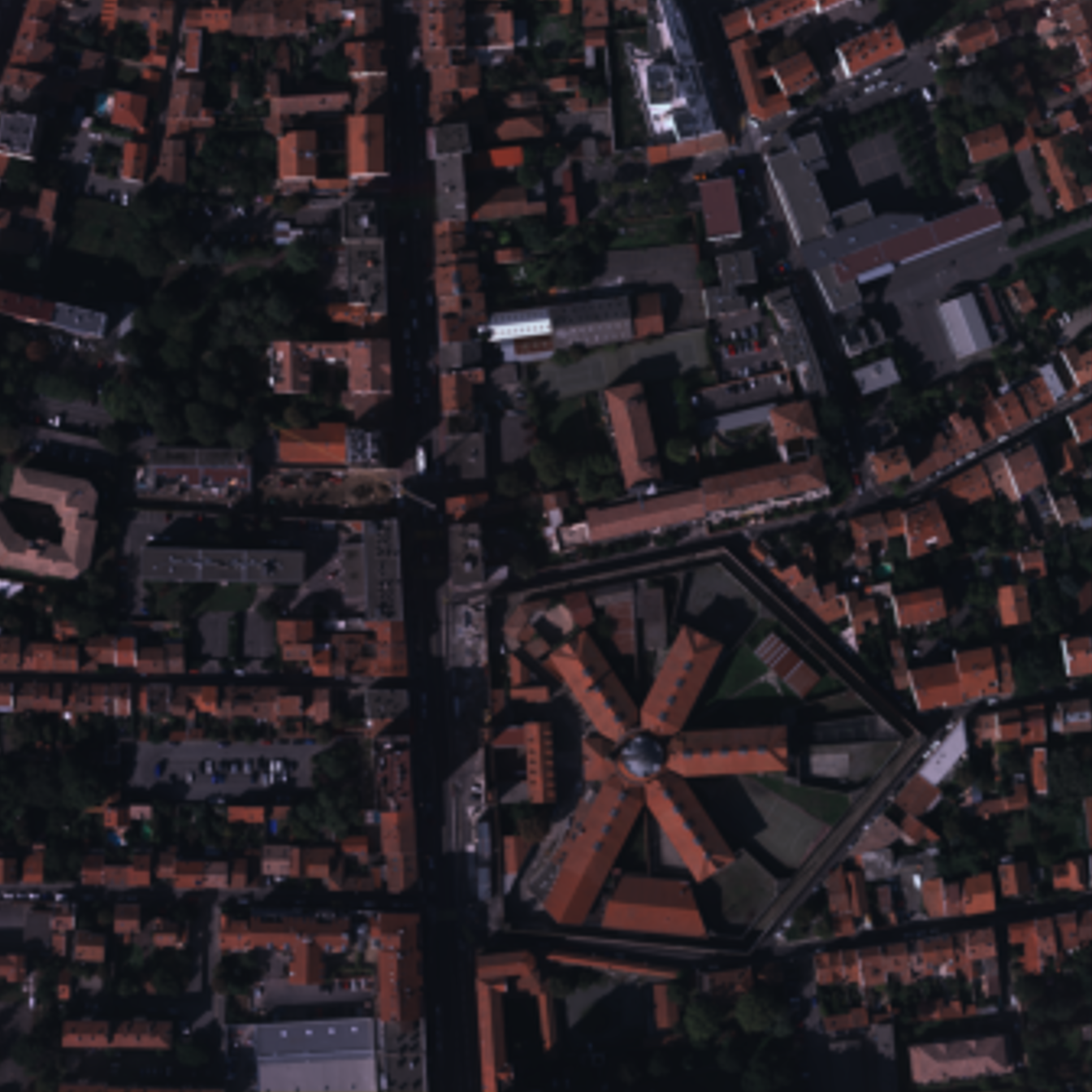} &
  \includegraphics[trim= 30.5cm 30cm 12.5cm 13cm, clip=true, width=0.32\textwidth]{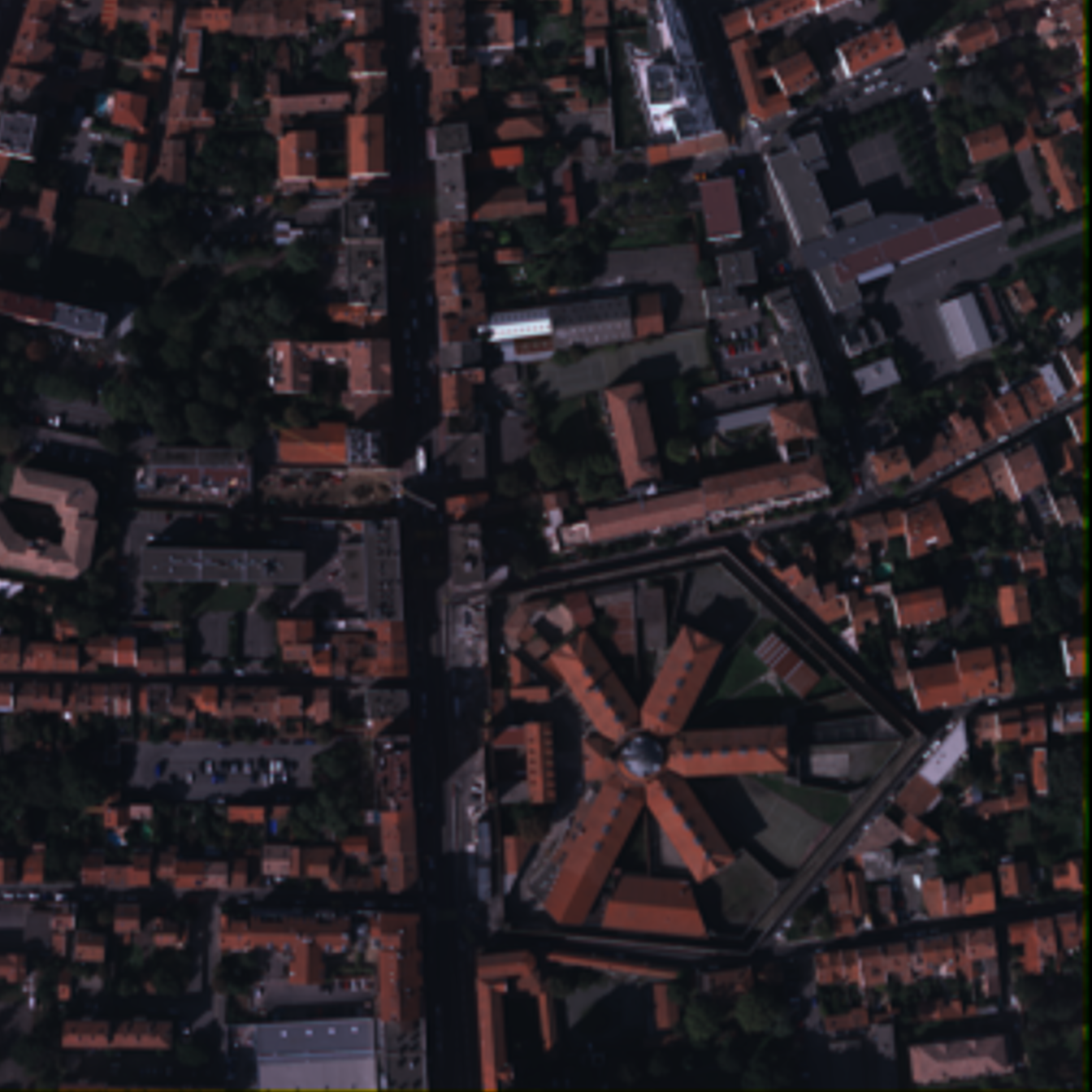} \\
  \footnotesize Reference & \footnotesize Decimation & \footnotesize Co-registration + decimation
  \end{tabular}
\caption{The central picture illustrates the low-resolution data simulated directly from the reference image. By first translating the original red, green, and blue channels, applying the downsampling process to the misregistered components and superimposing them into a common geometry, one obtains the image shown in the righthand-side picture. Although aliasing is apparent in both cases, the artifacts have been considerably increased after co-registration. In the last case, we also observe a ``drooling effect'', that is, the colors of the objects exceed their contours.}
\label{fig:registration}
\end{figure}

Let us check if the aliasing in the low-resolution spectral components increases after co-registration. For that purpose, we carried out an experiment on an RGB aerial image at resolution of 30 cm per pixel, courtesy of CNES. On the one hand, we computed the low-resolution data directly from the original image by Gaussian filtering of standard deviation $1.5$ followed by subsampling of factor $4$. We intentionally used a lower-than-recommended standard deviation in order to introduce aliasing. The obtained data is displayed in the central picture of Figure \ref{fig:registration}. On the other hand, we also applied a translation by splines to the reference red, green, and blue channels and followed the same decimation process than before but now on the misregistered data. After that, we interpolated them back into a common geometry and obtained the image shown in the righthand-side picture of Figure \ref{fig:registration}. Although both images contain aliasing, we clearly observe that it has considerably increased after co-registration. In the last case, a ``drooling effect'', that is, the colors of the objects exceeding their contours, is also observable. We can thus expect to obtain better results using the original spectral components than the resampled ones.

\section{Experimental Results}\label{sec:experiments}

This section is devoted to a detailed performance comparison between the proposed model lying in the minimization of the energy functional \eqref{eq:functional2}, hereafter denoted by NLVD, and some classical and state-of-the-art pansharpening techniques. We chose some of the best methods according to the recent review by Vivone {\it et al.} \cite{VivoneAlparoneChanussot2015} being representative of the two main pansharpening classes described in the introduction. Whenever possible, we compare with the CS-based methods PCA \cite{ChavezSides1991}, Brovey \cite{GillespieKahle1987}, BDSD \cite{GarzelliNenciniCapobianco2008}, Gram-Schmidt adaptive (GSA) \cite{AiazziBarontiSelva2007}, and PRACS \cite{ChoiYuKim2011}. We also compare with the MRA-based algorithms HPF with $5\times 5$ box filter \cite{ChavezSides1991}, smoothing filter-based intensity modulation (SFIM) \cite{Liu2000, WaldRanchin2002}, local mean and variance matching filter (LMVM) \cite{BethuneMuller1998}, additive {\it {\`a} trous} wavelet transform with unitary injection model (ATWT) \cite{VivoneRestaino2014},  additive wavelet luminance proportional (AWLP) \cite{OtazuGonzalezFors2005}, and generalized Laplacian pyramid with MTF-matched filter and high-pass modulation (GLP) \cite{AiazziAlparoneBaronti2006}. We further include the variational techniques P+XS \cite{BallesterCaselles2006} and its nonlocal variant (NLV) by Duran {\it et al.} \cite{DuranBuadesCollSbertSIIMS2014}, which consists in the minimization of the energy \eqref{eq:functional1} using the gradient descent method.

\begin{figure}[!t]
\centering \renewcommand{\arraystretch}{0.5}
\begin{tabular}{c@{\hskip 0.03in}c@{\hskip 0.03in}c}
  \includegraphics[trim= 23cm 9.1cm 12cm 25.9cm, clip=true, width=0.25\textwidth]{image1_rgb.png}&
  \includegraphics[trim= 2cm 32cm 33cm 3cm, clip=true, width=0.25\textwidth]{image1_rgb.png}&
  \includegraphics[trim= 26cm 23.5cm 9cm 11.5cm, clip=true, width=0.25\textwidth]{image1_rgb.png} \\
  \includegraphics[trim= 9cm 29cm 26cm 6cm, clip=true, width=0.25\textwidth]{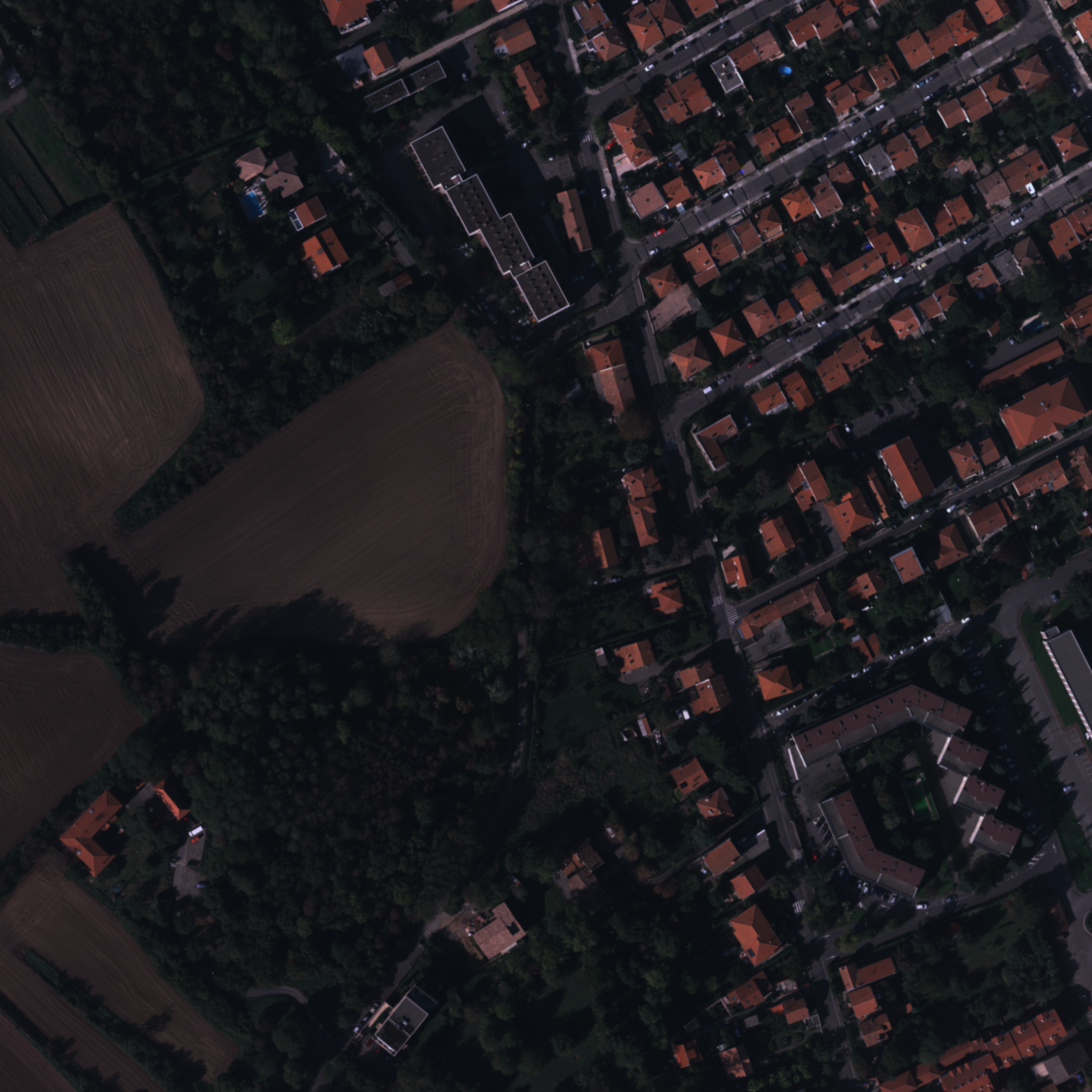}&
  \includegraphics[trim= 32.5cm 6cm 2.5cm 29cm, clip=true, width=0.25\textwidth]{image2_rgb.png}&
  \includegraphics[trim= 24cm 30cm 11cm 5cm, clip=true, width=0.25\textwidth]{image2_rgb.png} \\
\end{tabular}
\caption{Set of full-color aerial images at a resolution of 30 cm per pixel used in the experimental section. CNES provided us with blue, green, red, and near-infrared -- not displayed here -- bands at 10 cm. From these data, we built reference spectral channels at 30 and 60 cm by MTF filtering and decimation.}
\label{fig:dataset}
\end{figure}

We tested all pansharpening methods previously mentioned on data simulated from aerial images as well as on real Pl{\'e}iades imagery. CNES provided us with 4-band (blue, green, red, and near-infrared) images at a resolution of 10 cm per pixel collected by an aerial platform. From these data, we built reference spectral channels at resolutions of 30 and 60 cm by MTF filtering and subsampling. Therefore, we obtained the ground-truth images with which one can evaluate the fusion results in terms of several quality assessment indices. Figure \ref{fig:dataset} shows the set of full-color aerial images used in our experiments. CNES also furnished us with raw Pl{\'e}iades data consisting of a panchromatic at a resolution of 70 cm per pixel and blue, green, red, and near-infrared bands at a resolution of 2.8 m. The MTF for the panchromatic has a value of 0.15 at cut frequency, which avoids pretty much aliasing, while this value is greater than 0.26 for the spectral components thus leading to critical aliasing artifacts. Figure \ref{fig:datasetPleiades} displayed the satellite data of a Toulouse scene.

The variational methods P+XS, NLV, and NLVD were implemented in C/C++, and the corres\-ponding trade-off parameters were optimized in terms of the lowest error with respect to the ground truth available in simulated data. We also implemented LMVM using a $9\times 9$ window for the computation of the local mean and standard deviation at each pixel. The size of the local window was also optimized in terms of the lowest error. For all other techniques, we used the  source codes kindly provided by Dr.~Vivone \cite{VivoneAlparoneChanussot2015}. In all cases, we initialized with a simple interpolation of each spectral channel by bicubic splines.

\subsection{Quality Assessment Indices}

For simulated data, the spatial and spectral consistency of the fused products with respect to the ground-truth images are numerically evaluated by means of several quality assessment indices. Let $\u^R=\left(u_1^R, \ldots, u_C^R\right)$ be the high-resolution reference multispectral image and let us use the same notations than in Section \ref{sec:numerics}. We employ the following quality indices for the evaluation of the pansharpening techniques:

\begin{itemize}
\item The {\it Root Mean Squared Error} (RMSE) is one of the most popular measures that accounts for spatial distortion. It is computed as
$$
\text{RMSE}\left(u^R_k, u_k\right) = \sqrt{\dfrac{1}{|I|}\sum_{\p\in I} \left(u_k^R(\p) - u_k(\p)\right)}, \quad \forall \,k\in\{1,\ldots, C\},
$$
and its optimal value is zero. We shall calculate this measure for each band and average the results over all bands to obtain a global value.

\item The {\it Erreur Relative Globale Adimensionelle de Synth{\`e}se} (ERGAS) proposed by Ranchin and Wald \cite{RanchinWald2000} is an index that gives a global quality assessment of the fused product. It is defined as
$$
\text{ERGAS} = \dfrac{100}{s} \sqrt{\dfrac{1}{C} \sum_{k=1}^C \left( \dfrac{\text{RMSE}\left(u^R_k, u_k\right)}{\mu_{u_k^R}}\right)^2},
$$
where $s$ is the sampling factor and $\mu_{u_k^R}$ is the mean value of the $k$th spectral component of the reference image. Since the ERGAS is composed by a sum of RMSE, its optimal value is zero.

\item The {\it Spectral Angle Mapper} (SAM) introduced by Alparone {\it et al.} \cite{AlparoneBaronti2004} is a measure of spectral quality computed in the space defined by considering each channel as a coordinate axis. Mathematically, it is written as the absolute angle between the spectral vector of each pixel of the pansharpened image, $\u(\p)=(u_1(\p),\ldots, u_C(\p))$, and that of the reference image, $\u^R(\p)=(u_1^R(\p), \ldots, u_C^R(\p))$:
$$
\text{SAM}(\p)= \arccos\left(\dfrac{\langle \u^R(\p), \u(\p)\rangle}{\|\u^R(\p) \|_2  \|\u(\p) \|_2} \right),\quad \forall\p\in I,
$$
where $\langle \cdot, \cdot \rangle$ denotes the scalar product and $\|\cdot\|_2$, the vectorial $\ell^2$ norm. The global value of SAM for the whole image is obtained by averaging the single measures over all pixels. Its optimal value is zero, which means that there is no spectral distortion. 

\item The {\it Structural Similarity Index} (SSIM), which is also called {\it Universal Image Quality Index} (UIQI) or {\it Q-index}, was initially proposed by Wang and Bovik \cite{WangBovik2002} to model any image distortion as a combination of loss of correlation, luminance distortion, and contrast distortion. It is only applied to monochrome images as follows:
\begin{equation}\label{eq:ssim_def}
\text{SSIM}\left(u^R_k, u_k\right)  = \dfrac{4 \sigma_{u_k^R,u_k} \mu_{u_k^R} \mu_{u_k}}{\big(\sigma_{u^R_k}^2 + \sigma_{u_k}^2\big)\big(\mu^2_{u_k^R} + \mu^2_{u_k}\big)}, \quad\forall k\in\{1,\ldots, C\},
\end{equation}
where $ \sigma_{u_k^R,u_k}$ is the covariance between intensity values in the fused and reference channels, $\mu_{u_k^R}$ and $\sigma_{u^R_k}^2$ are the mean value and the variance of the reference band, and $\mu_{u_k}$ and $\sigma_{u_k}^2$ are those of the pansharpened band. All statistics are computed on $8\times 8$ image blocks and the resulting values are then averaged over the whole image. The SSIM index varies in the range $[-1,1]$, with one denoting the best fidelity to reference. Similar to RMSE, we shall calculate a global SSIM value as the average $\frac{1}{C} \sum_{k=1}^C\text{SSIM}\left(u^R_k, u_k\right)$.

\item The {\it Q$2^n$-index} introduced by Garzelli and Nencini \cite{GarzelliNencini2009} is an extension of the {\it Q$4$-index} \cite{AlparoneBaronti2004}, which in turn generalizes the idea of the universal SSIM measure to $4$-band images based on the theory of hypercomplex numbers. In this setting, each pixel in a multispectral image $\u$ is modelled as
$$
\u(\p)=u_1(\p) + u_2(\p)\vec{i}_1 + \cdots + u_C(\p)\vec{i}_{C-1},
$$
where $\vec{i}_1, \ldots, \vec{i}_{C-1}$ are the imaginary units. With this representation, the Q$2^n$-index can be calculated using \eqref{eq:ssim_def} for each pixel, that is, $\text{SSIM}\left(\u^R(\p), \u(\p)\right)$. Again, same as for \eqref{eq:ssim_def}, the statistics are computed on $8\times 8$ blocks and then averaged over the whole image to yield the global score index, the optimal value of which is one.
\end{itemize}

For real satellite data, the lack of reference images make more difficult a quantitative performance evaluation of pansharpening techniques at the original resolution. In such cases, the panchromatic and low-resolution components are somehow used to determine how much the spatial and spectral information is preserved during the fusion process. In this regard, we use the {\it Quality with No Reference} (QNR) index proposed by Alparone \cite{Alparone2008} in order to assess the quality of the results obtained on real satellite data. On the one hand, the spectral distortion in the fused product is computed as
$$
D_{\lambda} = \dfrac{1}{C(C-1)} \sum_{k=1}^C \sum_{l=1, l\neq k}^C \left| \text{SSIM}\left(\widetilde{u}_k, \widetilde{u}_l\right) - \text{SSIM}\left(u_k,u_l\right)\right|,
$$
where $\widetilde{u}_k$ is the $k$th spectral band upsampled at full resolution by bicubic interpolation and SSIM is defined in \eqref{eq:ssim_def}. On the other hand, the spatial distortion is estimated by
$$
D_S = \dfrac{1}{C} \sum_{k=1}^C \left| \text{SSIM}\left(P, u_k\right) - \text{SSIM}\big(\widetilde{P}_k,\widetilde{u}_k\big)\right|,
$$
where $\widetilde{P}_k$ is the low-resolution panchromatic at the same scale of $\widetilde{u}_k$ computed as in \eqref{eq:spatialratioall}. The QNR index is finally defined as the combination of the two previous measures:
$$
\text{QNR} = \left(1-D_{\lambda}\right)\left(1-D_S\right).
$$
The optimal value of QNR is one, which is obtained when both spectral and spatial distortions are equal to zero.

In general, a low performance in all the above indices also entails a rejection by a human visual inspection. In spite of this, any numerical criterion cannot fully replace human evaluation, which still is an important criterion for judging the performance of pansharpening algorithms particularly in satellite imagery.

\subsection{Performance Comparison on Simulated Data from Aerial Images}

In this subsection, we present a comprehensive quality assessment of the pansharpening methods under comparison on data simulated from the reference multispectral images displayed in Figure \ref{fig:dataset} at resolutions of 30 cm and 60 cm per pixel. We first take ideal conditions according to which all bands are geometrically aligned and the constraint \eqref{eq:panconstraint} that writes the panchromatic as a linear combination of the spectral components applies and can be taken advantage of. From this starting point, we move towards more and more realistic conditions. In this regard, we simulate misregistered data by translating the ground-truth components using a different transformation per channel and computing then each low-resolution spectral component. In the last set of experimental tests, the panchro-spectral constraint is also disabled leading to the most realistic scenario.

The choice of the trade-off parameters $\mu\geq 0$ and $\delta\geq 0$ in \eqref{eq:functional2} is an important issue since they balance the contribution of each term to the total energy. Furthermore, it is also important to fix correctly the filtering parameter $h>0$ in \eqref{eq:weights} because it controls how fast the weights decay with increasing dissimilarity of patches in the panchromatic image. In this setting, the research and comparison windows for the computation of the weight distribution in \eqref{eq:weightsdiscret} were fixed to $7\times 7$ and $3\times 3$, respectively. As far as the experiments on simulated data is concerned, $\mu$, $\delta$, and $h$ were estimated by trying different combinations of some preset values on the dataset displayed in Figure \ref{fig:dataset} and determining those at which the lowest RMSE was obtained. We use different conditions in the simulation of the data, namely considering RGB as well as 4-band images, the panchromatic being computed as a linear combination of spectral channels with different mixing coefficients, the low-resolution spectral components being and not being co-registered, and introducing several degrees of aliasing. Finally, we set $\mu=50$, $\delta=6.21$, and $h=1.25$ for all experiments. 

All fused products were saved in $8$-bit values relative to the intensity range $[0,255]$. In order to help the visual analysis, we display along this subsection the difference images by linearly mapping the range $[-20,20]$ to $[0,255]$ and saturating values outside this range. This is a manual linear stretching with constant minimum and maximum values that avoids favoring any of the methods under comparison.

\subsubsection{Registered Bands and Panchro-Spectral Constraint Fulfilled}

First, the simulated panchromatic and chromatic components are co-registered and the panchro-spectral constraint applies. The panchromatic images at 30 cm and 60 cm were obtained by linear combination of blue, green, and red channels with mixing coefficients $\alpha_B=\alpha_G=\alpha_R=\frac{1}{3}$. The spectral bands, with respective resolutions of 1.2 m and 2.4 m per pixel, were computed by filtering the ground-truth components with Gaussian kernel followed by subsampling of factor $s = 4$. In order to incorporate different degrees of aliasing, we used several standard deviations for the Gaussian, namely $\sigma\in\{1.3,1.7\}$.

The quantitative results obtained in RGB coordinates are reported in Tables \ref{table_30cm_RGB_regist_linear} and \ref{table_60cm_RGB_regist_linear}. In these experiments, Brovey, BDSD, GSA, PRACS, AWLP, P+XS, and NLV benefit from \eqref{eq:panconstraint} being fulfilled. Despite this {\it a priori} unfavourable situation, NLVD is superior to all pansharpening techniques under comparison except the former model NLV, which exhibits the best performance in terms of all indices. Note also that P+XS beats NLVD when the aliasing is not as apparent ($\sigma=1.7$), but its effectiveness is severely compromised if the aliasing in the data increases. In general terms, the variational methods outperform CS and MRA families since they are able to combine the advantages of both while reducing their drawbacks. Interestingly, LMVM behaves pretty well in terms of the SAM index, in fact it is the best non-variational method in this regard, but gets the worst RMSE results. Finally, it is worth underlining that GSA, ATWT, AWLP, GLP, and P+XS are the most affected by the aliasing effect. 

\begin{table}[!t]
\footnotesize
\centering
\begin{subtable}[h]{0.99\textwidth}
\centering
\begin{tabular}{|c|c|c|c|c|c|c|c|c|}
\hline
 & RMSE & ERGAS & SAM & SSIM & Q$2^n$  \\ \hline \hline
 Reference & 0 & 0 & 0 & 1 & 1 \\ \hline \hline
 PCA & 2.8923 & 2.1295 & 1.7090 & 0.9937 & 0.9692 \\ \hline
 Brovey & 2.5516 & 1.8790 & 1.4230 & 0.9947 & 0.9707 \\ \hline
 BDSD & 1.8640 & 1.3563 & 1.8767 & 0.9983 & 0.9907 \\ \hline
 GSA & 2.7280 & 1.9982 & 2.0335 & 0.9948 & 0.9758 \\ \hline
 PRACS & 2.1758 & 1.6018 & 1.3138 & 0.9981 & 0.9859 \\ \hline
 HPF & 2.7314 & 2.0206 & 1.3551 & 0.9964 & 0.9814 \\ \hline
 SFIM & 2.5841 & 1.9134 & 1.1889 & 0.9964 & 0.9815 \\ \hline
 LMVM & 3.0556 & 2.2619 & 1.0153 & 0.9972 & 0.9728 \\ \hline
 ATWT & 2.1074 & 1.5489 & 1.4156 & 0.9983 & 0.9892 \\ \hline
 AWLP & 2.0507 & 1.5223 & 1.3005 & 0.9985 & 0.9896 \\ \hline
 GLP & 2.1739 & 1.5937 & 1.2016 & 0.9980 & 0.9879 \\ \hline
 P+XS & 1.1031 & 0.7988 & 0.9034 & 0.9998 & 0.9946 \\ \hline
 NLV & {\bf 0.9395} & {\bf 0.6794} & {\bf 0.7543} & {\bf 0.9999} & {\bf 0.9954} \\ \hline
 NLVD & 1.1546 & 0.8377 & 0.9651 & 0.9997 & 0.9934 \\ \hline
\end{tabular}
\caption{Numerical results for $\sigma=1.7$.}
 \label{table_30cm_s17_RGB_regist_linear}
\end{subtable}

\medskip

\begin{subtable}{0.99\textwidth}
\centering
\begin{tabular}[h]{|c|c|c|c|c|c|c|c|c|}
\hline
 & RMSE & ERGAS & SAM & SSIM & Q$2^n$  \\ \hline \hline
Reference & 0 & 0 & 0 & 1 & 1 \\ \hline \hline
 PCA & 2.5703 & 1.8913 & 1.5554 & 0.9953 & 0.9745 \\ \hline
 Brovey & 2.1917 & 1.6125 & 1.3189 & 0.9963 & 0.9766 \\ \hline
 BDSD & 1.9834 & 1.4462 & 1.8614 & 0.9981 & 0.9894 \\ \hline
 GSA & 3.3798 & 2.4839 & 2.3475 & 0.9909 & 0.9647 \\ \hline
 PRACS & 1.9734 & 1.4493 & 1.2192 & 0.9988 & 0.9901 \\ \hline
 HPF & 2.5252 & 1.8615 & 1.3087 & 0.9972 & 0.9843 \\ \hline
 SFIM & 2.3760 & 1.7528 & 1.1284 & 0.9973 & 0.9847 \\ \hline
 LMVM & 3.1803 & 2.3472 & 1.1628 & 0.9953 & 0.9737 \\ \hline
 ATWT & 2.4381 & 1.7899 & 1.4544 & 0.9970 & 0.9861 \\ \hline
 AWLP & 2.4301 & 1.7887 & 1.2924 & 0.9971 & 0.9866 \\ \hline
 GLP & 2.9548 & 2.1714 & 1.2009 & 0.9955 & 0.9820 \\ \hline
 P+XS & 1.2795 & 0.9267 & 1.1080 & 0.9997 & 0.9917 \\ \hline
 NLV & {\bf 0.9780} & {\bf 0.7068} & {\bf 0.7620} & {\bf 0.9998} & {\bf 0.9952} \\ \hline
 NLVD & 1.1743 & 0.8531 & 0.9528 & 0.9996 & 0.9934 \\ \hline
\end{tabular}
\caption{Numerical results for $\sigma=1.3$.}
 \label{table_30cm_s13_RGB_regist_linear}
\end{subtable}
 \caption{Quantitative evaluation of the fused products on simulated data from RGB aerial images at resolution of 30 cm per pixel. For these experiments, the low-resolution spectral components were co-registered and the panchro-spectral constraint fulfilled with $\alpha_B=\alpha_G=\alpha_R=\frac{1}{3}$. In this ideal setting, NLV provides the best numerical results although the proposed NLVD model is the closest to it. Only P+XS beats NLVD when the aliasing is not as apparent (Table \ref{table_30cm_s17_RGB_regist_linear}). Interestingly, we observe that PCA, Brovey, PRACS, HPF, SFIM, and NLVD seem to work almost independent of the amount of aliasing in the data, since the quality of their performances with respect to all metrics even increase as $\sigma$ decreases.}
 \label{table_30cm_RGB_regist_linear}
\end{table}

\begin{table}[!t]
\footnotesize
\centering
\begin{subtable}[h]{0.99\textwidth}
\centering
\begin{tabular}{|c|c|c|c|c|c|c|c|c|}
\hline
 & RMSE & ERGAS & SAM & SSIM & Q$2^n$  \\ \hline \hline
 Reference & 0 & 0 & 0 & 1 & 1 \\ \hline \hline
 PCA & 3.4458 & 2.5340 & 2.0513 & 0.9860 & 0.9688 \\ \hline
 Brovey & 3.2170 & 2.3657 & 1.8400 & 0.9874 & 0.9697 \\ \hline
 BDSD & 2.3134 & 1.6871 & 2.5309 & 0.9932 & 0.9917 \\ \hline
 GSA & 3.3321 & 2.4379 & 2.6557 & 0.9864 & 0.9748 \\ \hline
 PRACS & 2.4670 & 1.8101 & 1.5914 & 0.9915 & 0.9871 \\ \hline
 HPF & 3.0962 & 2.2851 & 1.6416 & 0.9832 & 0.9799 \\ \hline
 SFIM & 2.9203 & 2.1572 & 1.4398 & 0.9839 & 0.9801 \\ \hline
 LMVM & 3.3380 & 2.4693 & 1.3545 & 0.9822 & 0.9745 \\ \hline
 ATWT & 2.3707 & 1.7413 & 1.6827 & 0.9909 & 0.9890 \\ \hline
 AWLP & 2.2580 & 1.6780 & 1.5476 & 0.9915 & 0.9894 \\ \hline
 GLP & 2.3643 & 1.7336 & 1.4226 & 0.9907 & 0.9884 \\ \hline
 P+XS & 1.3133 & 0.9495 & 1.1756 & 0.9973 & 0.9955 \\ \hline
 NLV & {\bf 1.1664} & {\bf 0.8414} & {\bf 0.9998} & {\bf 0.9979} & {\bf 0.9961} \\ \hline
 NLVD & 1.3671 & 0.9898 & 1.3293 & 0.9967 & 0.9945 \\ \hline
\end{tabular}
\caption{Numerical results for $\sigma=1.7$.}
 \label{table_60cm_s17_RGB_regist_linear}
\end{subtable}

\medskip

\begin{subtable}[h]{0.99\textwidth}
\centering
\begin{tabular}{|c|c|c|c|c|c|c|c|c|}
\hline
 & RMSE & ERGAS & SAM & SSIM & Q$2^n$  \\ \hline \hline
 Reference & 0 & 0 & 0 & 1 & 1 \\ \hline \hline
 PCA & 3.0206 & 2.2200 & 1.8641 & 0.9892 & 0.9757 \\ \hline
 Brovey & 2.7525 & 2.0224 & 1.6922 & 0.9905 & 0.9766 \\ \hline
 BDSD & 2.4637 & 1.8000 & 2.5402 & 0.9924 & 0.9905 \\ \hline
 GSA & 4.2735 & 3.1365 & 3.1862 & 0.9787 & 0.9602 \\ \hline
 PRACS & 2.2222 & 1.6271 & 1.4644 & 0.9932 & 0.9905 \\ \hline
 HPF & 2.8226 & 2.0771 & 1.5766 & 0.9865 & 0.9839 \\ \hline
 SFIM & 2.6557 & 1.9558 & 1.3513 & 0.9873 & 0.9844 \\ \hline
 LMVM & 3.4839 & 2.5695 & 1.5672 & 0.9811 & 0.9746 \\ \hline
 ATWT & 2.6433 & 1.9398 & 1.7476 & 0.9889 & 0.9867 \\ \hline
 AWLP & 2.6047 & 1.9196 & 1.5407 & 0.9895 & 0.9872 \\ \hline
 GLP & 3.1375 & 2.3051 & 1.4284 & 0.9860 & 0.9831 \\ \hline
 P+XS & 1.4588 & 1.0546 & 1.3259 & 0.9967 & 0.9940 \\ \hline
 NLV & {\bf 1.2073} & {\bf 0.8702} & {\bf 1.0079} & {\bf 0.9977} & {\bf 0.9959} \\ \hline
 NLVD & 1.3580 & 0.9823 & 1.3062 & 0.9967 & 0.9945 \\ \hline
\end{tabular}
\caption{Numerical results for $\sigma=1.3$.}
 \label{table_60cm_s13_RGB_regist_linear}
\end{subtable}
 \caption{Quantitative evaluation of the fused products on simulated data from RGB aerial images at resolution of 60 cm per pixel. For these experiments, the low-resolution spectral components were registered and the panchro-spectral constraint fulfilled with $\alpha_B=\alpha_G=\alpha_R=\frac{1}{3}$. The numerical results are less competitive than with data at 30 cm, which was highly expected. In general terms, the same conclusions than in Table \ref{table_30cm_RGB_regist_linear} can be drawn for almost all metrics. However, the increasing in spatial distortion because of the reduction in resolution is more noticeable in CS-based methods.}
 \label{table_60cm_RGB_regist_linear}
\end{table}

\begin{figure}[!p]
\footnotesize
\centering
\renewcommand{\arraystretch}{0.5}
\begin{tabular}{c@{\hskip 0.02in}c@{\hskip 0.02in}c@{\hskip 0.02in}c}
  \includegraphics[trim= 12cm 4cm 6cm 14cm, clip=true, width=0.243\textwidth]{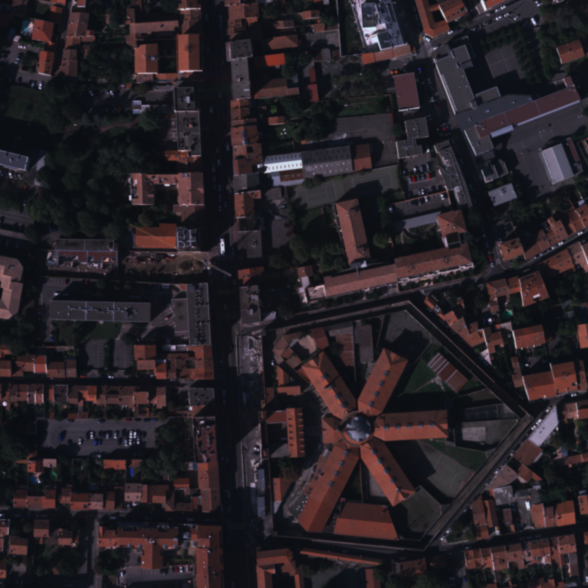} &
  \includegraphics[trim= 12cm 4cm 6cm 14cm, clip=true, width=0.243\textwidth]{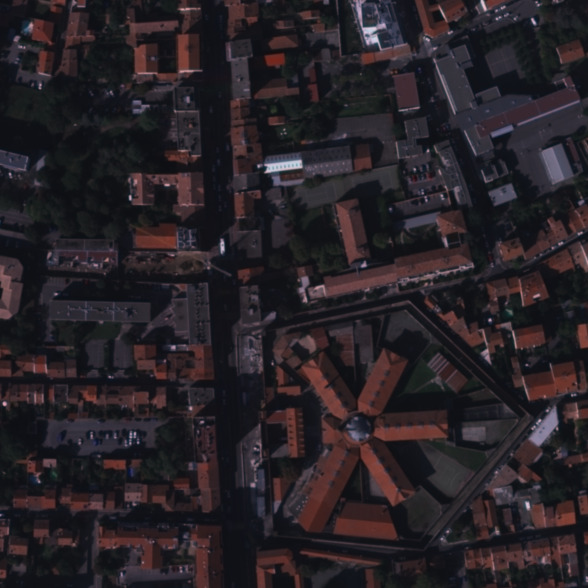} &
  \includegraphics[trim= 12cm 4cm 6cm 14cm, clip=true, width=0.243\textwidth]{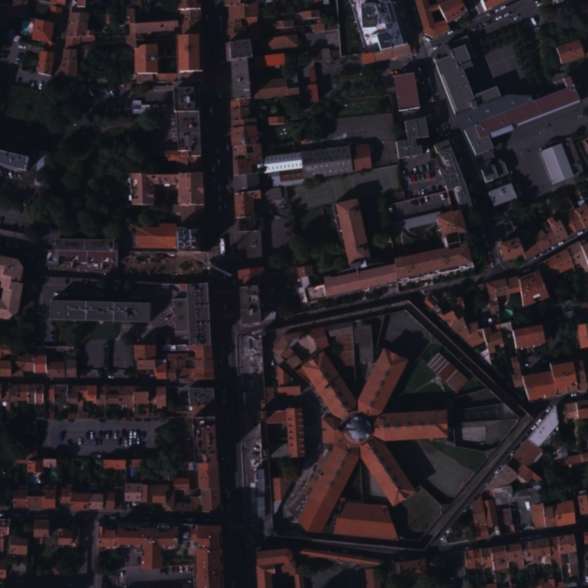} &
  \includegraphics[trim= 12cm 4cm 6cm 14cm, clip=true, width=0.243\textwidth]{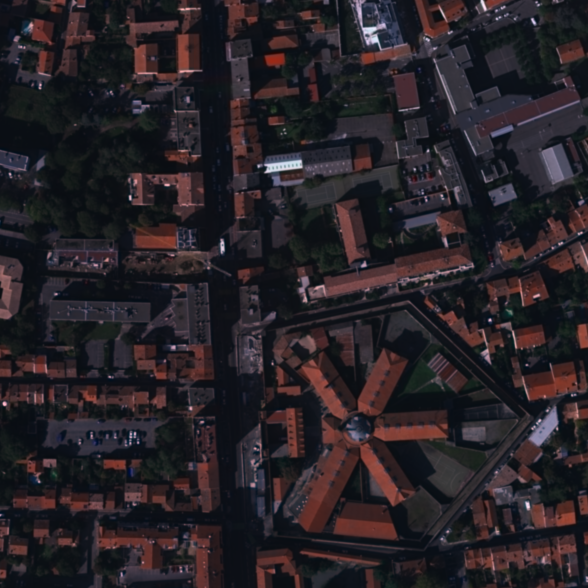} \\
  Reference & PCA & Brovey & BDSD \\
  \includegraphics[trim= 12cm 4cm 6cm 14cm, clip=true, width=0.243\textwidth]{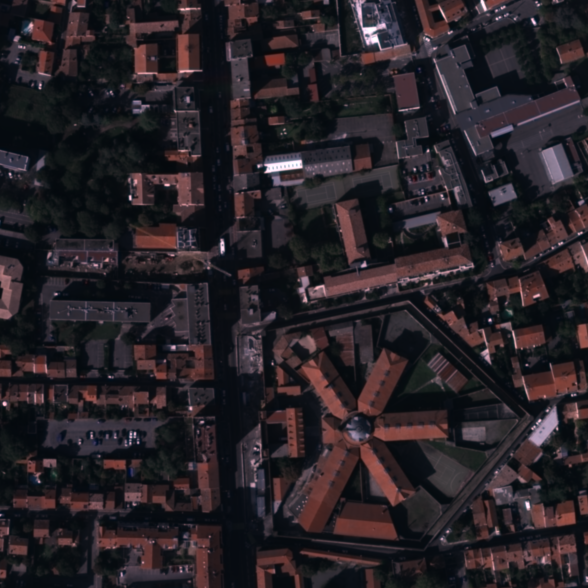} &
  \includegraphics[trim= 12cm 4cm 6cm 14cm, clip=true, width=0.243\textwidth]{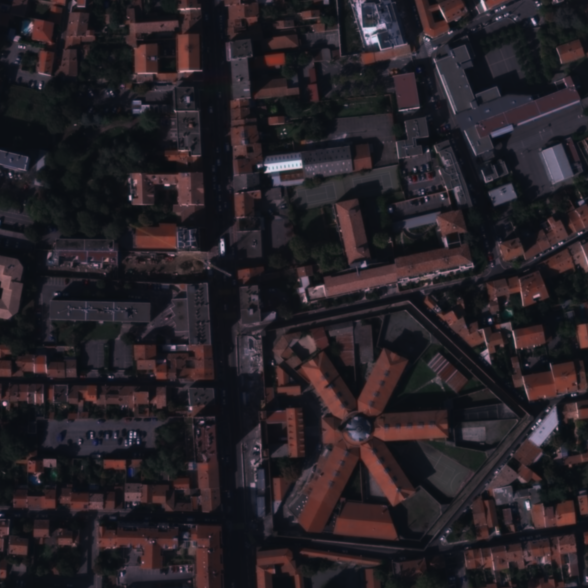} &
  \includegraphics[trim= 12cm 4cm 6cm 14cm, clip=true, width=0.243\textwidth]{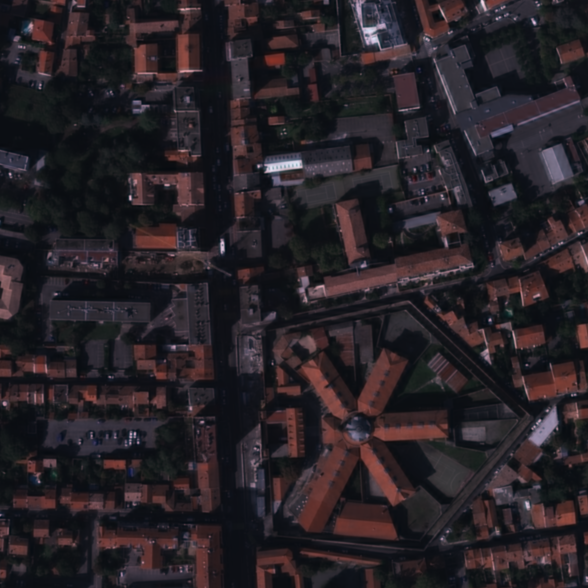} &
  \includegraphics[trim= 12cm 4cm 6cm 14cm, clip=true, width=0.243\textwidth]{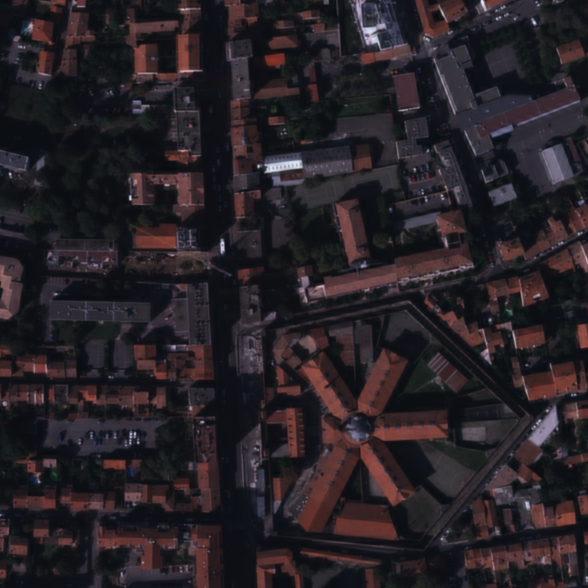} \\
  GSA & PRACS & HPF & SFIM \\
  \includegraphics[trim= 12cm 4cm 6cm 14cm, clip=true, width=0.243\textwidth]{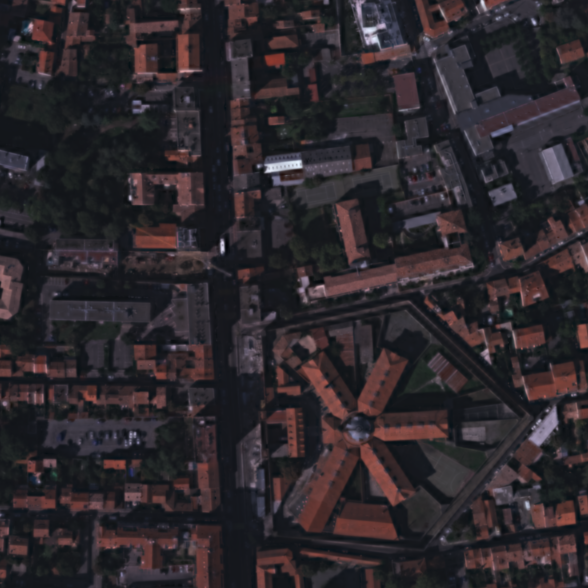} &
  \includegraphics[trim= 12cm 4cm 6cm 14cm, clip=true, width=0.243\textwidth]{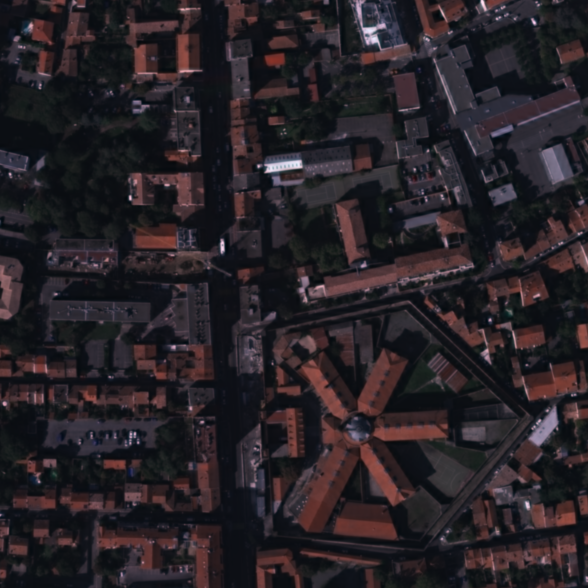} &
  \includegraphics[trim= 12cm 4cm 6cm 14cm, clip=true, width=0.243\textwidth]{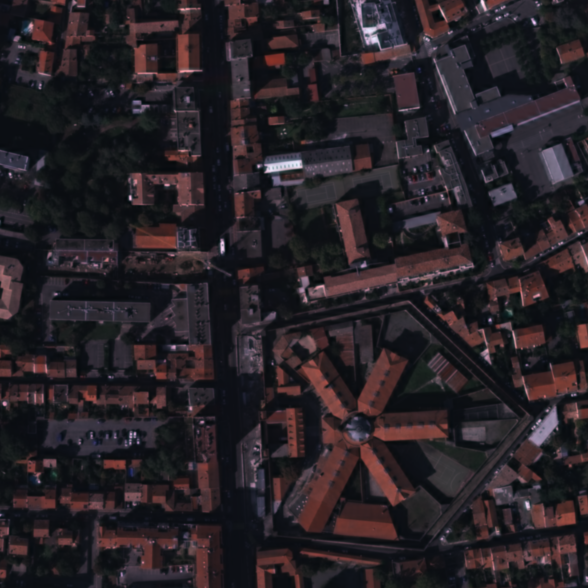} &
  \includegraphics[trim= 12cm 4cm 6cm 14cm, clip=true, width=0.243\textwidth]{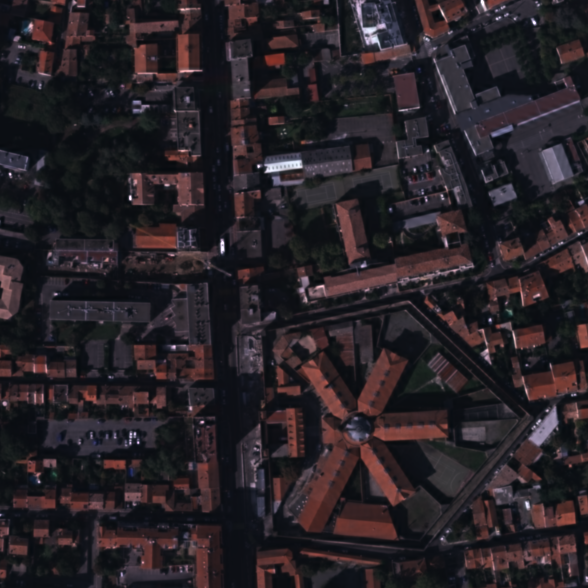} \\
  LMVM & ATWT & AWLP & GLP\\
\end{tabular}
\begin{tabular}{c@{\hskip 0.02in}c@{\hskip 0.02in}c}
  \includegraphics[trim= 12cm 4cm 6cm 14cm, clip=true, width=0.243\textwidth]{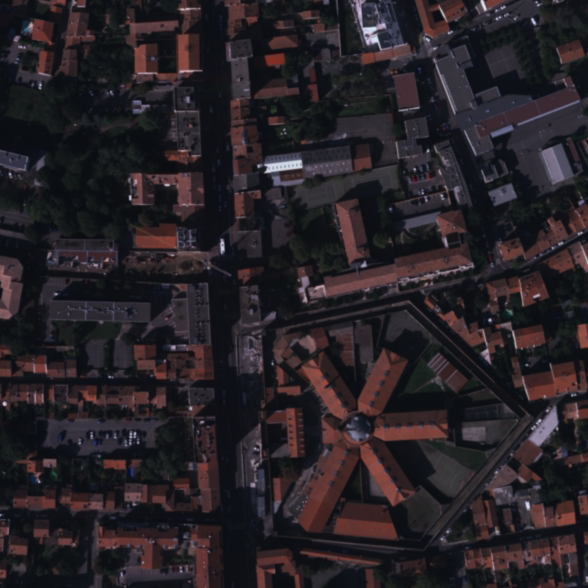} &
  \includegraphics[trim= 12cm 4cm 6cm 14cm, clip=true, width=0.243\textwidth]{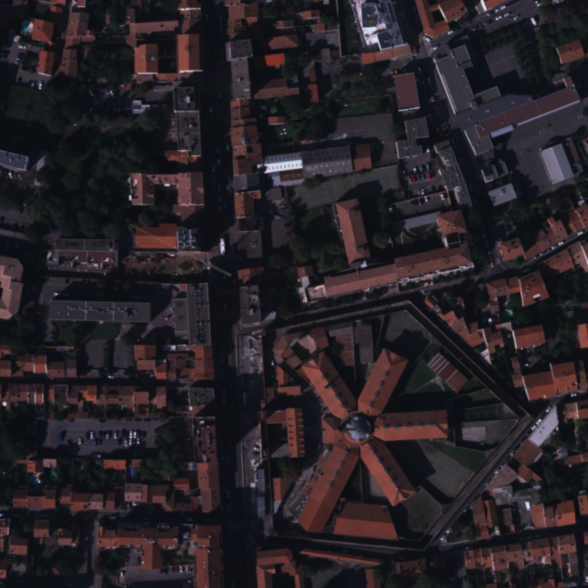} &
  \includegraphics[trim= 12cm 4cm 6cm 14cm, clip=true, width=0.243\textwidth]{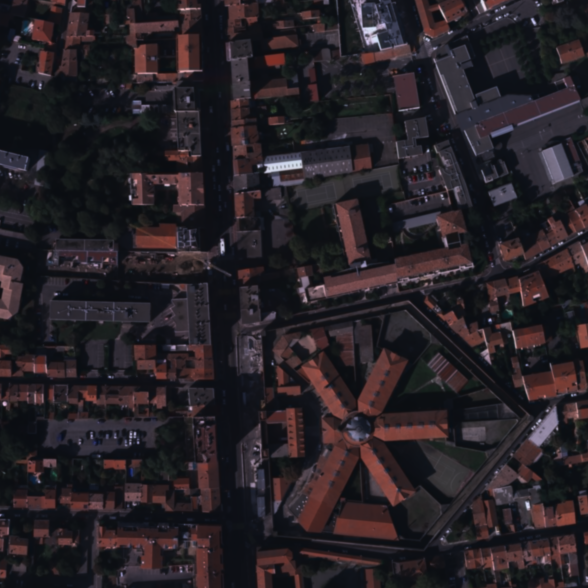} \\
  P+XS & NLV & NLVD\\
\end{tabular}
\caption{Close-ups of the reference RGB image at a resolution of 60 cm per pixel and of the fusion products provided by all methods under comparison. The Gaussian standard deviation used for the simulation of the low-resolution spectral components was $\sigma=1.3$. For these experiments, the data were registered and the panchro-spectral constraint fulfilled with $\alpha_B=\alpha_G=\alpha_R=\frac{1}{3}$. NLV and NLVD obtain convincing results in terms of spatial and spectral quality, being no significant differences between both. All pansharpening techniques except the two previous ones cause color distortions. Indeed, see that most of the objects in the scene, such as the blue fireplaces on the roofs, become almost grayish because of a reduction of the saturation of the chromatic components. This phenomenon is much apparent in CS-based than MRA-based fusion. Although being slightly better in terms of spectral quality, all multiresolution strategies severely compromise the geometry of the fusion products. In this regard, note how the contours of the buildings are partially damaged.}
\label{fig_60cm_s13_RGB_regist_linear}
\end{figure}

\begin{figure}[!p]
\footnotesize
\centering
\renewcommand{\arraystretch}{0.5}
\begin{tabular}{c@{\hskip 0.02in}c@{\hskip 0.02in}c@{\hskip 0.02in}c}
  \includegraphics[trim= 12cm 4cm 6cm 14cm, clip=true, width=0.243\textwidth]{RGBreglin_true.png} &
  \includegraphics[trim= 12cm 4cm 6cm 14cm, clip=true, width=0.243\textwidth]{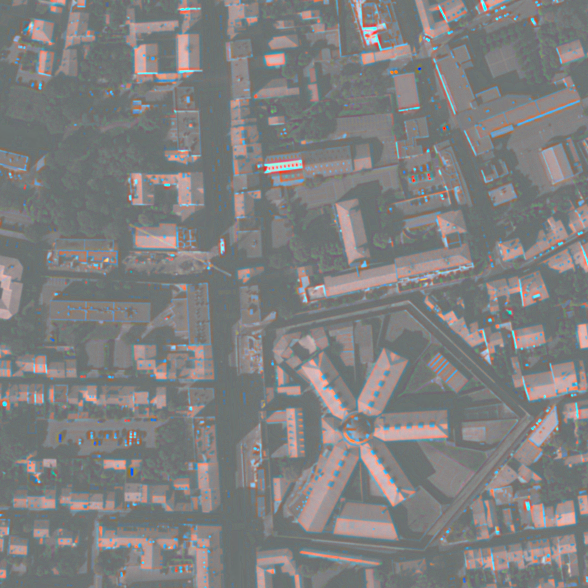} &
  \includegraphics[trim= 12cm 4cm 6cm 14cm, clip=true, width=0.243\textwidth]{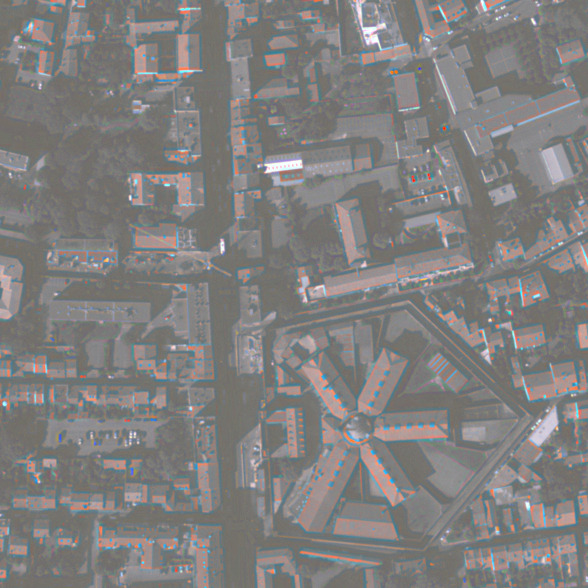} &
  \includegraphics[trim= 12cm 4cm 6cm 14cm, clip=true, width=0.243\textwidth]{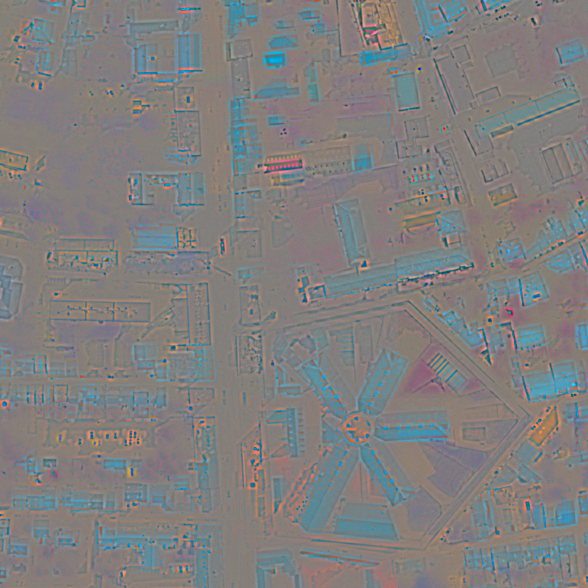} \\
  Reference & PCA & Brovey & BDSD \\
  \includegraphics[trim= 12cm 4cm 6cm 14cm, clip=true, width=0.243\textwidth]{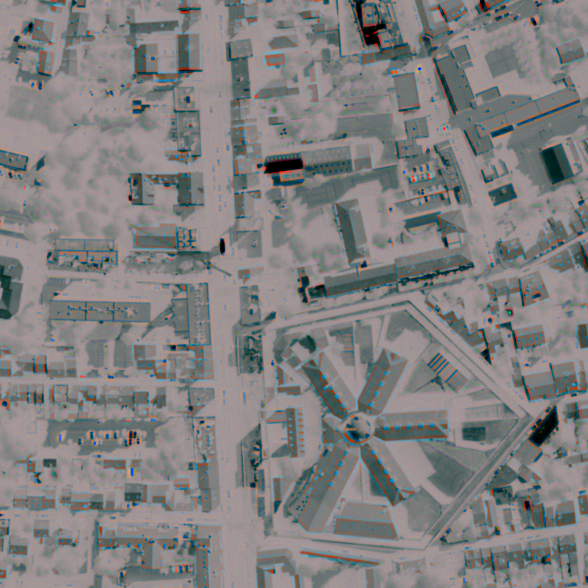} &
  \includegraphics[trim= 12cm 4cm 6cm 14cm, clip=true, width=0.243\textwidth]{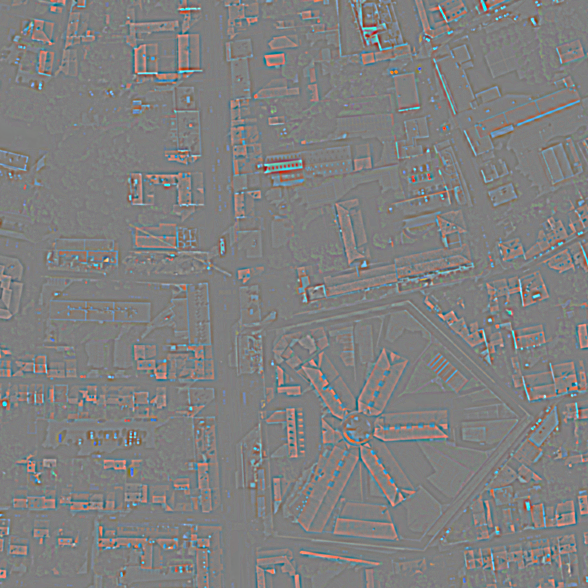} &
  \includegraphics[trim= 12cm 4cm 6cm 14cm, clip=true, width=0.243\textwidth]{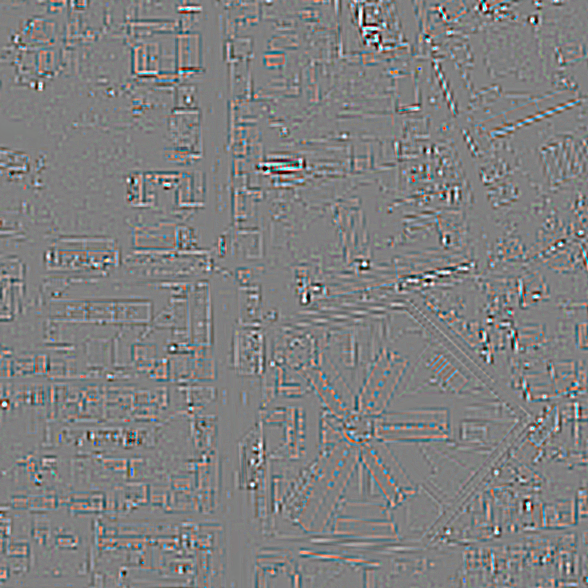} &
  \includegraphics[trim= 12cm 4cm 6cm 14cm, clip=true, width=0.243\textwidth]{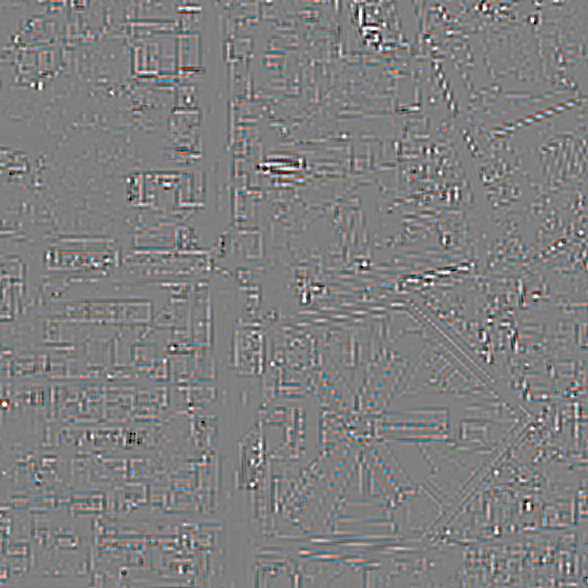} \\
  GSA & PRACS & HPF & SFIM \\
  \includegraphics[trim= 12cm 4cm 6cm 14cm, clip=true, width=0.243\textwidth]{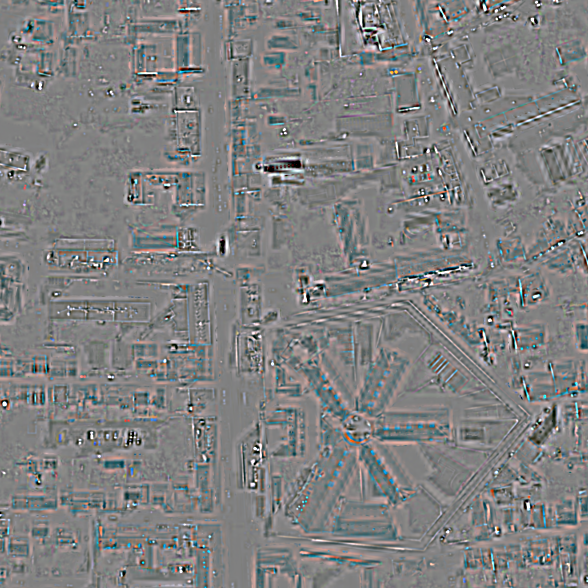} &
  \includegraphics[trim= 12cm 4cm 6cm 14cm, clip=true, width=0.243\textwidth]{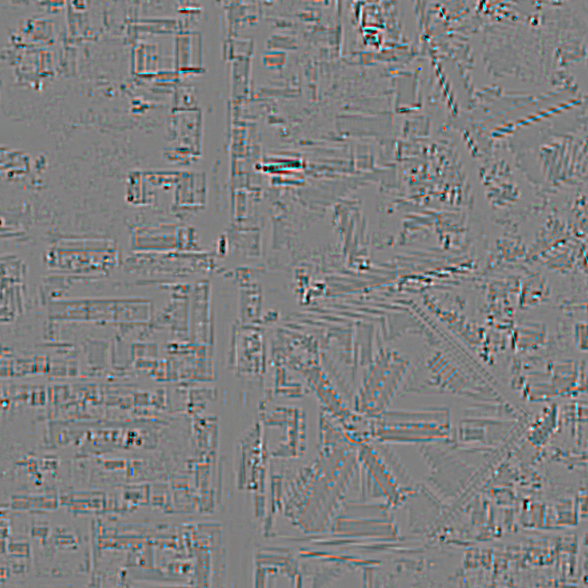} &
  \includegraphics[trim= 12cm 4cm 6cm 14cm, clip=true, width=0.243\textwidth]{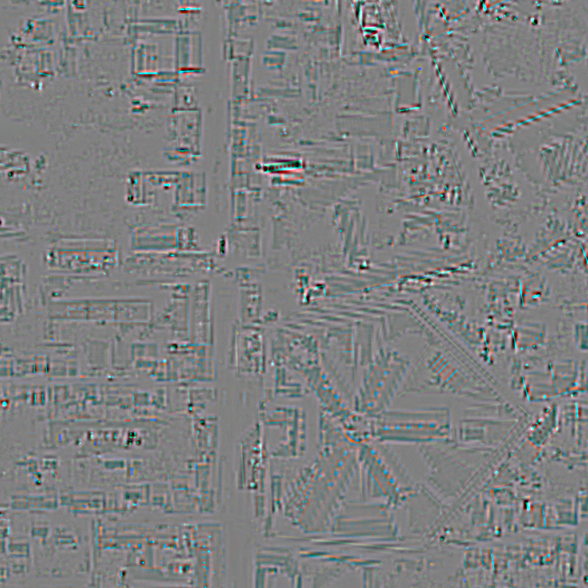} &
  \includegraphics[trim= 12cm 4cm 6cm 14cm, clip=true, width=0.243\textwidth]{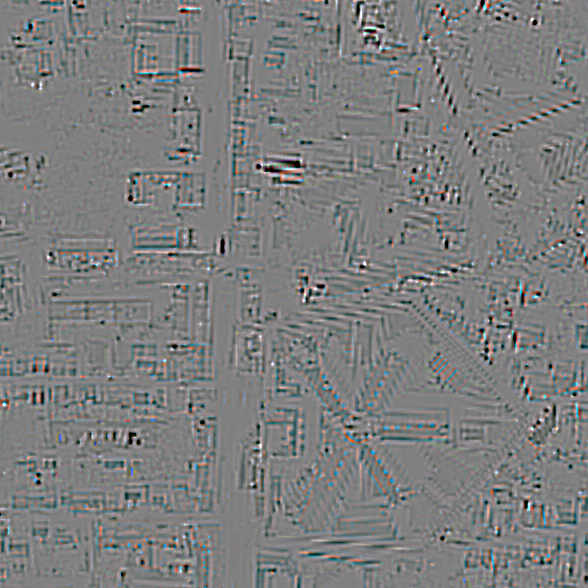} \\
  LMVM & ATWT & AWLP & GLP \\
\end{tabular}
\begin{tabular}{c@{\hskip 0.02in}c@{\hskip 0.02in}c} 
  \includegraphics[trim= 12cm 4cm 6cm 14cm, clip=true, width=0.243\textwidth]{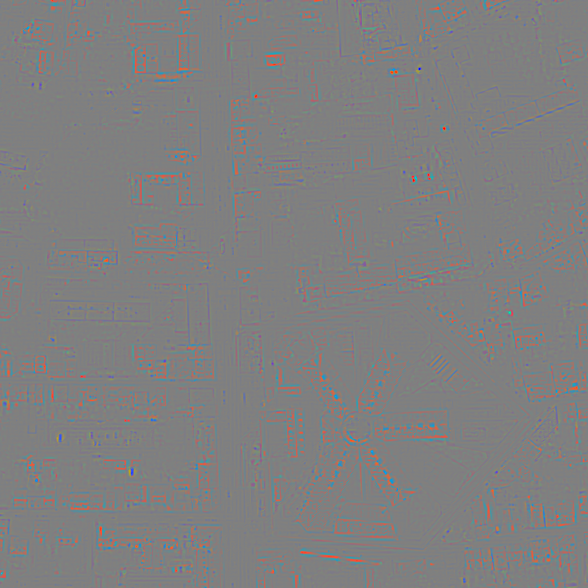} &
  \includegraphics[trim= 12cm 4cm 6cm 14cm, clip=true, width=0.243\textwidth]{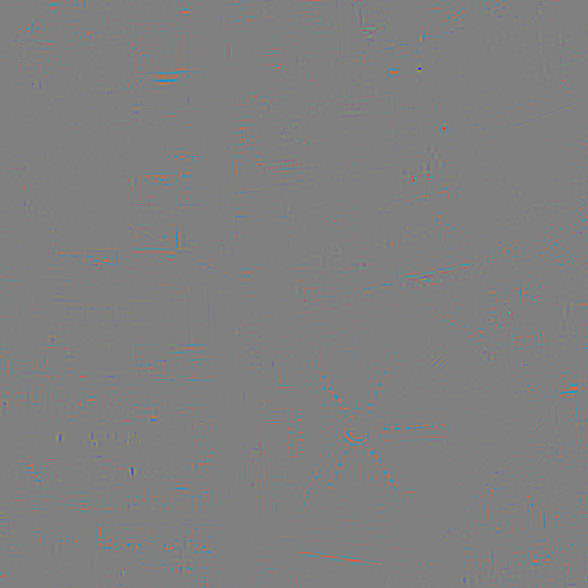} &
  \includegraphics[trim= 12cm 4cm 6cm 14cm, clip=true, width=0.243\textwidth]{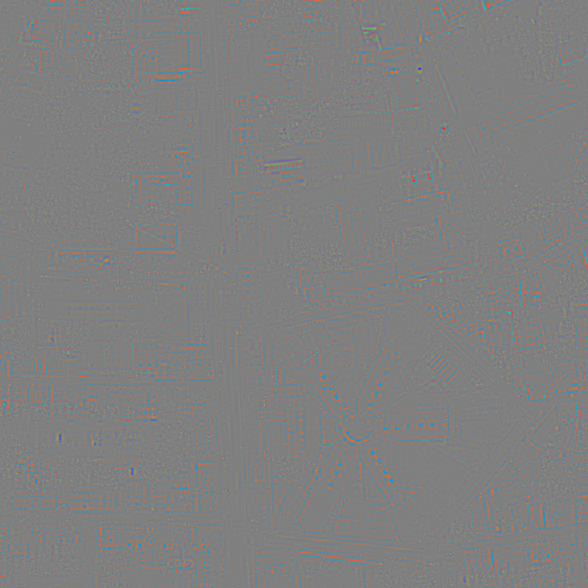} \\
  P+XS & NLV & NLVD\\
\end{tabular}
\caption{Close-ups of the reference RGB image at a resolution of 60 cm per pixel and of the difference images associated to the fusion products displayed in Figure \ref{fig_60cm_s13_RGB_regist_linear}. For visualization purposes, the intensity values have been linearly rescaled from $[-20,20]$ to $[0,255]$. The first conclusion that can be drawn is the superiority of NLV and NLVD methods since the corresponding difference images contain much less amount of information. Despite being numerically close to them, P+XS introduces annoying color spots. In general terms, CS and MRA families lead to greater spectral and spatial quality loss than variational models. On the one hand, CS-based techniques mainly suffer from color distortions since the chromatic components are almost in the difference images. In particular, observe the poor quality of the results provided by BDSD and GSA -- in the latter case, the outstanding dark gray means that the distortion has the same magnitude in all channels.  On the other hand, the difference images associated to MRA-based techniques contain jagged edges because of aliasing, which implies that the geometry of the fusion products has been damaged.}
\label{fig_60cm_s13_RGB_regist_linear_dif}
\end{figure}

Figure \ref{fig_60cm_s13_RGB_regist_linear} displays close-ups of the fusion products on the first image from the proposed dataset at a resolution of 60 cm per pixel. The Gaussian standard deviation used for the simulation of the low-resolution spectral components was $\sigma=1.3$. In order to help the visual analysis, Figure \ref{fig_60cm_s13_RGB_regist_linear_dif} shows the difference images between the reference one and the result provided by each method. The first conclusion that can be drawn from the visual inspection is the superiority of the variational models under comparison, particularly NLV and NLVD, over CS-based as well as MRA-based techniques. Although P+XS performed well from a quantitive point of view, the pansharpened image shows some annoying color spots. We further observe that CS and MRA families lead to greater spectral and spatial quality loss than NLV and NLVD. For example, see how the color of the fireplaces in the fused products provided by these two classes of approaches is not preserved or how the contours of the buildings are in the difference images from Figure \ref{fig_60cm_s13_RGB_regist_linear_dif}, meaning that the geometry in the fused products has been partially distorted. Finally, let us emphasize that MRA-based fusion techniques are the most affected by aliasing. Indeed, see the jagged edges of the buildings in Figure \ref{fig_60cm_s13_RGB_regist_linear_dif}.

\subsubsection{Non-Registered Bands and Panchro-Spectral Constraint Fulfilled}

In view of the misregistration of satellite imagery, we modified the way the low-resolution data were simulated to make it more realistic. The panchromatic images were still obtained by linear combination of the spectral bands at resolutions of 30 cm and 60 cm per pixel. Before computing the low-resolution channels, we first applied a translation by splines -- different for each one -- to the ground-truth components. Then, we simulated each low-resolution spectral bands by filtering the corresponding high-resolution one after translation with Gaussian kernel followed by subsampling of factor $s=4$. We also used here different standard deviations, $\sigma\in\{1.3, 1.7\}$, to introduce different amounts of aliasing. 

Channel-decoupled pansharpening models, such as HPF, SFIM, LMVM, ATWT, GLP, and the new-proposed NLVD can be applied to each component independently after superimposing the panchromatic, which hardly contains aliasing, into the reference of each spectral band. For visualization purposes, the inferred high-resolution channels are then registered into a common geometry using the inverse translation. However, all other techniques under comparison require spectral components to be co-registered. In these cases, the low-resolution bands are first co-registered and then the methods applied. Since HPF, SFIM, LMVM, ATWT, GLP, and NLVD also work on co-registered data, we report the numerical results obtained in both chains. However, we only display visually the pansharpened image provided by the variant with better quality indices. Once again, it is important to keep in mind that Brovey, BDSD, GSA, PRACS, AWLP, P+XS, and NLV make use of the panchro-spectral constraint.

\medskip

\noindent {\bf RGB images.} We first test all methods on RGB color images with the same weight per channel in the linear combination assumption, that is, $\alpha_B=\alpha_G=\alpha_R=\frac{1}{3}$. 

Tables \ref{table_30cm_RGB_nonregist_linear} and \ref{table_60cm_RGB_nonregist_linear} display the quantitative results in RGB coordinates generated by each method. First of all, we observe that NLVD, even though not taking advantage of the validity of the panchro-spectral constraint, outperforms all other methods in terms of any of the metrics used in the compared quality assessment. This superiority is even more clear when $\sigma$ decreases. Another important issue that deserves to be highlighted is that HPF, SFIM, and particularly NLVD improve their numerical results when pansharpening takes place before co-registration of the spectral components. The opposite happens with techniques using Laplacian-pyramid or wavelet decomposition strategies such as ATWT and GLP. Halfway between both cases, LMVM obtains better results on misregistered data for $\sigma=1.7$, but the spectral distortion quantified by the SAM index is lower on co-registered spectral components if $\sigma=1.3$ is considered. By comparing the results in Tables  \ref{table_30cm_RGB_nonregist_linear} and \ref{table_60cm_RGB_nonregist_linear} with those reported in Tables \ref{table_30cm_RGB_regist_linear} and \ref{table_60cm_RGB_regist_linear}, we realize that only NLVD is able to get similar competitive quality indices, which demonstrates the robustness of the proposed model to misregistration and aliasing. It is also worth noticing that P+XS is not competitive any more in terms of spectral quality since the corresponding SAM values are now one of the highest. 

\begin{table}[!p]
\footnotesize
\centering
\begin{subtable}[h]{0.99\textwidth}
\centering
\begin{tabular}{|c|c|c|c|c|c|c|c|c|c|}
\hline
 & Methods & RMSE & ERGAS & SAM & SSIM & Q$2^n$  \\ \hline \hline
 & Reference & 0 & 0 & 0 & 1 & 1 \\ \hline \hline
 \multirow{14}{*}{\rotatebox[origin=c]{90}{Registered}} & PCA & 3.0039 & 2.2055 & 1.8849 & 0.9932 & 0.9673 \\ \cline{2-7}
& Brovey & 2.7954 & 2.0489 & 1.7668 & 0.9937 & 0.9679 \\ \cline{2-7}
& BDSD & 1.9628 & 1.4337 & 2.0636 & 0.9979 & 0.9895 \\ \cline{2-7}
& GSA & 2.6904 & 1.9767 & 1.9991 & 0.9951 & 0.9775 \\ \cline{2-7}
& PRACS & 2.3621 & 1.7260 & 1.5384 & 0.9976 & 0.9837 \\ \cline{2-7}
& HPF & 3.0622 & 2.2510 & 1.6067 & 0.9951 & 0.9777 \\ \cline{2-7}
& SFIM & 2.9550 & 2.1718 & 1.4619 & 0.9950 & 0.9775 \\ \cline{2-7}
& LMVM & 3.2791 & 2.4233 & 1.3920 & 0.9958 & 0.9698 \\ \cline{2-7}
& ATWT & 2.2277 & 1.6317 & 1.5484 & 0.9980 & 0.9881 \\ \cline{2-7}
& AWLP & 2.0889 & 1.5470 & 1.4481 & 0.9982 & 0.9885 \\ \cline{2-7}
& GLP & 2.0919 & 1.5369 & 1.3451 & 0.9980 & 0.9881 \\ \cline{2-7}
& P+XS & 1.9259 & 1.3960 & 1.7746 & 0.9985 & 0.9904 \\ \cline{2-7}
& NLV & 1.6488 & 1.1991 & 1.3790 & 0.9987 & 0.9923 \\ \cline{2-7}
& NLVD & 1.7785 & 1.2821 & 1.5380 & 0.9984 & 0.9901 \\ \hline\hline
 \multirow{6}{*}{\rotatebox[origin=c]{90}{Misregistered}} & HPF & 2.7936 & 2.0656 & 1.5229 & 0.9964 & 0.9795 \\ \cline{2-7}
& SFIM & 2.6506 & 1.9612 & 1.3734 & 0.9963 & 0.9794 \\ \cline{2-7}
& LMVM & 3.1284 & 2.3142 & 1.3300 & 0.9972 & 0.9714 \\ \cline{2-7}
& ATWT & 2.1991 & 1.6158 & 1.5669 & 0.9983 & 0.9871 \\ \cline{2-7}
& GLP & 2.2191 & 1.6279 & 1.5535 & 0.9980 & 0.9862 \\ \cline{2-7}
& NLVD & {\bf 1.2539} & {\bf 0.9096} & {\bf 0.9379} & {\bf 0.9997} & {\bf 0.9924} \\ \hline
\end{tabular}
\caption{Numerical results for $\sigma=1.7$.}
 \label{table_30cm_s17_RGB_nonregist_linear}
 \end{subtable}
 
\medskip

\begin{subtable}[h]{0.99\textwidth}
\centering
\begin{tabular}{|c|c|c|c|c|c|c|c|c|}
\hline
 & & RMSE & ERGAS & SAM & SSIM & Q$2^n$  \\ \hline \hline
 & Reference & 0 & 0 & 0 & 1 & 1 \\ \hline \hline
\multirow{14}{*}{\rotatebox[origin=c]{90}{Registered}} & PCA & 2.8007 & 2.0542 & 1.8913 & 0.9942 & 0.9710 \\ \cline{2-7}
 & Brovey & 2.6037 & 1.9038 & 1.8320 & 0.9948 & 0.9722 \\ \cline{2-7}
 & BDSD & 2.2150 & 1.6258 & 2.2283 & 0.9971 & 0.9870 \\ \cline{2-7}
 & GSA & 3.2939 & 2.4315 & 2.3180 & 0.9916 & 0.9677 \\ \cline{2-7}
 & PRACS & 2.2815 & 1.6614 & 1.6095 & 0.9980 & 0.9866 \\ \cline{2-7}
 & HPF & 2.7942 & 2.0488 & 1.6760 & 0.9964 & 0.9815 \\ \cline{2-7}
 & SFIM & 2.6753 & 1.9616 & 1.5343 & 0.9964 & 0.9818 \\ \cline{2-7}
 & LMVM & 3.2563 & 2.3997 & 1.5485 & 0.9955 & 0.9718 \\ \cline{2-7}
 & ATWT & 2.3512 & 1.7268 & 1.6523 & 0.9973 & 0.9865 \\ \cline{2-7}
 & AWLP & 2.2490 & 1.6606 & 1.5143 & 0.9974 & 0.9870 \\ \cline{2-7}
 & GLP & 2.5726 & 1.9054 & 1.4290 & 0.9965 & 0.9844 \\ \cline{2-7}
 & P+XS & 2.2787 & 1.6481 & 2.0610 & 0.9980 & 0.9864 \\ \cline{2-7}
 & NLV & 1.8613 & 1.3531 & 1.4922 & 0.9984 & 0.9915 \\ \cline{2-7}
 & NLVD & 2.0456 & 1.4674 & 1.6055 & 0.9978 & 0.9886 \\ \hline\hline
 \multirow{6}{*}{\rotatebox[origin=c]{90}{Misregistered}} & HPF & 2.5747 & 1.8976 & 1.6611 & 0.9973 & 0.9826 \\ \cline{2-7}
 & SFIM & 2.4197 & 1.7845 & 1.5091 & 0.9973 & 0.9831 \\ \cline{2-7}
 & LMVM & 3.2381 & 2.3894 & 1.7085 & 0.9954 & 0.9726 \\ \cline{2-7}
 & ATWT & 2.4977 & 1.8342 & 1.7665 & 0.9970 & 0.9844 \\ \cline{2-7}
 & GLP & 2.9509 & 2.1709 & 1.7641 & 0.9956 & 0.9808 \\ \cline{2-7}
 & NLVD & {\bf 1.2685} & {\bf 0.9211} & {\bf 0.9153} & {\bf 0.9997} & {\bf 0.9924} \\ \hline
\end{tabular}
\caption{Numerical results for $\sigma=1.3$.}
 \label{table_30cm_s13_RGB_nonregist_linear}
 \end{subtable}
 \caption{Quantitative evaluation of the fused products on simulated data from RGB aerial images at resolution of 30 cm per pixel. For these experiments, the low-resolution spectral components were non registered but the panchro-spectral constraint fulfilled with $\alpha_B=\alpha_G=\alpha_R=\frac{1}{3}$. Note that NLVD outperforms all other techniques for any quality index. It is also worth underlining that the differences even increase as the amount of aliasing so does.}
 \label{table_30cm_RGB_nonregist_linear}
\end{table}

\begin{table}[!p]
\footnotesize
\centering
\begin{subtable}[h]{0.99\textwidth}
\centering
\begin{tabular}{|c|c|c|c|c|c|c|c|c|c|}
\hline
 & Methods & RMSE & ERGAS & SAM & SSIM & Q$2^n$  \\ \hline \hline
 & Reference & 0 & 0 & 0 & 1 & 1 \\ \hline \hline
 \multirow{14}{*}{\rotatebox[origin=c]{90}{Registered}} & PCA & 3.6253 & 2.6568 & 2.3229 & 0.9847 & 0.9664 \\ \cline{2-7}
& Brovey & 3.5296 & 2.5844 & 2.3043 & 0.9841 & 0.9665 \\ \cline{2-7}
& BDSD & 2.3694 & 1.7329 & 2.6764 & 0.9921 & 0.9904 \\ \cline{2-7}
& GSA & 3.2179 & 2.3634 & 2.5458 & 0.9868 & 0.9768 \\ \cline{2-7}
& PRACS & 2.7105 & 1.9739 & 1.9125 & 0.9897 & 0.9849 \\ \cline{2-7}
& HPF & 3.5210 & 2.5823 & 2.0088 & 0.9785 & 0.9756 \\ \cline{2-7}
& SFIM & 3.3798 & 2.4790 & 1.8330 & 0.9790 & 0.9756 \\ \cline{2-7}
& LMVM & 3.6546 & 2.6913 & 1.7220 & 0.9783 & 0.9712 \\ \cline{2-7}
& ATWT & 2.5947 & 1.8969 & 1.8936 & 0.9892 & 0.9874 \\ \cline{2-7}
& AWLP & 2.3909 & 1.7675 & 1.7757 & 0.9899 & 0.9879 \\ \cline{2-7}
& GLP & 2.3448 & 1.7189 & 1.6470 & 0.9902 & 0.9880 \\ \cline{2-7}
& P+XS & 2.2741 & 1.6505 & 2.3471 & 0.9914 & 0.9913 \\ \cline{2-7}
& NLV & 1.9506 & 1.4171 & 1.8454 & 0.9937 & 0.9930 \\ \cline{2-7}
& NLVD & 2.1042 & 1.5172 & 2.0948 & 0.9922 & 0.9916 \\ \hline\hline
 \multirow{6}{*}{\rotatebox[origin=c]{90}{Misregistered}} & HPF & 3.1970 & 2.3588 & 1.8820 & 0.9826 & 0.9779 \\ \cline{2-7}
& SFIM & 3.0253 & 2.2336 & 1.6982 & 0.9833 & 0.9782 \\ \cline{2-7}
& LMVM & 3.4377 & 2.5428 & 1.7329 & 0.9817 & 0.9732 \\ \cline{2-7}
& ATWT & 2.5071 & 1.8405 & 1.9055 & 0.9904 & 0.9869 \\ \cline{2-7}
& GLP & 2.4511 & 1.7959 & 1.8927 & 0.9903 & 0.9866 \\ \cline{2-7}
& NLVD & {\bf 1.4781} & {\bf 1.0706} & {\bf 1.2735} & {\bf 0.9966} & {\bf 0.9937} \\ \hline
\end{tabular}
\caption{Numerical results for $\sigma=1.7$.}
 \label{table_60cm_s17_RGB_nonregist_linear}
 \end{subtable}
 
\medskip

\begin{subtable}[h]{0.99\textwidth}
\centering
\begin{tabular}{|c|c|c|c|c|c|c|c|c|}
\hline
 & & RMSE & ERGAS & SAM & SSIM & Q$2^n$  \\ \hline \hline
 & Reference & 0 & 0 & 0 & 1 & 1 \\ \hline \hline
\multirow{14}{*}{\rotatebox[origin=c]{90}{Registered}} & PCA & 3.3452 & 2.4490 & 2.3465 & 0.9864 & 0.9714 \\ \cline{2-7}
& Brovey & 3.2555 & 2.3792 & 2.3757 & 0.9857 & 0.9716 \\ \cline{2-7}
& BDSD & 2.6646 & 1.9578 & 2.9046 & 0.9896 & 0.9877 \\ \cline{2-7}
& GSA & 4.0106 & 2.9580 & 3.0052 & 0.9800 & 0.9649 \\ \cline{2-7}
& PRACS & 2.6074 & 1.8945 & 1.9996 & 0.9903 & 0.9874 \\ \cline{2-7}
& HPF & 3.1970 & 2.3392 & 2.0950 & 0.9827 & 0.9806 \\ \cline{2-7}
& SFIM & 3.0481 & 2.2311 & 1.9165 & 0.9834 & 0.9811 \\ \cline{2-7}
& LMVM & 3.5565 & 2.6205 & 1.9991 & 0.9794 & 0.9739 \\ \cline{2-7}
& ATWT & 2.6364 & 1.9316 & 2.0308 & 0.9886 & 0.9868 \\ \cline{2-7}
& AWLP & 2.4729 & 1.8237 & 1.8555 & 0.9894 & 0.9874 \\ \cline{2-7}
& GLP & 2.7410 & 2.0254 & 1.7519 & 0.9878 & 0.9852 \\ \cline{2-7}
& P+XS & 2.6392 & 1.9099 & 2.6357 & 0.9891 & 0.9885 \\ \cline{2-7}
& NLV & 2.2017 & 1.5999 & 2.0246 & 0.9925 & 0.9919 \\ \cline{2-7}
& NLVD & 2.4038 & 1.7267 & 2.1797 & 0.9903 & 0.9900 \\ \hline\hline
 \multirow{6}{*}{\rotatebox[origin=c]{90}{Misregistered}} & HPF & 2.9146 & 2.1448 & 2.0657 & 0.9861 & 0.9822 \\ \cline{2-7}
& SFIM & 2.7407 & 2.0181 & 1.8726 & 0.9869 & 0.9829 \\ \cline{2-7}
& LMVM & 3.5595 & 2.6253 & 2.2290 & 0.9807 & 0.9738 \\ \cline{2-7}
& ATWT & 2.7393 & 2.0095 & 2.1843 & 0.9885 & 0.9851 \\ \cline{2-7}
& GLP & 3.1447 & 2.3085 & 2.1732 & 0.9859 & 0.9819 \\ \cline{2-7}
& NLVD & {\bf 1.4682} & {\bf 1.0631} & {\bf 1.2419} & {\bf 0.9966} & {\bf 0.9936} \\ \hline
\end{tabular}
\caption{Numerical results for $\sigma=1.3$.}
 \label{table_60cm_s13_RGB_nonregist_linear}
 \end{subtable}
 \caption{Quantitative evaluation of the fused products on simulated data from RGB aerial images at resolution of 60 cm per pixel. For these experiments, the low-resolution spectral components were non registered but the panchro-spectral constraint fulfilled with $\alpha_B=\alpha_G=\alpha_R=\frac{1}{3}$. NLVD is the best method from all metrics and the differences with respect to the other techniques is more outstanding than in Table \ref{table_30cm_RGB_nonregist_linear}.}
 \label{table_60cm_RGB_nonregist_linear}
\end{table}

\begin{figure}[!p]
\footnotesize
\centering
\renewcommand{\arraystretch}{0.5}
\begin{tabular}{c@{\hskip 0.02in}c@{\hskip 0.02in}c@{\hskip 0.02in}c}
  \includegraphics[trim= 23cm 35cm 15.5cm 3.5cm, clip=true, width=0.243\textwidth]{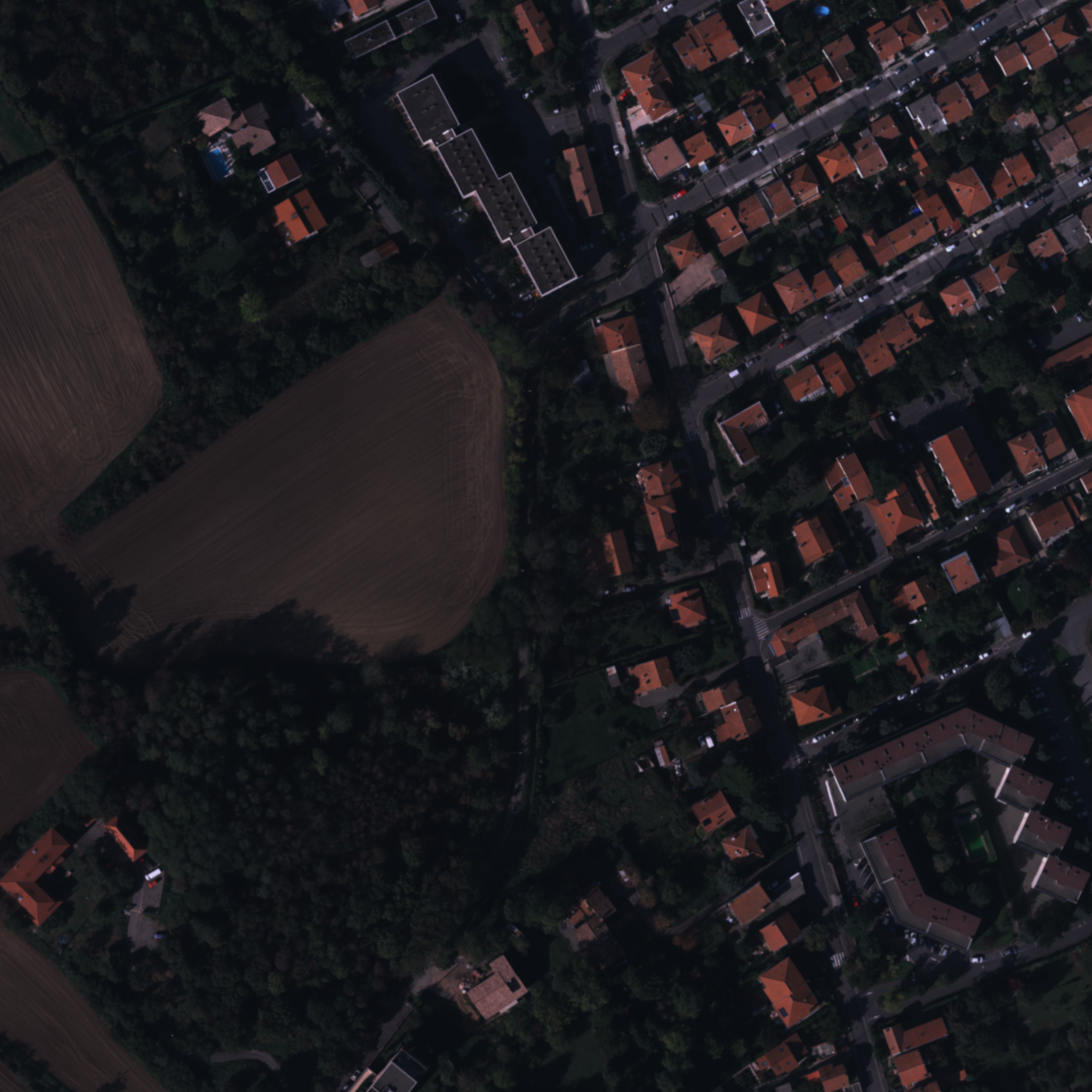} &
  \includegraphics[trim= 23cm 35cm 15.5cm 3.5cm, clip=true, width=0.243\textwidth]{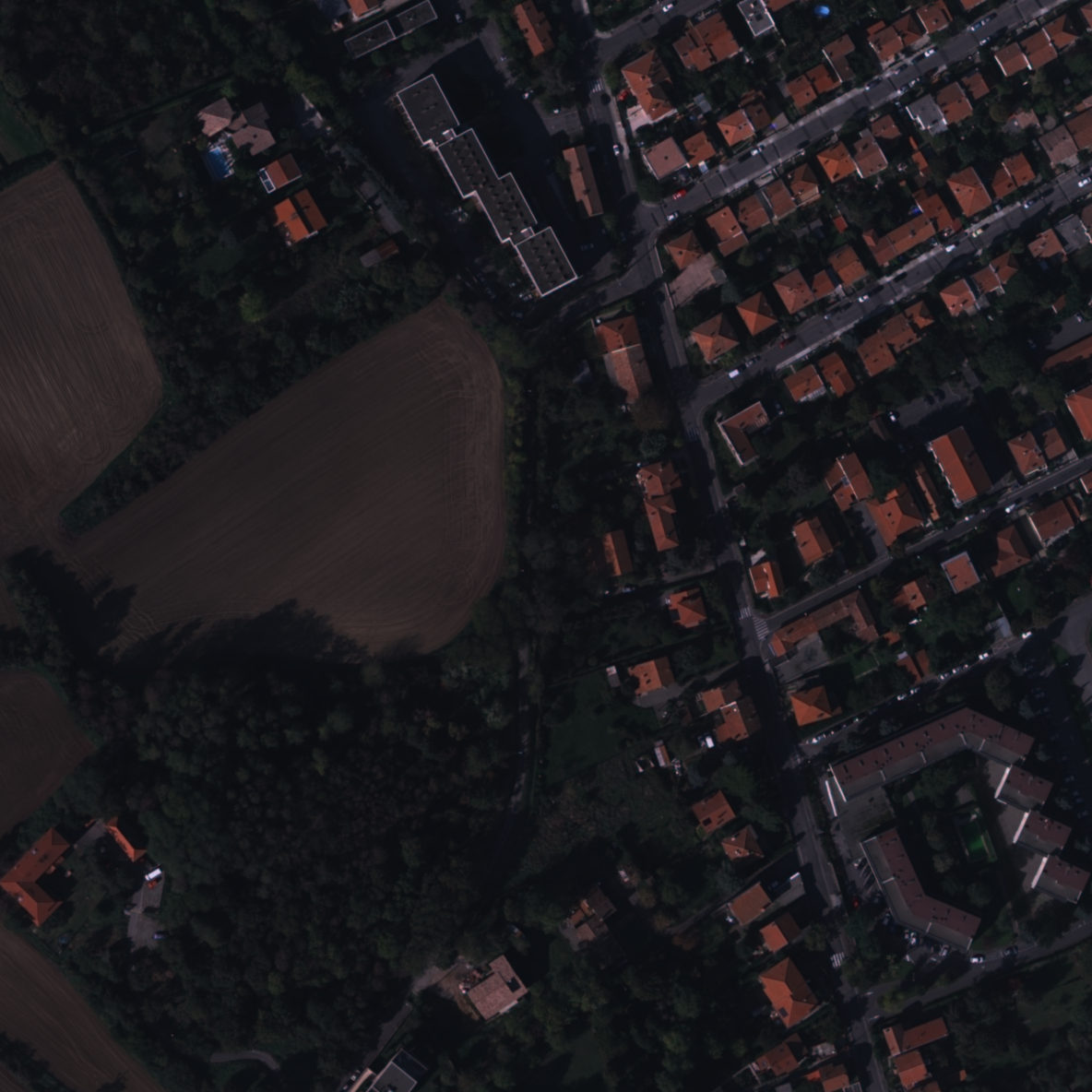} &
  \includegraphics[trim= 23cm 35cm 15.5cm 3.5cm, clip=true, width=0.243\textwidth]{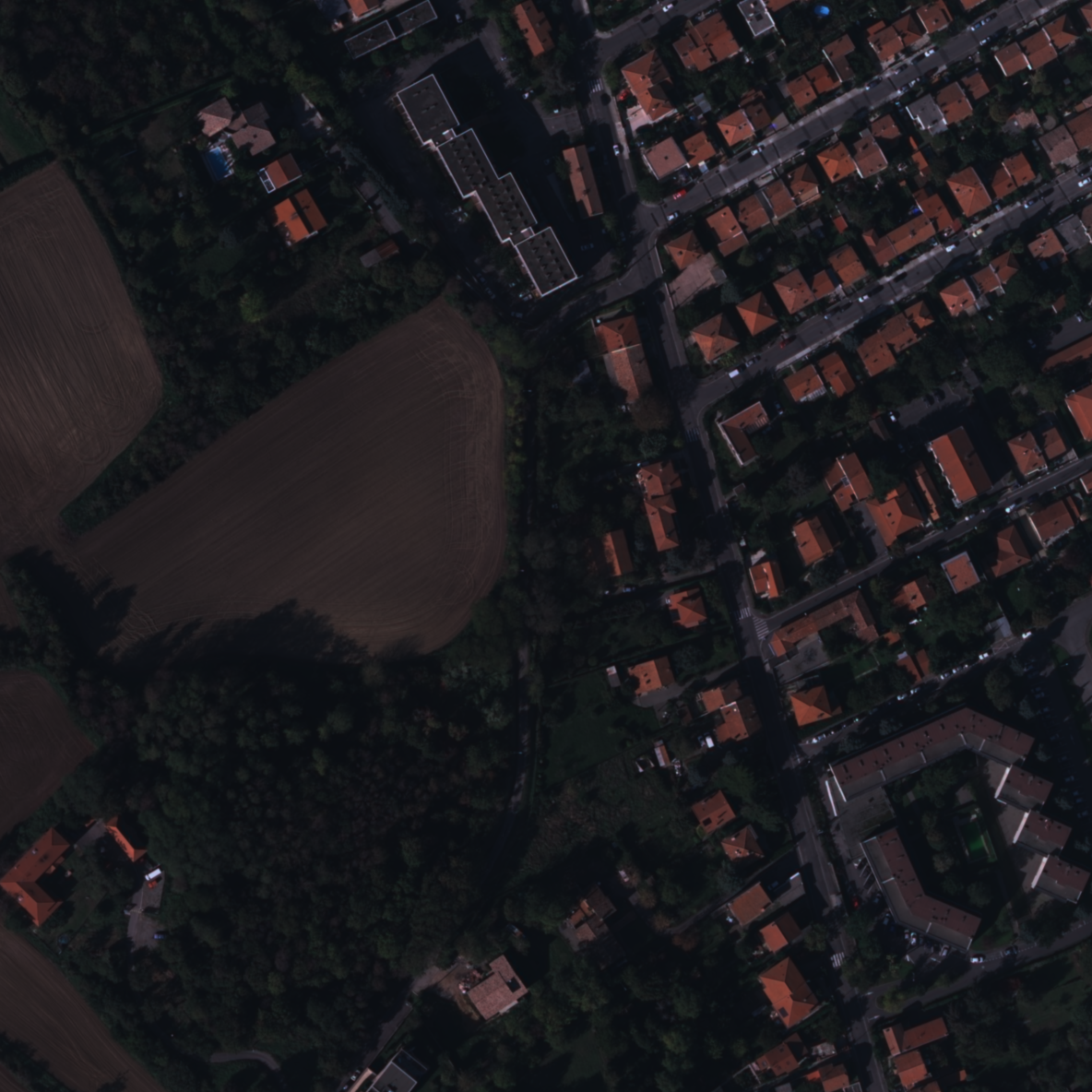} &
  \includegraphics[trim= 23cm 35cm 15.5cm 3.5cm, clip=true, width=0.243\textwidth]{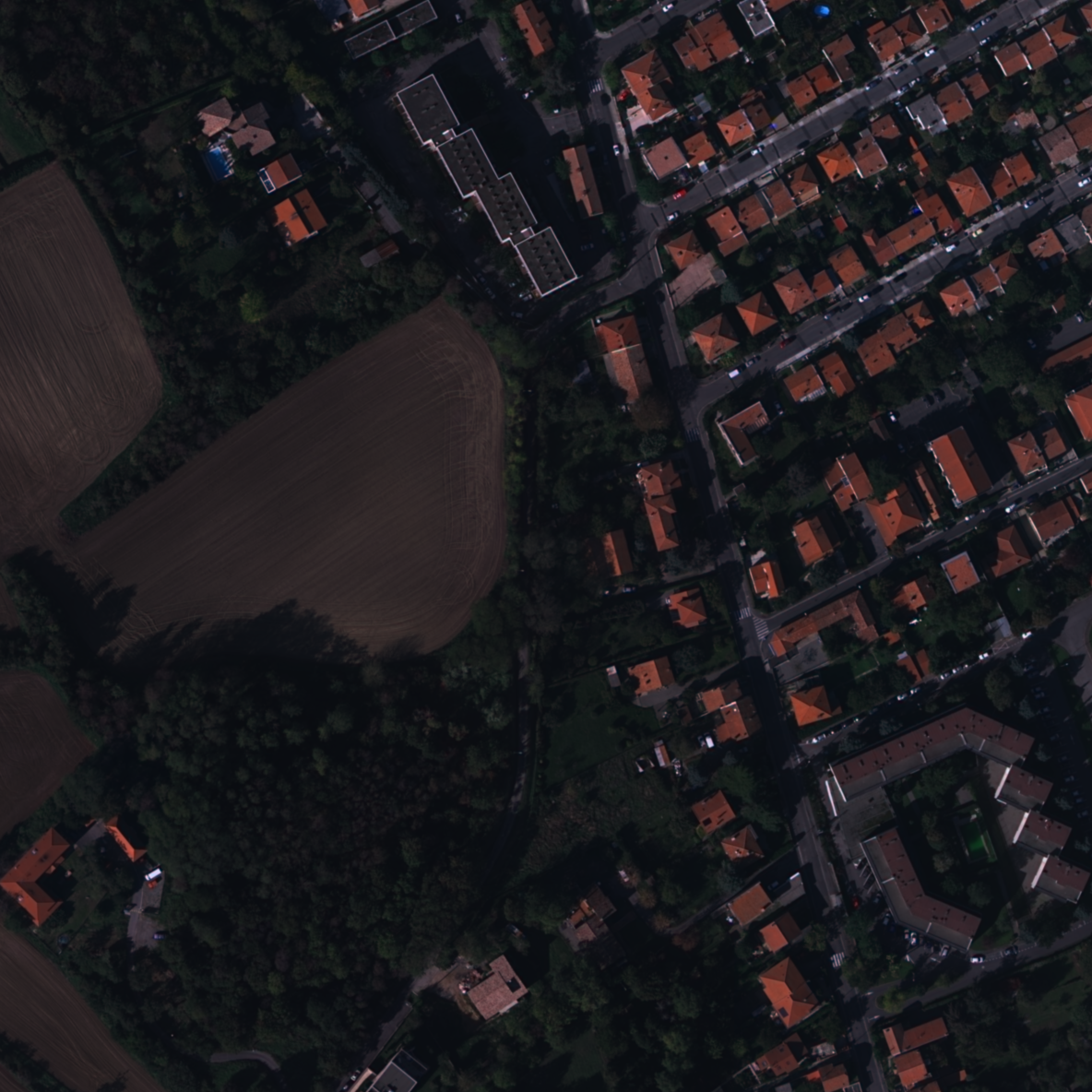} \\
  Reference & PCA & Brovey & BDSD\\ 
  \includegraphics[trim= 23cm 35cm 15.5cm 3.5cm, clip=true, width=0.243\textwidth]{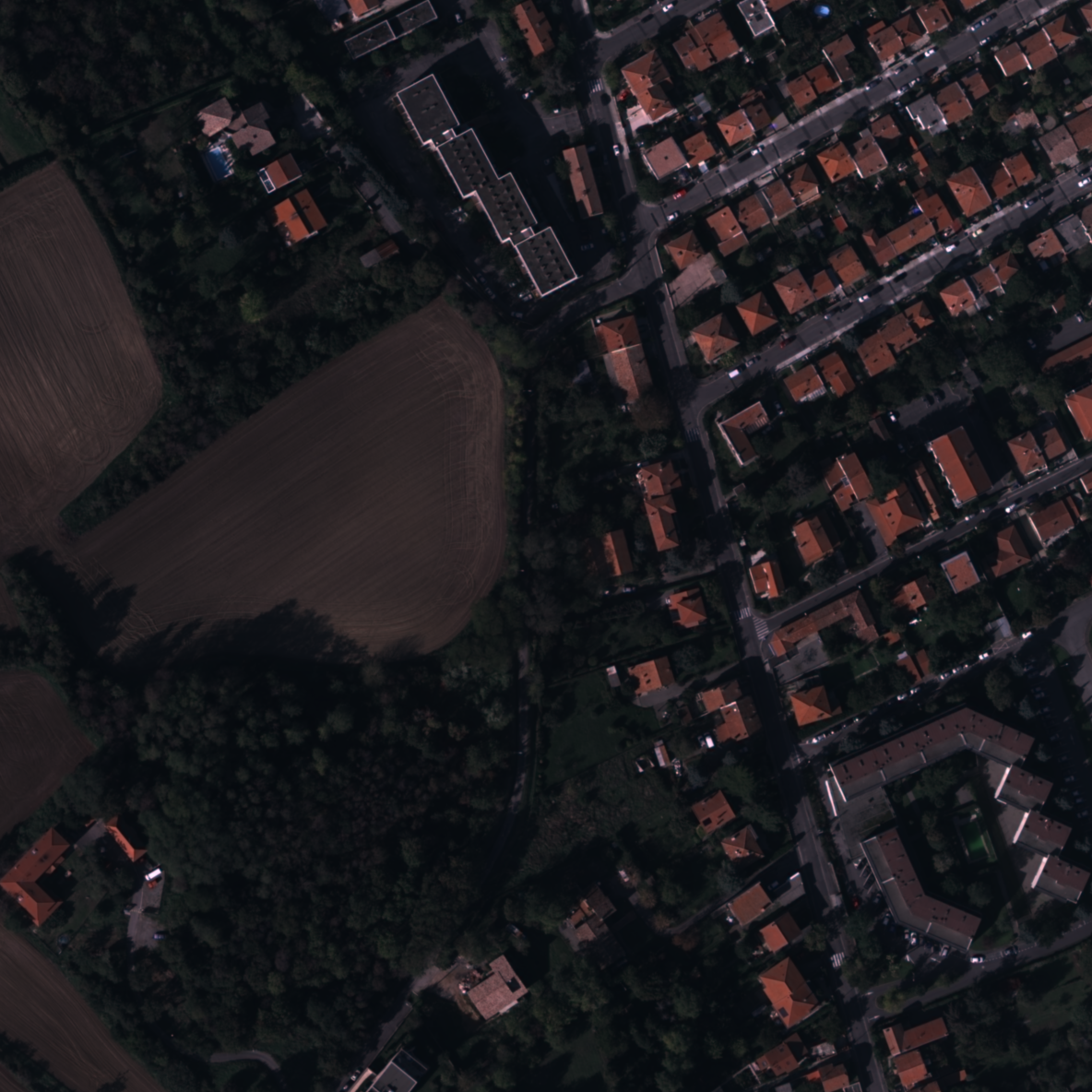} &
  \includegraphics[trim= 23cm 35cm 15.5cm 3.5cm, clip=true, width=0.243\textwidth]{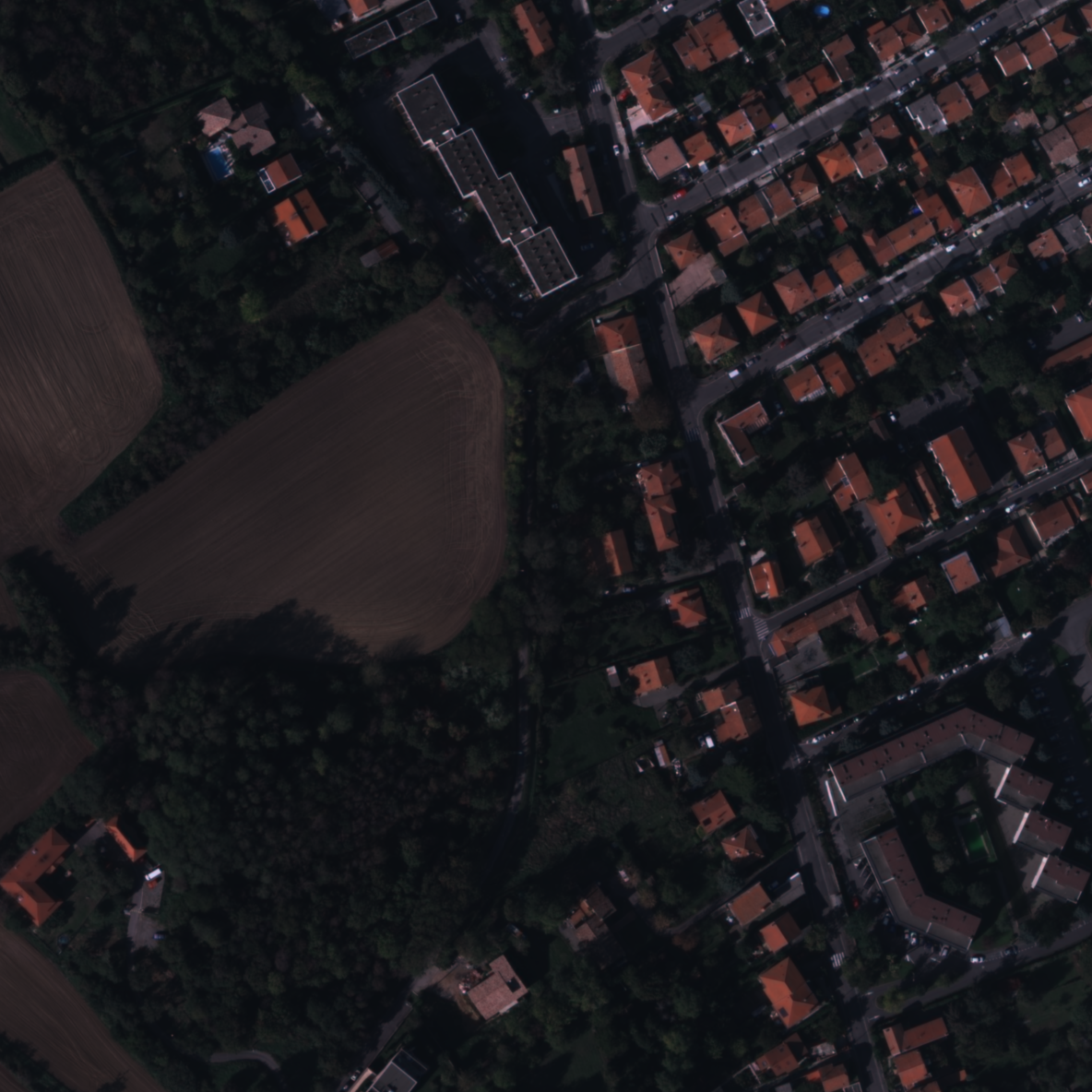} &
  \includegraphics[trim= 23cm 35cm 15.5cm 3.5cm, clip=true, width=0.243\textwidth]{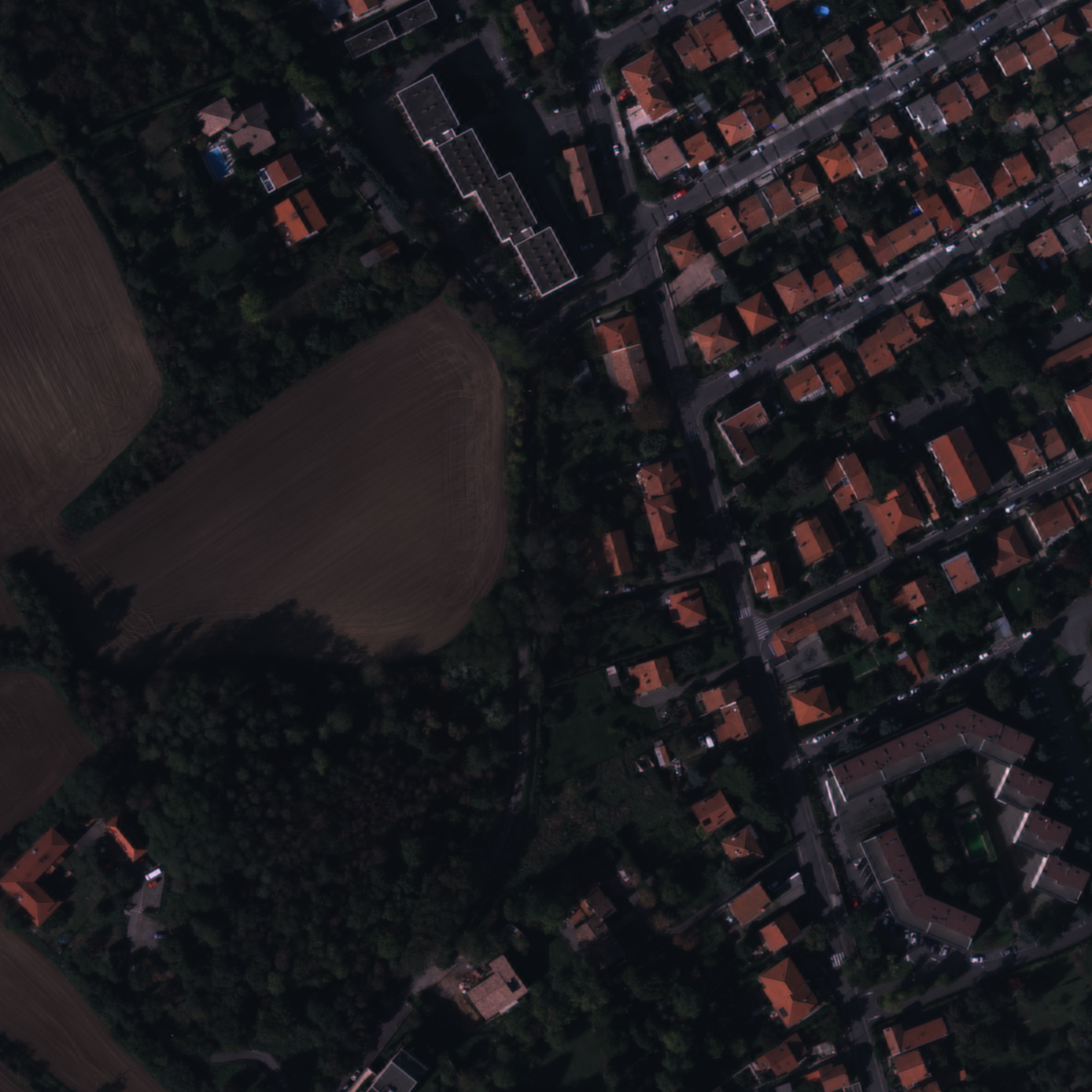} &
  \includegraphics[trim= 23cm 35cm 15.5cm 3.5cm, clip=true, width=0.243\textwidth]{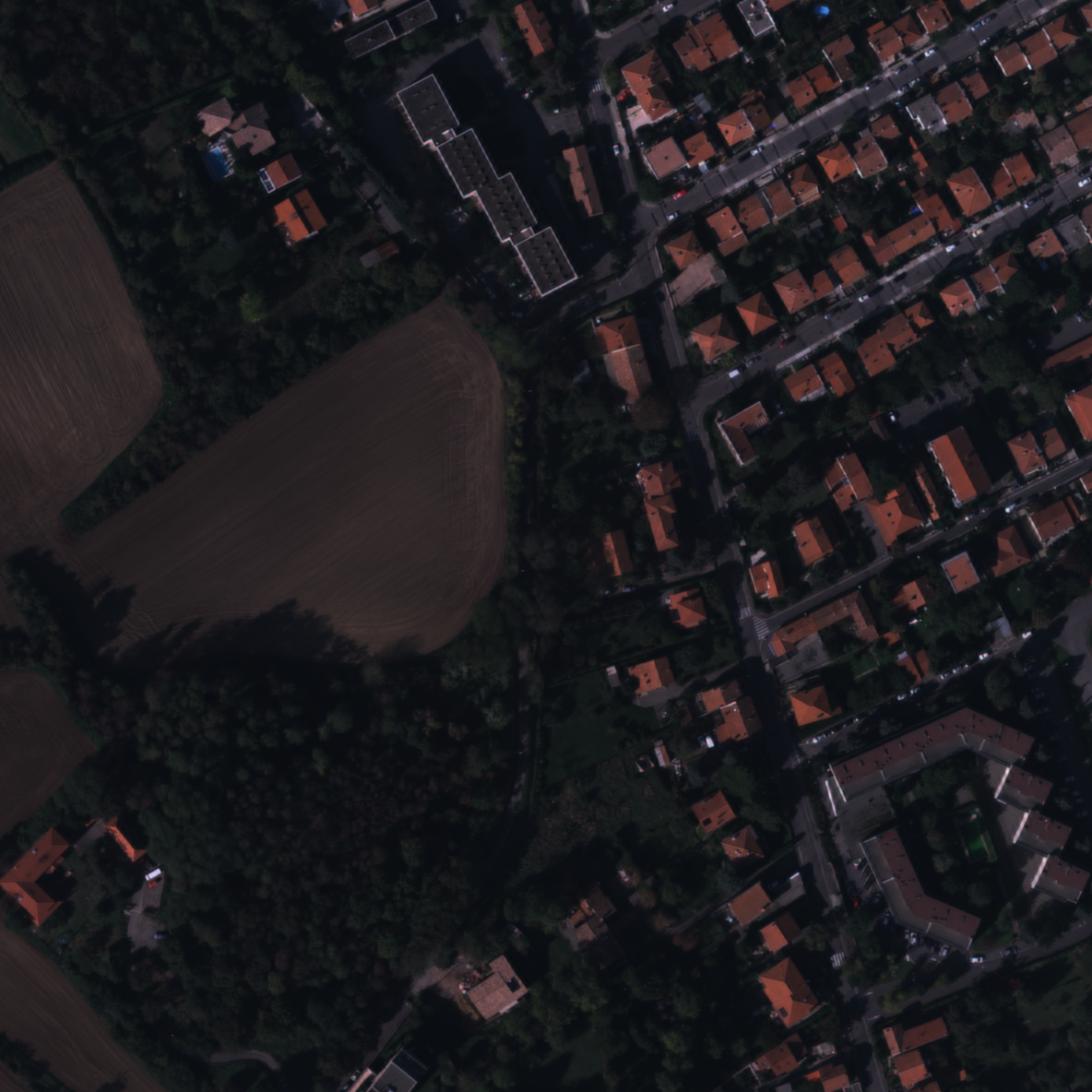} \\
  GSA & PRACS & HPF & SFIM\\
  \includegraphics[trim= 23cm 35cm 15.5cm 3.5cm, clip=true, width=0.243\textwidth]{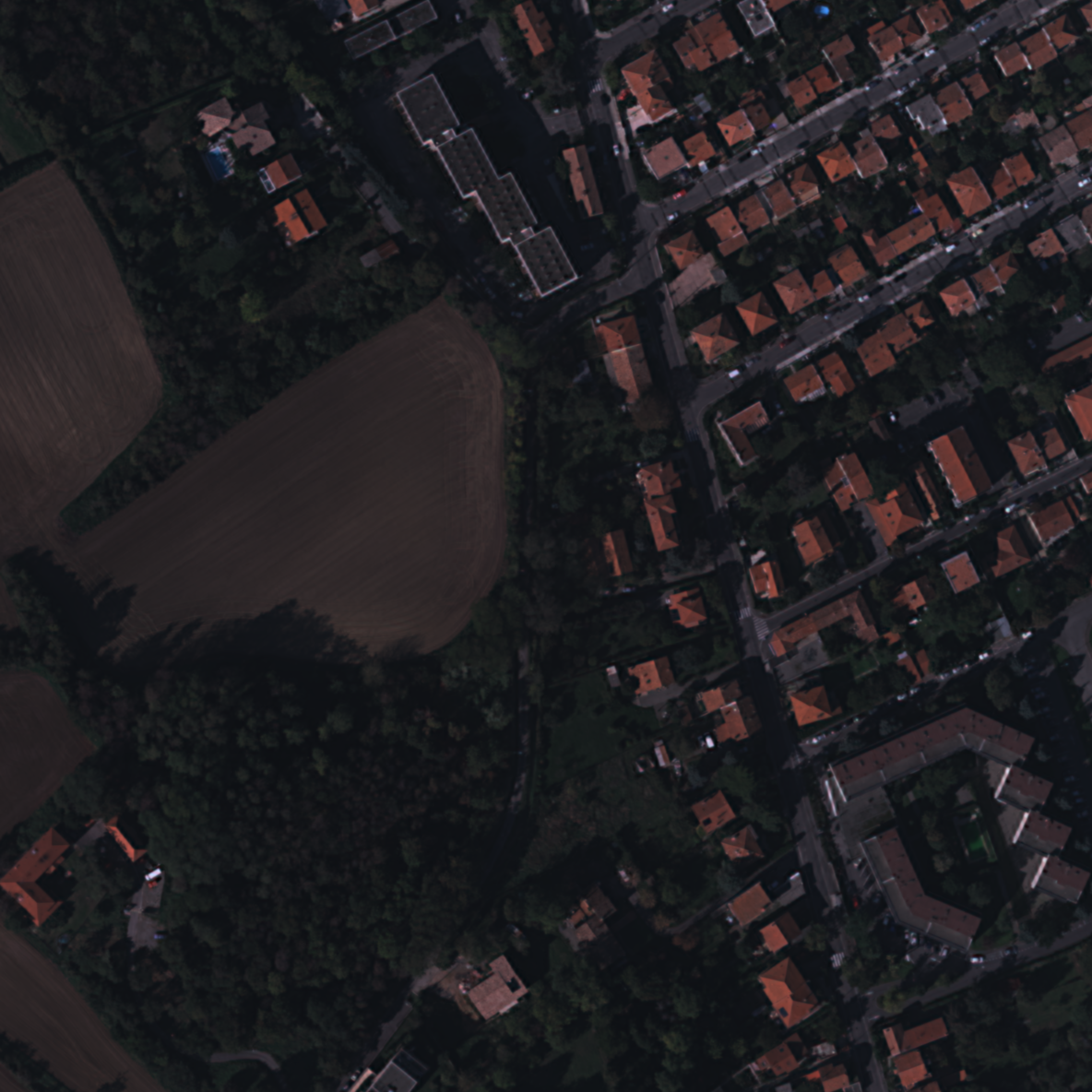} &
  \includegraphics[trim= 23cm 35cm 15.5cm 3.5cm, clip=true, width=0.243\textwidth]{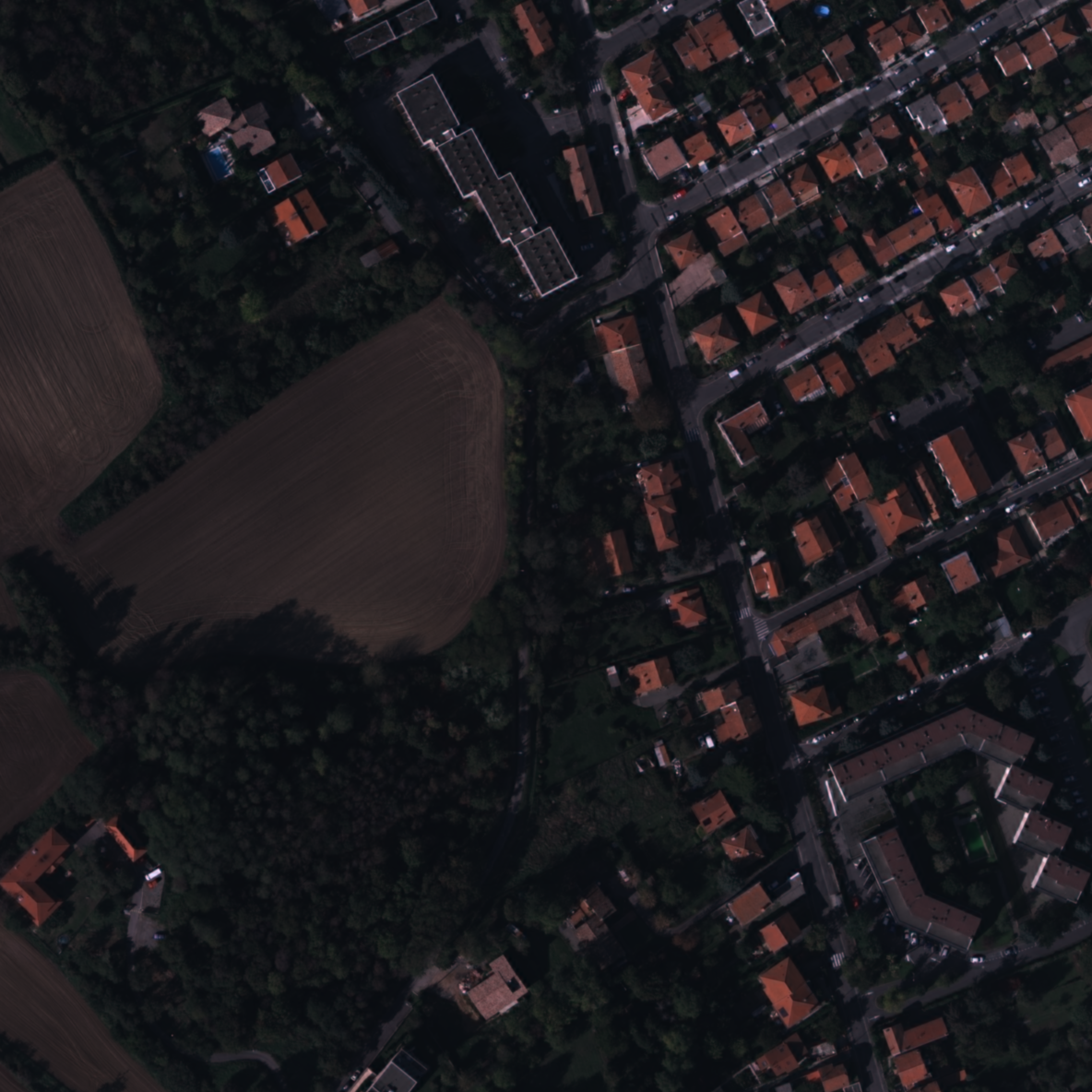} &
  \includegraphics[trim= 23cm 35cm 15.5cm 3.5cm, clip=true, width=0.243\textwidth]{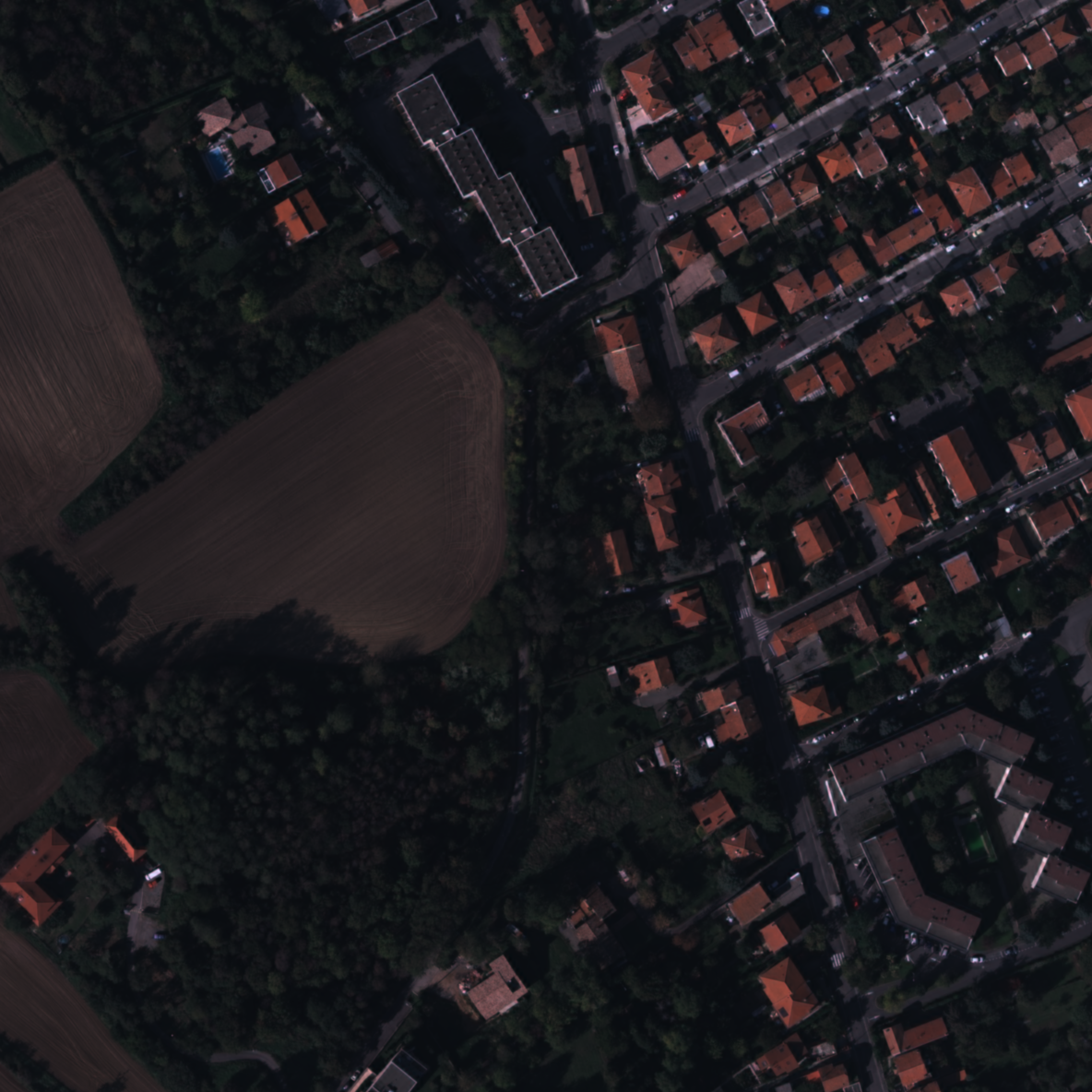} &
  \includegraphics[trim= 23cm 35cm 15.5cm 3.5cm, clip=true, width=0.243\textwidth]{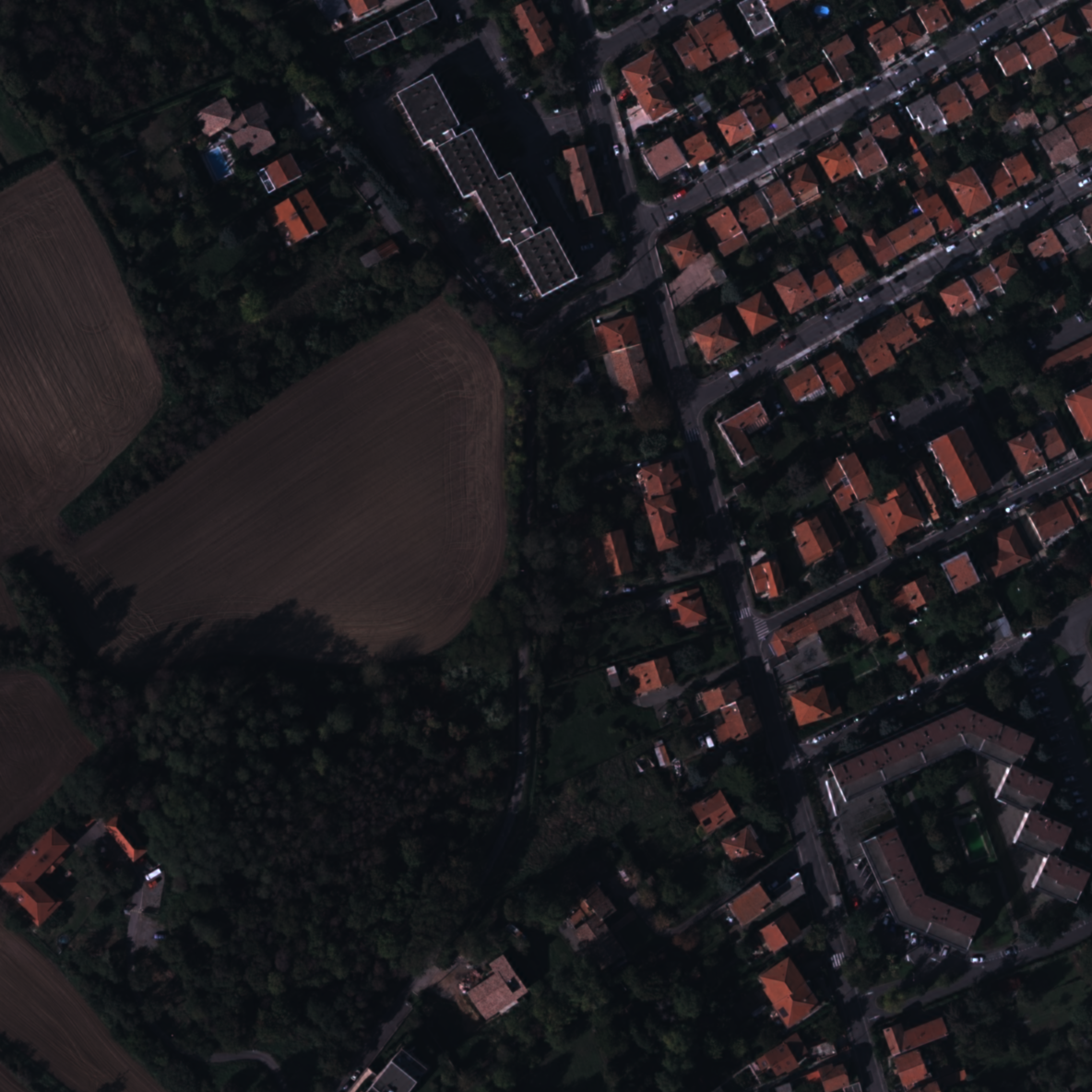} \\
 LMVM & ATWT & AWLP & GLP \\
\end{tabular}
\begin{tabular}{c@{\hskip 0.02in}c@{\hskip 0.02in}c}
  \includegraphics[trim= 23cm 35cm 15.5cm 3.5cm, clip=true, width=0.243\textwidth]{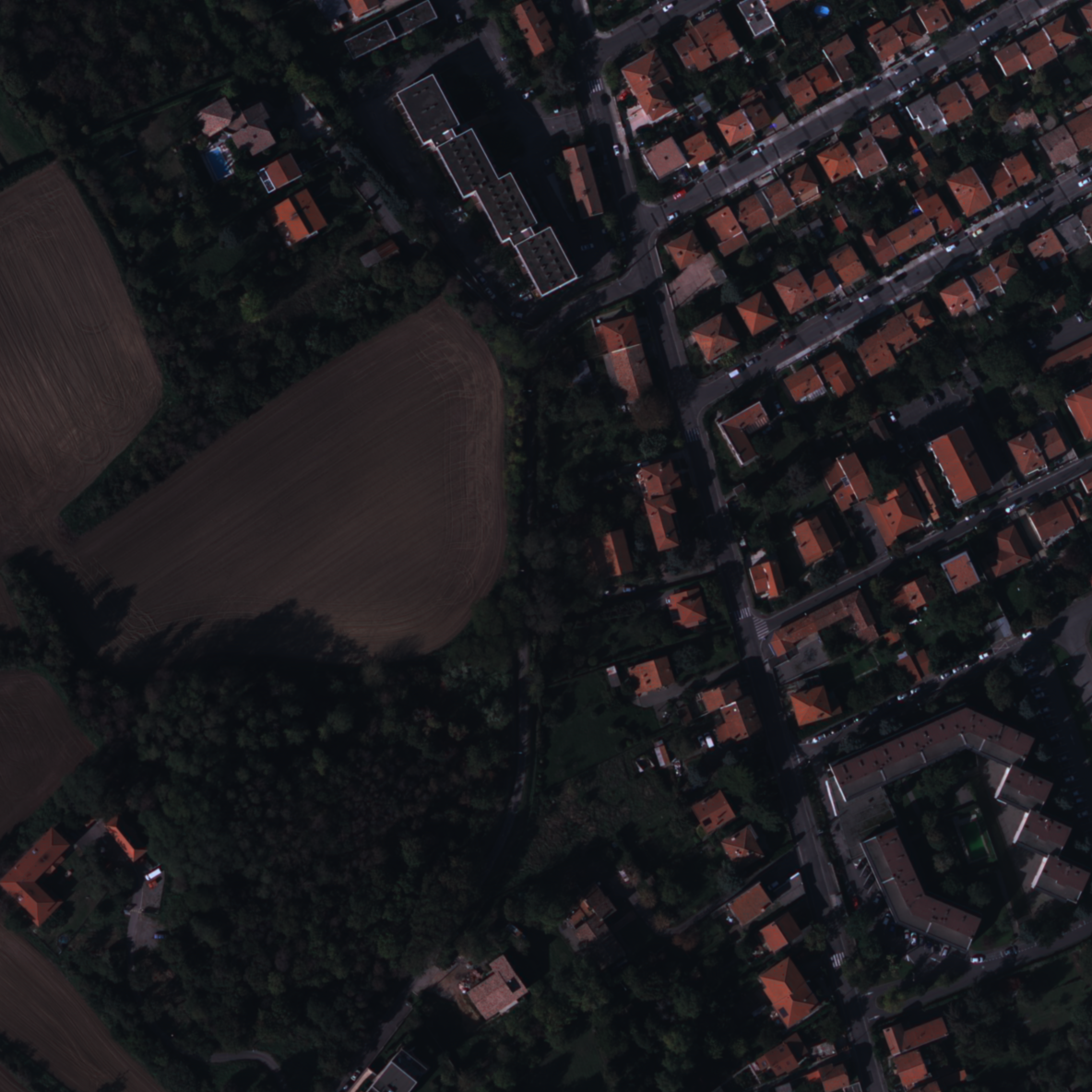} &
  \includegraphics[trim= 23cm 35cm 15.5cm 3.5cm, clip=true, width=0.243\textwidth]{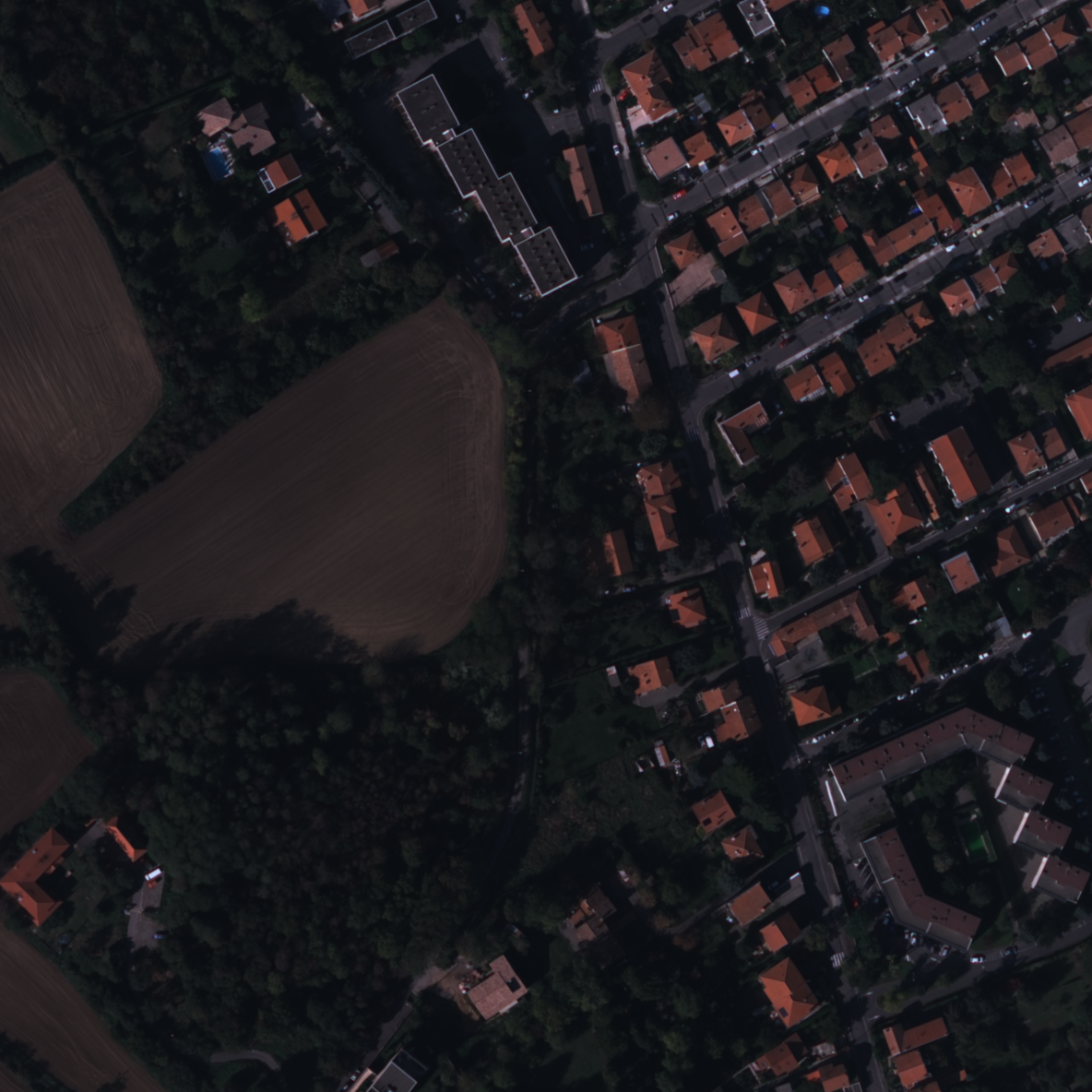} &
  \includegraphics[trim= 23cm 35cm 15.5cm 3.5cm, clip=true, width=0.243\textwidth]{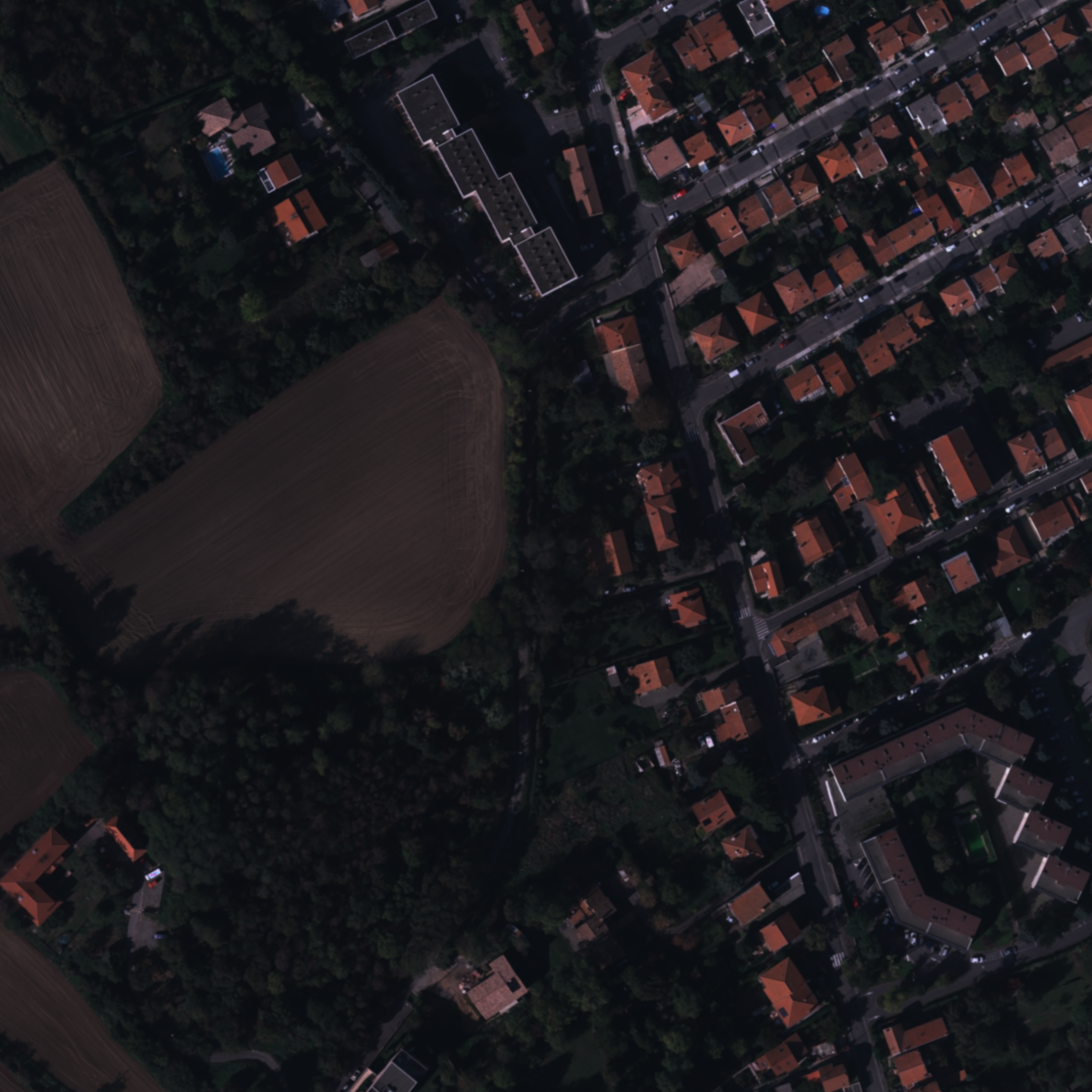} \\
  P+XS & NLV & NLVD\\
\end{tabular}
\caption{Close-ups of the reference full-color image at a resolution of 30 cm per pixel and of the fusion products provided by all methods under comparison. The Gaussian standard deviation used for the simulation of the low-resolution spectral components was $\sigma=1.3$. For these experiments, the data were non registered and the panchro-spectral constraint fulfilled with $\alpha_B=\alpha_G=\alpha_R=\frac{1}{3}$. All pansharpening techniques except ours cause annoying color distortions because of aliasing. Indeed, see how the artifacts on the hood of the white car that stand out in all other results are suppressed by the proposed model. With regard to the comparison between NLV and NLVD, one concludes that NLVD leads to considerably better visual quality. As an example, the red car at the top of the pictures is more blurred and grayish in the pansharpened image provided by NLV. Such phenomenon can also be detected in all results by CS-based algorithms.}
\label{fig_30cm_s13_RGB_noregist_linear}
\end{figure}

\begin{figure}[!p]
\footnotesize
\centering
\renewcommand{\arraystretch}{0.5}
\begin{tabular}{c@{\hskip 0.02in}c@{\hskip 0.02in}c@{\hskip 0.02in}c}
  \includegraphics[trim= 23cm 35cm 15.5cm 3.5cm, clip=true, width=0.243\textwidth]{RGBnoreglin_true.png} &
  \includegraphics[trim= 23cm 35cm 15.5cm 3.5cm, clip=true, width=0.243\textwidth]{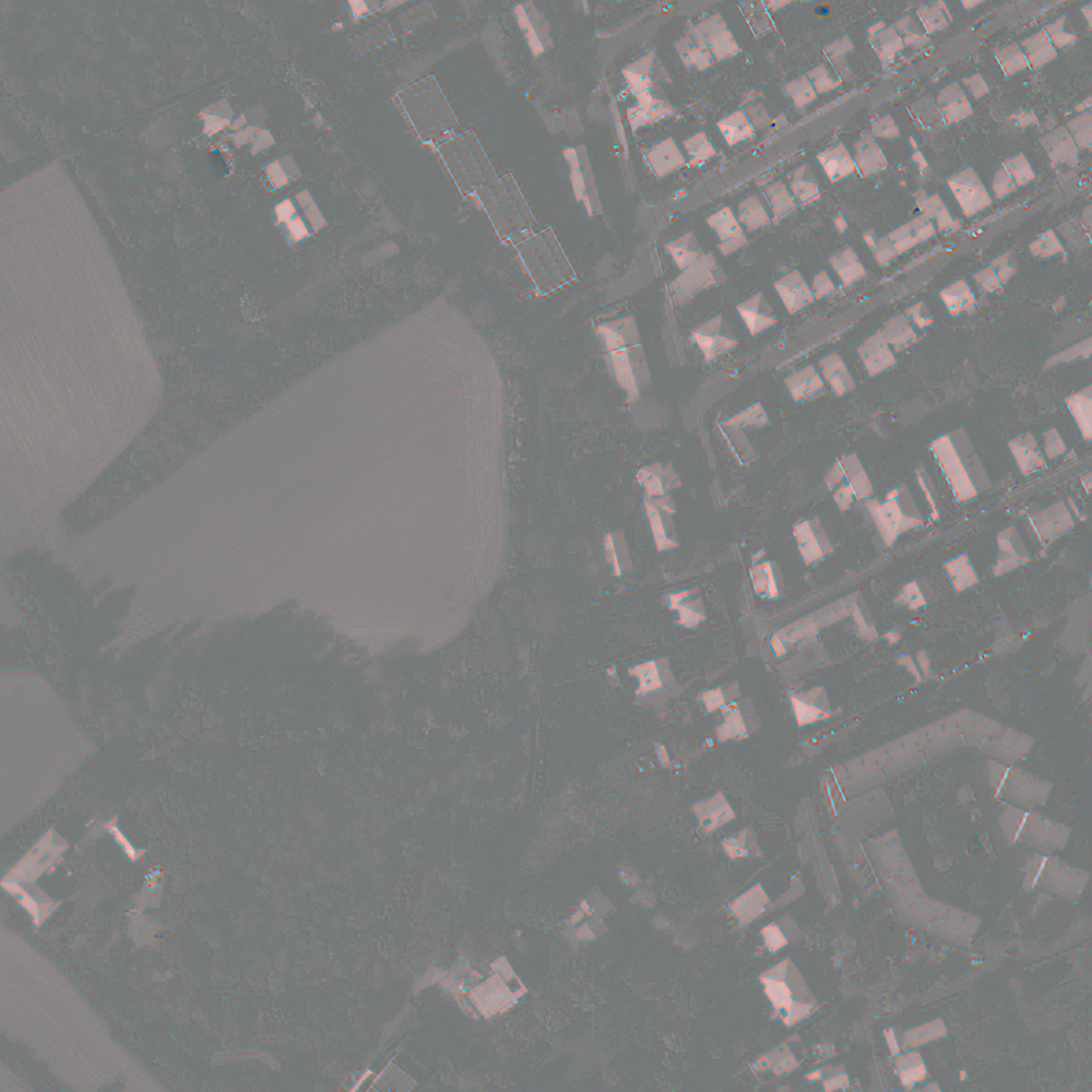} &
  \includegraphics[trim= 23cm 35cm 15.5cm 3.5cm, clip=true, width=0.243\textwidth]{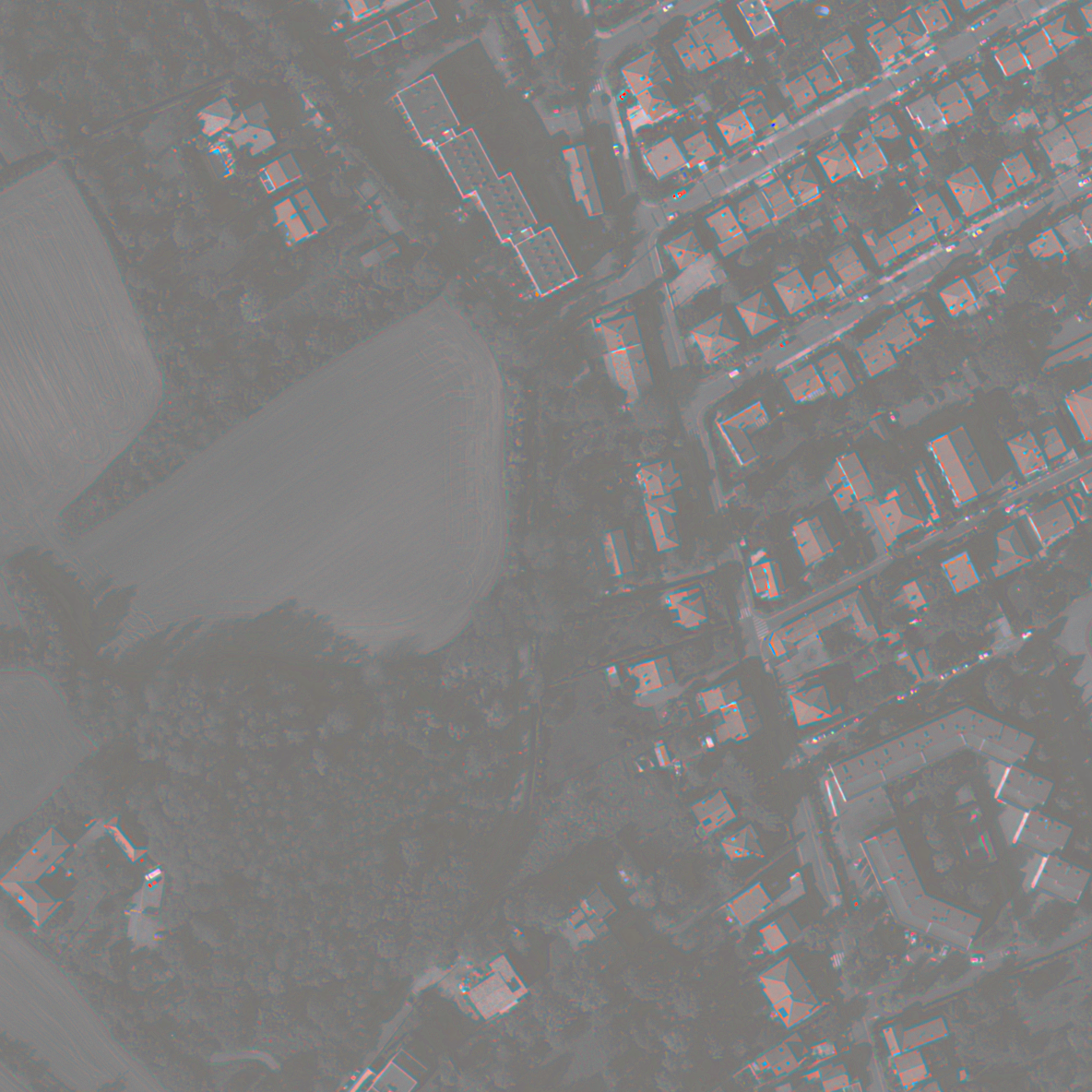} &
  \includegraphics[trim= 23cm 35cm 15.5cm 3.5cm, clip=true, width=0.243\textwidth]{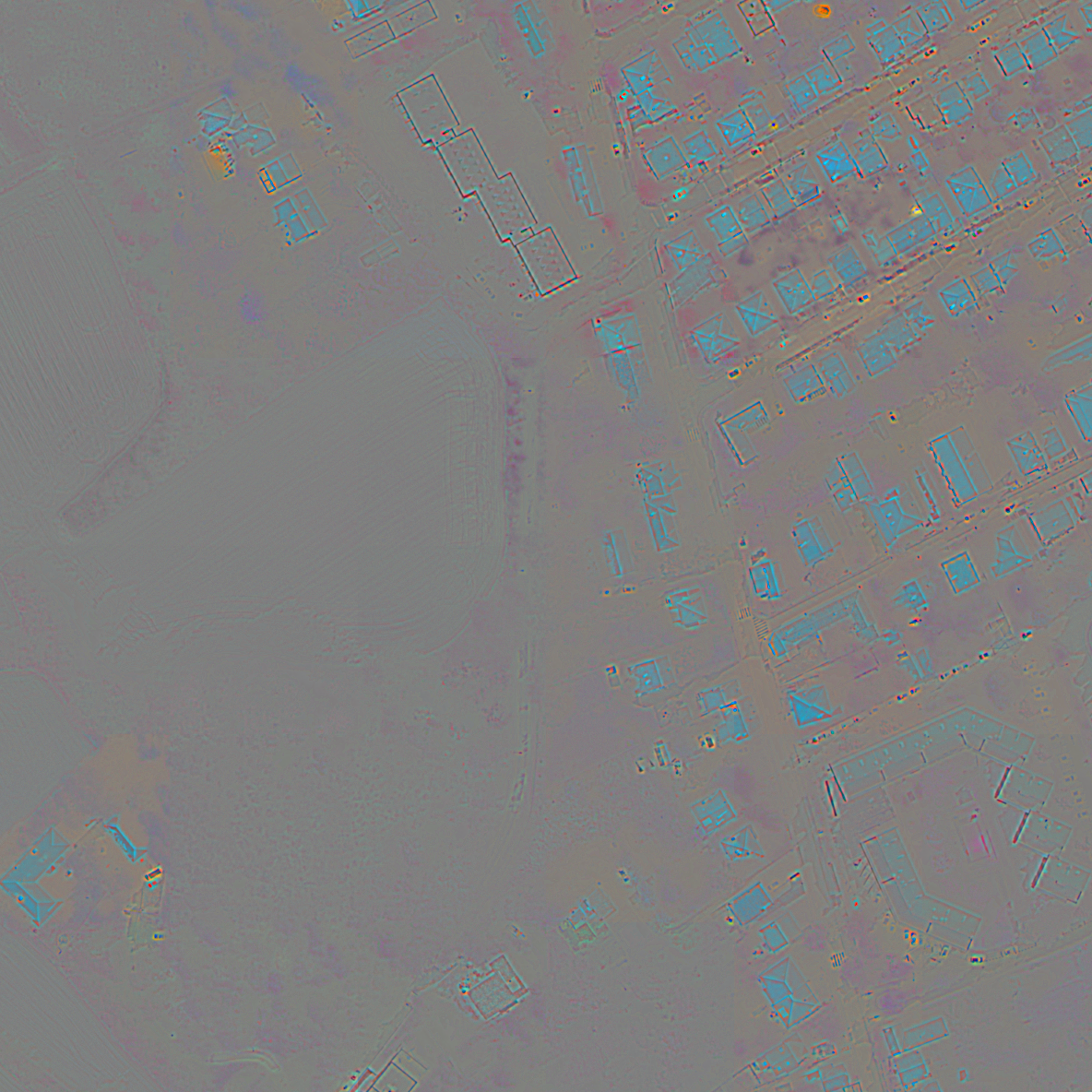} \\
  Reference & PCA & Brovey & BDSD\\ 
  \includegraphics[trim= 23cm 35cm 15.5cm 3.5cm, clip=true, width=0.243\textwidth]{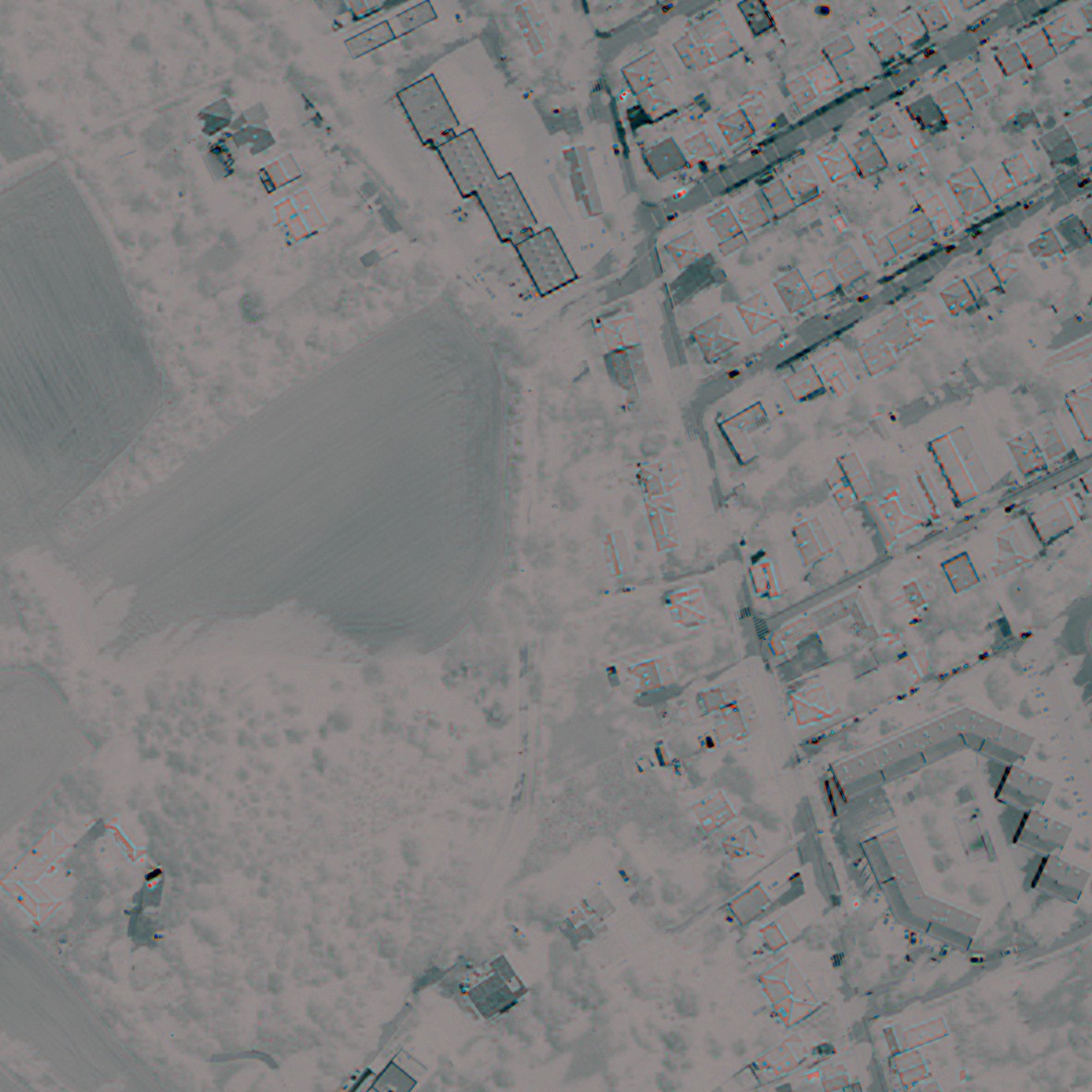} &
  \includegraphics[trim= 23cm 35cm 15.5cm 3.5cm, clip=true, width=0.243\textwidth]{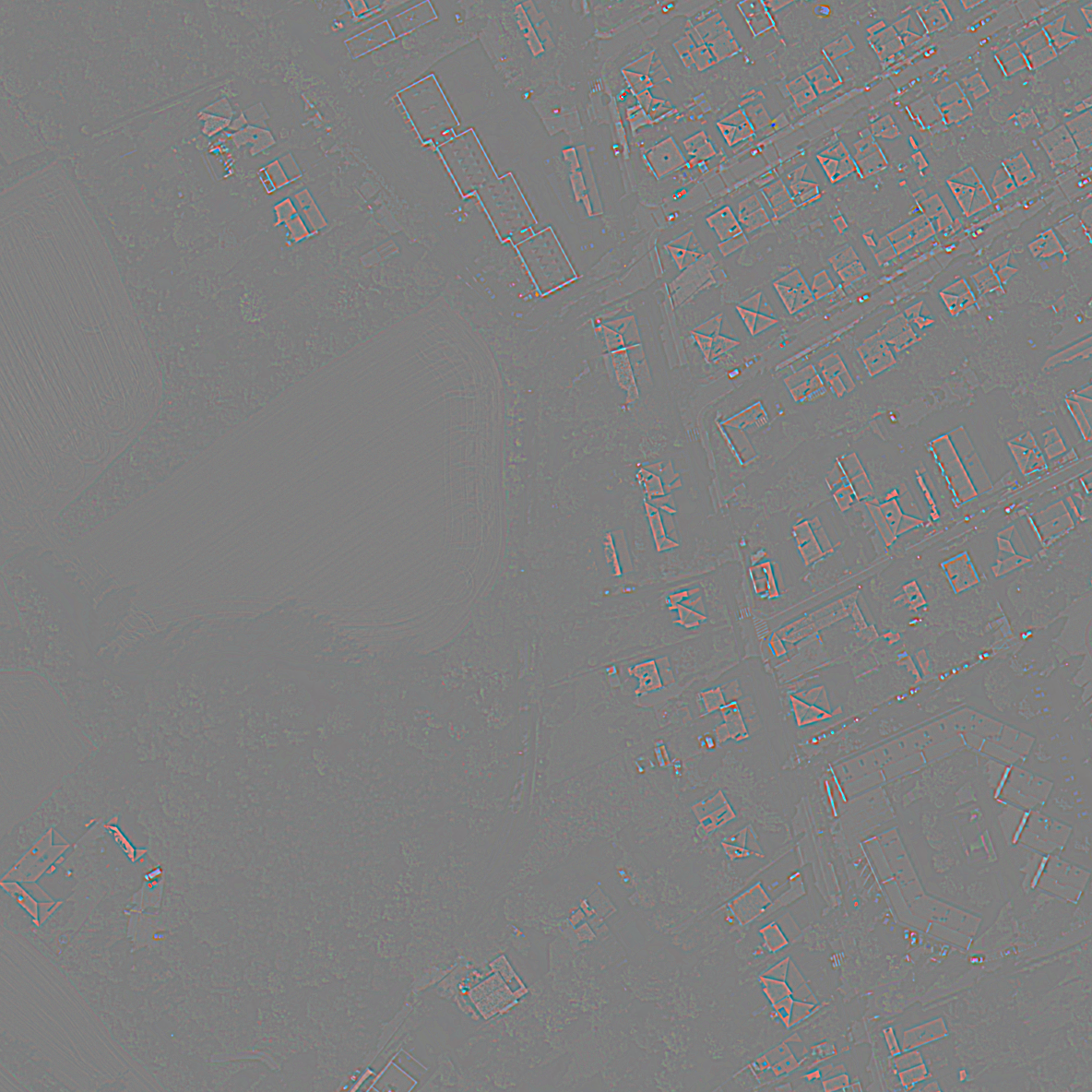} &
  \includegraphics[trim= 23cm 35cm 15.5cm 3.5cm, clip=true, width=0.243\textwidth]{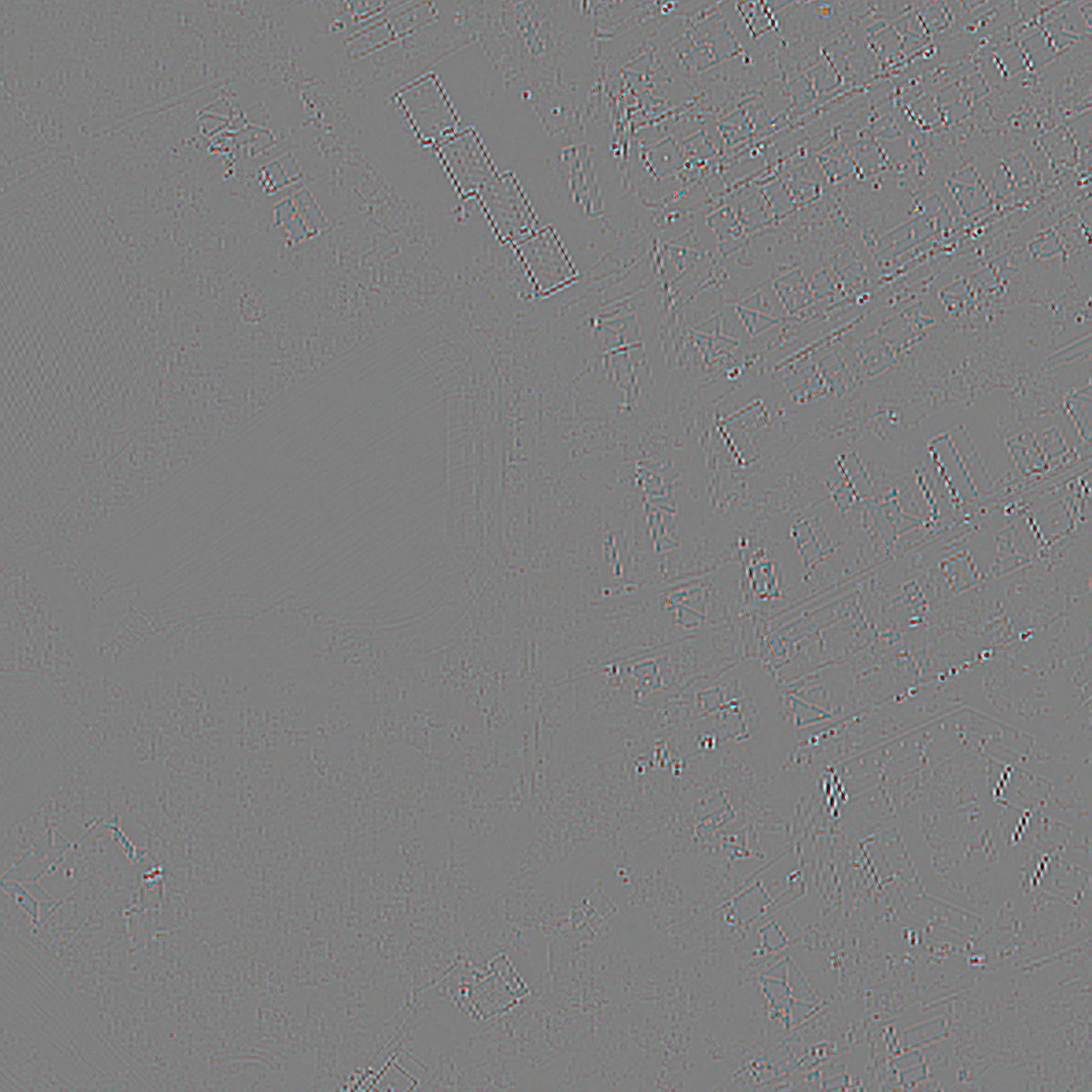} &
  \includegraphics[trim= 23cm 35cm 15.5cm 3.5cm, clip=true, width=0.243\textwidth]{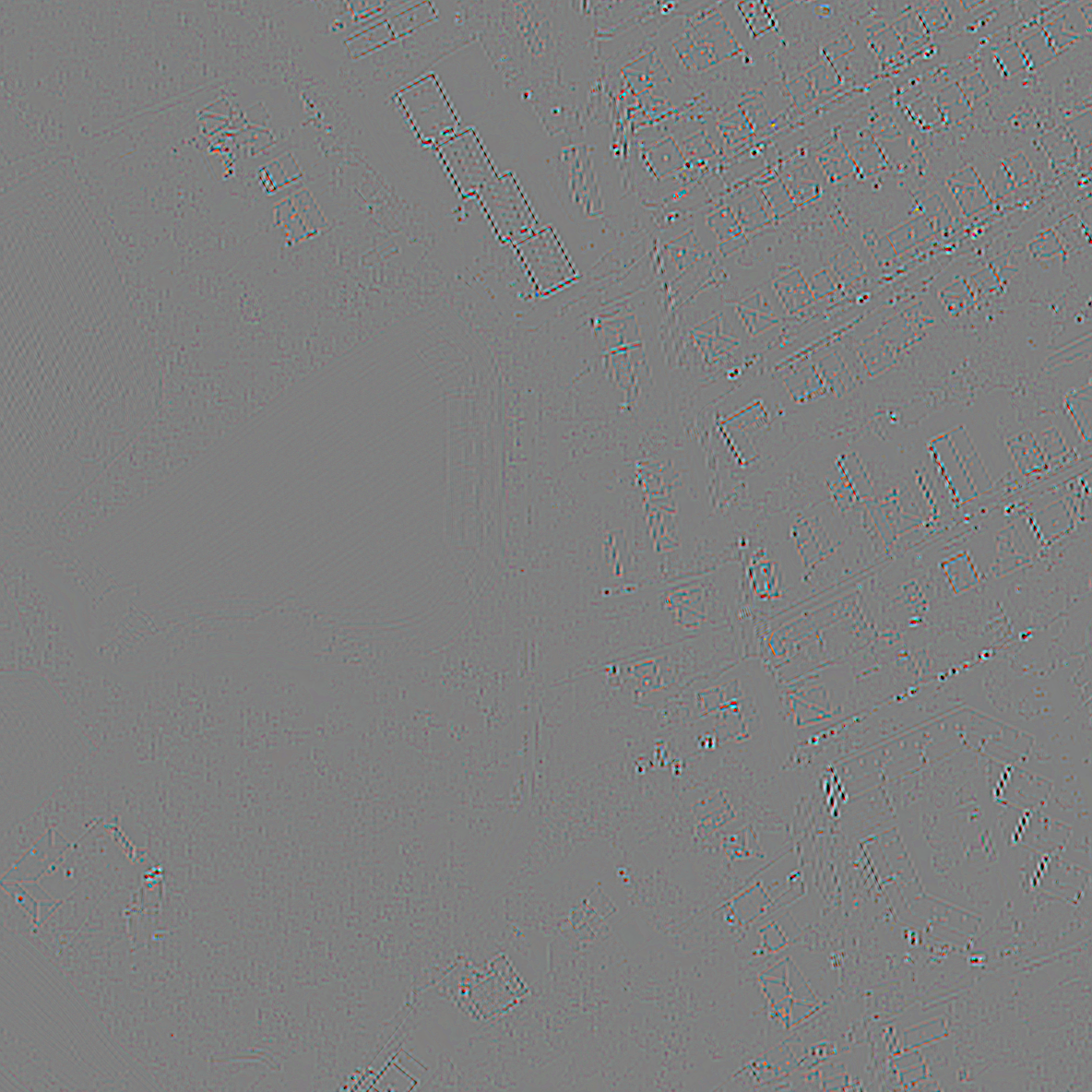} \\
  GSA & PRACS & HPF & SFIM\\
  \includegraphics[trim= 23cm 35cm 15.5cm 3.5cm, clip=true, width=0.243\textwidth]{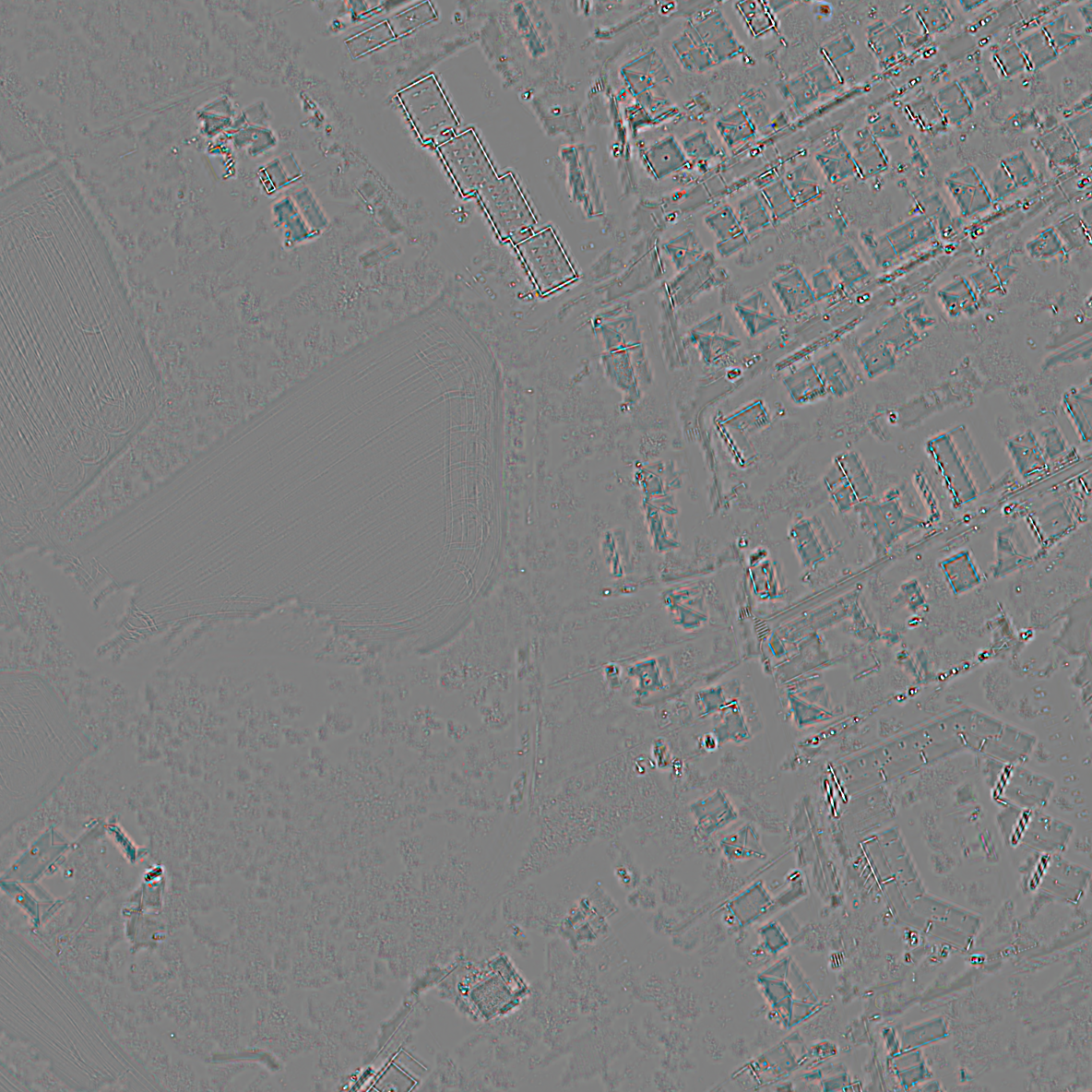} &
  \includegraphics[trim= 23cm 35cm 15.5cm 3.5cm, clip=true, width=0.243\textwidth]{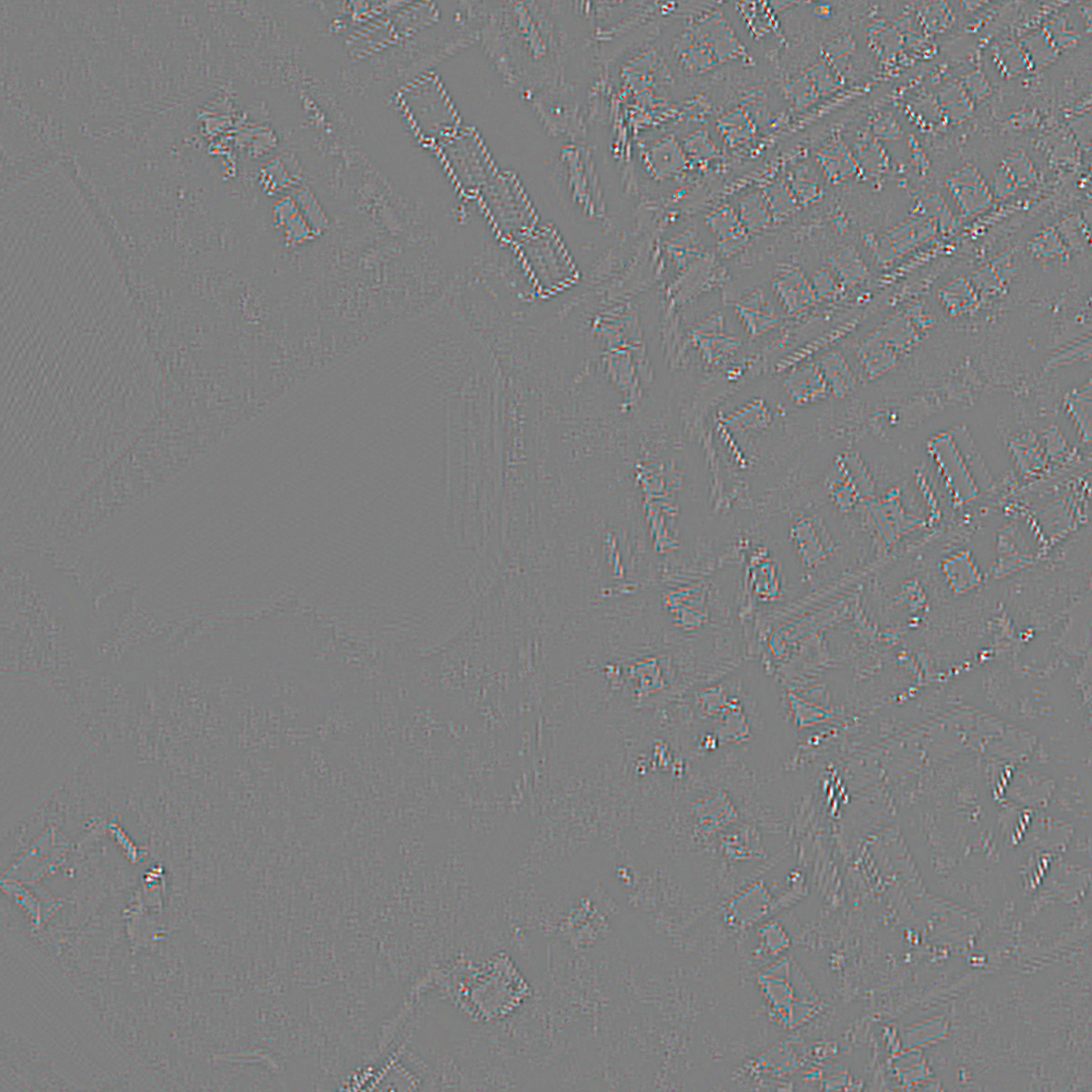} &
  \includegraphics[trim= 23cm 35cm 15.5cm 3.5cm, clip=true, width=0.243\textwidth]{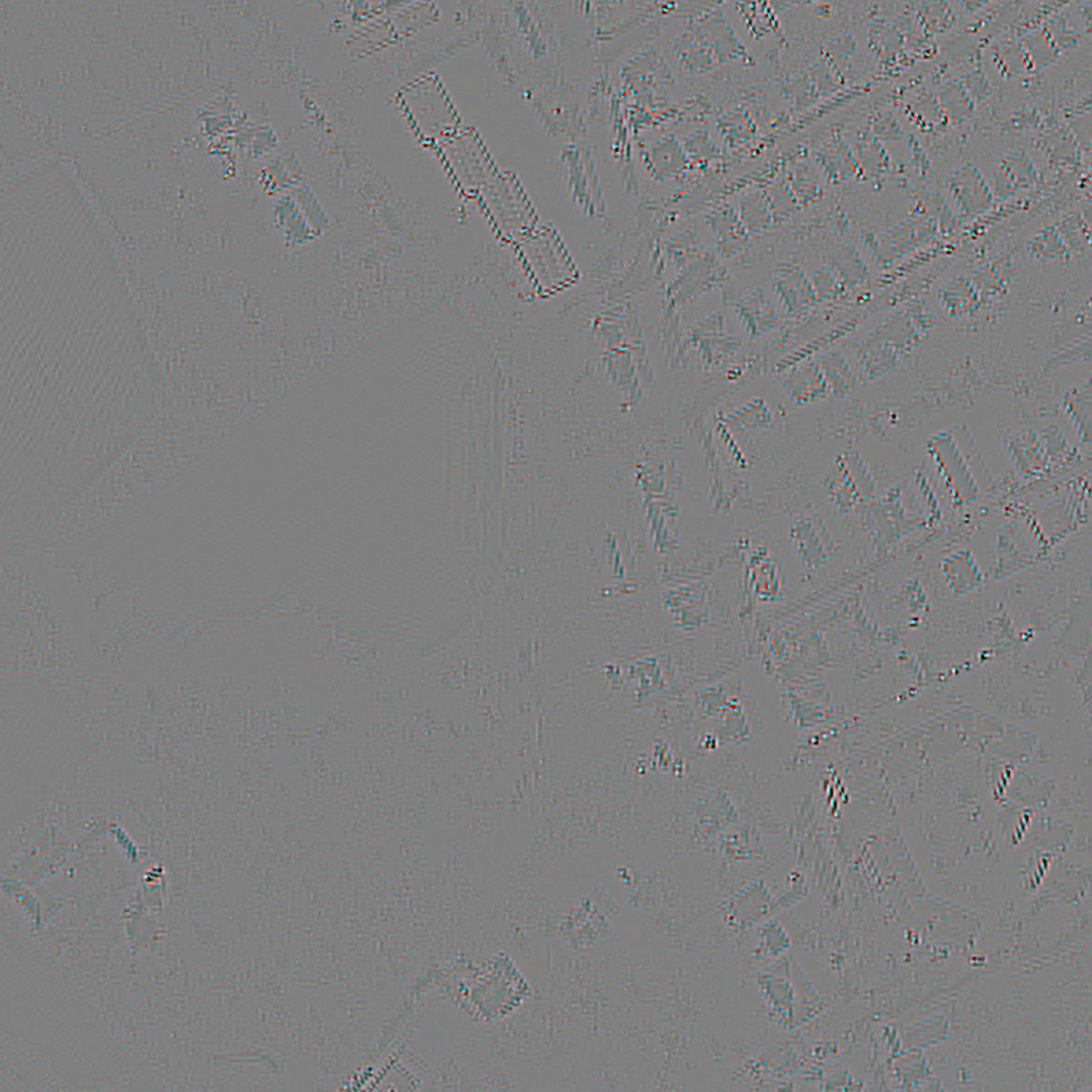} &
  \includegraphics[trim= 23cm 35cm 15.5cm 3.5cm, clip=true, width=0.243\textwidth]{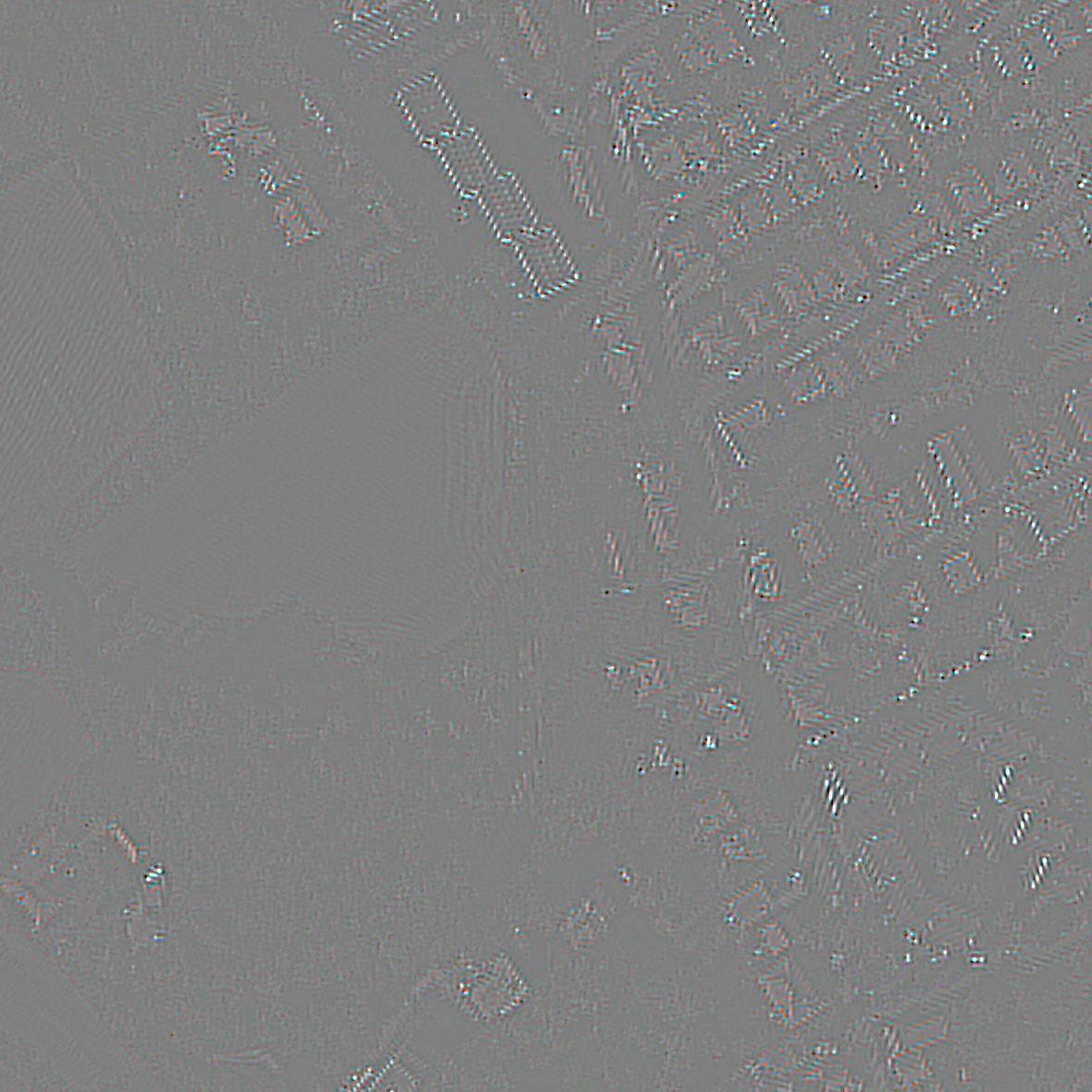} \\
  LMVM & ATWT & AWLP & GLP \\
\end{tabular}
\begin{tabular}{c@{\hskip 0.02in}c@{\hskip 0.02in}c}
  \includegraphics[trim= 23cm 35cm 15.5cm 3.5cm, clip=true, width=0.243\textwidth]{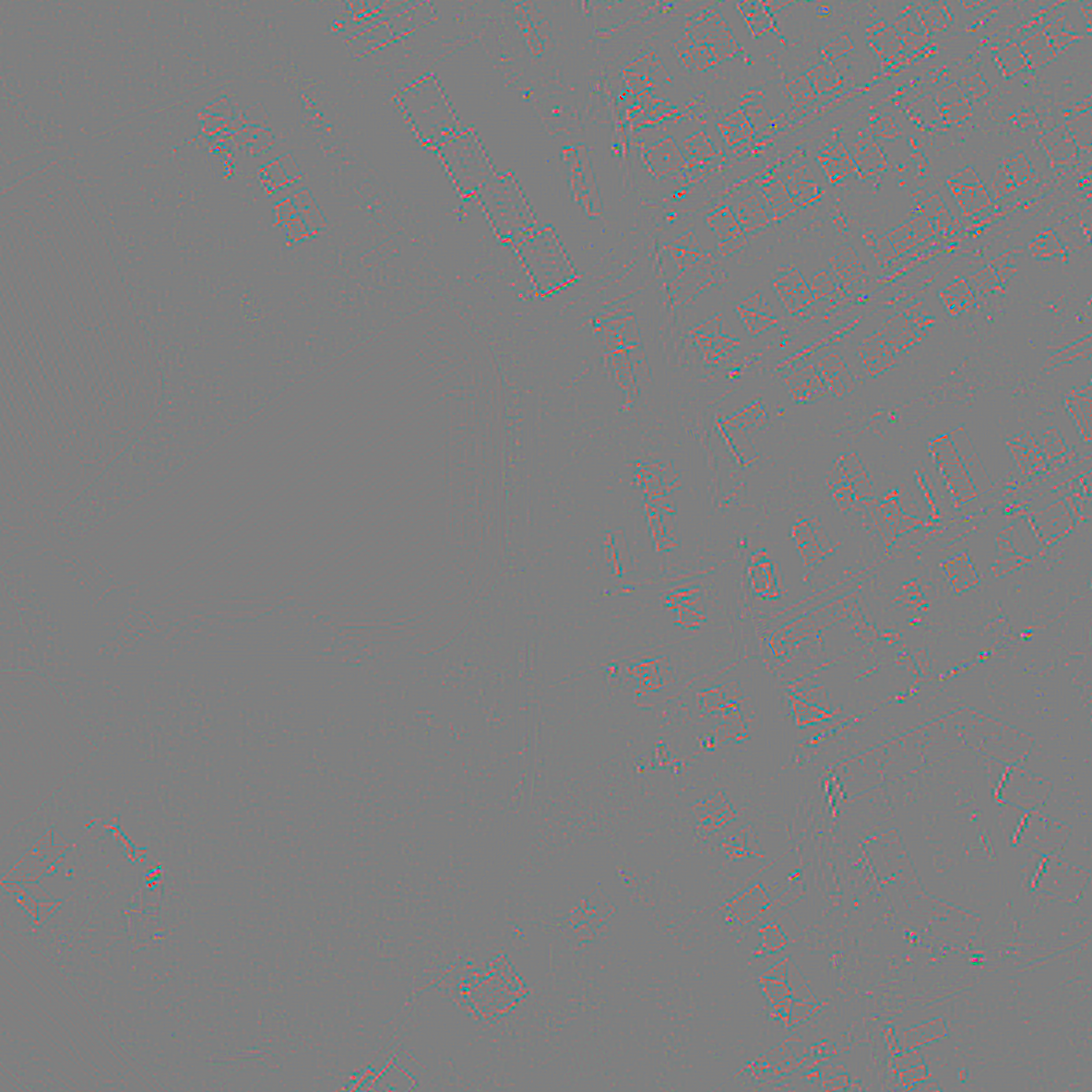} &
  \includegraphics[trim= 23cm 35cm 15.5cm 3.5cm, clip=true, width=0.243\textwidth]{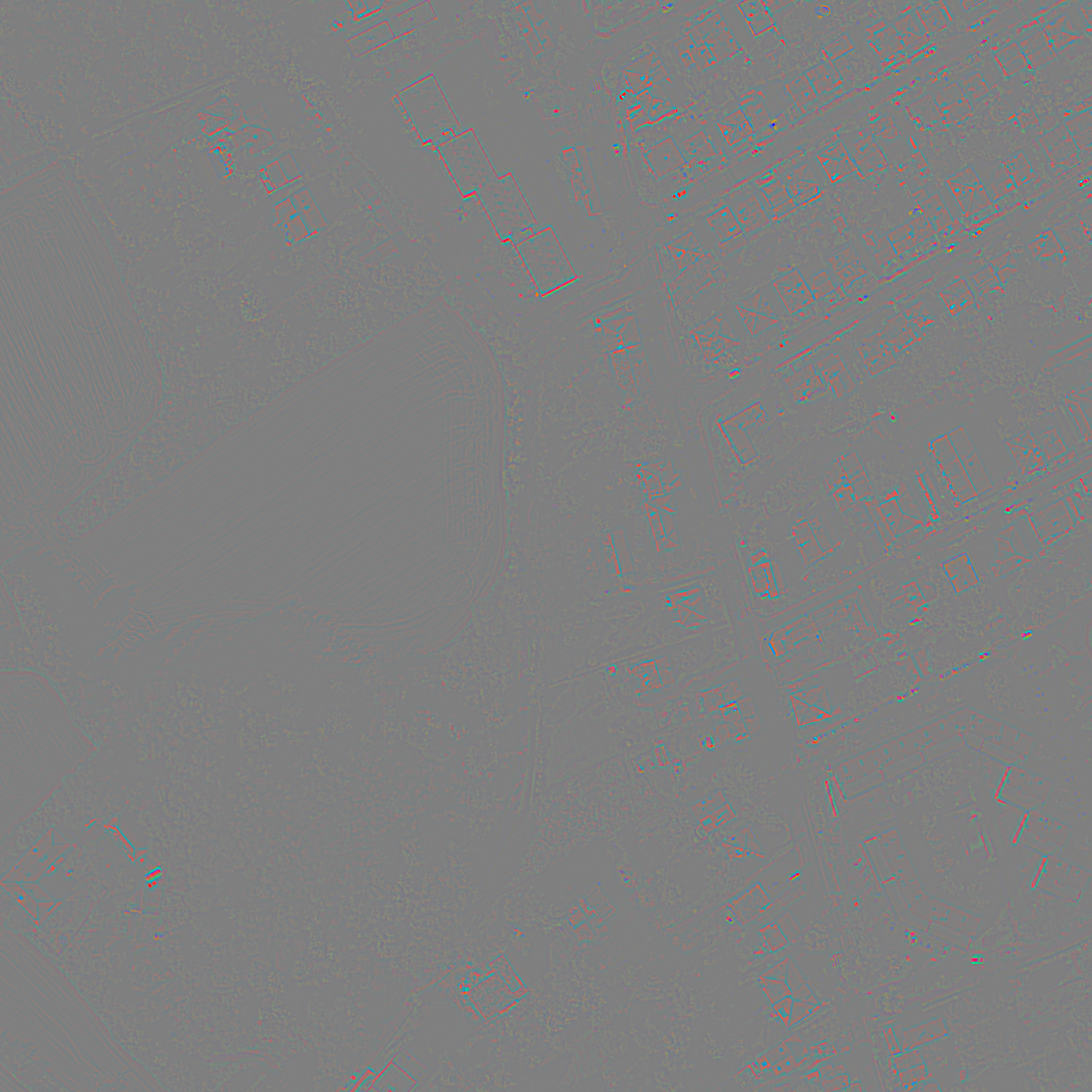} &
  \includegraphics[trim= 23cm 35cm 15.5cm 3.5cm, clip=true, width=0.243\textwidth]{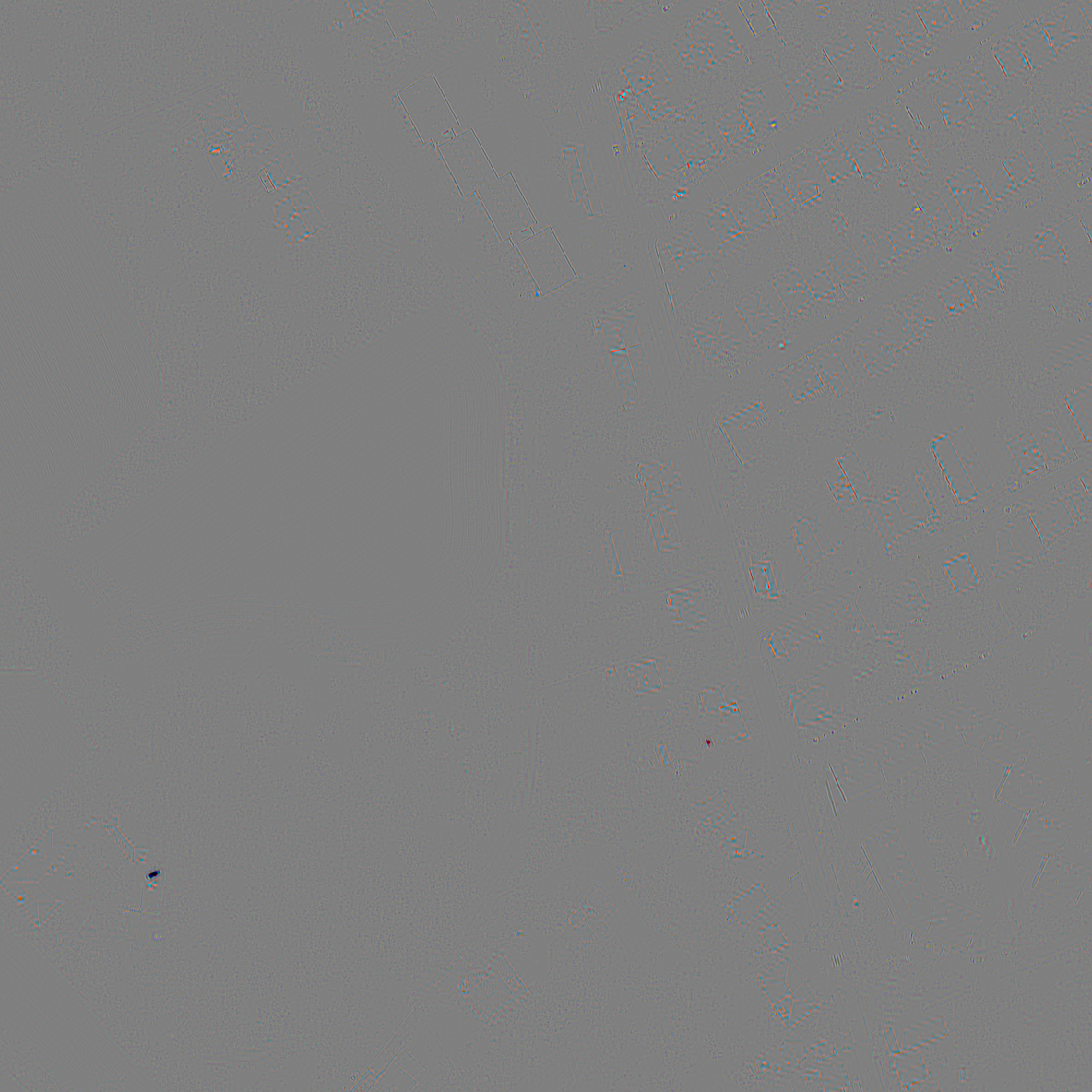} \\
  P+XS & NLV & NLVD\\
\end{tabular}
\caption{Close-ups of the reference RGB image at a resolution of 30 cm per pixel and of the difference images associated to the fusion products displayed in Figure \ref{fig_30cm_s13_RGB_noregist_linear}. For visualization purposes, the intensity values have been linearly rescaled from $[-20,20]$ to $[0,255]$. First, note the superiority of NLVD with respect to the other techniques since the corresponding difference image contains less amount of information. On the one hand, CS-based techniques suffer from spectral distortion since a lot of chromaticity is kept in the difference images. On the other hand, observe the ripples at the contours of the buildings introduced by MRA-based methods, thus leading to poor spatial quality. In the end, only the proposed NLVD model is able to almost suppress aliasing artifacts. Importantly, NLV is more affected by aliasing than NLVD, especially at edges, and it also induces a loss of spectral intensity in smooth areas such as the roofs of the buildings, which is avoided by NLVD.}
\label{fig_30cm_s13_RGB_noregist_linear_dif}
\end{figure}

In Figures \ref{fig_30cm_s13_RGB_noregist_linear} and \ref{fig_30cm_s13_RGB_noregist_linear_dif}, one can find close-ups of the results as well as of the corresponding difference images obtained from the last picture of the dataset at a resolution of 30 cm per pixel. The Gaussian standard deviation used for the simulation of the low-resolution spectral bands was $\sigma=1.3$. In general terms, the visual inspection matches with the numerical results and NLVD achieves the best performance appearance of the final product. For instance, observe the color artifacts on the hood of the white car in all images from Figure \ref{fig_30cm_s13_RGB_noregist_linear} except ours. Furthermore, Figure \ref{fig_30cm_s13_RGB_noregist_linear_dif} confirms that strong aliasing severely compromises the performances of all other techniques under comparison since the difference images contain much more amount of structural geometry and chromaticity. In particular, let us remark the visual improvement by NLVD with respect to the former model NLV.

\medskip

\noindent {\bf Four-band images.} The performance comparison of the pansharpening methods is carried out now on $4$-band images with blue, green, red, and near-infrared components. The mixing coefficients in the panchro-spectral constraint \eqref{eq:panconstraint} were fixed to $\alpha_B=0.1$, $\alpha_G=0.4$, $\alpha_R=0.25$, and $\alpha_I=0.25$. These values are more realistic than using the same weight per channel as it can be checked in Figure \ref{fig:satellite_responses} by comparing the sensitivity to different wavelengths of light of the panchromatic sensor with that of each spectral band.

We report in Tables \ref{table_30cm_RGBNIR_nonregist_linear} and \ref{table_60cm_RGBNIR_nonregist_linear} the quantitative results obtained on the $4$-band dataset. As in Tables \ref{table_30cm_RGB_nonregist_linear} and \ref{table_60cm_RGB_nonregist_linear}, all indices demonstrate the superiority of NLVD. However, the improvement in accuracy over all methods under comparison is even more important in these tests.  Another special feature of the numerical results displayed here is the fact that MRA-based techniques such as AWLP and GLP outperform the variational models P+XS and NLV in terms of RMSE and ERGAS. This superiority also extends to SFIM and LMVM if the SAM index is evaluated. In general, pansharpening techniques using the panchro-spectral constraint, which are those from CS-based and variational -- except NLVD -- fusion families, are less competitive than before. Finally, we want to emphasize that NLVD achieves pretty better SSIM as well as Q$2^n$ values compared with all other methods, which confirms that the proposed model better avoids the loss of correlation and the luminance and contrast distortions in the final fusion products.

\begin{table}[!p]
\footnotesize
\centering
\begin{subtable}[h]{0.99\textwidth}
\centering
\begin{tabular}{|c|c|c|c|c|c|c|c|c|c|}
\hline
 & Methods & RMSE & ERGAS & SAM & SSIM & Q$2^n$  \\ \hline \hline
 & Reference & 0 & 0 & 0 & 1 & 1 \\ \hline \hline
 \multirow{14}{*}{\rotatebox[origin=c]{90}{Registered}} & PCA & 3.8756 & 2.6238 & 3.0294 & 0.9906 & 0.9510 \\ \cline{2-7}
& Brovey & 3.3976 & 2.3213 & 2.4678 & 0.9916 & 0.9531 \\ \cline{2-7}
& BDSD & 3.4849 & 2.2999 & 3.8224 & 0.9929 & 0.9682 \\ \cline{2-7}
& GSA & 3.7002 & 2.5175 & 3.5101 & 0.9886 & 0.9500 \\ \cline{2-7}
& PRACS & 3.8925 & 2.6006 & 2.5167 & 0.9943 & 0.9691 \\ \cline{2-7}
& HPF & 3.7910 & 2.5727 & 2.7987 & 0.9939 & 0.9700 \\ \cline{2-7}
& SFIM & 3.3277 & 2.2710 & 2.2662 & 0.9946 & 0.9758 \\ \cline{2-7}
& LMVM & 3.7228 & 2.5779 & 2.2481 & 0.9941 & 0.9656 \\ \cline{2-7}
& ATWT & 3.3266 & 2.2254 & 2.9246 & 0.9945 & 0.9732 \\ \cline{2-7}
& AWLP & 2.7237 & 1.8491 & 2.1697 & 0.9967 & 0.9820 \\ \cline{2-7}
& GLP & 2.7938 & 1.8935 & 2.2287 & 0.9961 & 0.9790 \\ \cline{2-7}
& P+XS & 3.1962 & 2.2277 & 2.7540 & 0.9973 & 0.9756 \\ \cline{2-7}
& NLV & 3.1139 & 2.1514 & 2.7081 & 0.9971 & 0.9778 \\ \cline{2-7}
& NLVD & 2.4059 & 1.6283 & 2.0902 & 0.9978 & 0.9838 \\ \hline\hline
 \multirow{6}{*}{\rotatebox[origin=c]{90}{Misregistered}} & HPF & 3.6107 & 2.4400 & 2.7235 & 0.9947 & 0.9720 \\ \cline{2-7}
& SFIM & 3.1316 & 2.1256 & 2.2503 & 0.9955 & 0.9773 \\ \cline{2-7}
& LMVM & 3.5302 & 2.4450 & 2.0809 & 0.9958 & 0.9678 \\ \cline{2-7}
& ATWT & 3.2945 & 2.2047 & 2.8964 & 0.9947 & 0.9729 \\ \cline{2-7}
& GLP & 2.8838 & 1.9628 & 2.4530 & 0.9960 & 0.9775 \\ \cline{2-7}
& NLVD & {\bf 1.9242} & {\bf 1.2817} & {\bf 1.5990} & {\bf 0.9992} & {\bf 0.9877} \\ \hline
\end{tabular}
\caption{Numerical results for $\sigma=1.7$.}
 \label{table_30cm_s17_RGBNIR_nonregist_linear}
 \end{subtable}
 
\medskip

\begin{subtable}[h]{0.99\textwidth}
\centering
\begin{tabular}{|c|c|c|c|c|c|c|c|c|}
\hline
 & & RMSE & ERGAS & SAM & SSIM & Q$2^n$  \\ \hline \hline
 & Reference & 0 & 0 & 0 & 1 & 1 \\ \hline \hline
\multirow{14}{*}{\rotatebox[origin=c]{90}{Registered}} & PCA & 3.5695 & 2.4165 & 2.9152 & 0.9926 & 0.9583 \\ \cline{2-7}
& Brovey & 3.1142 & 2.1312 & 2.4127 & 0.9934 & 0.9607 \\ \cline{2-7}
& BDSD & 3.6321 & 2.4053 & 3.9108 & 0.9920 & 0.9654 \\ \cline{2-7}
& GSA & 4.0778 & 2.8092 & 3.7328 & 0.9848 & 0.9380 \\ \cline{2-7}
& PRACS & 3.6478 & 2.4421 & 2.4511 & 0.9957 & 0.9743 \\ \cline{2-7}
& HPF & 3.5266 & 2.3889 & 2.7637 & 0.9952 & 0.9723 \\ \cline{2-7}
& SFIM & 3.0394 & 2.0765 & 2.2149 & 0.9960 & 0.9787 \\ \cline{2-7}
& LMVM & 3.7214 & 2.5611 & 2.3588 & 0.9941 & 0.9680 \\ \cline{2-7}
& ATWT & 3.4145 & 2.2915 & 2.9350 & 0.9935 & 0.9701 \\ \cline{2-7}
& AWLP & 2.9597 & 1.9904 & 2.1617 & 0.9956 & 0.9795 \\ \cline{2-7}
& GLP & 3.2229 & 2.2127 & 2.1966 & 0.9943 & 0.9739 \\ \cline{2-7}
& P+XS & 3.8497 & 2.6899 & 3.3857 & 0.9960 & 0.9502 \\ \cline{2-7}
& NLV & 3.1887 & 2.2045 & 2.7274 & 0.9969 & 0.9769 \\ \cline{2-7}
& NLVD & 2.5630 & 1.7492 & 2.1581 & 0.9973 & 0.9826 \\ \hline\hline
 \multirow{6}{*}{\rotatebox[origin=c]{90}{Misregistered}} & HPF & 3.3743 & 2.2777 & 2.7293 & 0.9957 & 0.9737 \\ \cline{2-7}
& SFIM & 2.8816 & 1.9606 & 2.2504 & 0.9966 & 0.9796 \\ \cline{2-7}
& LMVM & 3.7112 & 2.5532 & 2.4623 & 0.9940 & 0.9687 \\ \cline{2-7}
& ATWT & 3.4737 & 2.3407 & 2.9671 & 0.9932 & 0.9688 \\ \cline{2-7}
& GLP & 3.4799 & 2.4055 & 2.5229 & 0.9935 & 0.9710 \\ \cline{2-7}
& NLVD & {\bf 1.9106} & {\bf 1.2719} & {\bf 1.5698} & {\bf 0.9991} & {\bf 0.9877} \\ \hline
\end{tabular}
\caption{Numerical results for $\sigma=1.3$.}
 \label{table_30cm_s13_RGBNIR_nonregist_linear}
 \end{subtable}
 \caption{Quantitative evaluation of the fused products on simulated data from 4-band (blue, green, red, and near-infrared) aerial images at resolution of 30 cm per pixel. For these experiments, the low-resolution spectral components were non registered but the panchro-spectral constraint fulfilled with $\alpha_B=0.1$, $\alpha_G=0.4$, $\alpha_R=0.25$, and $\alpha_I=0.25$. All indices exhibit the superiority of NLVD. Getting the best SSIM and Q$2^n$ values demonstrates that NLVD is more likely to avoid loss of correlation as well as luminance and contrast distortions. In general terms, CS-based methods, P+XS, and NLV are significantly less competitive against MRA on 4-band data.}
 \label{table_30cm_RGBNIR_nonregist_linear}
\end{table}

\begin{table}[!p]
\footnotesize
\centering
\begin{subtable}[h]{0.99\textwidth}
\centering
\begin{tabular}{|c|c|c|c|c|c|c|c|c|c|}
\hline
 & Methods & RMSE & ERGAS & SAM & SSIM & Q$2^n$  \\ \hline \hline
 & Reference & 0 & 0 & 0 & 1 & 1 \\ \hline \hline
 \multirow{14}{*}{\rotatebox[origin=c]{90}{Registered}}& PCA & 4.9159 & 3.3160 & 3.9751 & 0.9654 & 0.9385 \\ \cline{2-7}
& Brovey & 4.4534 & 3.0250 & 3.3984 & 0.9725 & 0.9466 \\ \cline{2-7}
& BDSD & 4.2339 & 2.8040 & 5.0496 & 0.9734 & 0.9673 \\ \cline{2-7}
& GSA & 4.6683 & 3.1937 & 4.8755 & 0.9583 & 0.9486 \\ \cline{2-7}
& PRACS & 4.6497 & 3.0636 & 3.4306 & 0.9686 & 0.9656 \\ \cline{2-7}
& HPF & 4.5418 & 3.0479 & 3.6885 & 0.9668 & 0.9658 \\ \cline{2-7}
& SFIM & 4.0828 & 2.7452 & 3.1101 & 0.9727 & 0.9716 \\ \cline{2-7}
& LMVM & 4.3227 & 2.9573 & 2.9164 & 0.9699 & 0.9644 \\ \cline{2-7}
& ATWT & 3.9925 & 2.6537 & 3.7917 & 0.9724 & 0.9717 \\ \cline{2-7}
& AWLP & 3.3246 & 2.2412 & 2.8979 & 0.9814 & 0.9799 \\ \cline{2-7}
& GLP & 3.4375 & 2.2964 & 3.0390 & 0.9791 & 0.9777 \\ \cline{2-7}
& P+XS & 3.7925 & 2.6223 & 3.7186 & 0.9779 & 0.9749 \\ \cline{2-7}
& NLV & 3.6570 & 2.5022 & 3.6127 & 0.9785 & 0.9767 \\ \cline{2-7}
& NLVD & 3.0509 & 2.0501 & 2.9537 & 0.9840 & 0.9825 \\ \hline\hline
 \multirow{6}{*}{\rotatebox[origin=c]{90}{Misregistered}} & HPF & 4.3392 & 2.8963 & 3.5974 & 0.9700 & 0.9682 \\ \cline{2-7}
& SFIM & 3.8663 & 2.5815 & 3.0818 & 0.9758 & 0.9737 \\ \cline{2-7}
& LMVM & 4.1876 & 2.8557 & 2.9360 & 0.9726 & 0.9660 \\ \cline{2-7}
& ATWT & 3.9376 & 2.6146 & 3.7678 & 0.9732 & 0.9716 \\ \cline{2-7}
& GLP & 3.5081 & 2.3498 & 3.3037 & 0.9789 & 0.9767 \\ \cline{2-7}
& NLVD & {\bf 2.5023} & {\bf 1.6527} & {\bf 2.3301} & {\bf 0.9898} & {\bf 0.9863} \\ \hline
\end{tabular}
\caption{Numerical results for $\sigma=1.7$.}
 \label{table_60cm_s17_RGBNIR_nonregist_linear}
 \end{subtable}
 
\medskip

\begin{subtable}[h]{0.99\textwidth}
\centering
\begin{tabular}{|c|c|c|c|c|c|c|c|c|}
\hline
 & & RMSE & ERGAS & SAM & SSIM & Q$2^n$  \\ \hline \hline
 & Reference & 0 & 0 & 0 & 1 & 1 \\ \hline \hline
\multirow{14}{*}{\rotatebox[origin=c]{90}{Registered}} & PCA & 4.5020 & 3.0357 & 3.8317 & 0.9705 & 0.9499 \\ \cline{2-7}
& Brovey & 4.0622 & 2.7628 & 3.3038 & 0.9763 & 0.9562 \\ \cline{2-7}
& BDSD & 4.4427 & 2.9565 & 5.2301 & 0.9713 & 0.9642 \\ \cline{2-7}
& GSA & 5.2046 & 3.6123 & 5.2658 & 0.9513 & 0.9346 \\ \cline{2-7}
& PRACS & 4.3534 & 2.8740 & 3.3244 & 0.9731 & 0.9710 \\ \cline{2-7}
& HPF & 4.2026 & 2.8199 & 3.6338 & 0.9715 & 0.9699 \\ \cline{2-7}
& SFIM & 3.7160 & 2.5032 & 3.0121 & 0.9776 & 0.9763 \\ \cline{2-7}
& LMVM & 4.3267 & 2.9458 & 3.2403 & 0.9709 & 0.9664 \\ \cline{2-7}
& ATWT & 3.9926 & 2.6677 & 3.8018 & 0.9717 & 0.9700 \\ \cline{2-7}
& AWLP & 3.4567 & 2.3163 & 2.8516 & 0.9805 & 0.9786 \\ \cline{2-7}
& GLP & 3.7497 & 2.5405 & 2.9646 & 0.9762 & 0.9738 \\ \cline{2-7}
& P+XS & 4.3881 & 3.0454 & 4.2395 & 0.9705 & 0.9569 \\ \cline{2-7}
& NLV & 3.8551 & 2.6366 & 3.7644 & 0.9764 & 0.9749 \\ \cline{2-7}
& NLVD & 3.2148 & 2.1829 & 3.0485 & 0.9827 & 0.9813 \\ \hline\hline
 \multirow{6}{*}{\rotatebox[origin=c]{90}{Misregistered}} & HPF & 4.0256 & 2.6886 & 3.6031 & 0.9740 & 0.9717 \\ \cline{2-7}
& SFIM & 3.5358 & 2.3679 & 3.0568 & 0.9800 & 0.9777 \\ \cline{2-7}
& LMVM & 4.3325 & 2.9489 & 3.3960 & 0.9721 & 0.9667 \\ \cline{2-7}
& ATWT & 4.0346 & 2.7032 & 3.8681 & 0.9714 & 0.9688 \\ \cline{2-7}
& GLP & 4.0062 & 2.7326 & 3.3692 & 0.9744 & 0.9712 \\ \cline{2-7}
& NLVD & {\bf 2.4909} & {\bf 1.6444} & {\bf 2.2824} & {\bf 0.9898} & {\bf 0.9863} \\ \hline
\end{tabular}
\caption{Numerical results for $\sigma=1.3$.}
 \label{table_60cm_s13_RGBNIR_nonregist_linear}
 \end{subtable}
 \caption{Quantitative evaluation of the fused products on simulated data from 4-band (blue, green, red, and near-infrared) aerial images at resolution of 60 cm per pixel. For these experiments, the low-resolution spectral components were non registered but the panchro-spectral constraint fulfilled with $\alpha_B=0.1$, $\alpha_G=0.4$, $\alpha_R=0.25$, and $\alpha_I=0.25$. Although the results are obviously worse than on data at a resolution of 30 cm, NLVD is the best method in terms of all quality indexes.}
 \label{table_60cm_RGBNIR_nonregist_linear}
\end{table}

\begin{figure}[!p]
\footnotesize
\centering
\renewcommand{\arraystretch}{0.5}
\begin{tabular}{c@{\hskip 0.02in}c@{\hskip 0.02in}c@{\hskip 0.02in}c}
  \includegraphics[trim= 28.5cm 27.6cm 10.5cm 11.4cm, clip=true, width=0.243\textwidth]{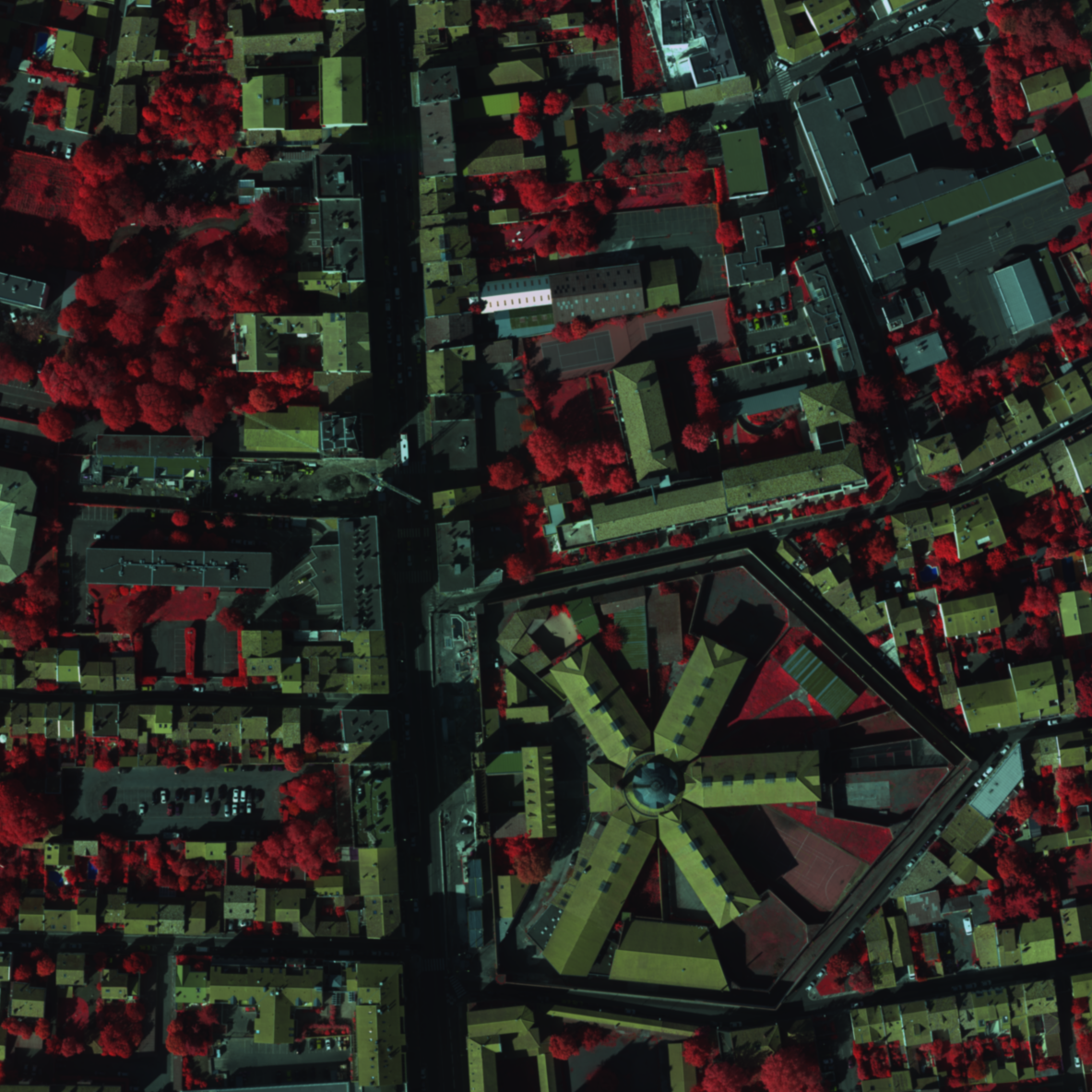} &
  \includegraphics[trim= 28.5cm 27.6cm 10.5cm 11.4cm, clip=true, width=0.243\textwidth]{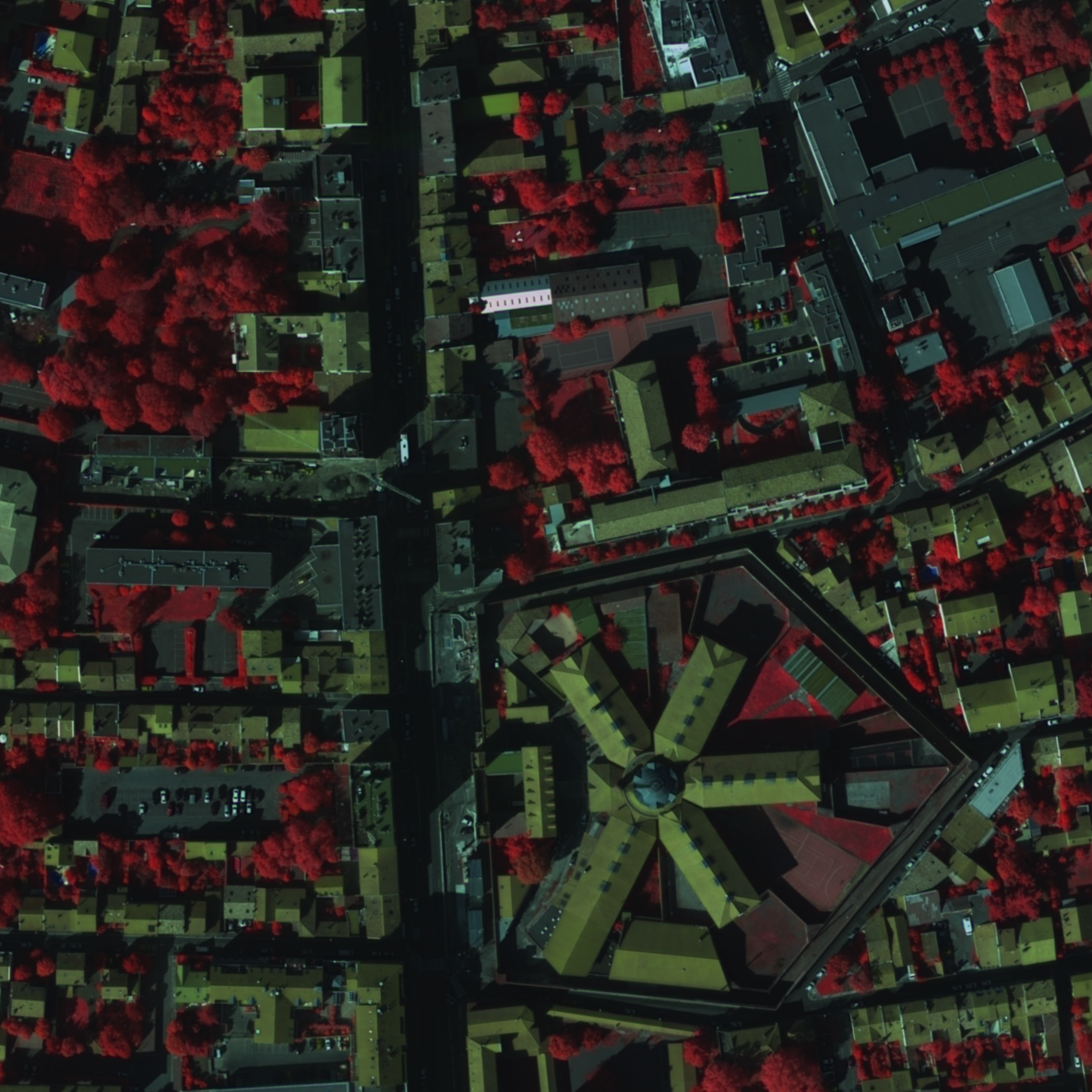} &
  \includegraphics[trim= 28.5cm 27.6cm 10.5cm 11.4cm, clip=true, width=0.243\textwidth]{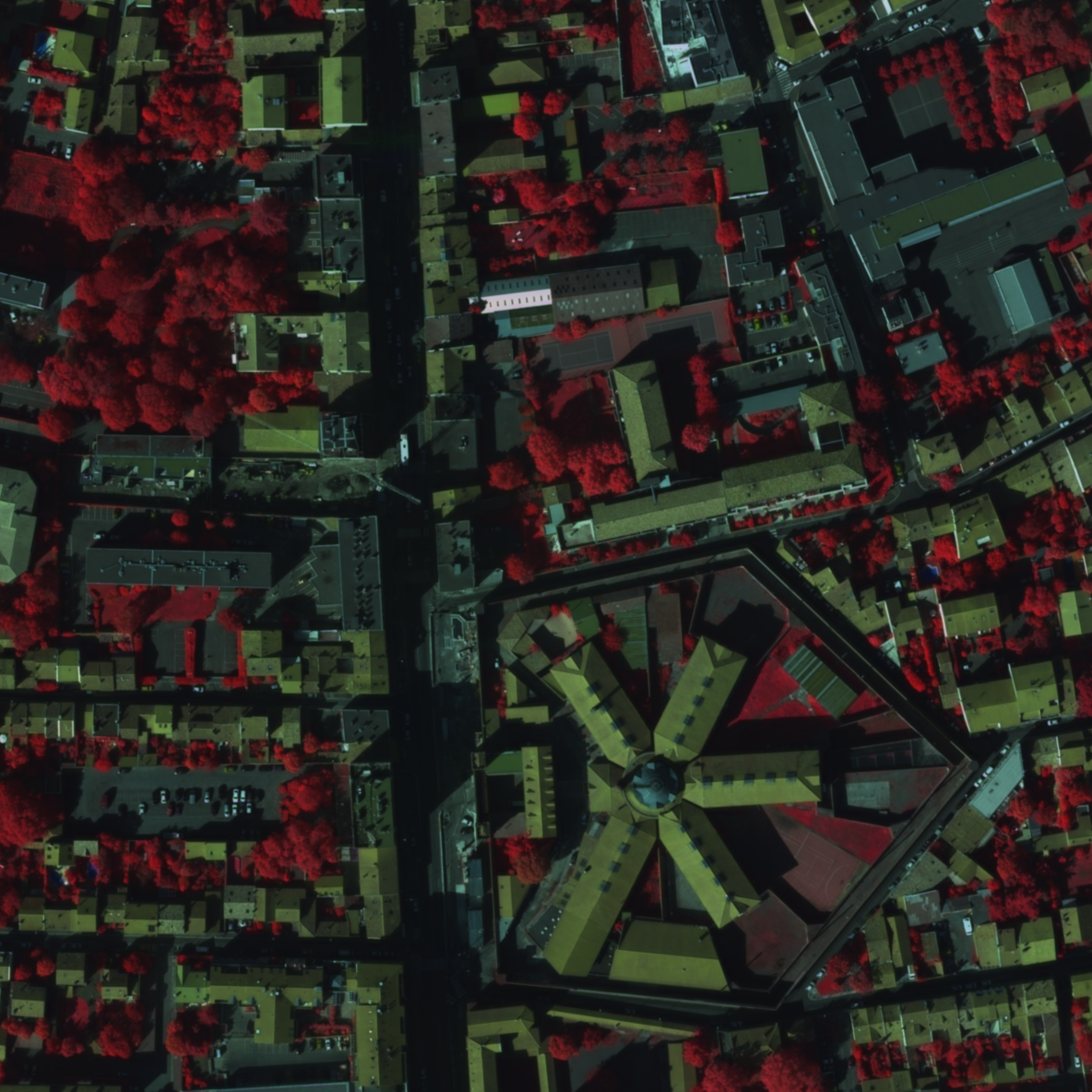} &
  \includegraphics[trim= 28.5cm 27.6cm 10.5cm 11.4cm, clip=true, width=0.243\textwidth]{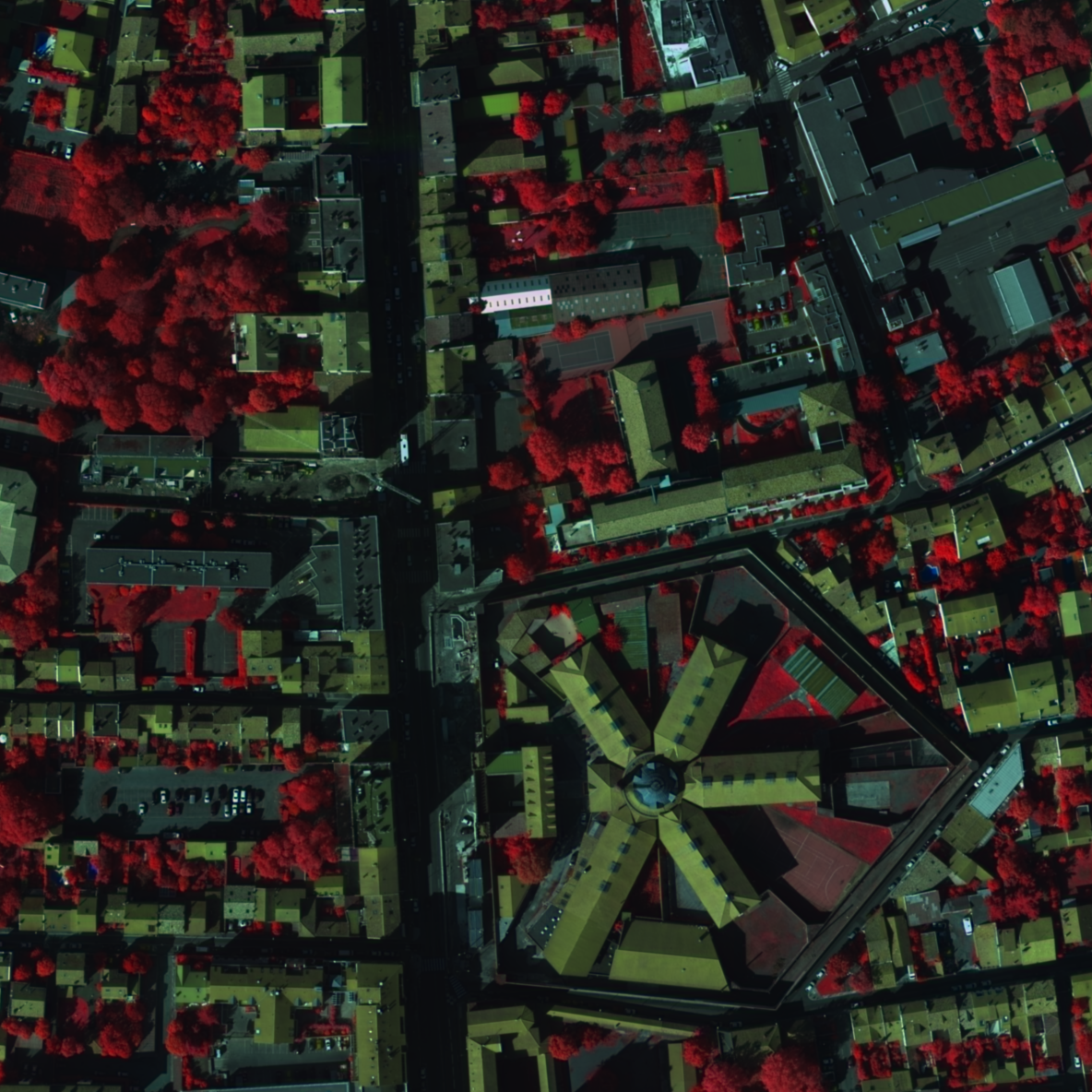} \\
  Reference & PCA & Brovey & BDSD\\ 
  \includegraphics[trim= 28.5cm 27.6cm 10.5cm 11.4cm, clip=true, width=0.243\textwidth]{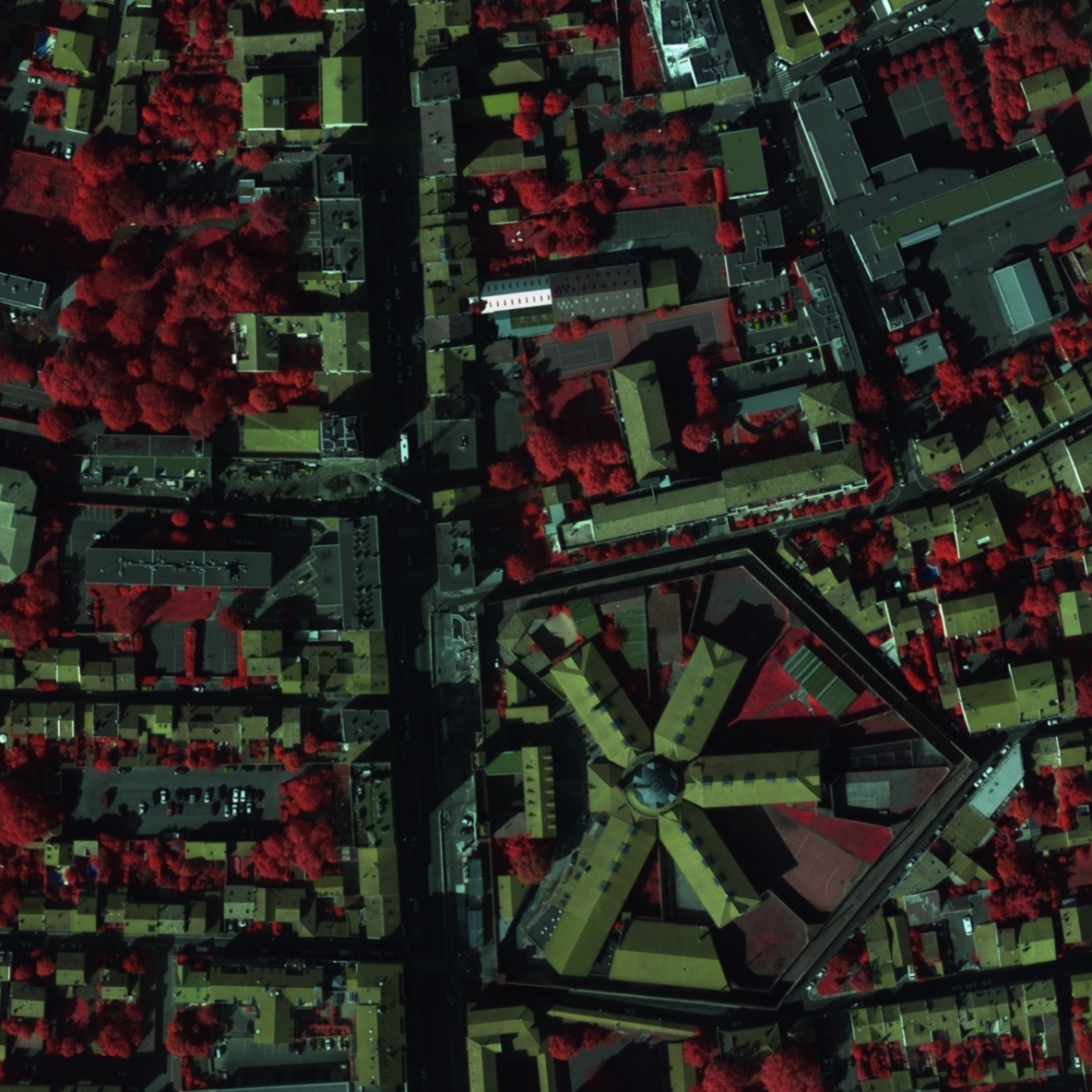} &
  \includegraphics[trim= 28.5cm 27.6cm 10.5cm 11.4cm, clip=true, width=0.243\textwidth]{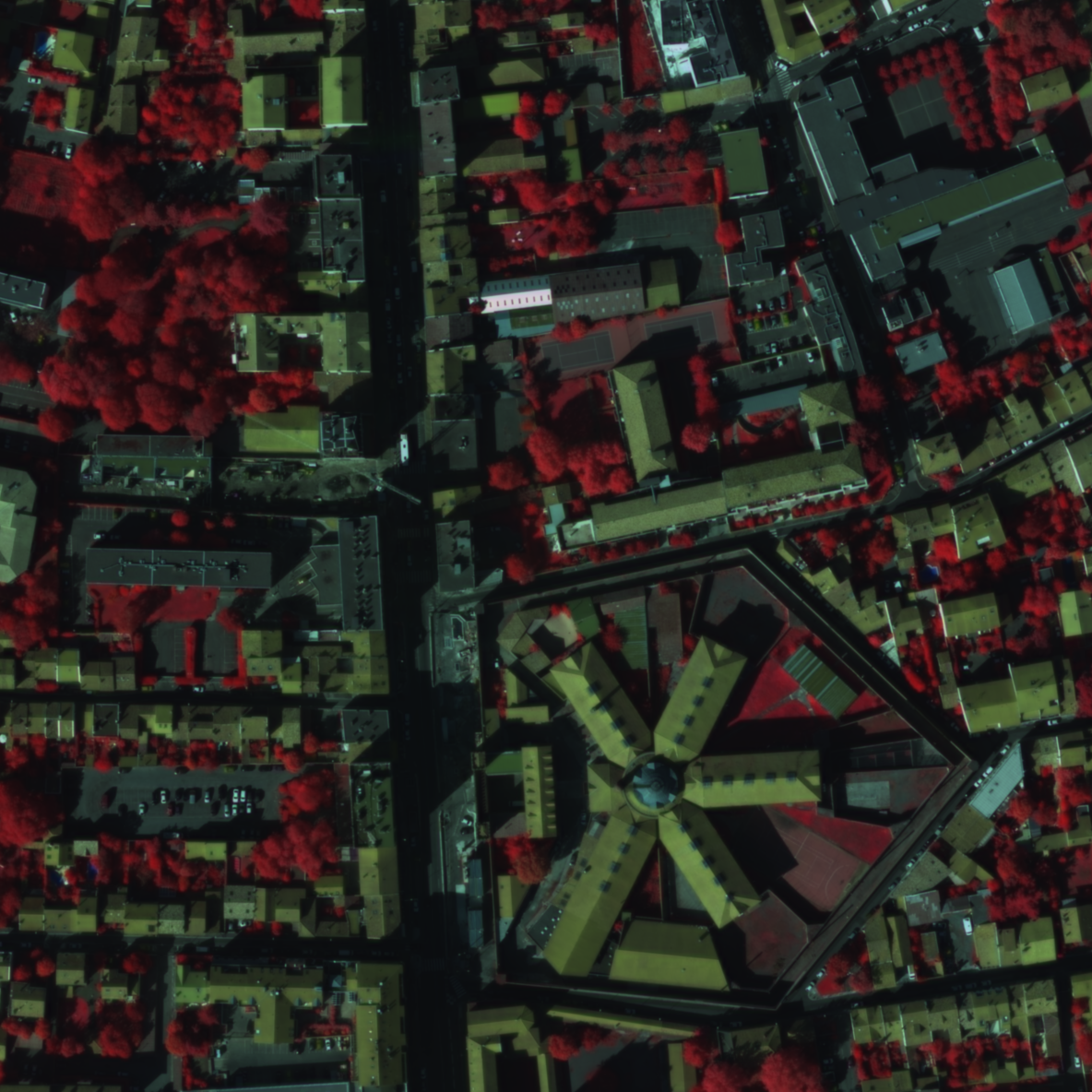} &
  \includegraphics[trim= 28.5cm 27.6cm 10.5cm 11.4cm, clip=true, width=0.243\textwidth]{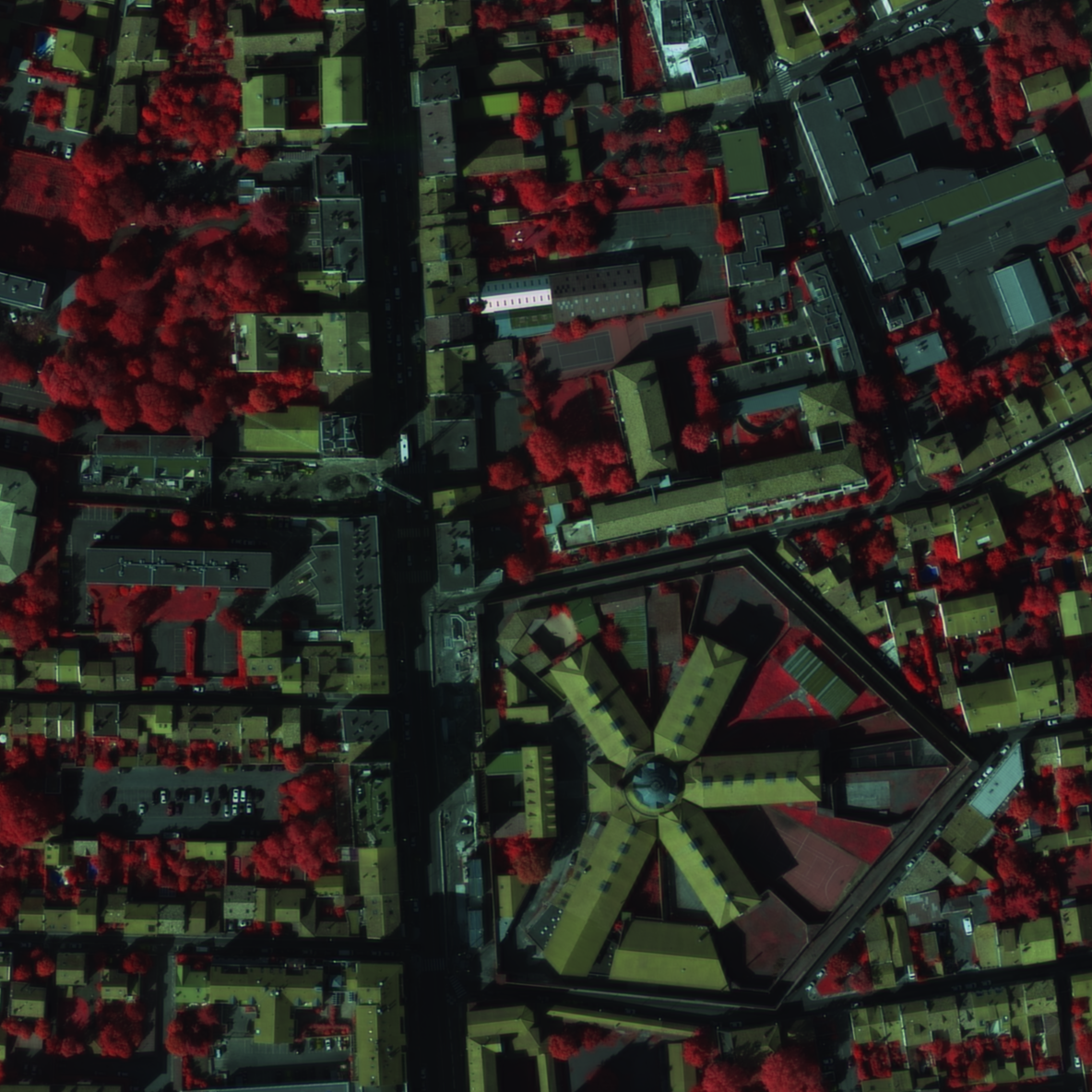} &
  \includegraphics[trim= 28.5cm 27.6cm 10.5cm 11.4cm, clip=true, width=0.243\textwidth]{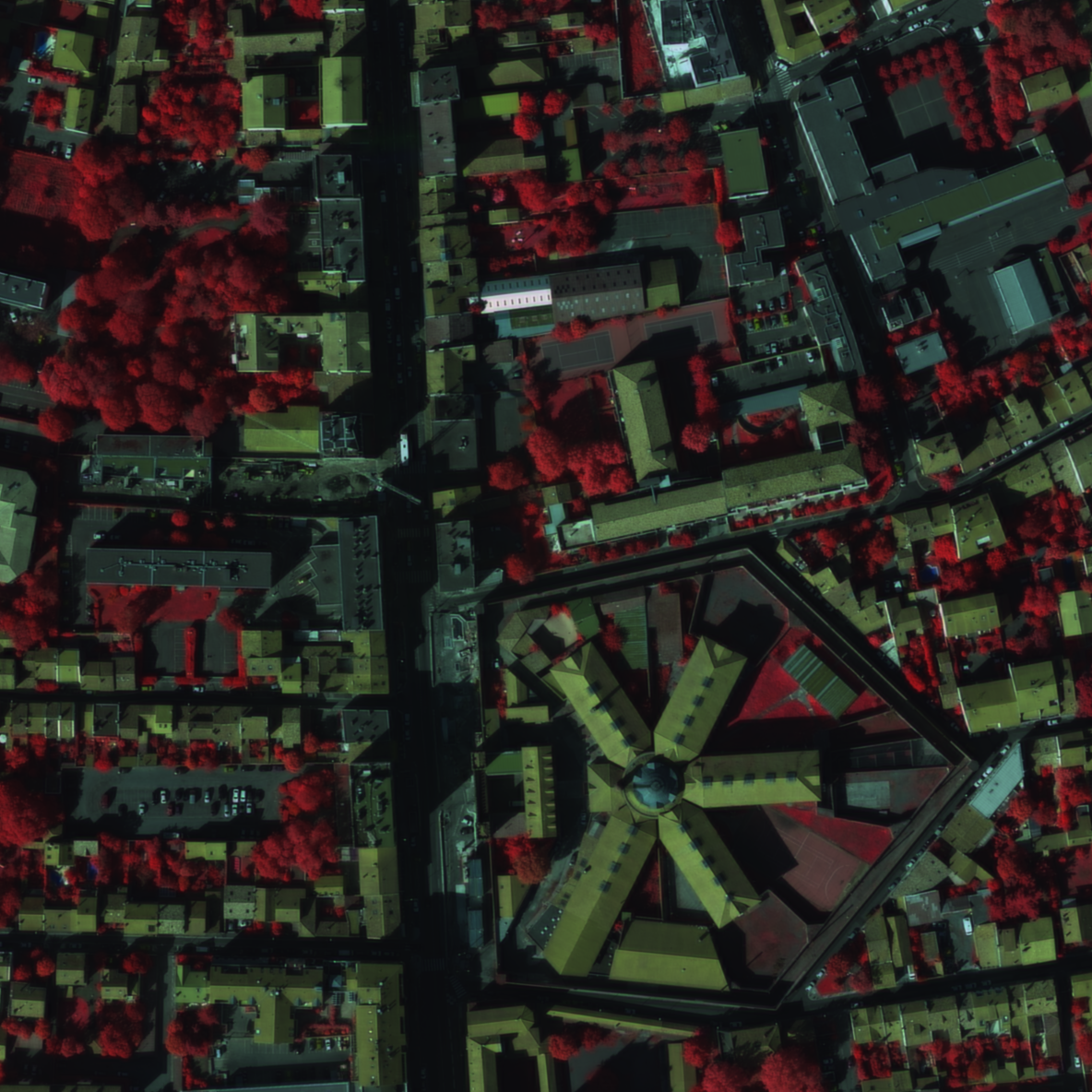} \\
  GSA & PRACS & HPF & SFIM\\
  \includegraphics[trim= 28.5cm 27.6cm 10.5cm 11.4cm, clip=true, width=0.243\textwidth]{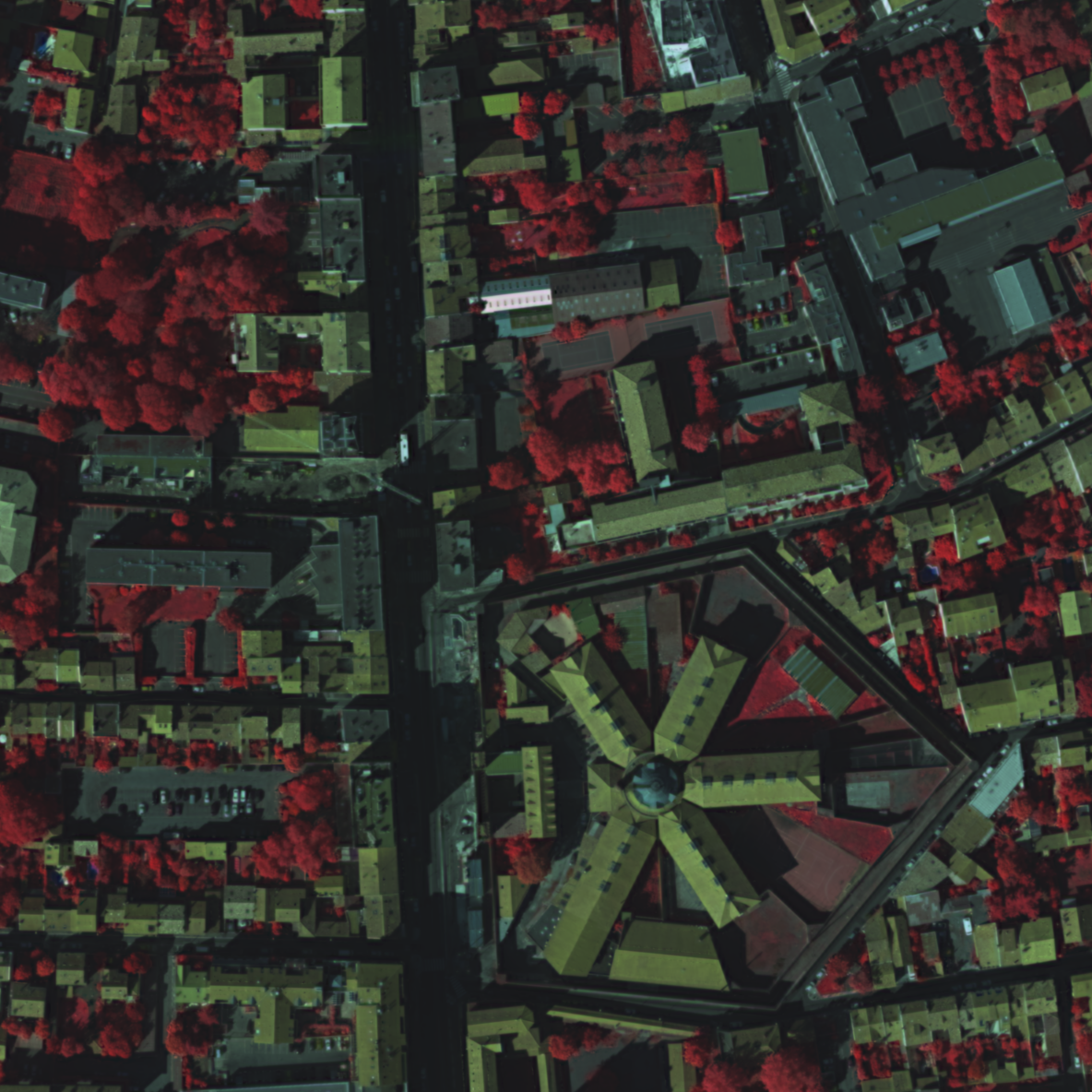} & 
  \includegraphics[trim= 28.5cm 27.6cm 10.5cm 11.4cm, clip=true, width=0.243\textwidth]{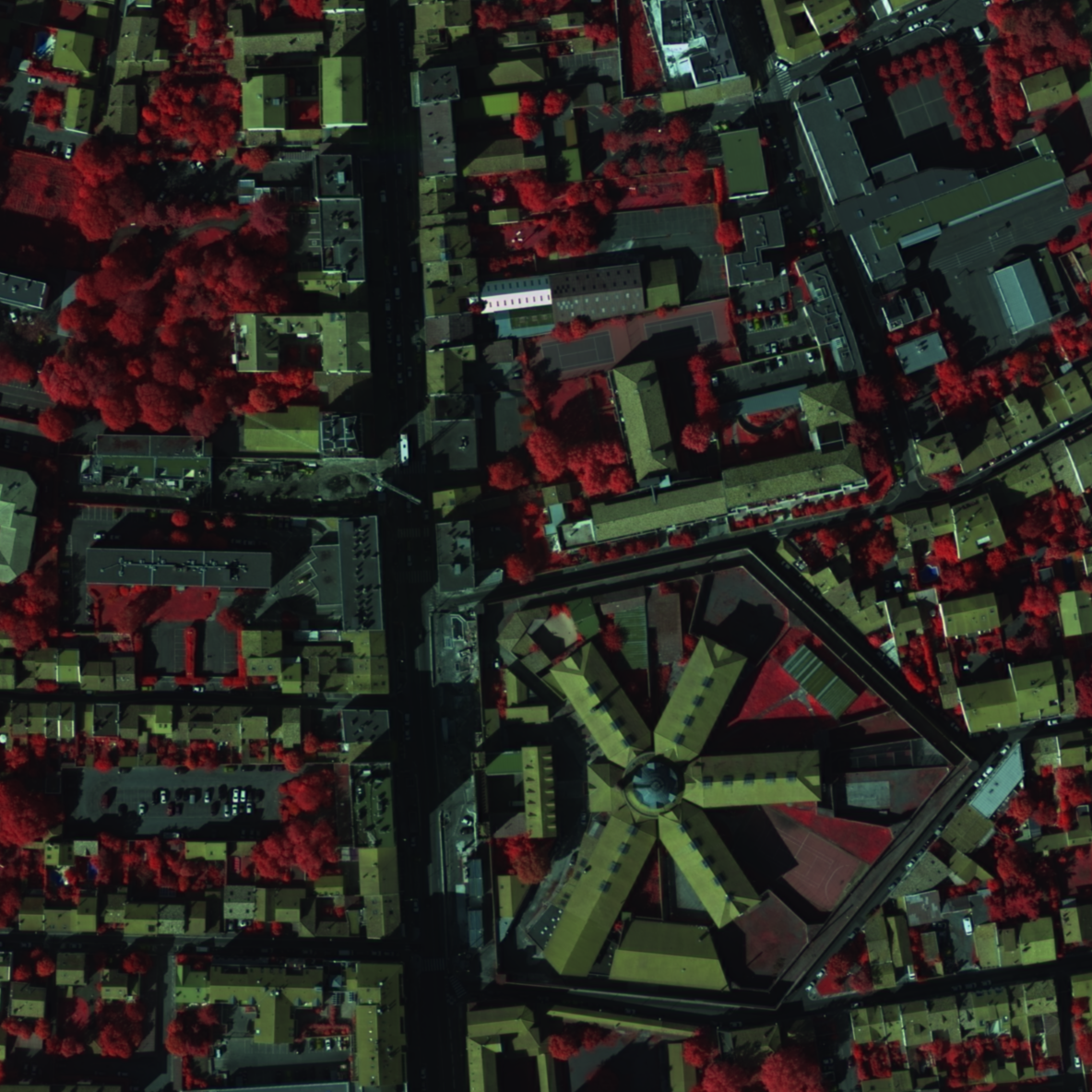} &
  \includegraphics[trim= 28.5cm 27.6cm 10.5cm 11.4cm, clip=true, width=0.243\textwidth]{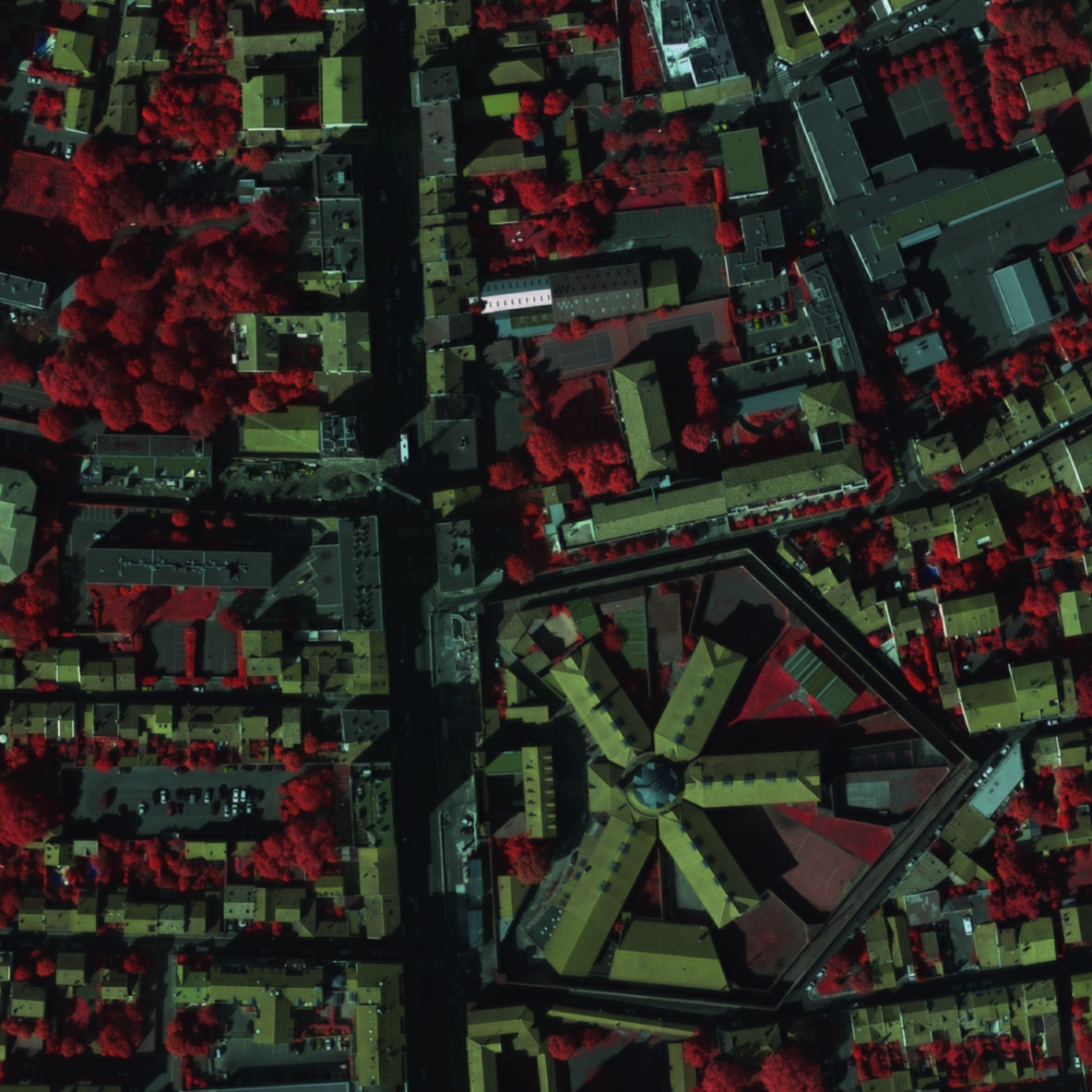} &
  \includegraphics[trim= 28.5cm 27.6cm 10.5cm 11.4cm, clip=true, width=0.243\textwidth]{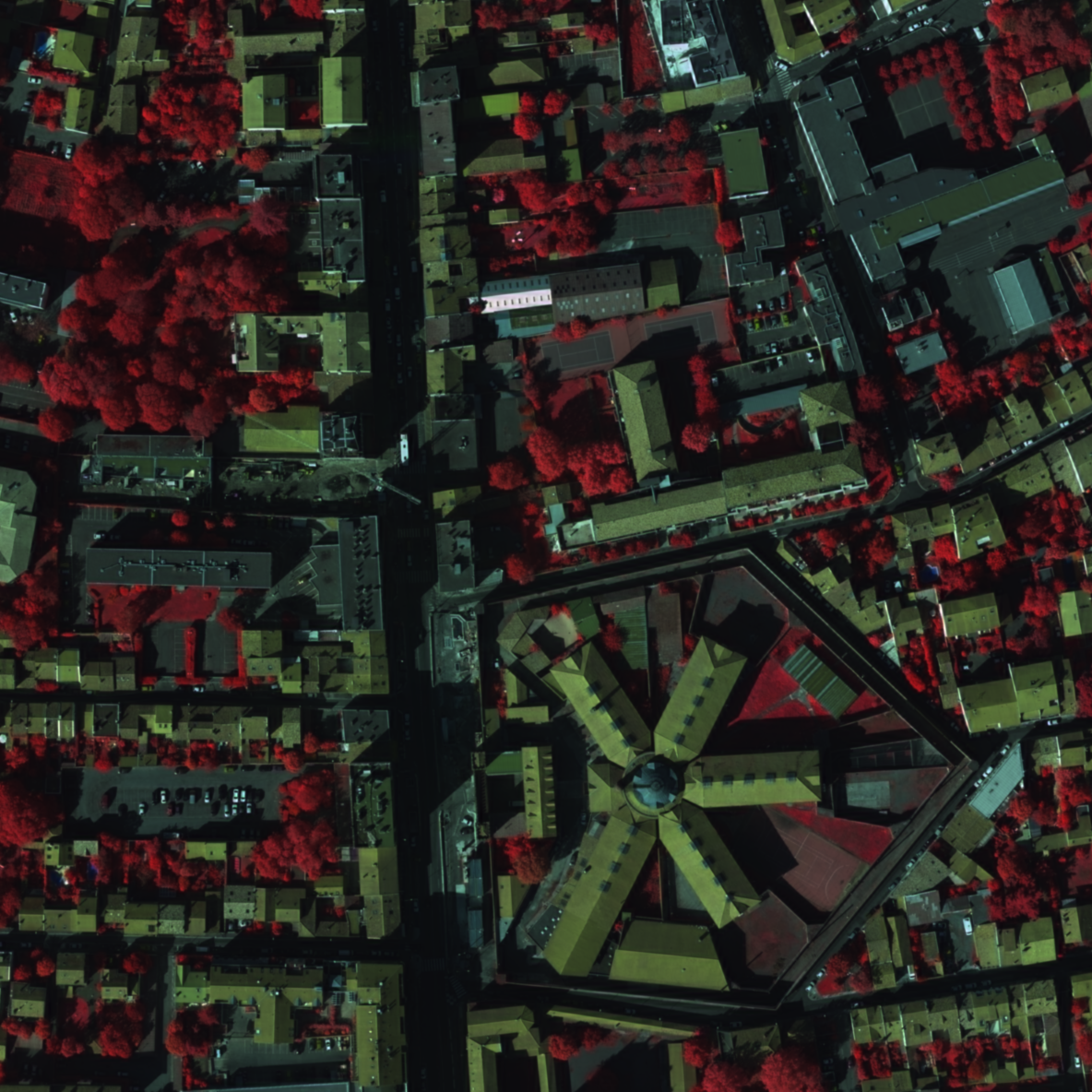} \\
  LMVM & ATWT & AWLP & GLP\\
\end{tabular}
\begin{tabular}{c@{\hskip 0.02in}c@{\hskip 0.02in}c}
  \includegraphics[trim= 28.5cm 27.6cm 10.5cm 11.4cm, clip=true, width=0.243\textwidth]{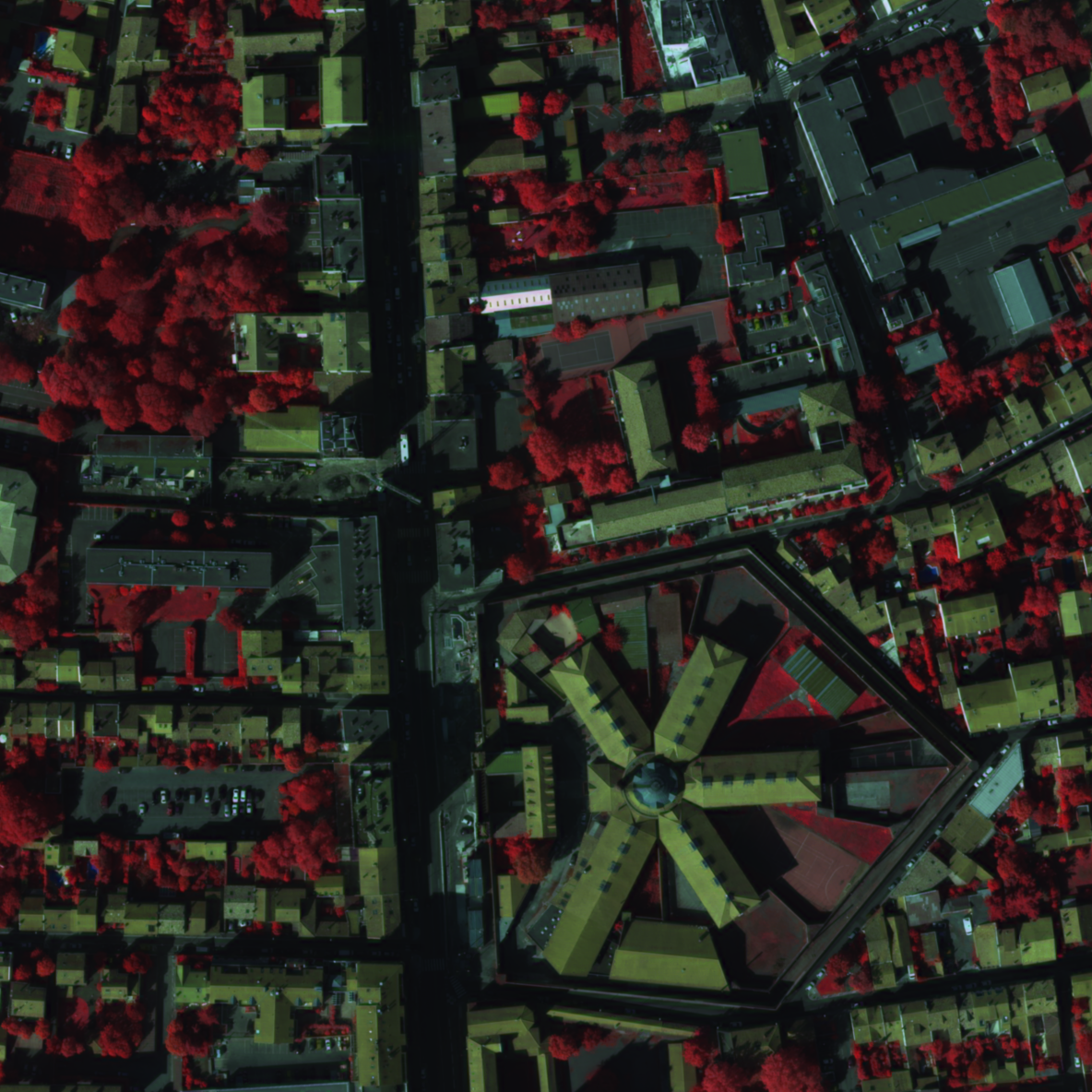}  &
  \includegraphics[trim= 28.5cm 27.6cm 10.5cm 11.4cm, clip=true, width=0.243\textwidth]{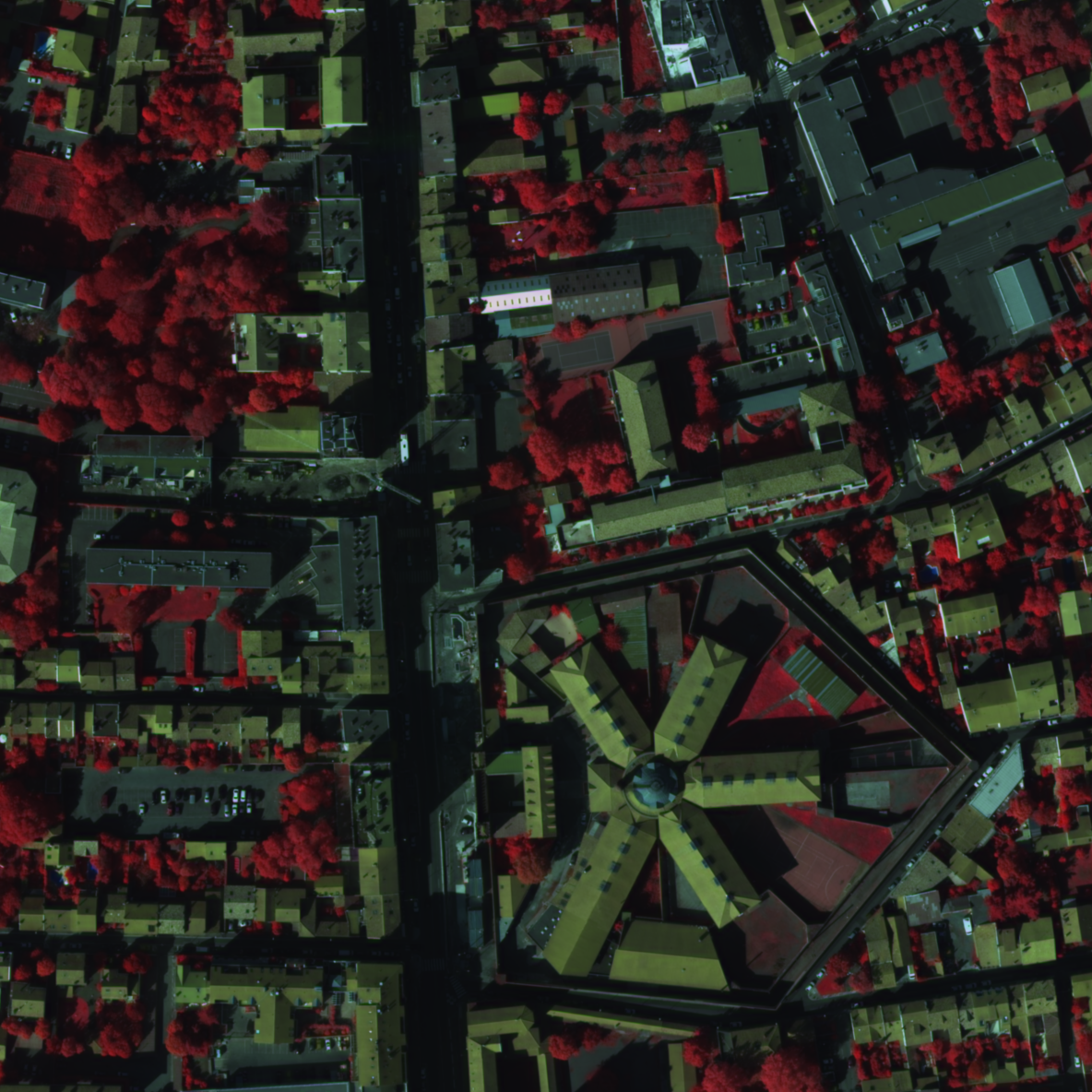} &
  \includegraphics[trim= 28.5cm 27.6cm 10.5cm 11.4cm, clip=true, width=0.243\textwidth]{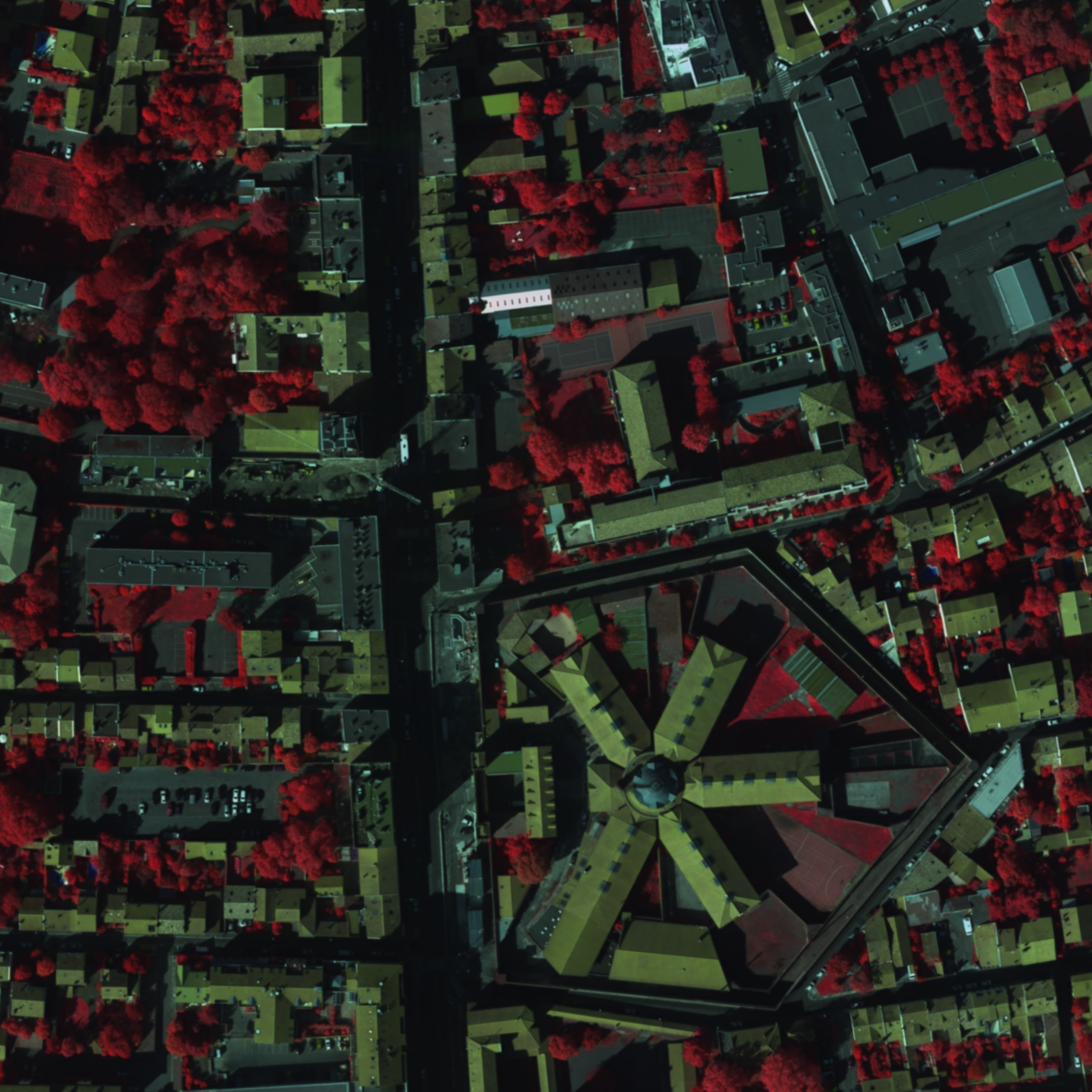} \\
  P+XS & NLV & NLVD\\
\end{tabular}
\caption{Close-ups of the reference false-color image involving infra-red, red, and green bands in place of the usual RGB at a resolution of 30 cm per pixel and of the fusion products provided by all methods under comparison. The Gaussian standard deviation used for the simulation of the low-resolution spectral components was $\sigma=1.7$. For these experiments, the data were non registered and the panchro-spectral constraint fulfilled with $\alpha_B=0.1$, $\alpha_G=0.4$, $\alpha_R=0.25$, and $\alpha_I=0.25$. To a lesser or greater extent, strong aliasing severely damages the pansharpened images provided by all techniques except ours. In particular, observe how the annoying color artifacts on the white cars and along the wall appearing in all other results are almost avoided by NLVD. Because of not being predictable to such an extent from the quantitative evaluation, it is really impressive the low visual quality of the fusion products provided by P+XS and NLV, since not only the spectral distortion is prominent but there is also a clear loss of sharpness.}
\label{fig_30cm_s17_RGBNIR_noregist_linear}
\end{figure}

\begin{figure}[!p]
\footnotesize
\centering
\renewcommand{\arraystretch}{0.5}
\begin{tabular}{c@{\hskip 0.02in}c@{\hskip 0.02in}c@{\hskip 0.02in}c}
  \includegraphics[trim= 28.5cm 27.6cm 10.5cm 11.4cm, clip=true, width=0.243\textwidth]{RGBNIRnoreglin_true.png} &
  \includegraphics[trim= 28.5cm 27.6cm 10.5cm 11.4cm, clip=true, width=0.243\textwidth]{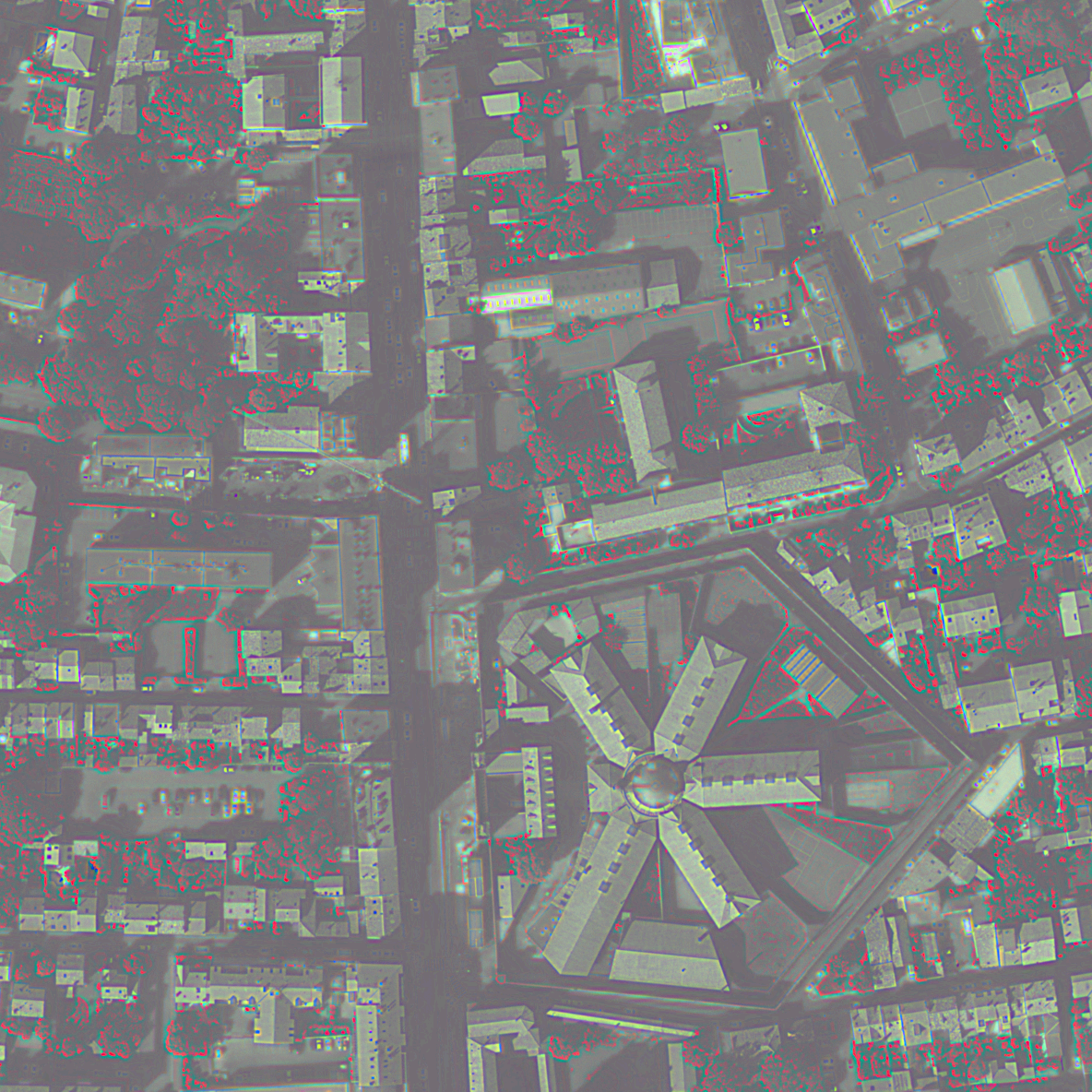} &
  \includegraphics[trim= 28.5cm 27.6cm 10.5cm 11.4cm, clip=true, width=0.243\textwidth]{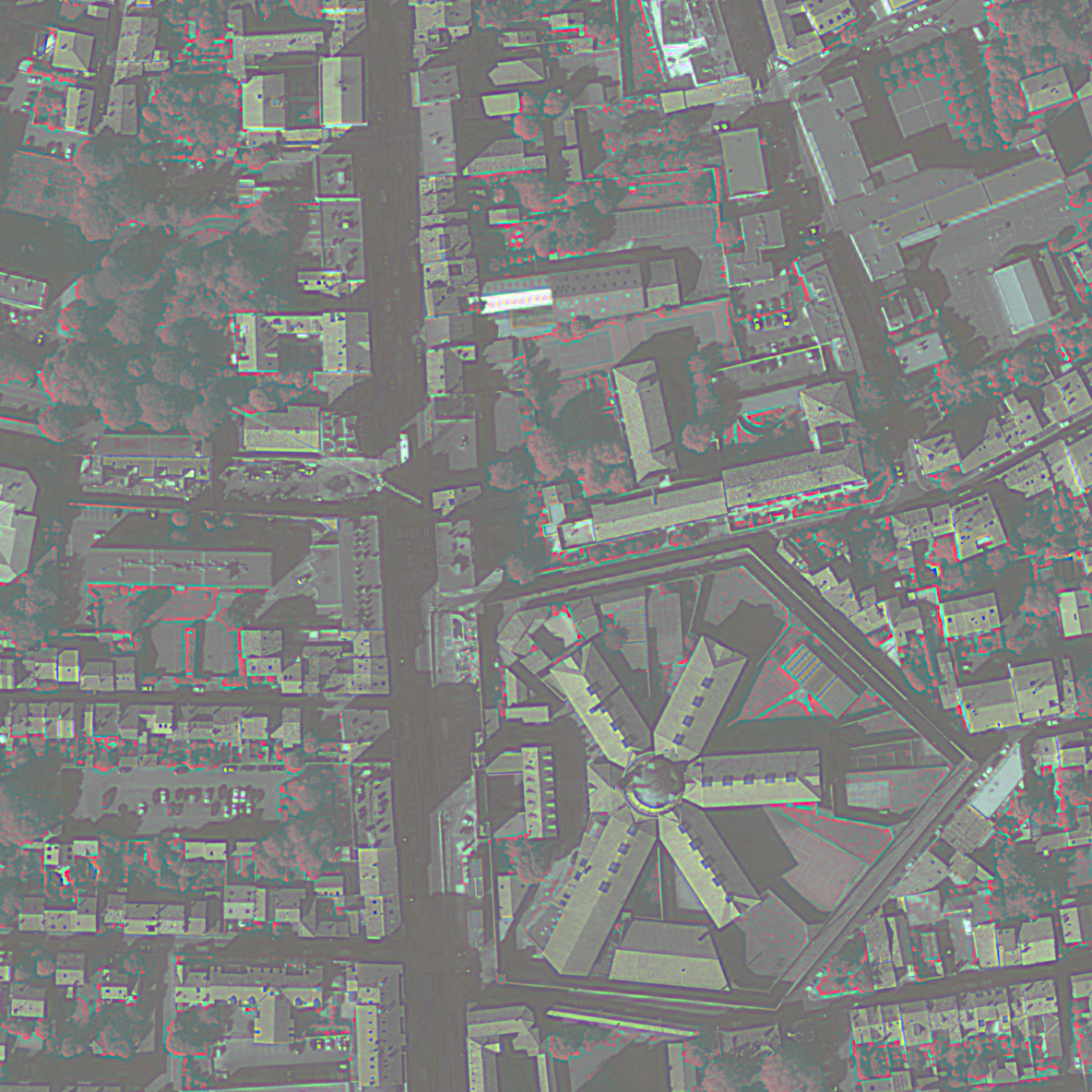} &
  \includegraphics[trim= 28.5cm 27.6cm 10.5cm 11.4cm, clip=true, width=0.243\textwidth]{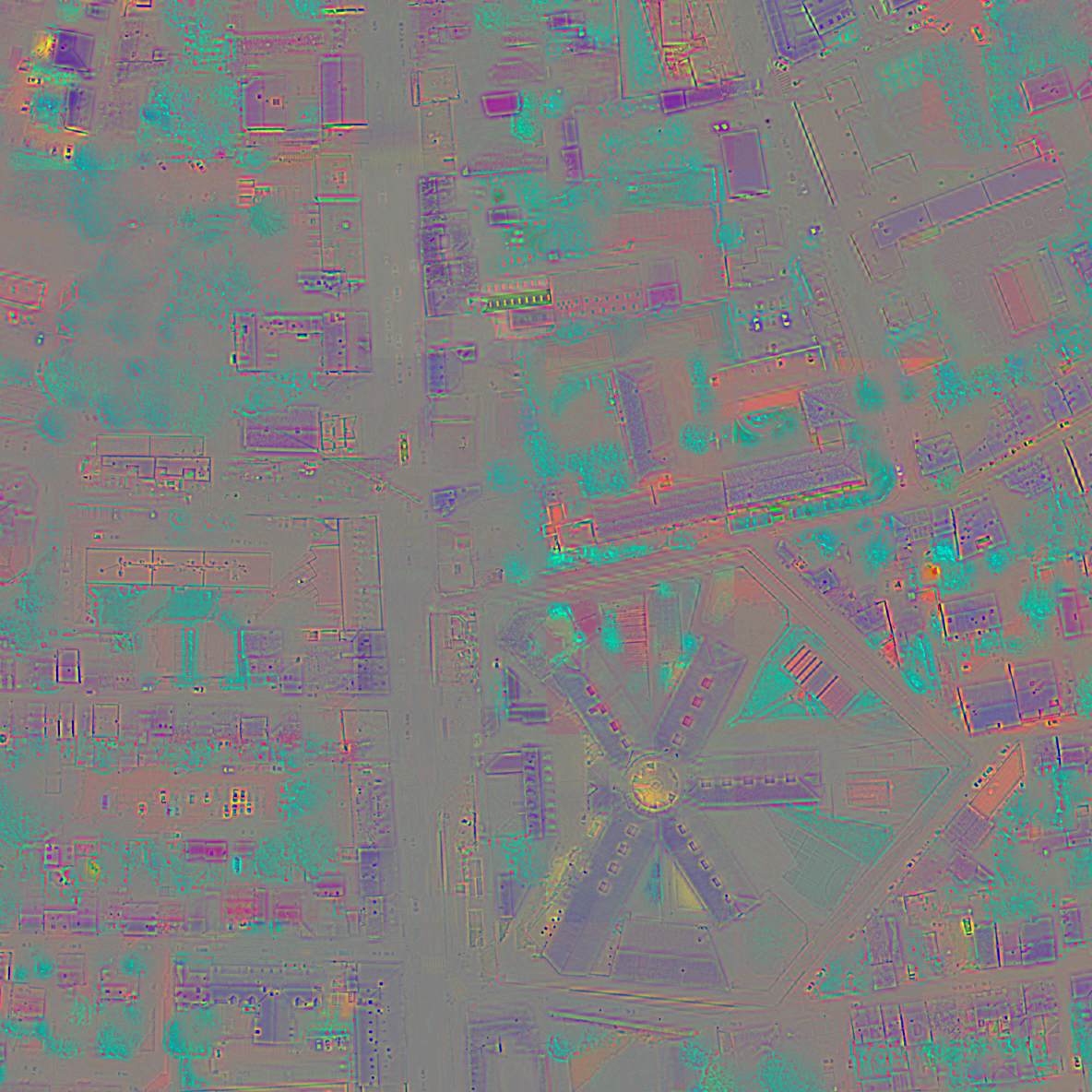} \\
  Reference & PCA & Brovey & BDSD\\ 
  \includegraphics[trim= 28.5cm 27.6cm 10.5cm 11.4cm, clip=true, width=0.243\textwidth]{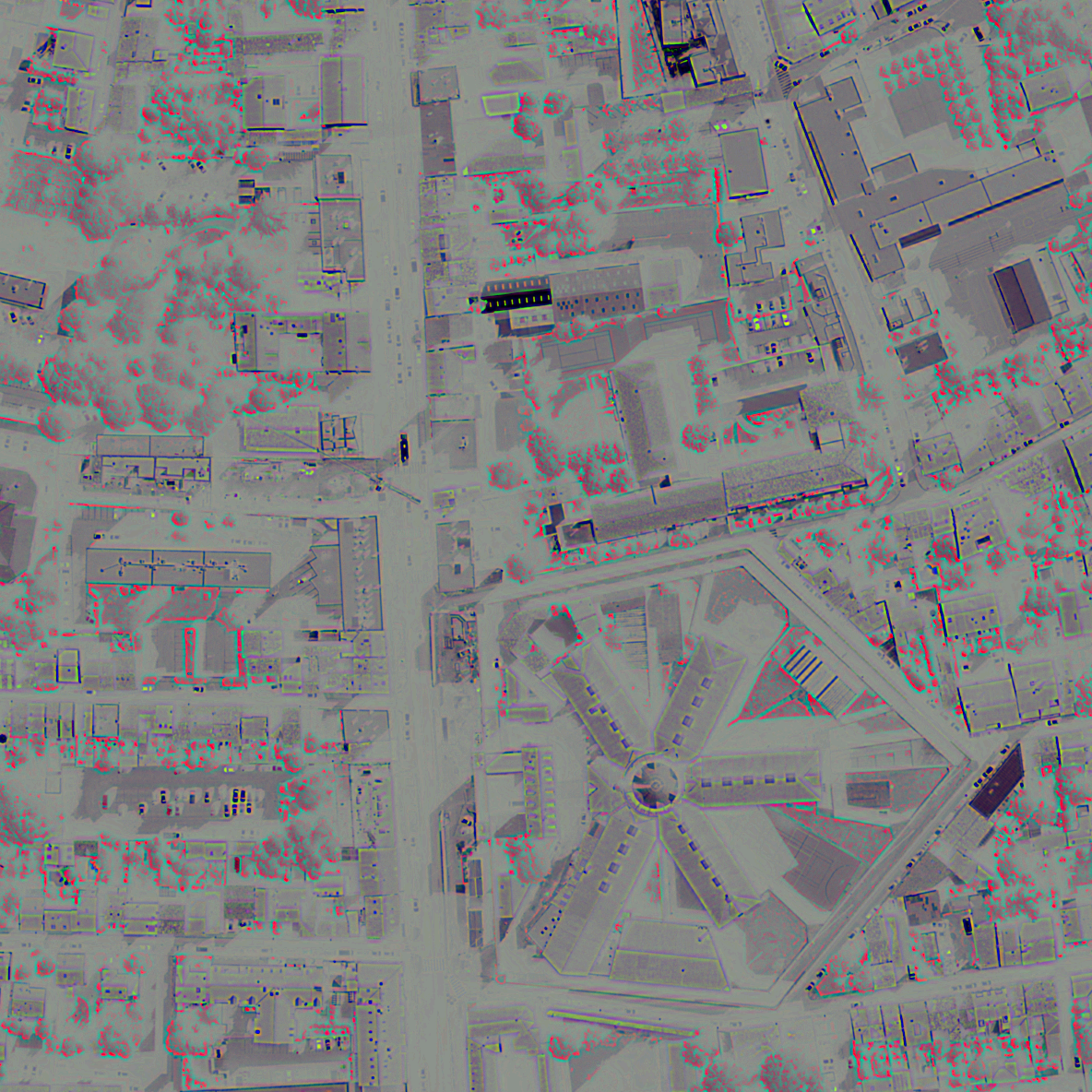} &
  \includegraphics[trim= 28.5cm 27.6cm 10.5cm 11.4cm, clip=true, width=0.243\textwidth]{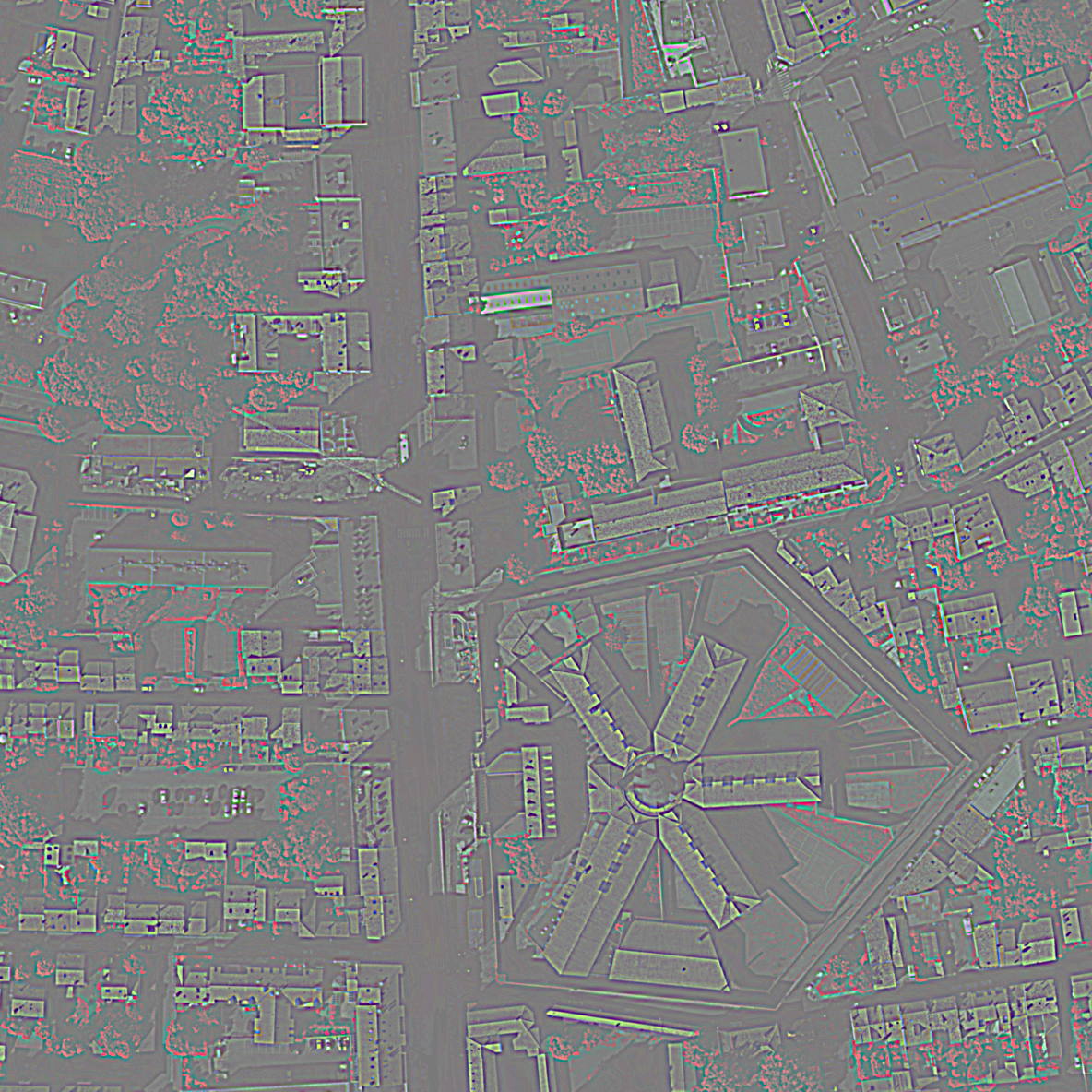} &
  \includegraphics[trim= 28.5cm 27.6cm 10.5cm 11.4cm, clip=true, width=0.243\textwidth]{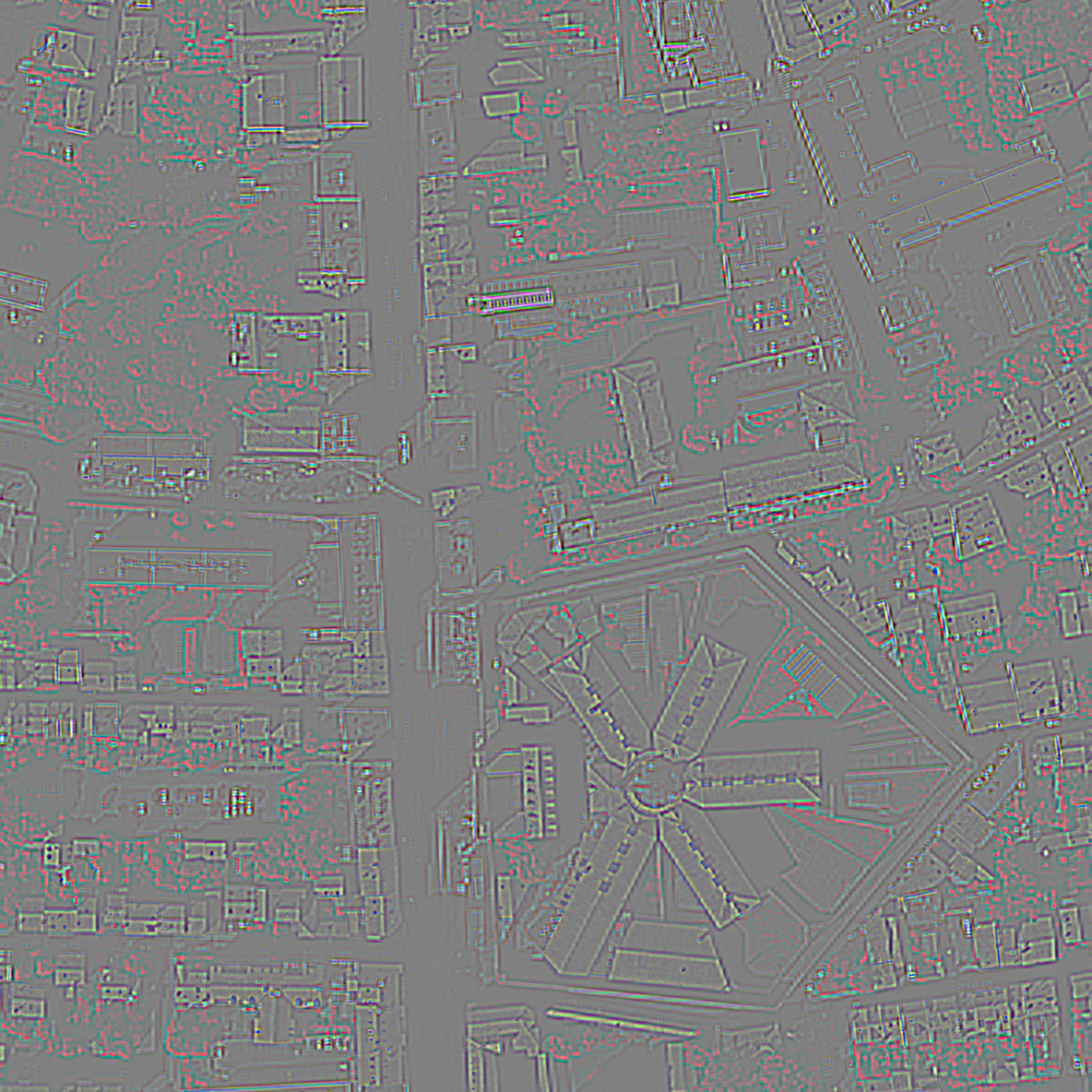} &
  \includegraphics[trim= 28.5cm 27.6cm 10.5cm 11.4cm, clip=true, width=0.243\textwidth]{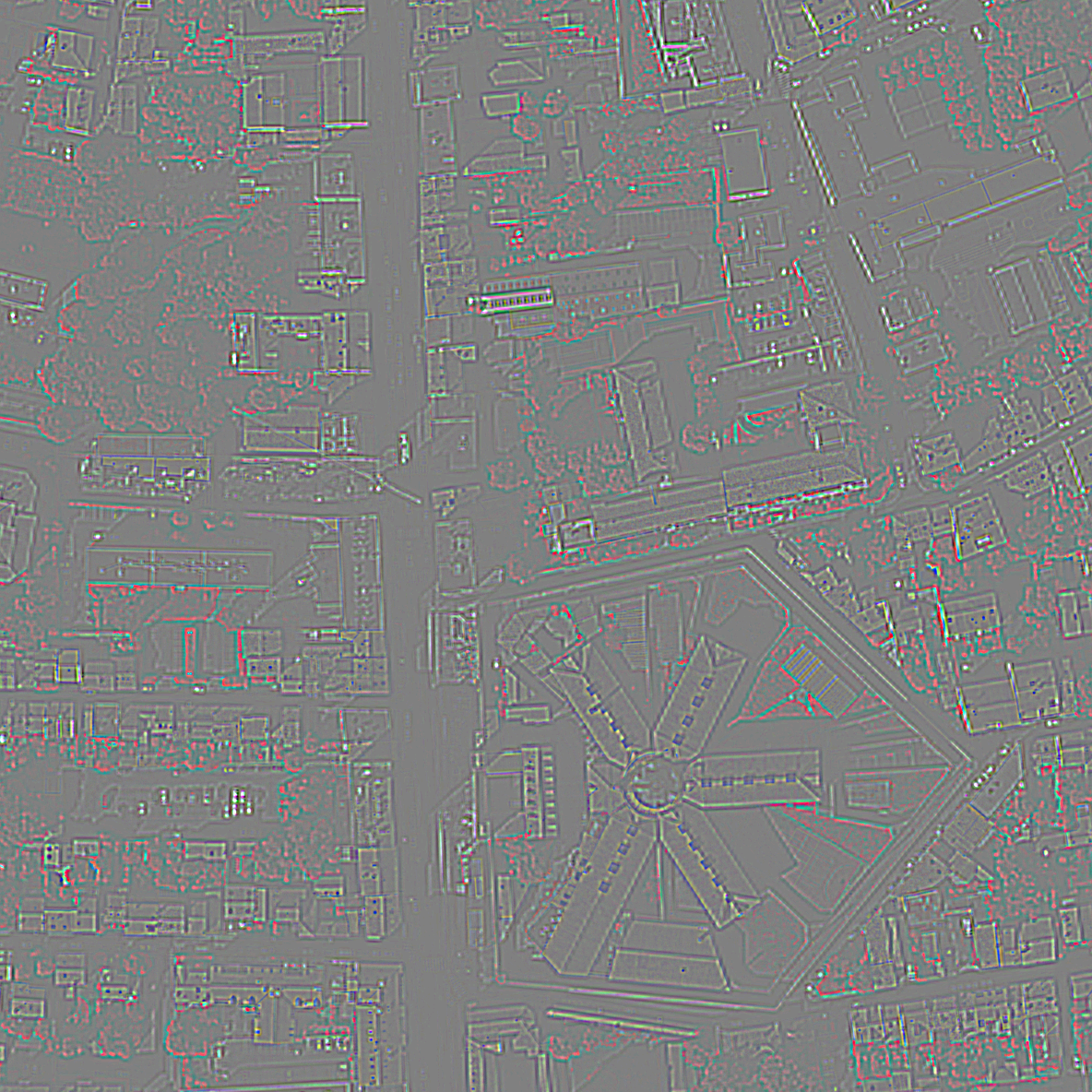} \\
  GSA & PRACS & HPF & SFIM\\
  \includegraphics[trim= 28.5cm 27.6cm 10.5cm 11.4cm, clip=true, width=0.243\textwidth]{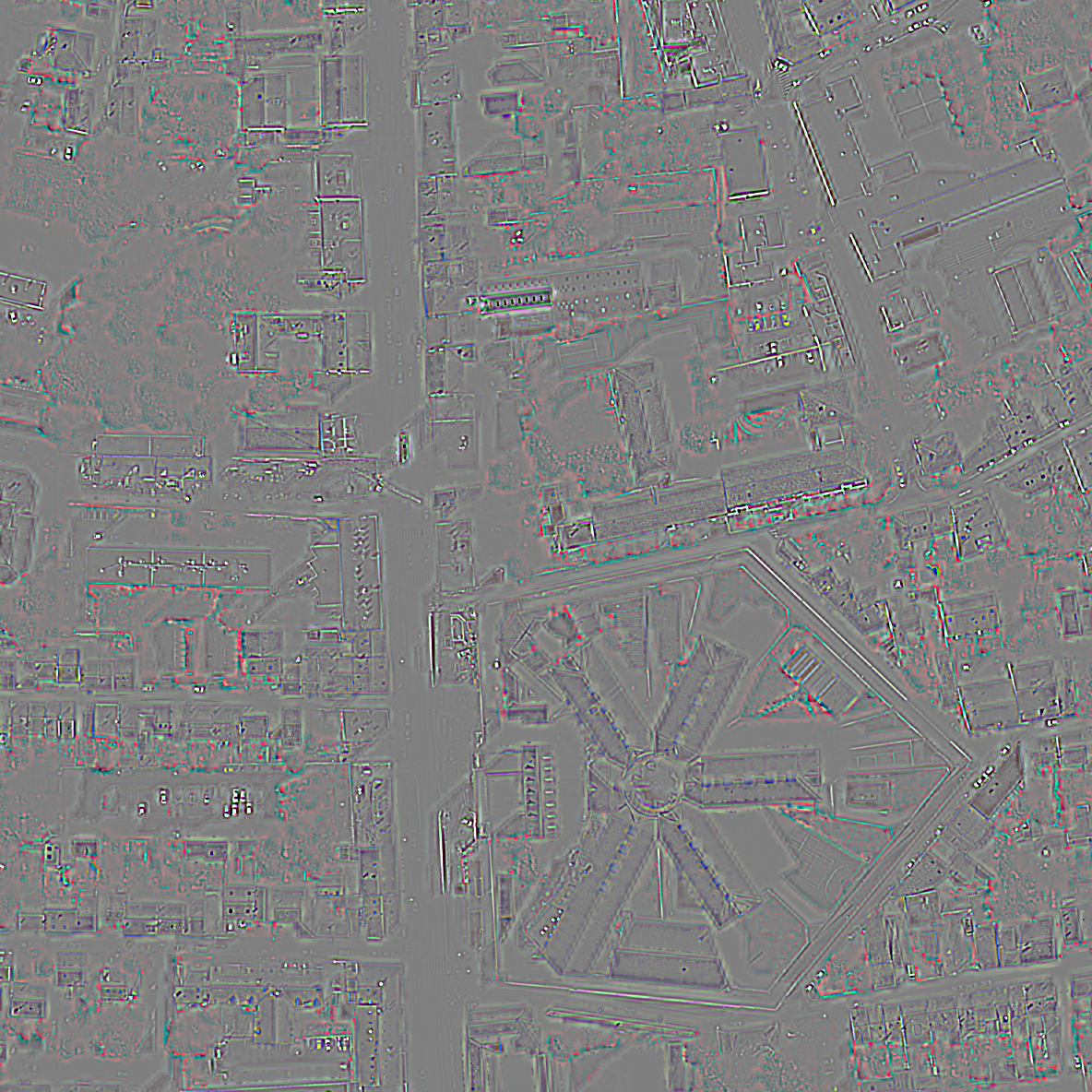} &
  \includegraphics[trim= 28.5cm 27.6cm 10.5cm 11.4cm, clip=true, width=0.243\textwidth]{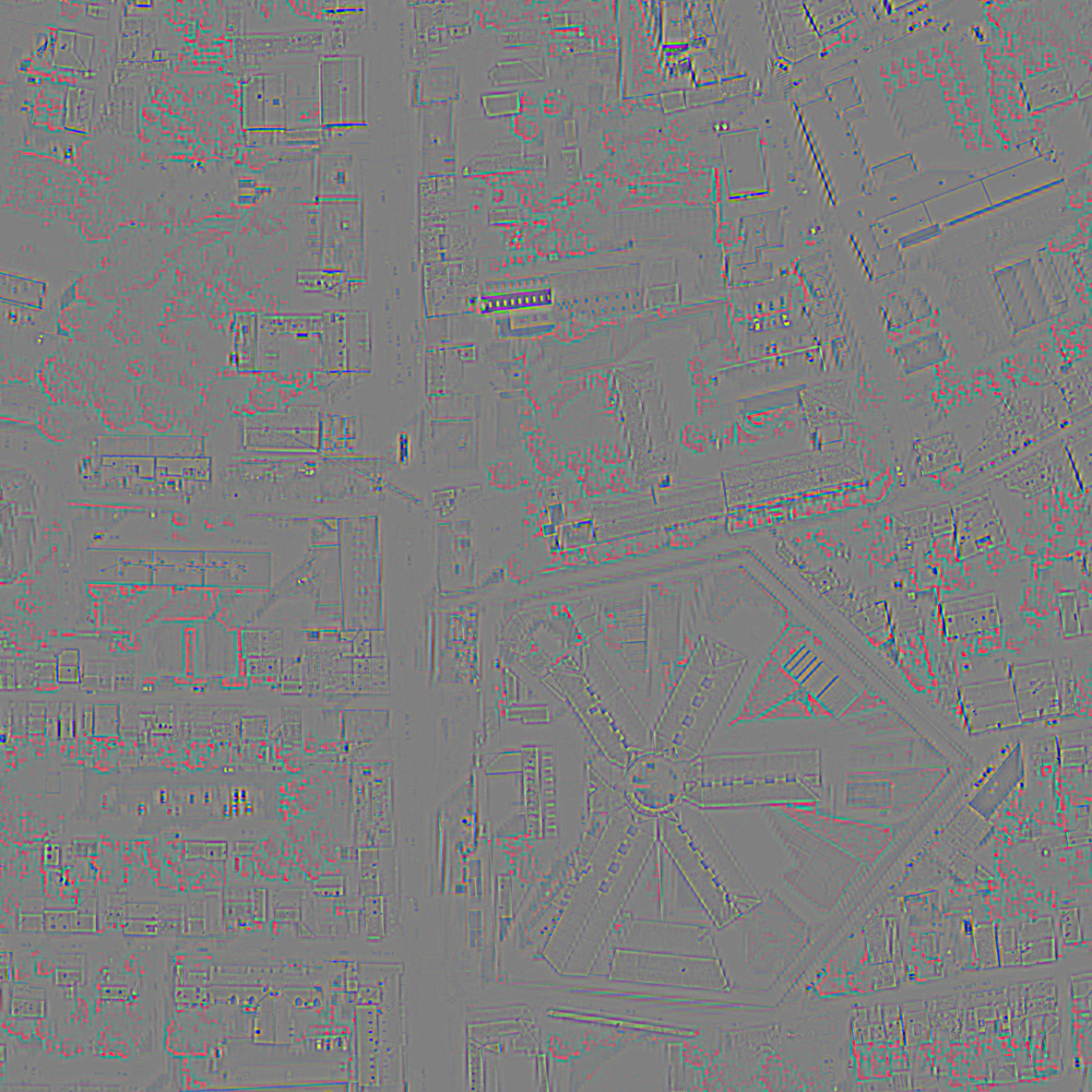} &
  \includegraphics[trim= 28.5cm 27.6cm 10.5cm 11.4cm, clip=true, width=0.243\textwidth]{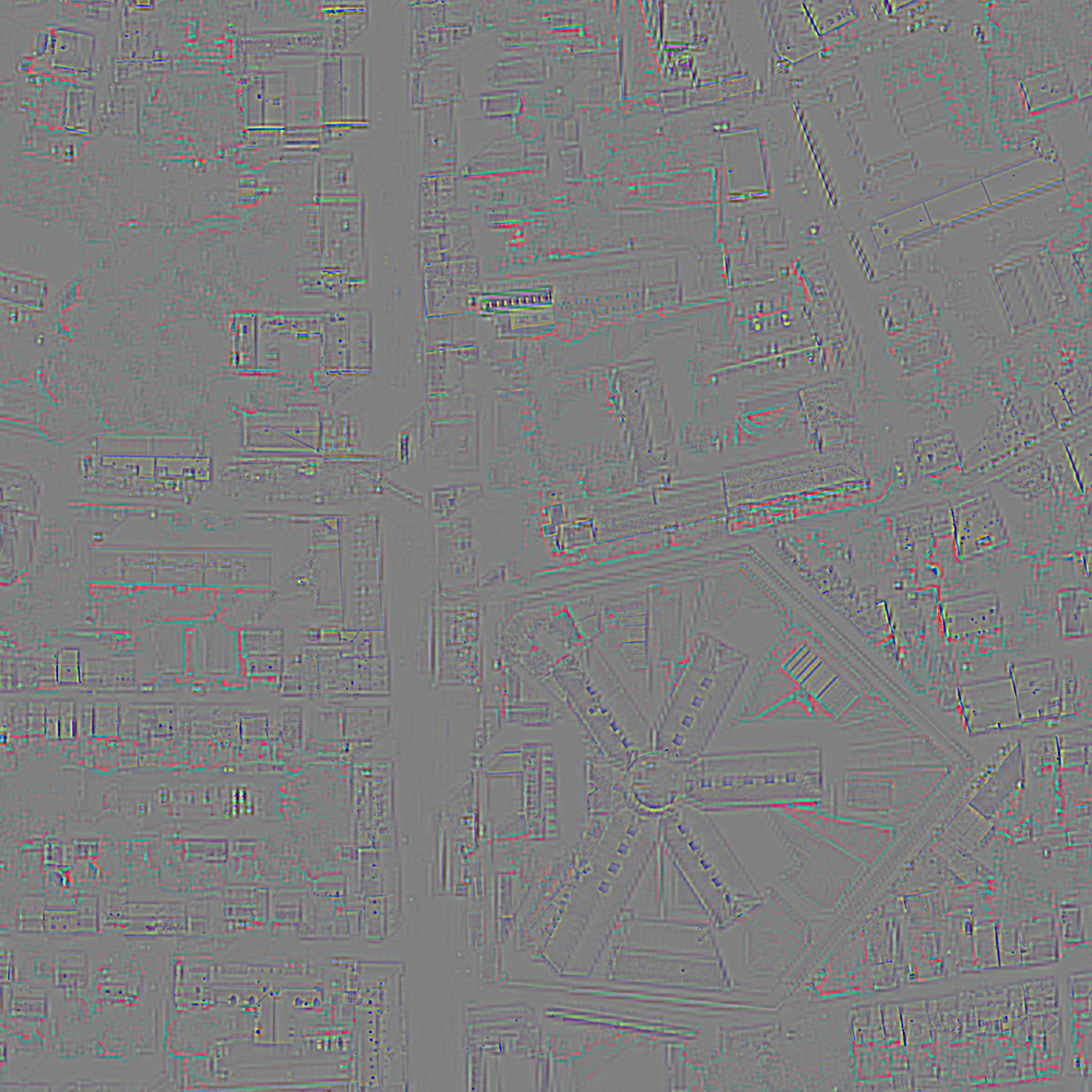} &
  \includegraphics[trim= 28.5cm 27.6cm 10.5cm 11.4cm, clip=true, width=0.243\textwidth]{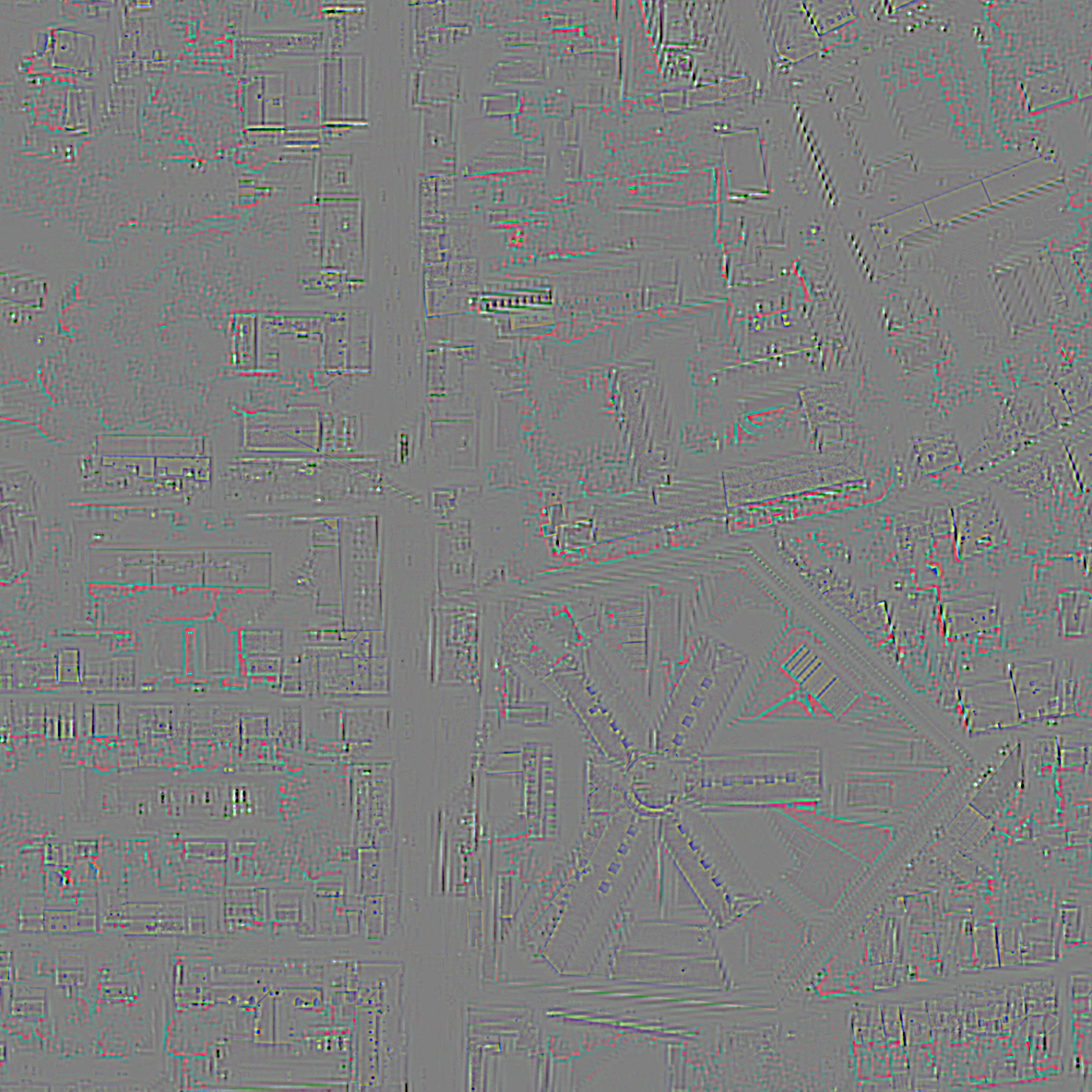} \\
  LMVM & ATWT & AWLP & GLP \\
\end{tabular}
\begin{tabular}{c@{\hskip 0.02in}c@{\hskip 0.02in}c}
  \includegraphics[trim= 28.5cm 27.6cm 10.5cm 11.4cm, clip=true, width=0.243\textwidth]{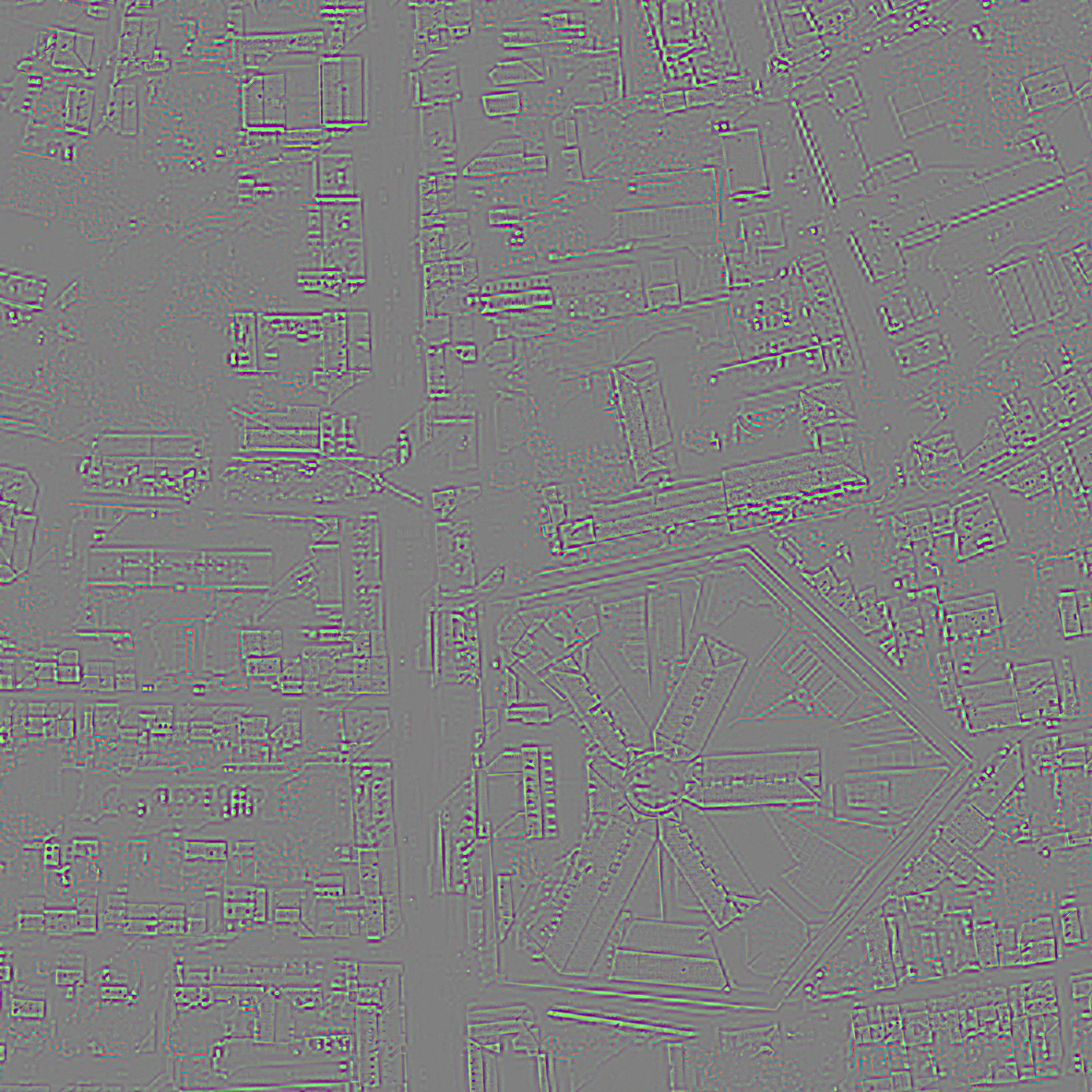} &
  \includegraphics[trim= 28.5cm 27.6cm 10.5cm 11.4cm, clip=true, width=0.243\textwidth]{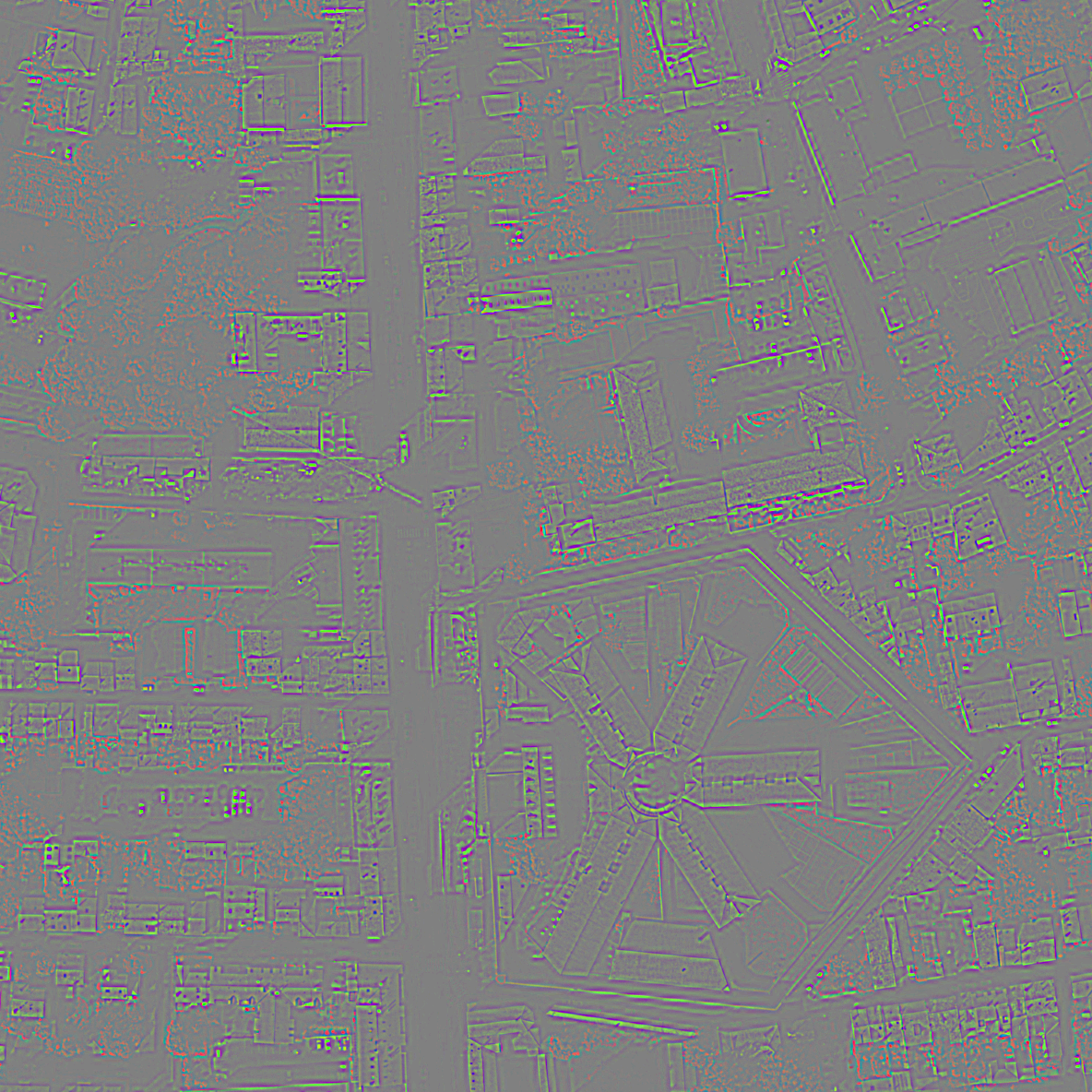} &
  \includegraphics[trim= 28.5cm 27.6cm 10.5cm 11.4cm, clip=true, width=0.243\textwidth]{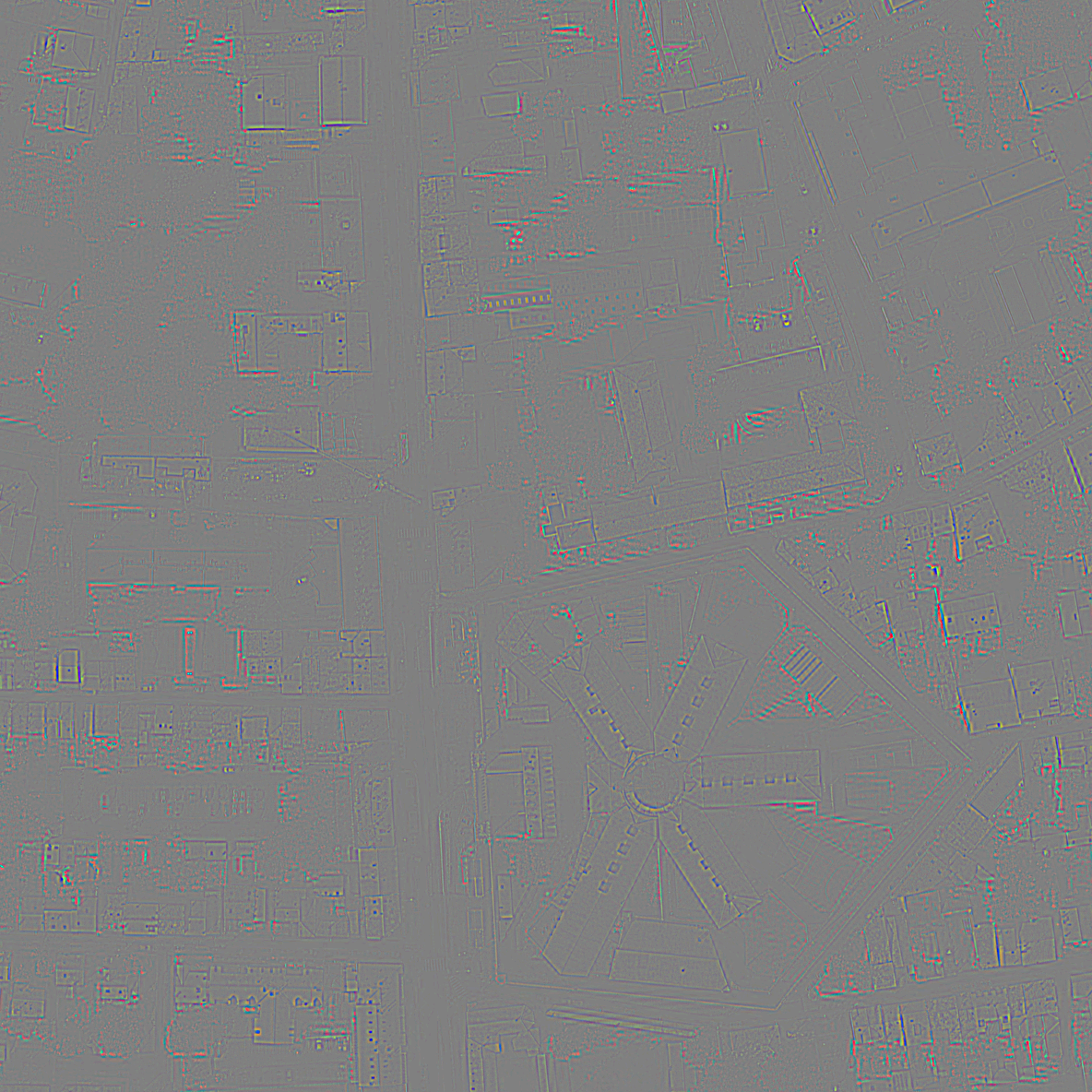} \\
  P+XS & NLV & NLVD\\
\end{tabular}
\caption{Close-ups of the reference false-color image involving infra-red, red, and green bands in place of the usual RGB at a resolution of 30 cm per pixel and of the difference images associated to the fusion products displayed in Figure \ref{fig_30cm_s17_RGBNIR_noregist_linear}. For visualization purposes, the intensity values have been linearly rescaled from $[-20,20]$ to $[0,255]$. The first conclusion that can be drawn is the superiority of the proposed model in preserving the spatial and spectral information from the panchromatic and low-resolution bands, respectively. Indeed, the fact that the difference image associated to NLVD contains less geometry and chromaticity than the others is evidence of that. As observed along the experimental section, CS-based methods seem to mainly suffer from spectral distortion, whereas MRA-based techniques are less accurate in reproducing the geometry of the scene. Furthermore, the difference images also illustrate the poor performances of both P+XS and NLV.}
\label{fig_30cm_s17_RGBNIR_noregist_linear_dif}
\end{figure}

For a visual quality assessment, Figure \ref{fig_30cm_s17_RGBNIR_noregist_linear} shows close-ups of the false-color images at a resolution of 30 cm per pixel involving near-infrared, red, and green components in place of the usual RGB associated to the results on the third picture of the proposed dataset (Figure \ref{fig:dataset})  The Gaussian standard deviation used for the simulation of the low-resolution spectral bands was $\sigma=1.7$. As previously done, we also display in Figure \ref{fig_30cm_s17_RGBNIR_noregist_linear_dif} the difference images between the reference one and each result. We observe that strong aliasing severely compromises the visual performances of all methods under comparison except ours. Indeed, see the color artifacts on the white cars and, in particular, the aliased pattern along the wall separating the two parking areas.

\medskip

The results displayed in this subsection, particularly those on 4-band images, illustrate that NLVD applied on registered spectral components also leads to convincing results. Indeed, note that in Tables \ref{table_30cm_RGBNIR_nonregist_linear} and \ref{table_60cm_RGBNIR_nonregist_linear} it becomes the second best performing method for almost all metrics -- only AWLP in Table \ref{table_60cm_RGBNIR_nonregist_linear} and GLP in Table \ref{table_60cm_s13_RGBNIR_nonregist_linear} lead to better SAM values. Even if we propose the use of NLVD with the original misregistered low-resolution spectral data, the quality indices show that, in the case that we do not dispose of them, NLVD applied on the re-interpolated components still provides fused products with pretty good spatial and spectral quality. Apart from the analysis of the effects caused by aliasing and misregistration, NLVD outperforming classical and state-of-the-art methods in both chains implies that the proposed model, which is based on the minimization of the energy \eqref{eq:functional2}, describes the satellite image fusion problem better than P+XS and NLV whenever blue, green, red, and near-infrared bands are involved and the mixing coefficients in \eqref{eq:panconstraint} become more and more realistic.

\subsubsection{Non-Registered Bands and Pancho-Spectral Constraint Not Fulfilled}

As we have discussed and experimentally checked in Section \ref{sec:linearity}, the panchro-spectral constraint \eqref{eq:panconstraint} does not follow for real satellite data. Accordingly, we move towards the more realistic scenario in which none of the methods based on it can be applied. Therefore, we compare here the proposed model with PCA, HPF, SFIM, LMVM, ATWT, and GLP. Even if the panchro-spectral constraint is not fulfilled, we are forced to simulate the panchromatic image and the linear combination is still the simplest way. In this case, the mixing coefficients were fixed to $\alpha_B=0$, $\alpha_G=0.4$, $\alpha_R=0.35$, and $\alpha_I=0.25$. Since the blue band of any instrument usually falls almost outside the panchromatic band to avoid selective scattering effects from the atmosphere, we set $\alpha_B=0$. The low-resolution channels were simulated after translation of the high-resolution ones by Gaussian filtering of standard deviations $\sigma\in\{1.3, 1.7\}$ and subsampling of factor $s=4$.

Tables \ref{table_30cm_RGBNIR_nonregist_nolinear} and \ref{table_60cm_RGBNIR_nonregist_nolinear} reveal the quality indexes obtained by each of the methods on the data from Figure \ref{fig:dataset} at resolutions of 30 cm and 60 cm per pixel, respectively. We observe that the numerical results displayed in these tables are quite similar to those from Tables \ref{table_30cm_RGBNIR_nonregist_linear} and \ref{table_60cm_RGBNIR_nonregist_linear}, for which the blue band had a small influence in the simulation of the panchromatic. This is entirely understandable since all pansharpening methods under comparison in this subsection do not depend on the panchro-spectral constraint so their performances are not supposed to be affected by changing the mixing coefficients. In general terms, the current results strengthen the idea that better quality measures are obtained with the proposed model if the satellite image fusion takes place before co-registration of the low-resolution spectral components. Another important point to note is that PCA, HPF, SFIM, and NLVD behaves almost independent of the amount of aliasing in the provided data, which is highly desirable in remote sensing.

\begin{table}[!t]
\footnotesize
\centering
\begin{subtable}[h]{0.99\textwidth}
\centering
\begin{tabular}{|c|c|c|c|c|c|c|c|c|c|}
\hline
 & Methods & RMSE & ERGAS & SAM & SSIM & Q$2^n$  \\ \hline \hline
 & Reference & 0 & 0 & 0 & 1 & 1 \\ \hline \hline
 \multirow{7}{*}{\rotatebox[origin=c]{90}{Registered}} & PCA & 3.9434 & 2.6747 & 3.0181 & 0.9900 & 0.9512 \\ \cline{2-7}
& HPF & 3.8244 & 2.6034 & 2.7653 & 0.9937 & 0.9700 \\ \cline{2-7}
& SFIM & 3.3639 & 2.2981 & 2.2541 & 0.9945 & 0.9753 \\ \cline{2-7}
& LMVM & 3.7391 & 2.5905 & 2.2386 & 0.9940 & 0.9654 \\ \cline{2-7}
& ATWT & 3.3418 & 2.2453 & 2.8791 & 0.9946 & 0.9737 \\ \cline{2-7}
& GLP & 2.8024 & 1.9009 & 2.2086 & 0.9961 & 0.9792 \\ \cline{2-7}
& NLVD & 2.4639 & 1.6870 & 2.1040 & 0.9978 & 0.9828 \\ \hline\hline
 \multirow{6}{*}{\rotatebox[origin=c]{90}{Misregistered}} & HPF & 3.6465 & 2.4726 & 2.6870 & 0.9946 & 0.9719 \\ \cline{2-7}
& SFIM & 3.1718 & 2.1559 & 2.2354 & 0.9954 & 0.9769 \\ \cline{2-7}
& LMVM & 3.5459 & 2.4570 & 2.0719 & 0.9958 & 0.9676 \\ \cline{2-7}
& ATWT & 3.3076 & 2.2228 & 2.8473 & 0.9947 & 0.9734 \\ \cline{2-7}
& GLP & 2.8923 & 1.9701 & 2.4302 & 0.9960 & 0.9778 \\ \cline{2-7}
& NLVD & {\bf 1.9405} & {\bf 1.3010} & {\bf 1.5880} & {\bf 0.9991} & {\bf 0.9873} \\ \hline
\end{tabular}
\caption{Numerical results for $\sigma=1.7$.}
 \label{table_30cm_s17_RGBNIR_nonregist_nolinear}
 \end{subtable}
 
\medskip

\begin{subtable}[h]{0.99\textwidth}
\centering
\begin{tabular}{|c|c|c|c|c|c|c|c|c|c|}
\hline
 & Methods & RMSE & ERGAS & SAM & SSIM & Q$2^n$  \\ \hline \hline
 & Reference & 0 & 0 & 0 & 1 & 1 \\ \hline \hline
 \multirow{7}{*}{\rotatebox[origin=c]{90}{Registered}} & PCA & 3.6511 & 2.4779 & 2.9088 & 0.9920 & 0.9569 \\ \cline{2-7}
& HPF & 3.5464 & 2.4104 & 2.7299 & 0.9951 & 0.9725 \\ \cline{2-7}
& SFIM & 3.0666 & 2.0972 & 2.2020 & 0.9959 & 0.9785 \\ \cline{2-7}
& LMVM & 3.7347 & 2.5715 & 2.3516 & 0.9940 & 0.9678 \\ \cline{2-7}
& ATWT & 3.4040 & 2.2925 & 2.8895 & 0.9937 & 0.9709 \\ \cline{2-7}
& GLP & 3.2023 & 2.1987 & 2.1756 & 0.9944 & 0.9745 \\ \cline{2-7}
& NLVD & 2.6714 & 1.8381 & 2.1971 & 0.9972 & 0.9814 \\ \hline\hline
 \multirow{6}{*}{\rotatebox[origin=c]{90}{Misregistered}} & HPF & 3.3939 & 2.2991 & 2.6922 & 0.9957 & 0.9739 \\ \cline{2-7}
& SFIM & 2.9114 & 1.9834 & 2.2348 & 0.9965 & 0.9795 \\ \cline{2-7}
& LMVM & 3.7247 & 2.5637 & 2.4541 & 0.9940 & 0.9686 \\ \cline{2-7}
& ATWT & 3.4586 & 2.3378 & 2.9180 & 0.9934 & 0.9697 \\ \cline{2-7}
& GLP & 3.4590 & 2.3913 & 2.4987 & 0.9936 & 0.9716 \\ \cline{2-7}
& NLVD & {\bf 1.9266} & {\bf 1.2909} & {\bf 1.5573} & {\bf 0.9991} & {\bf 0.9872} \\ \hline
\end{tabular}
\caption{Numerical results for $\sigma=1.3$.}
 \label{table_30cm_s13_RGBNIR_nonregist_nolinear}
 \end{subtable}
\caption{Quantitative evaluation of the fused products on simulated data from 4-band (blue, green, red, and near-infrared) aerial images at resolution of 30 cm per pixel. The panchromatic images were computed as a weighted average of the reference spectral components with mixing coefficients $\alpha_B=0$, $\alpha_G=0.4$, $\alpha_R=0.35$, and $\alpha_I=0.25$. For these experiments, the low-resolution data were non registered and the panchro-spectral constraint not fulfilled. All quality indices strengthen the superiority of NLVD for solving the pansharpening problem in a real sceario, where misregistration and aliasing are widely present.}
 \label{table_30cm_RGBNIR_nonregist_nolinear}
\end{table}

\begin{table}[!t]
\footnotesize
\centering
\begin{subtable}[h]{0.99\textwidth}
\centering
\begin{tabular}{|c|c|c|c|c|c|c|c|c|c|}
\hline
 & Methods & RMSE & ERGAS & SAM & SSIM & Q$2^n$  \\ \hline \hline
 & Reference & 0 & 0 & 0 & 1 & 1 \\ \hline \hline
 \multirow{7}{*}{\rotatebox[origin=c]{90}{Registered}} & PCA & 4.9474 & 3.3347 & 3.9511 & 0.9649 & 0.9423 \\ \cline{2-7}
& HPF & 4.5719 & 3.0728 & 3.6506 & 0.9664 & 0.9655 \\ \cline{2-7}
& SFIM & 4.1198 & 2.7712 & 3.0973 & 0.9721 & 0.9710 \\ \cline{2-7}
& LMVM & 4.3371 & 2.9677 & 2.9034 & 0.9697 & 0.9641 \\ \cline{2-7}
& ATWT & 4.0172 & 2.6753 & 3.7393 & 0.9723 & 0.9717 \\ \cline{2-7}
& GLP & 3.4612 & 2.3119 & 3.0173 & 0.9789 & 0.9776 \\ \cline{2-7}
& NLVD & 3.0917 & 2.0992 & 2.9921 & 0.9835 & 0.9818 \\ \hline\hline
 \multirow{6}{*}{\rotatebox[origin=c]{90}{Misregistered}} & HPF & 4.3718 & 2.9231 & 3.5546 & 0.9696 & 0.9678 \\ \cline{2-7}
& SFIM & 3.9071 & 2.6104 & 3.0648 & 0.9752 & 0.9731 \\ \cline{2-7}
& LMVM & 4.2021 & 2.8661 & 2.9210 & 0.9724 & 0.9658 \\ \cline{2-7}
& ATWT & 3.9615 & 2.6353 & 3.7094 & 0.9732 & 0.9717 \\ \cline{2-7}
& GLP & 3.5315 & 2.3649 & 3.2771 & 0.9787 & 0.9766 \\ \cline{2-7}
& NLVD & {\bf 2.5151} & {\bf 1.6692} & {\bf 2.3160} & {\bf 0.9893} & {\bf 0.9858} \\ \hline
\end{tabular}
\caption{Numerical results for $\sigma=1.7$.}
 \label{table_60cm_s17_RGBNIR_nonregist_nolinear}
 \end{subtable}
 
\medskip

\begin{subtable}[h]{0.99\textwidth}
\centering
\begin{tabular}{|c|c|c|c|c|c|c|c|c|c|}
\hline
 & Methods & RMSE & ERGAS & SAM & SSIM & Q$2^n$  \\ \hline \hline
 & Reference & 0 & 0 & 0 & 1 & 1 \\ \hline \hline
 \multirow{7}{*}{\rotatebox[origin=c]{90}{Registered}} & PCA & 4.5426 & 3.0620 & 3.8123 & 0.9700 & 0.9533 \\ \cline{2-7}
& HPF & 4.2227 & 2.8382 & 3.5942 & 0.9713 & 0.9699 \\ \cline{2-7}
& SFIM & 3.7449 & 2.5233 & 2.9984 & 0.9772 & 0.9759 \\ \cline{2-7}
& LMVM & 4.3390 & 2.9549 & 3.2275 & 0.9708 & 0.9662 \\ \cline{2-7}
& ATWT & 3.9987 & 2.6759 & 3.7477 & 0.9718 & 0.9703 \\ \cline{2-7}
& GLP & 3.7458 & 2.5361 & 2.9408 & 0.9763 & 0.9741 \\ \cline{2-7}
& NLVD & 3.3101 & 2.2610 & 3.0769 & 0.9817 & 0.9802 \\ \hline\hline
 \multirow{6}{*}{\rotatebox[origin=c]{90}{Misregistered}} & HPF & 4.1987 & 2.7774 & 3.5002 & 0.9740 & 0.9719 \\ \cline{2-7}
& SFIM & 3.8504 & 2.5429 & 3.0686 & 0.9780 & 0.9761 \\ \cline{2-7}
& LMVM & 4.3444 & 2.9574 & 3.3814 & 0.9719 & 0.9665 \\ \cline{2-7}
& ATWT & 4.2034 & 2.7813 & 3.7172 & 0.9730 & 0.9705 \\ \cline{2-7}
& GLP & 4.2608 & 2.8615 & 3.3472 & 0.9741 & 0.9713 \\ \cline{2-7}
& NLVD & {\bf 2.5034} & {\bf 1.6608} & {\bf 2.2657} & {\bf 0.9893} & {\bf 0.9857} \\ \hline
\end{tabular}
\caption{Numerical results for $\sigma=1.3$.}
 \label{table_60cm_s13_RGBNIR_nonregist_nolinear}
 \end{subtable}
\caption{Quantitative evaluation of the fused products on simulated data from 4-band (blue, green, red, and near-infrared) aerial images at resolution of 60 cm per pixel. The panchromatic images were computed as a weighted average of the reference spectral components with mixing coefficients $\alpha_B=0$, $\alpha_G=0.4$, $\alpha_R=0.35$, and $\alpha_I=0.25$. For these experiments, the low-resolution bands were non registered and the panchro-spectral constraint not fulfilled. In general terms, the same conclusions than in Table \ref{table_30cm_RGBNIR_nonregist_nolinear} can be drawn. Indeed, NLVD outperforms all techniques under comparison for any quality index.}
\label{table_60cm_RGBNIR_nonregist_nolinear}
\end{table}

\begin{figure}[!t]
\footnotesize
\centering
\renewcommand{\arraystretch}{0.5}
\begin{tabular}{c@{\hskip 0.02in}c@{\hskip 0.02in}c@{\hskip 0.02in}c}
  \includegraphics[trim= 15.5cm 3.5cm 2.5cm 14.5cm, clip=true, width=0.243\textwidth]{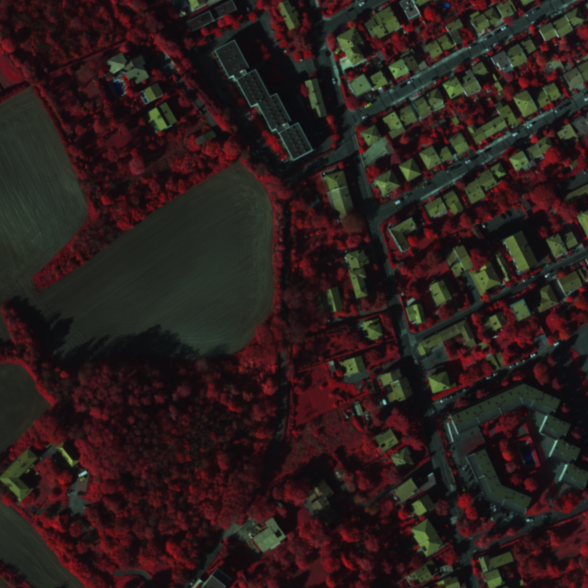} &
  \includegraphics[trim= 15.5cm 3.5cm 2.5cm 14.5cm, clip=true, width=0.243\textwidth]{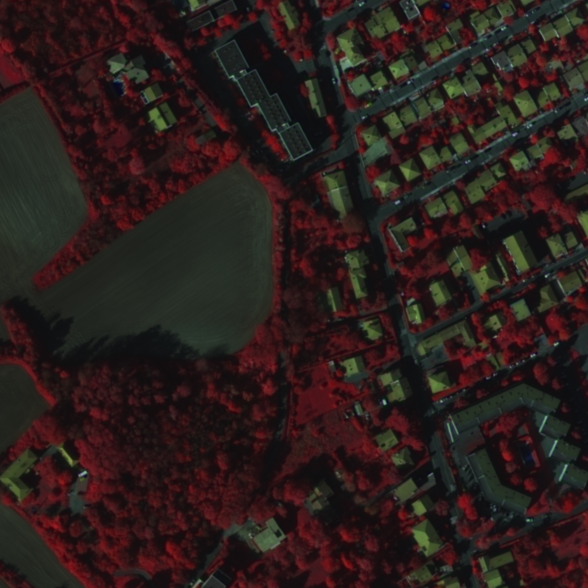} &
  \includegraphics[trim= 15.5cm 3.5cm 2.5cm 14.5cm 11cm, clip=true, width=0.243\textwidth]{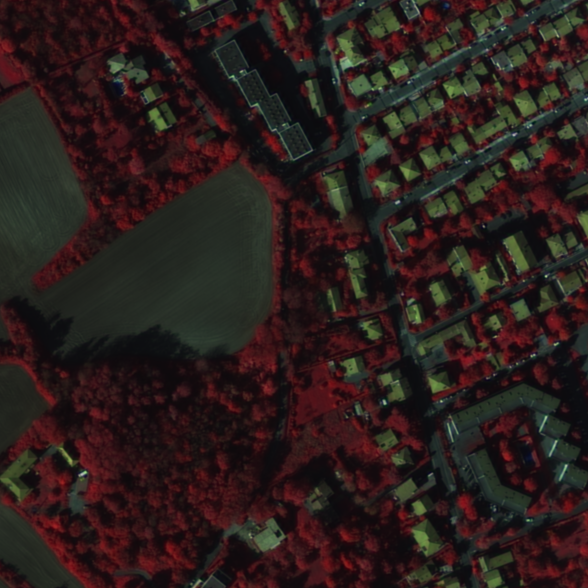} &
  \includegraphics[trim= 15.5cm 3.5cm 2.5cm 14.5cm 11cm, clip=true, width=0.243\textwidth]{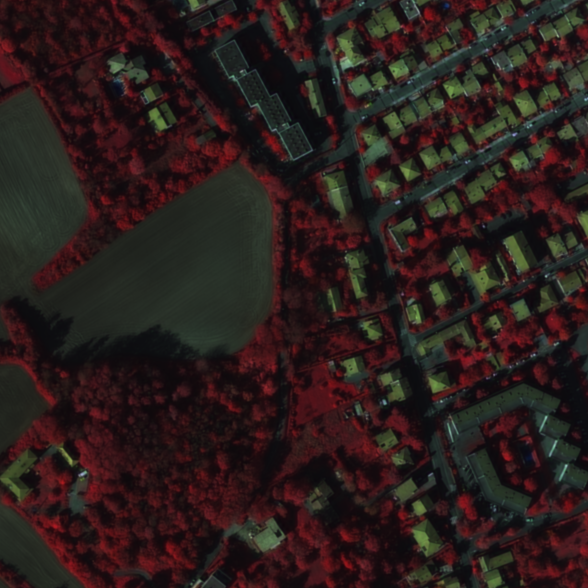} \\
  Reference & PCA & HPF & SFIM\\
  \includegraphics[trim= 15.5cm 3.5cm 2.5cm 14.5cm, clip=true, width=0.243\textwidth]{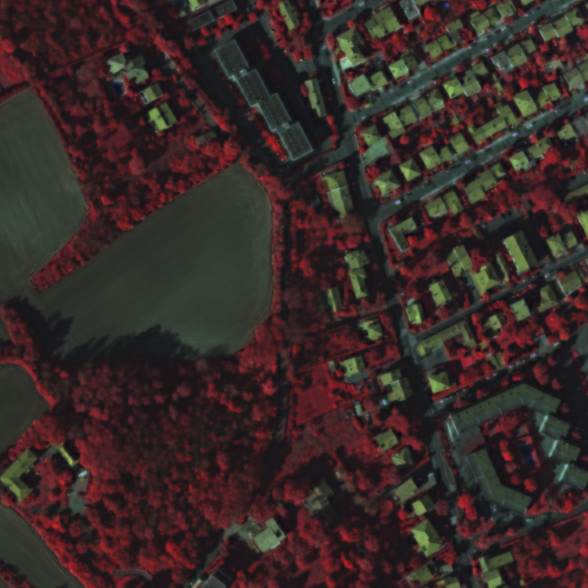} &
  \includegraphics[trim= 15.5cm 3.5cm 2.5cm 14.5cm, clip=true, width=0.243\textwidth]{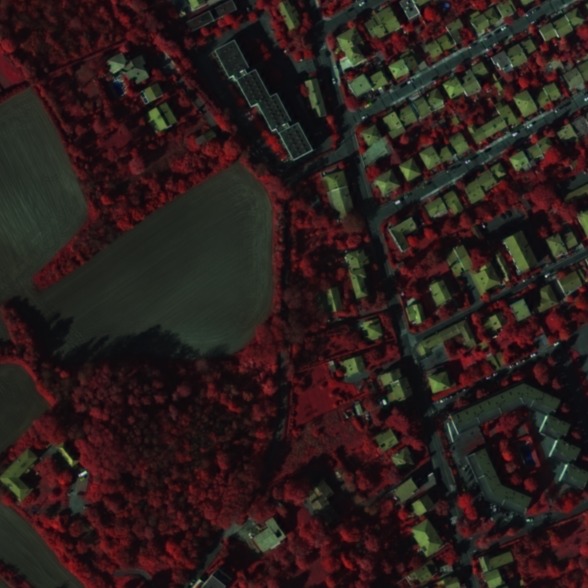} &
  \includegraphics[trim= 15.5cm 3.5cm 2.5cm 14.5cm, clip=true, width=0.243\textwidth]{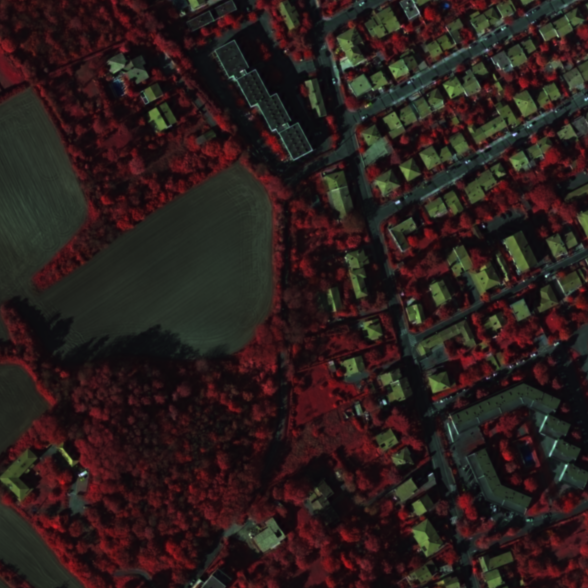} &
  \includegraphics[trim= 15.5cm 3.5cm 2.5cm 14.5cm, clip=true, width=0.243\textwidth]{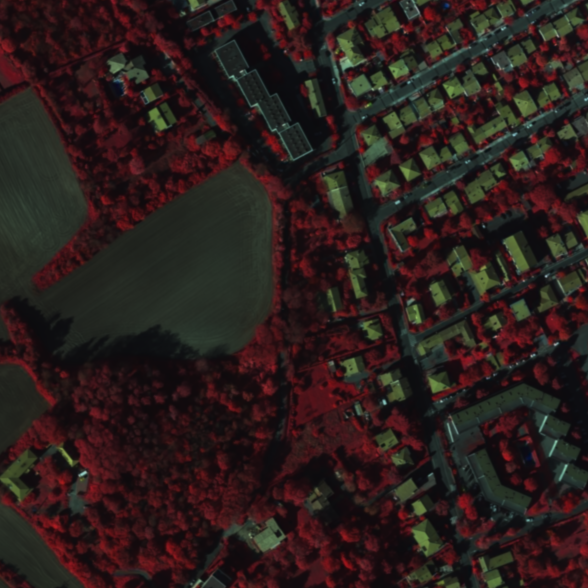} \\
  LMVM & ATWT & GLP & NLVD\\
\end{tabular}
\caption{Close-ups of the reference false-color image involving infra-red, red, and green bands in place of the usual RGB at a resolution of 60 cm per pixel and of the fusion products provided by all methods under comparison. The Gaussian standard deviation used for the simulation of the low-resolution spectral components was $\sigma=1.3$. For these experiments, the data were non registered and the panchro-spectral constraint not fulfilled. Obeserve the aliasing on the white roof at the upper left corner of all fused products except the one provided by NLVD. We conclude that the proposed model is the less affected by aliasing.}
\label{fig_60cm_s13_RGBNIR_noregist_nolinear}
\end{figure}

\begin{figure}[!h]
\footnotesize
\centering
\renewcommand{\arraystretch}{0.5}
\begin{tabular}{c@{\hskip 0.02in}c@{\hskip 0.02in}c@{\hskip 0.02in}c}
  \includegraphics[trim= 15.5cm 3.5cm 2.5cm 14.5cm, clip=true, width=0.243\textwidth]{RGBNIRnoregnolin_true.png} &
  \includegraphics[trim= 15.5cm 3.5cm 2.5cm 14.5cm, clip=true, width=0.243\textwidth]{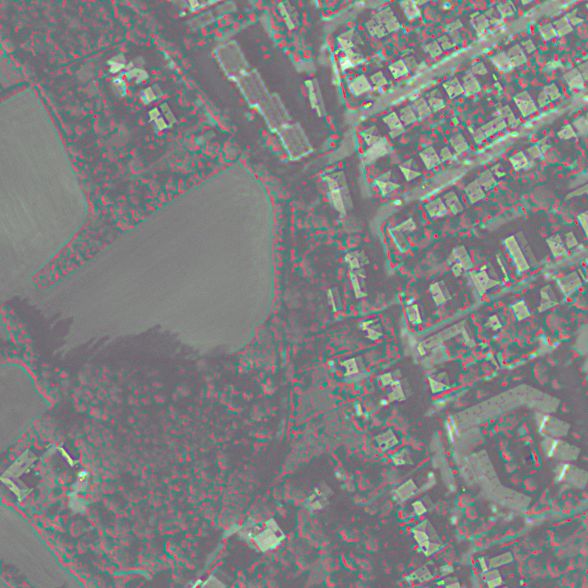} &
  \includegraphics[trim= 15.5cm 3.5cm 2.5cm 14.5cm 11cm, clip=true, width=0.243\textwidth]{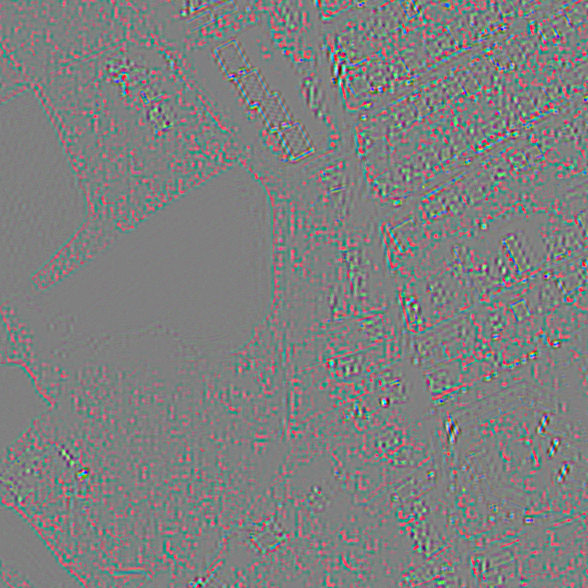} &
  \includegraphics[trim= 15.5cm 3.5cm 2.5cm 14.5cm 11cm, clip=true, width=0.243\textwidth]{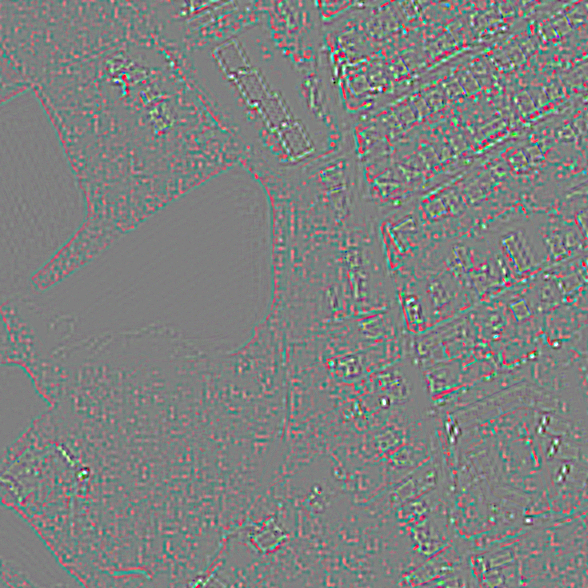} \\
  Reference & PCA & HPF & SFIM\\
  \includegraphics[trim= 15.5cm 3.5cm 2.5cm 14.5cm, clip=true, width=0.243\textwidth]{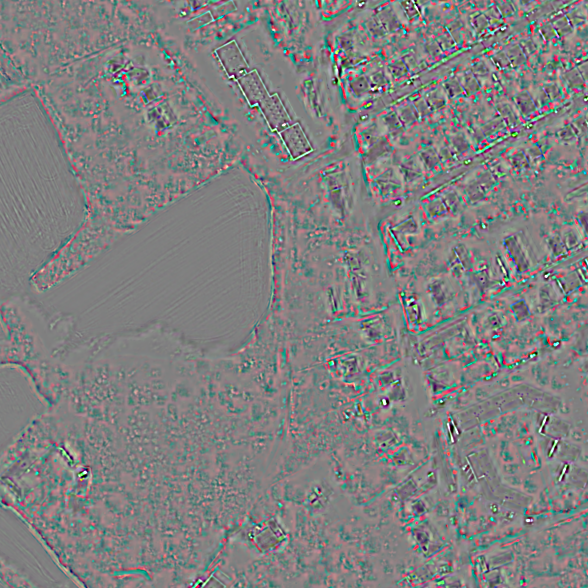} &
  \includegraphics[trim= 15.5cm 3.5cm 2.5cm 14.5cm, clip=true, width=0.243\textwidth]{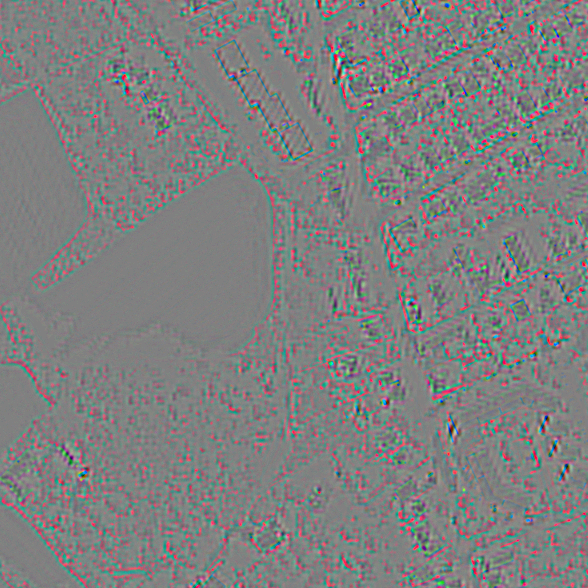} &
  \includegraphics[trim= 15.5cm 3.5cm 2.5cm 14.5cm, clip=true, width=0.243\textwidth]{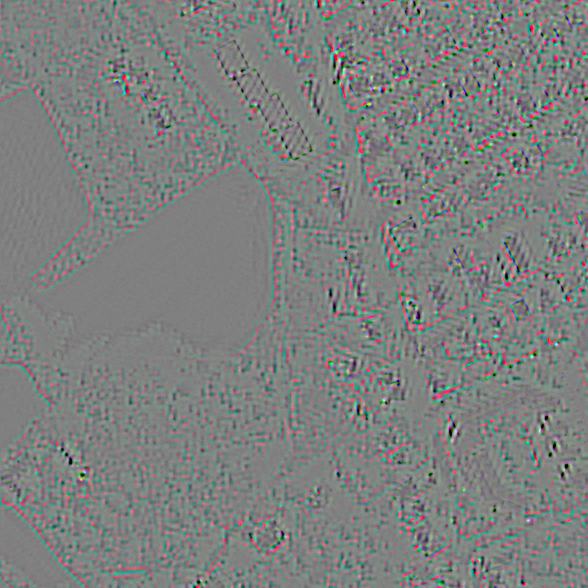} &
  \includegraphics[trim= 15.5cm 3.5cm 2.5cm 14.5cm, clip=true, width=0.243\textwidth]{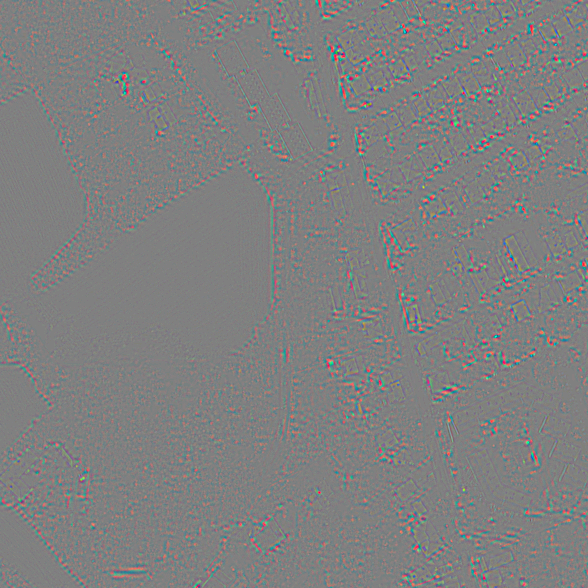} \\
  LMVM & ATWT & GLP & NLVD\\
\end{tabular}
\caption{Close-ups of the reference false-color image involving infra-red, red, and green bands in place of the usual RGB at a resolution of 60 cm per pixel and of the difference images associated to the fusion products displayed in Figure \ref{fig_60cm_s13_RGBNIR_noregist_nolinear}. For visualization purposes, the intensity values have been linearly rescaled from $[-20,20]$ to $[0,255]$. Since the difference image associated to NLVD contains less geometry and chromaticity than the others, one can conclude that the proposed model gives rise to results with the best spectral and spatial quality.}
\label{fig_60cm_s13_RGBNIR_noregist_nolinear_dif}
\end{figure}

Figure \ref{fig_60cm_s13_RGBNIR_noregist_nolinear} displays close-ups of the false-color images associated to the ground truth and to the pansharpened products provided by each method on the fifth picture of the proposed dataset (Figure \ref{fig:dataset}) at a resolution of 60 cm per pixel. The Gaussian standard deviation used for the simulation of the low-resolution spectral bands was $\sigma=1.3$. We also show in Figure \ref{fig_60cm_s13_RGBNIR_noregist_nolinear_dif} the corresponding difference images. In general, all results except that obtained from NLVD suffer from strong aliasing, as can be observed on the white roof at the upper left corner of the pictures in Figure \ref{fig_60cm_s13_RGBNIR_noregist_nolinear}. This phenomenon can be assessed at a glance in Figure \ref{fig_60cm_s13_RGBNIR_noregist_nolinear_dif}. It is worth noticing that PCA leads to a loss of spectral quality, whereas more spatial details are damaged with MRA-based methods. In particular, observe the jagged edges of the fused products by HPF, SFIM, LMVM, ATWT, and GLP. Finally, NLVD provides the best pleasant visual result in Figure \ref{fig_60cm_s13_RGBNIR_noregist_nolinear} and the fact that the associated difference image in Figure \ref{fig_60cm_s13_RGBNIR_noregist_nolinear_dif} contains less amount of information means that the proposed model provides the best spatial and spectral quality among methods under comparison.

\subsection{Application to Pl{\'e}iades Imagery}

We finally tested the performance of the proposed NLVD model for pansharpening Pl{\'e}iades imagery. Let us recall that Pl{\'e}iades produces a panchromatic image at spatial resolution of 70 cm per pixel and four spectral bands (blue, green, red, and near-infrared) at resolution of 2.8 m per pixel. The MTF for the panchromatic has a value of 0.15 at cut frequency (low aliasing), while this value is greater than 0.26 (strong aliasing) for the spectral components. The panchromatic being slightly aliased can be resampled into the reference of any spectral band, thus permitting the fusion of each band separately as proposed for Pl{\'e}iades images by Latry {\it et al.} \cite{LatryBlanchet2013}.

We compare NLVD with the methods listed in the previous subsection that apply on misregistered spectral data and do not use the panchro-spectral constraint, which are HPF, SFIM, LMVM, ATWT, and GLP. We further incorporate the results of the fusion technique proposed by Latry {\it et al.} \cite{LatryBlanchet2013}, which we call LBF and consists in applying the relation \eqref{eq:MRAterm1}. Note that LBF is closely related to the local mean matching filtering described by De B{\'e}thune {\it et al.} \cite{BethuneMuller1998}.  Since all these methods are band-decoupled, we warped the panchromatic into the reference of each spectral component using the transformations furnished to us by CNES and solved there the fusion problem. Each high-resolution band obtained from the minimization was finally transformed into a common reference, where all channels can be super-imposed for visualization purposes. In this scenario, we cannot estimate the optimal parameters of our model in terms of the lowest error since reference images are no more available. Consequently, and in order to be fair to all techniques under comparison, we used the values estimated previously for simulated data. Furthermore, the standard deviation used in the spectral preserving term as well in the computation of the low-resolution panchromatic images has to be estimated according to the amount of aliasing in the spectral data. Since this is unknown, we fixed it visually to $\sigma=1.3$. The data to which each pansharpening method applies was given in Figure \ref{fig:datasetPleiades}.
 
Table \ref{table_pleiades} displays the indices that measure how much spatial and spectral information each method is able to preserve from the panchromatic and low-resolution bands, respectively. We realize that LMVM, LBF, and NLVD outperform all other techniques, particularly in terms of avoiding the loss of spectral quality quantified by $D_{\lambda}$. For any of the quality assessment indices, the model proposed in this paper gives rise to the best values, although being closely followed by LBF.

\begin{table}[!t]
\footnotesize
\centering
\begin{tabular}{|c|c|c|c|}
\hline
Methods & $D_{\lambda}$ & $D_S$ & QNR  \\ \hline \hline
Reference & 0 & 0 & 1 \\ \hline \hline
HPF & 0.3480 & 0.0720 & 0.6051 \\ \hline
SFIM & 0.3475 & 0.0661 & 0.6094 \\ \hline 
LMVM & 0.0892 & 0.0991 & 0.8205 \\ \hline 
ATWT & 0.3534 & 0.0978 & 0.5833 \\ \hline 
GLP & 0.3537 & 0.0943  & 0.5853 \\ \hline 
LBF & 0.0363 & 0.0686 & 0.8976 \\ \hline
NLVD & {\bf 0.0305} & {\bf 0.0685} & {\bf 0.9031} \\ \hline 
\end{tabular}
\caption{Quantitative evaluation of the pansharpened images obtained from the Pl{\'e}iades products displayed in Figure \ref{fig:datasetPleiades}. All quality indices strengthen the superiority of NLVD in preserving the spatial and spectral information from the provided data, although LBF gives numerical results close to ours.}
 \label{table_pleiades}
\end{table}

\begin{figure}[!t]
\footnotesize
\centering
\renewcommand{\arraystretch}{0.5}
\begin{tabular}{c@{\hskip 0.03in}c@{\hskip 0.03in}c@{\hskip 0.03in}c}
  \includegraphics[trim= 7cm 0cm 19cm 26cm, clip=true, width=0.243\textwidth]{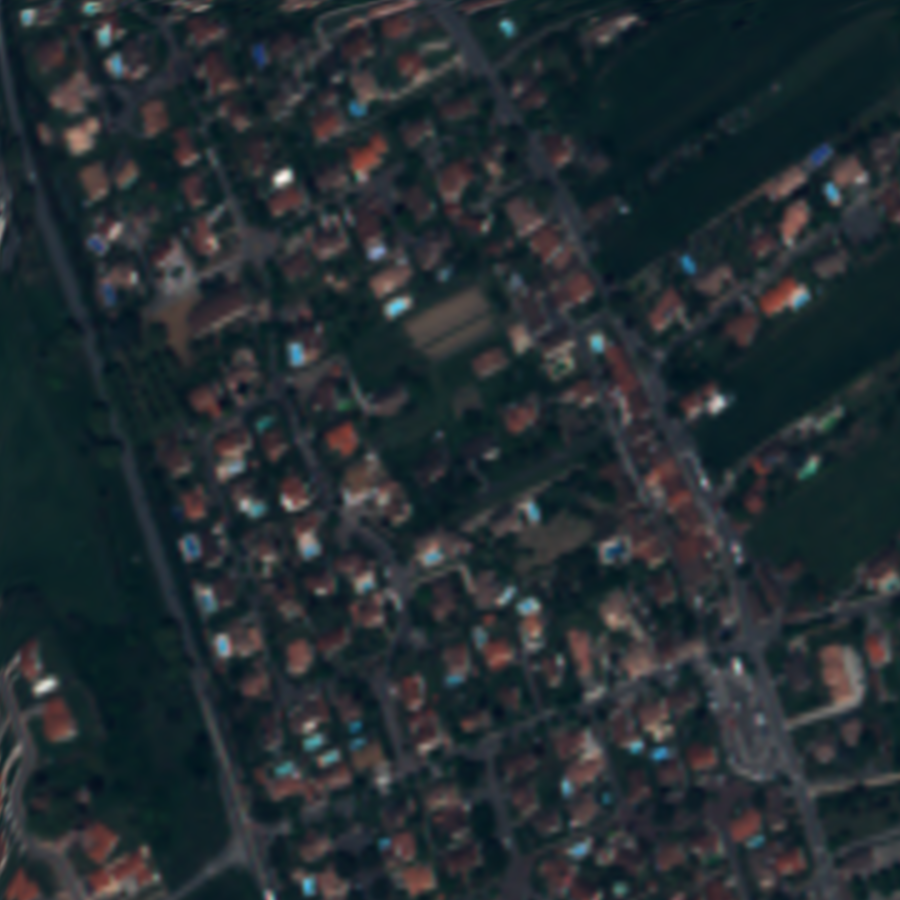} & 
  \includegraphics[trim= 7cm 0cm 19cm 26cm, clip=true, width=0.243\textwidth]{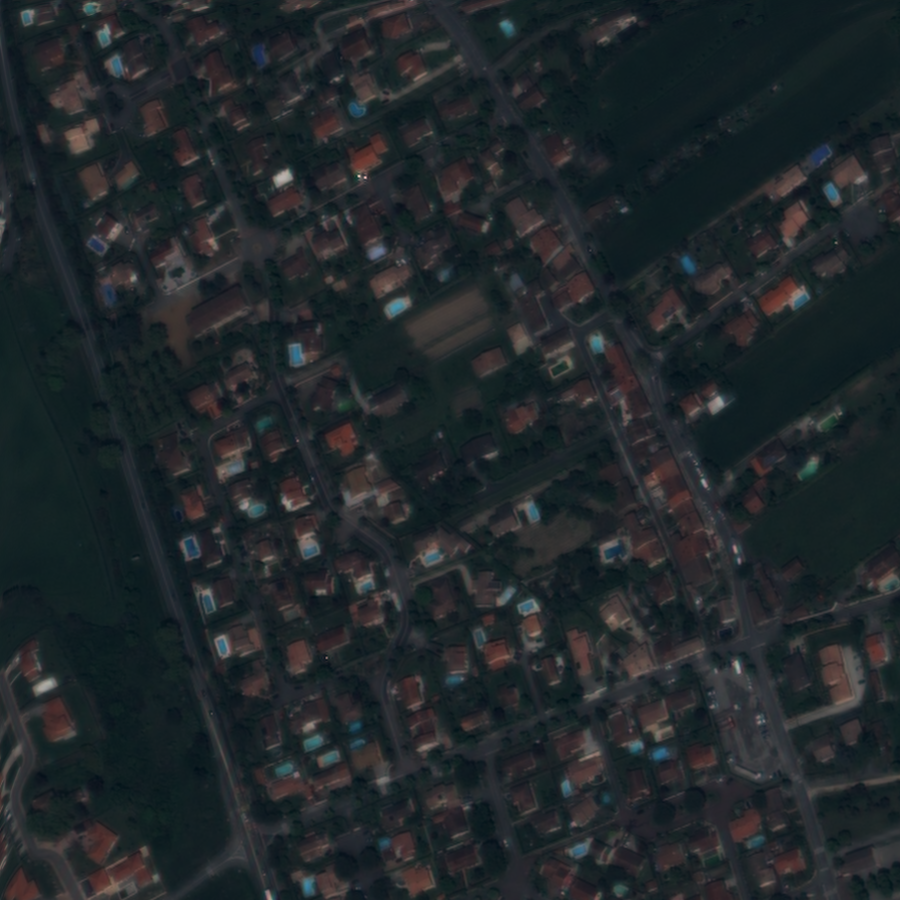} &
  \includegraphics[trim= 7cm 0cm 19cm 26cm, clip=true, width=0.243\textwidth]{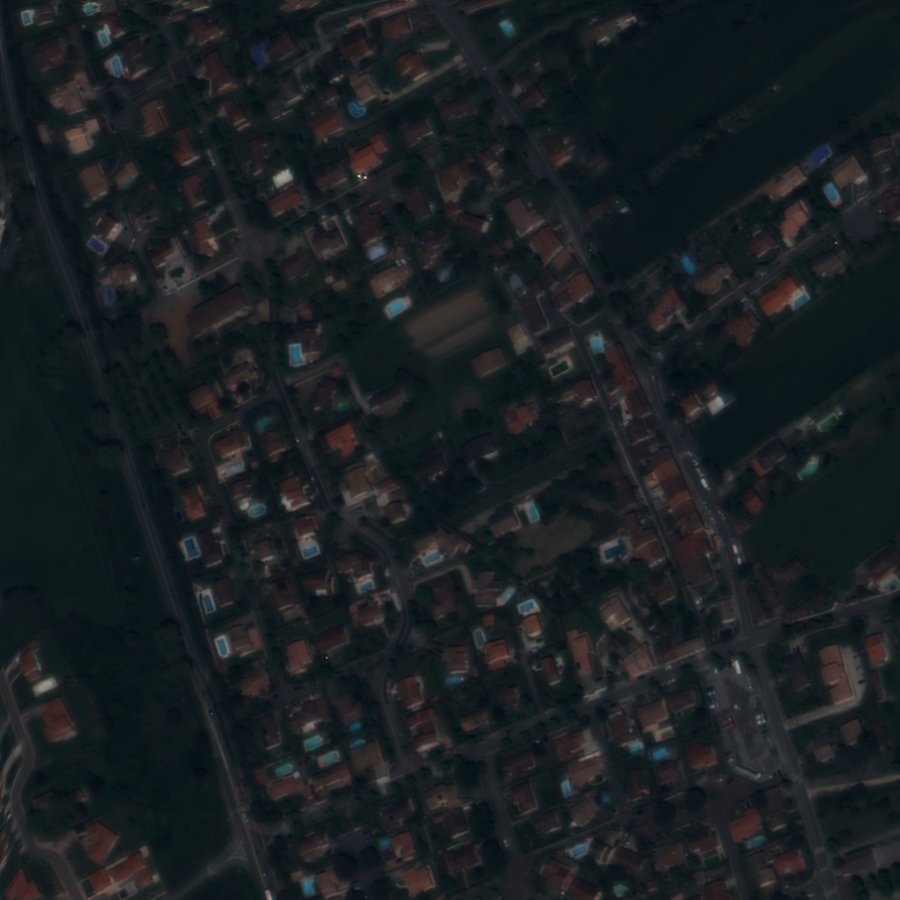} &
  \includegraphics[trim= 7cm 0cm 19cm 26cm, clip=true, width=0.243\textwidth]{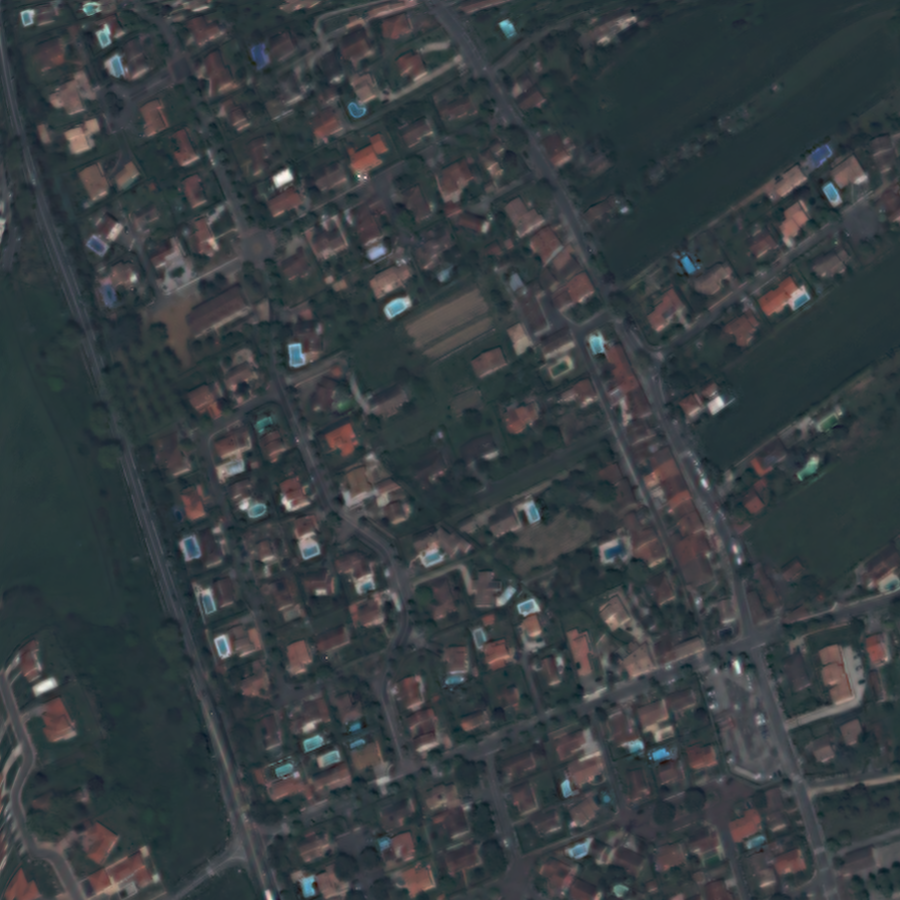} \\
  Interpolated & HPF & SFIM & LMVM \\
  \includegraphics[trim= 7cm 0cm 19cm 26cm, clip=true, width=0.243\textwidth]{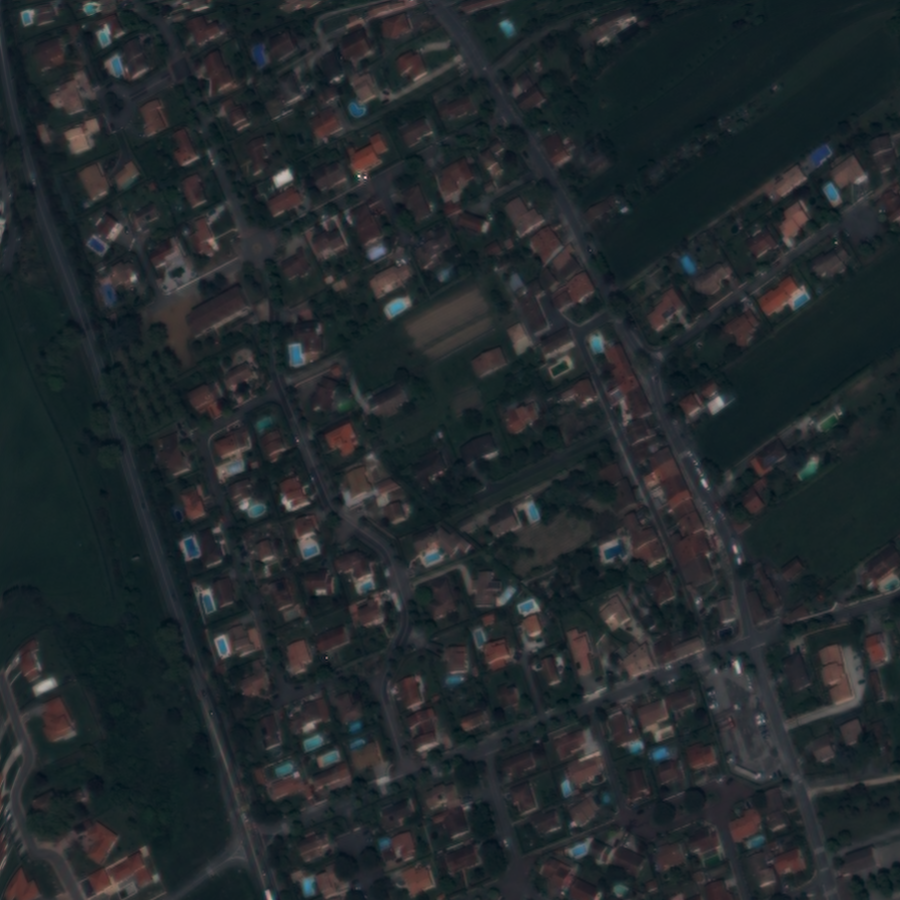} &
  \includegraphics[trim= 7cm 0cm 19cm 26cm, clip=true, width=0.243\textwidth]{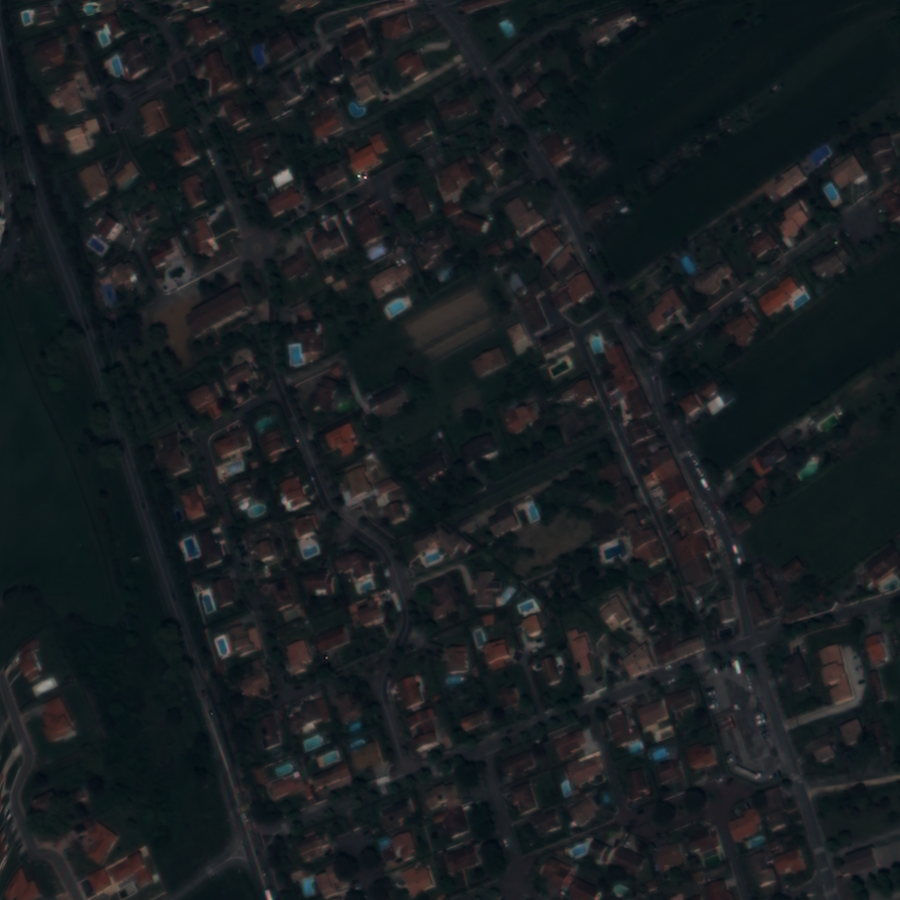} &
  \includegraphics[trim= 7cm 0cm 19cm 26cm, clip=true, width=0.243\textwidth]{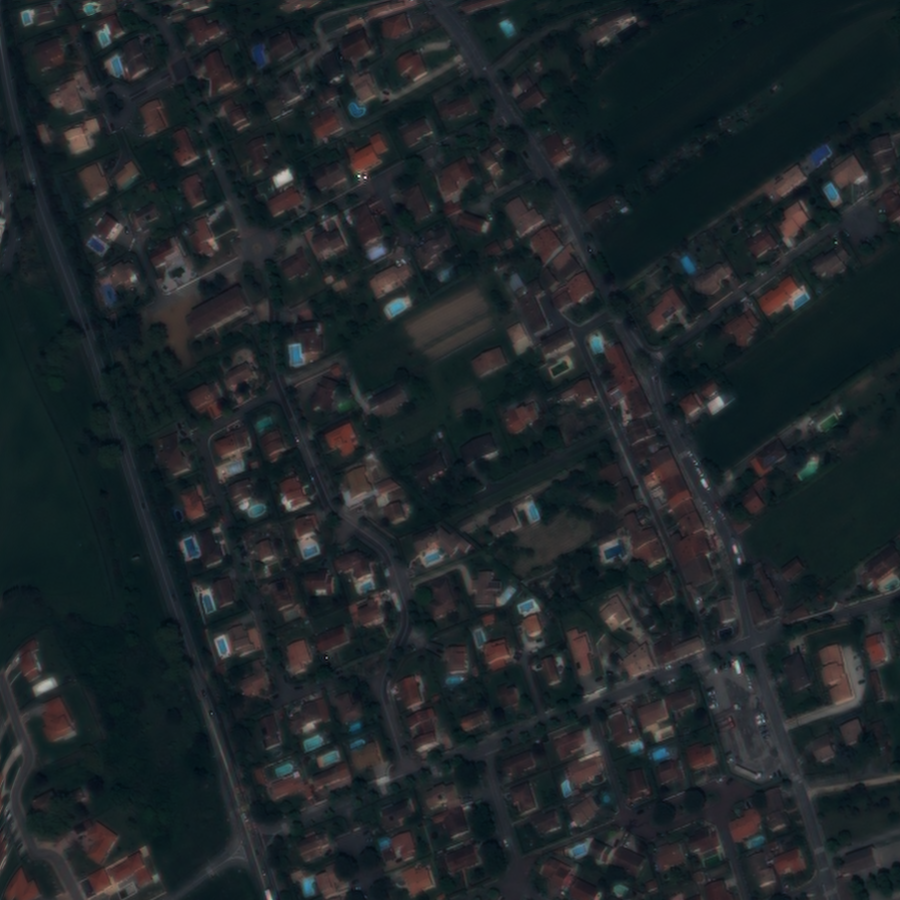} &
  \includegraphics[trim= 7cm 0cm 19cm 26cm, clip=true, width=0.243\textwidth]{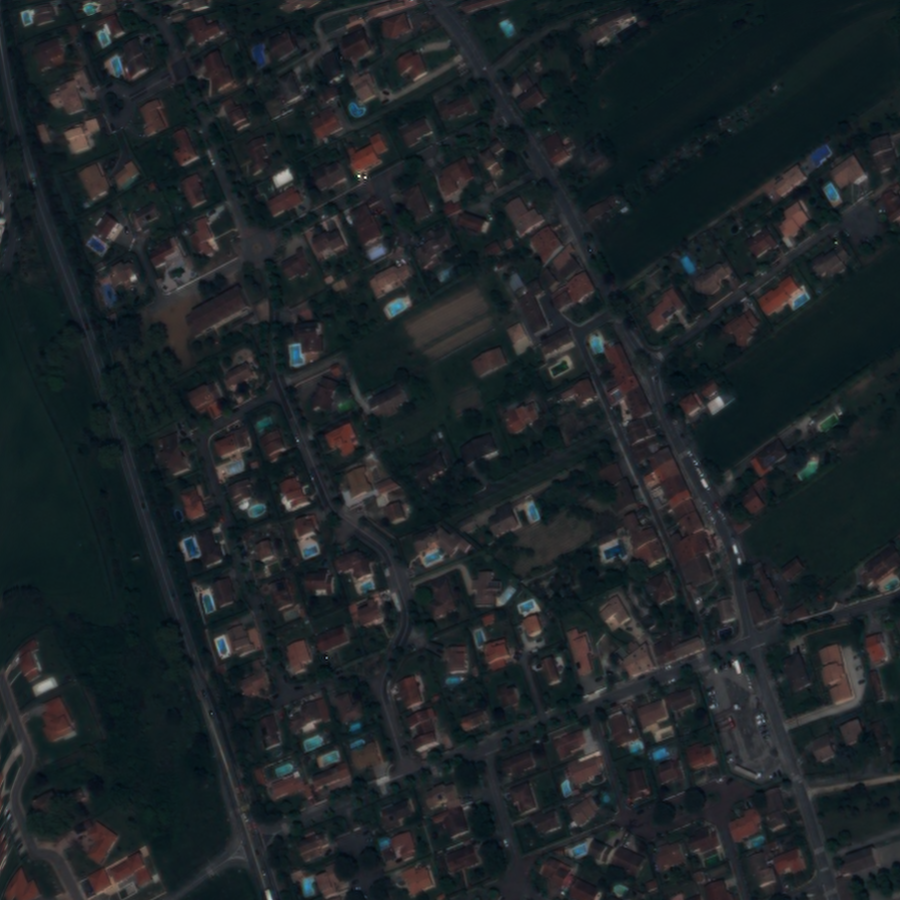} \\
  ATWT & GLP & LBF & NLVD
\end{tabular}
\caption{Close-ups of the results provided by each method on the Pl{\'e}iades data displayed in Figure \ref{fig:datasetPleiades}. All results except ours contain strong aliasing, which mainly concentrates throughout the main road. NLVD considerably suppresses the color artifacts due to aliasing, especially in the saturated areas such as the white truck at the bottom of the picture.}
\label{fig:pleiades1}
\end{figure}

\begin{figure}[!t]
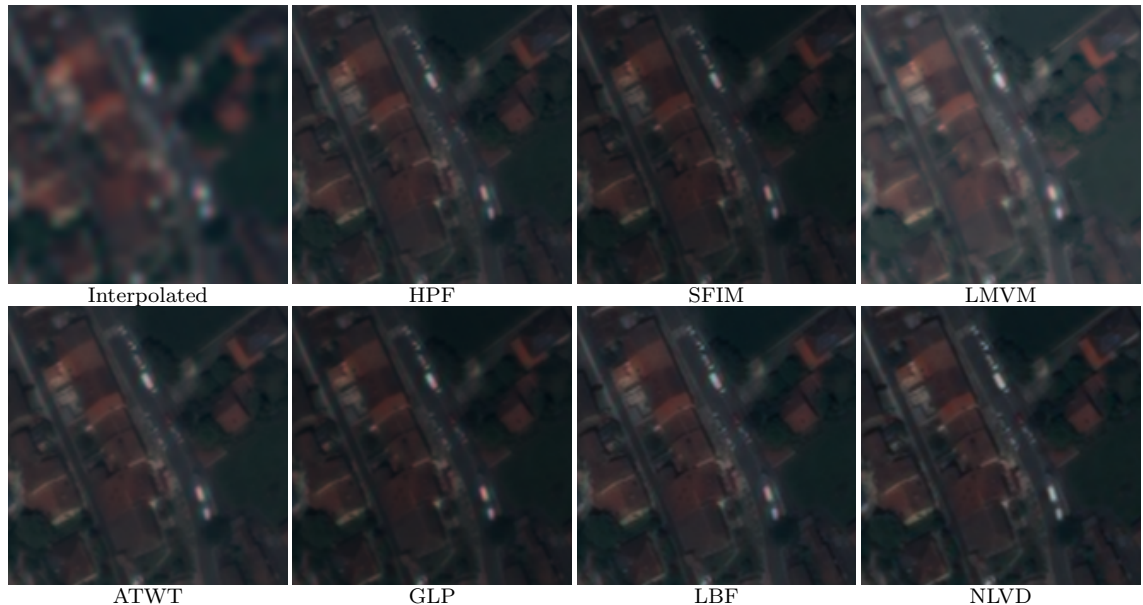

\footnotesize
\centering
\renewcommand{\arraystretch}{0.5}
\begin{tabular}{c@{\hskip 0.03in}c@{\hskip 0.03in}c@{\hskip 0.03in}c}
  \includegraphics[trim= 22cm 10.5cm 4cm 15.5cm, clip=true, width=0.243\textwidth]{pleiades_int.png} & 
  \includegraphics[trim= 22cm 10.5cm 4cm 15.5cm, clip=true, width=0.243\textwidth]{pleiades_HPF.png} &
  \includegraphics[trim= 22cm 10.5cm 4cm 15.5cm, clip=true, width=0.243\textwidth]{pleiades_SFIM.png} &
  \includegraphics[trim= 22cm 10.5cm 4cm 15.5cm, clip=true, width=0.243\textwidth]{pleiades_LMVM.png} \\
  Interpolated & HPF & SFIM & LMVM \\
  \includegraphics[trim= 22cm 10.5cm 4cm 15.5cm, clip=true, width=0.243\textwidth]{pleiades_ATWT.png} &
  \includegraphics[trim= 22cm 10.5cm 4cm 15.5cm, clip=true, width=0.243\textwidth]{pleiades_GLP.png} &
  \includegraphics[trim= 22cm 10.5cm 4cm 15.5cm, clip=true, width=0.243\textwidth]{pleiades_LBF.png} &
  \includegraphics[trim= 22cm 10.5cm 4cm 15.5cm, clip=true, width=0.243\textwidth]{pleiades_NLVD.png} \\
  ATWT & GLP & LBF & NLVD
\end{tabular}
\caption{Close-ups of the fused products for the Pl{\'e}iades imagery displayed in Figure \ref{fig:datasetPleiades}. Just as in the case of Figure \ref{fig:pleiades1}, any of the methods except ours is able to suppress convincingly the strong aliasing from the input low-resolution data. Indeed, see the color artifacts surrounding the white cars in the scene. The new-proposed variational model is able to eliminate practically the effects of the aliasing and leads to a final product with better spectral and spatial resolution.}
\label{fig:pleiades2}
\end{figure}

In order to check the validity of the quantitative measures, Figures \ref{fig:pleiades1} and \ref{fig:pleiades2} show some close-ups on the results obtained by each method. For visualization purposes, we display the red, green, and blue channels as color images. In both cases, the proposed variational model better incorporates the high frequencies of the panchromatic into the inferred high-resolution spectral components. Although the radiometric constraint \eqref{eq:MRAterm}, variants of which are used by almost all other techniques under comparison, takes care of providing a result with high-spatial resolution, the nonlocal regularization term we used in \eqref{eq:functional2} computes the weight distribution on the panchromatic and, thus, helps to transfer the geometry to the fused product. It is also important to note that the inherent false frequency alias of the low-resolution spectral bands prevails in the pansharpened images provided by all methods except ours. Indeed, observe the aliasing that concentrates throughout the main road in Figure \ref{fig:pleiades1} as well as several color distortions that appear in Figure \ref{fig:pleiades2} surrounding saturated objects like white cars. The proposed NLVD method is able to noticeably reduce these artifacts, although some aliasing still remains suggesting that better parameters in the energy minimization could have been used. Since most of the compared techniques apply \eqref{eq:MRAterm1}, the visual quality assessment points out that this constraint is not enough to overcome the drawbacks because of aliasing by itself. Consequently, the nonlocal regularization term and the data-fidelity term have a positive influence in avoiding the creation of false frequencies.

\section{Conclusions}\label{sec:conclusions}

We have introduced a new band-decoupled variational method  for pansharpening. This new method does not need the initial spectral data to be co-registered and suppresses any assumption on the linear dependence of the spectral and panchromatic modalities.  This makes the algorithm suitable for application on real satellite imagery.

We have showed that, in general, it is better to deal with the original data. The presence of aliasing in the spectral bands make not recommendable to re-interpolate them for co-registration. Most methods that can be applied with registered and non-registered spectral components perform better when dealing with the non re-interpolated components. 

The proposed method applied independently to each band performs the best on simulated data compared to classical and state-of-the-art techniques. The difference in performance is smaller if we only have access to the already co-registered data, but still it is performing the best. Finally, we have also applied the proposed algorithm to real satellite data acquired by Pl{\'e}iades and furnished to us by CNES, illustrating the good performance of the method in terms of spatial and spectral quality and its ability to suppress the aliasing artifacts which are intrinsic to the low-resolution spectral bands.

\bibliographystyle{siam}
\bibliography{references}

\begin{thebibliography}{10}

\bibitem{AiazziAlparone2002}
{\sc B.~Aiazzi, L.~Alparone, S.~Baronti, and A.~Garzelli}, {\em Context-driven
  fusion of high spatial and spectral resolution images based on oversampled
  multiresolution analysis}, IEEE Trans. Geosci. Remote Sens., 40 (2002),
  pp.~2300--2312.

\bibitem{AiazziAlparoneBaronti2006}
{\sc B.~Aiazzi, L.~Alparone, S.~Baronti, A.~Garzelli, and M.~Selva}, {\em
  {{MTF}}-tailored multiscale fusion of high-resolution {{MS}} and {{Pan}}
  imagery}, Photogramm. Eng. Remote Sens., 72 (2006), pp.~591--596.

\bibitem{AiazziBarontiSelva2007}
{\sc B.~Aiazzi, S.~Baronti, and M.~Selva}, {\em Improving component
  substitution pansharpening through multivariate regression of {{MS+Pan}}
  data}, IEEE Trans. Geosci. Remote Sens., 45 (2007), pp.~3230--3239.

\bibitem{Alparone2008}
{\sc L.~Alparone}, {\em Multispectral and panchromatic data fusion assessment
  without reference}, Photogramm. Eng. Remote Sens., 74 (2008), pp.~193--200.

\bibitem{AlparoneBaronti2004}
{\sc L.~Alparone, S.~Baronti, A.~Garzelli, and F.~Nencini}, {\em A global
  quality measurement of pan-sharpened multispectral imagery}, IEEE Geosci.
  Remote Sens. Lett., 1 (2004), pp.~313--317.

\bibitem{AlySharma2014}
{\sc H.~Aly and G.~Sharma}, {\em A regularized model-based optimization
  framework for pan-sharpening}, IEEE Trans. Image Process., 23 (2014),
  pp.~2596--2608.

\bibitem{AmroMateosVega2011}
{\sc I.~Amro, J.~Mateos, M.~Vega, R.~Molina, and A.~Katsaggelos}, {\em A survey
  of classical methods and new trends in pansharpening of multispectral
  images}, EURASIP J. Adv. Sig. Process., 2011 (2011), pp.~1--22.

\bibitem{arias2011variational}
{\sc P.~Arias, G.~Facciolo, V.~Caselles, and G.~Sapiro}, {\em A variational
  framework for exemplar-based image inpainting}, International journal of
  computer vision, 93 (2011), pp.~319--347.

\bibitem{BallesterCaselles2006}
{\sc C.~Ballester, V.~Caselles, L.~Igual, J.~Verdera, and B.~Roug{\'e}}, {\em A
  variational model for {{P+XS}} image fusion}, Int. J. Comput. Vis., 69
  (2006), pp.~43--58.

\bibitem{BarontiAiazzi2011}
{\sc S.~Baronti, B.~Aiazzi, M.~Selva, and A.~Garzelli}, {\em A theoretical
  analysis of the effects of aliasing and misregistration on pansharpened
  imagery}, IEEE J. Sel. Topics Signal Process., 5 (2011), pp.~446--453.

\bibitem{BuadesCollLisani2007}
{\sc A.~Buades, B.~Coll, J.-L. Lisani, and C.~Sbert}, {\em Conditional image
  diffusion}, Inverse Probl. Imag., 1 (2007), pp.~593--608.

\bibitem{BuadesCollMorell2011}
{\sc A.~Buades, B.~Coll, and J.-M. Morel}, {\em Self-similarity-based image
  denoising}, Comm. ACM, 54 (2011), pp.~109--117.

\bibitem{CarperLillesand1990}
{\sc J.~Carper, T.~Lillesand, and R.~Kiefer}, {\em The use of
  intensity-hue-saturation transformations for merging {{SPOT}} panchromatic
  and multispectral image data}, Photogramm. Eng. Remote Sens., 56 (1990),
  pp.~457--467.

\bibitem{ChavezKwarteng1989}
{\sc P.~Chavez and A.~Kwarteng}, {\em Extracting spectral contrast in
  {{Landsat}} thematic mapper image data using selective principal component
  analysis}, Photogramm. Eng. Remote Sens., 55 (1989), pp.~339--348.

\bibitem{ChavezSides1991}
{\sc P.~Chavez, S.~Sides, and J.~Anderson}, {\em Comparison of three different
  methods to merge multiresolution and multispectral data: {{Landsat TM}} and
  {{SPORT}} panchromatic}, Photogramm. Eng. Remote Sens., 57 (1991),
  pp.~295--303.

\bibitem{ChoiYuKim2011}
{\sc J.~Choi, K.~Yu, and Y.~Kim}, {\em A new adaptive component-substitution
  based satellite image fusion by using partial replacement}, IEEE Trans.
  Geosci. Remote Sens., 49 (2011), pp.~295--309.

\bibitem{Dacorogna2008}
{\sc B.~Dacorogna}, {\em Direct Methods in the Calculus of Variations}, vol.~78
  of Applied Mathematical Sciences, Springer-Verlag New York, second~ed., 2008.

\bibitem{BethuneMuller1998}
{\sc S.~De~B{\'e}thune, F.~Muller, and J.-P. Donnay}, {\em Fusion of
  multispectral and panchromatic images by local mean and variance matching
  filtering techniques}, in Proc. Fusion of Earth Data, Nice, France, 1998,
  pp.~31--37.

\bibitem{DuranBuadesTIP2014}
{\sc J.~Duran and A.~Buades}, {\em Self-similarity and spectral correlation
  adaptive algorithm for color demosaicking}, IEEE Trans. Image Process., 23
  (2014), pp.~4031--4040.

\bibitem{DuranBuadesCollSbertSIIMS2014}
{\sc J.~Duran, A.~Buades, B.~Coll, and C.~Sbert}, {\em A nonlocal variational
  model for pansharpening image fusion}, SIAM J. Imaging Sci., 7 (2014),
  pp.~761--796.

\bibitem{EkelandTemam1999}
{\sc I.~Ekeland and R.~Temam}, {\em Convex Analysis and Variational Problems},
  vol.~28 of Classics in Applied Mathematics, Society for Industrial and
  Applied Mathematics, 1999.

\bibitem{EladDatsenko2009}
{\sc M.~Elad and D.~Datsenko}, {\em Example-based regularization deployed to
  super-resolution reconstruction of a single image}, Computer J., 52 (2009),
  pp.~15--30.

\bibitem{GarzelliNencini2009}
{\sc A.~Garzelli and F.~Nencini}, {\em Hypercomplex quality assessment of
  multi/hyperspectral images}, IEEE Geosci. Remote Sens. Lett., 6 (2009),
  pp.~662--665.

\bibitem{GarzelliNenciniCapobianco2008}
{\sc A.~Garzelli, F.~Nencini, and L.~Capobianco}, {\em Optimal {{MMSE}} pan
  sharpening of very high resolution multispectral images}, IEEE Trans. Geosci.
  Remote Sens., 46 (2008), pp.~228--236.

\bibitem{GilboaOsher2007}
{\sc G.~Gilboa and S.~Osher}, {\em Nonlocal linear image regularization and
  supervised segmentation}, SIAM Multiscale Model. Simul., 6 (2007),
  pp.~595--630.

\bibitem{GilboaOsher2008}
\leavevmode\vrule height 2pt depth -1.6pt width 23pt, {\em Nonlocal operators
  with applications to image processing}, Multiscale Model. Simmul., 7 (2008),
  pp.~1005--1028.

\bibitem{GillespieKahle1987}
{\sc A.~Gillespie, A.~Kahle, and R.~Walker}, {\em Color enhancement of highly
  correlated images {II}. {{Channel}} ratio and ``chromacity'' transform
  techniques}, Remote Sens. Environ., 22 (1987), pp.~343--365.

\bibitem{HalladaCox1983}
{\sc W.~Hallada and S.~Cox}, {\em Image sharpening for mixed spatial and
  spectral resolution satellite systems}, in Proc. Int. Symp. Remote Sensing of
  Environment, Ann Arbor, MI, USA, 1983, pp.~1023--1032.

\bibitem{HeCondatBioucas2014}
{\sc X.~He, L.~Condat, J.~Bioucas-Dias, J.~Chanussot, and J.~Xia}, {\em A new
  pansharpening method based on spatial and spectral sparsity priors}, IEEE
  Trans. Image Process., 23 (2014), pp.~4160--4174.

\bibitem{HeCondatChanussot2012}
{\sc X.~He, L.~Condat, and J.~Chanussot}, {\em Pansharpening using total
  variation regularization}, in Proc. IEEE Int. Conf. Geoscience and Remote
  Sensing Symposium (IGARSS), Munich, Germany, 2012, pp.~166--169.

\bibitem{KhanAlparone2009}
{\sc M.~Khan, L.~Alparone, and J.~Chanussot}, {\em Pansharpening quality
  assessment using the modulation transfer functions of instruments}, IEEE
  Trans. Geosci. Remote Sens., 47 (2009), pp.~3880--3891.

\bibitem{KindermannOsher2005}
{\sc S.~Kindermann, S.~Osher, and P.~Jones}, {\em Deblurring and denoising of
  images by nonlocal functionals}, Multiscale Model. Simmul., 4 (2005),
  pp.~1091--1115.

\bibitem{LabenBrower2000}
{\sc C.~Laben and B.~Brower}, {\em Process for enhancing the spatial resolution
  of multispectral imagery using pan-sharpening}.
\newblock U.S. Patent 6011875, 2000.

\bibitem{LatryBlanchet2013}
{\sc C.~Latry, G.~Blanchet, and S.~Fourest}, {\em Chaine de fusion {{P+XS}} des
  images {{Pl{\'e}iades-HR}}}, in Proc. Colloquium Groupe d'{\'E}tudes du
  Traitement du Signal et des Images (GRETSI), Brest, France, 2013.

\bibitem{LeeLee2010}
{\sc J.~Lee and C.~Lee}, {\em Fast and efficient panchromatic sharpening}, IEEE
  Trans. Geosci. Remote Sens., 48 (2010), pp.~155--163.

\bibitem{LiYang2011}
{\sc S.~Li and B.~Yang}, {\em A new pan-sharpening method using a compressed
  sensing technique}, IEEE Trans. Geosci. Remote Sens., 49 (2011).

\bibitem{Liu2000}
{\sc J.~Liu}, {\em Smoothing filter-based intensity modulation: {{A}} spectral
  preserve image fusion technique for improving spatial details}, Int. J.
  Remote Sens., 21 (2000), pp.~3461--3472.

\bibitem{Mallat1989}
{\sc S.~Mallat}, {\em A theory for multiresolution signal decomposition:
  {{The}} wavelet representation}, IEEE Trans. Pattern Anal. Mach. Intell., 11
  (1989), pp.~674--693.

\bibitem{MoellerWittman2012}
{\sc M.~M{\"o}ller, T.~Wittman, A.~Bertozzi, and M.~Burger}, {\em A variational
  approach for sharpening high dimensional images}, SIAM J. Imaging Sci., 5
  (2012), pp.~150--178.

\bibitem{NenciniGarzelli2007}
{\sc F.~Nencini, A.~Garzelli, S.~Baronti, and L.~Alparone}, {\em Remote sensing
  image fusion using the curvelet transform}, Inf. Fusion, 8 (2007),
  pp.~143--156.

\bibitem{NunezOtazuFors1999}
{\sc J.~Nu{\~n}ez, X.~Otazu, O.~Fors, A.~Prades, V.~Pala, and R.~Arbiol}, {\em
  Multiresolution-based image fusion with additive wavelet decomposition}, IEEE
  Trans. Geosci. Remote Sens., 37 (1999), pp.~1204--1211.

\bibitem{OtazuGonzalezFors2005}
{\sc X.~Otazu, M.~Gonz{\'a}lez-Aud{\'\i}cana, O.~Fors, and J.~Nu{\~n}ez}, {\em
  Introduction of sensor spectral response into image fusion methods.
  application to wavelet-based methods}, IEEE Trans. Geosci. Remote Sens., 43
  (2005), pp.~2376--2385.

\bibitem{PalssonSveinsson2012}
{\sc F.~Palsson, J.~Sveinsson, M.~Ulfarsson, and J.~Benediktsson}, {\em A new
  pansharpening method using an explicit formation model regularized via total
  variation}, in Proc. IEEE Int. Conf. Geoscience and Remote Sensing Symposium
  (IGARSS), Munich, Germany, 2012, pp.~2288--2291.

\bibitem{PohlGenderen2015}
{\sc C.~Pohl and J.~van Genderen}, {\em Structuring contemporary remote sensing
  image fusion}, Int. J. Image Data Fusion, 6 (2015), pp.~3--21.

\bibitem{RanchinWald2000}
{\sc T.~Ranchin and L.~Wald}, {\em Fusion of high spatial and spectral
  resolution images: {{The ARSIS}} concept and its implementation}, Photogramm.
  Eng. Remote Sens., 66 (2000), pp.~49--61.

\bibitem{ROF1992}
{\sc L.~Rudin, S.~Osher, and E.~Fatemi}, {\em Nonlinear total variation based
  noise removal algorithms}, Physica D, 60 (1992), pp.~259--268.

\bibitem{Schowengerdt2006}
{\sc R.~Schowengerdt}, {\em Remote Sensing: {{Models}} and Methods for Image
  Processing}, Academic Press, third~ed., 2006.

\bibitem{ShahYounan2008}
{\sc V.~Shah, N.~Younan, and R.~King}, {\em An efficient pan-sharpening method
  via a combined adaptive-{{PCA}} approach and contourlets}, IEEE Trans.
  Geosci. Remote Sens., 46 (2008), pp.~1323--1335.

\bibitem{Shensa1992}
{\sc M.~Shensa}, {\em The discrete wavelet transform: {{Wedding}} the {\`a}
  trous and {{Mallat}} algorithms}, IEEE Trans. Signal Process., 40 (1992),
  pp.~2464--2482.

\bibitem{Shettigara1992}
{\sc V.~Shettigara}, {\em A generalized component substitution technique for
  spatial enhancement of multispectral images using a higher resoltuion data
  set}, Photogramm. Eng. Remote Sens., 58 (1992), pp.~561--567.

\bibitem{ThomasRanchinWald2008}
{\sc C.~Thomas, T.~Ranchin, L.~Wald, and J.~Chanussot}, {\em Synthesis of
  multispectral images to high spatial resolution: {{A}} critical review of
  fusion methods based on remote sensing physics}, IEEE Trans. Geosci. Remote
  Sens., 46 (2008), pp.~1301--1312.

\bibitem{TuHuangHungChang2004}
{\sc T.-M. Tu, P.~Huang, C.-L. Hung, and C.-P. Chang}, {\em A fast
  intensity-hue-saturation fusion technique with spectral adjustment for
  {{IKONOS}} imagery}, IEEE Geosci. Remote Sens. Lett., 1 (2004), pp.~309--312.

\bibitem{TuSuShyu2001}
{\sc T.-M. Tu, S.-C. Su, H.-C. Shyu, and P.~Huang}, {\em A new look at
  {{IHS}}-like image fusion methods}, Inf. Fusion, 2 (2001), pp.~177--186.

\bibitem{VivoneAlparoneChanussot2015}
{\sc G.~Vivone, L.~Alparone, J.~Chanussot, M.~Dalla~Mura, A.~Garzelli,
  R.~Restaino, G.~Licciardi, and L.~Wald}, {\em A critical comparison among
  pansharpening algorithms}, IEEE Trans. Geosci. Remote Sens., 53 (2015),
  pp.~2565--2586.

\bibitem{VivoneRestaino2014}
{\sc G.~Vivone, R.~Restaino, M.~Mura, G.~Licciardi, and J.~Chanussot}, {\em
  Contrast and error-based fusion schemes for multispectral image
  pansharpening}, IEEE Geosci. Remote Sens. Lett., 11 (2014), pp.~930--934.

\bibitem{WaldRanchin2002}
{\sc L.~Wald and T.~Ranchin}, {\em Comment: {{Liu}} 'smoothing filter-based
  intensity modulation: A spectral preserve image fusion technique for
  improving spatial details'}, Int. J. Remote Sens., 23 (2002), pp.~593--597.

\bibitem{WaldRanchin1997}
{\sc L.~Wald, T.~Ranchin, and M.~Mangolini}, {\em Fusion of satellite images of
  different spatial resolutions: {{Assessing}} the quality of resulting
  images}, Photogramm. Eng. Remote Sens., 63 (1997), pp.~691--699.

\bibitem{WangBovik2002}
{\sc Z.~Wang and A.~Bovik}, {\em A universal image quality index}, IEEE Signal
  Process. Lett., 9 (2002), pp.~81--84.

\bibitem{Yocky1995}
{\sc D.~Yocky}, {\em Image merging and data fusion by means of the discrete
  two-dimensional wavelet transform}, J. Opt. Soc. Amer. A, 12 (1995),
  pp.~1834--1841.

\bibitem{ZhangFang2015}
{\sc G.~Zhang, F.~Fang, A.~Zhou, and F.~Li}, {\em Pan-sharpening of
  multi-spectral images using a new variational model}, Int. J. Remote Sens.,
  36 (2015), pp.~1484--1508.

\bibitem{ZhuBamler2013}
{\sc X.~Zhu and R.~Bamler}, {\em A sparse image fusion algorithm with
  application to pan-sharpening}, IEEE Trans. Geosci. Remote Sens., 51 (2013),
  pp.~2827--2836.

\end{thebibliography}

\end{document}